\documentclass[10pt,twocolumn,letterpaper]{article}

 \usepackage{cvpr}              %

\usepackage{graphicx}
\usepackage{booktabs}
\usepackage[T1]{fontenc}   %
\usepackage{lmodern}       %
\usepackage{microtype}
\usepackage{subcaption}

\usepackage[accsupp]{axessibility}

\usepackage{array}      %
\usepackage{longtable}  %
\usepackage{multirow}   %
\usepackage[export]{adjustbox}

\definecolor{cvprblue}{rgb}{0.21,0.49,0.74}
\usepackage[pagebackref,breaklinks,colorlinks,allcolors=cvprblue]{hyperref}

\usepackage{amsmath}
\usepackage{amssymb}
\usepackage{mathtools}

\usepackage[utf8]{inputenc} %
\usepackage[T1]{fontenc}    %
\usepackage{url}            %
\usepackage{booktabs}       %
\usepackage{amsfonts}       %
\usepackage{nicefrac}       %
\usepackage{microtype}      %
\usepackage{colortbl, xcolor}         %
\usepackage{multirow}
\usepackage{graphicx}
\usepackage{subcaption}
\usepackage{caption}
\usepackage{array}
\usepackage{makecell}
\usepackage{amssymb}
\usepackage{xspace}
\usepackage{amsmath}
\usepackage[table]{xcolor}
\usepackage{bbm}
\usepackage{pifont}
\usepackage{makecell}
\usepackage{enumitem} %
\usepackage[export]{adjustbox}
\usepackage{tikz}
\usetikzlibrary{calc}
\usepackage{mwe} %
\usepackage{comment}

\usepackage[capitalize,noabbrev]{cleveref}

\newcommand{\method}[1]{{NewtPhys}}

\definecolor{darkmagenta}{RGB}{186, 0, 186}
\definecolor{slateblue}{RGB}{90, 70, 230}

\definecolor{todo}{HTML}{E15554}
\definecolor{author1}{HTML}{40A050}
\definecolor{author2}{HTML}{C33C54}
\definecolor{author3}{HTML}{E6AA68}
\definecolor{author4}{HTML}{0F4893}

\definecolor{material}{HTML}{2BAE27}
\definecolor{mechanics}{HTML}{FF5733}
\definecolor{spatial}{HTML}{3498DB}
\definecolor{permanence}{HTML}{F43FC7}
\definecolor{temporal}{HTML}{0DA792}
\definecolor{viewpoint}{HTML}{EEAC32}

\newcommand{\materialCol}[1]{\textcolor{material}{#1}}
\newcommand{\mechanicsCol}[1]{\textcolor{mechanics}{#1}}
\newcommand{\spatialCol}[1]{\textcolor{spatial}{#1}}
\newcommand{\permanenceCol}[1]{\textcolor{permanence}{#1}}
\newcommand{\temporalCol}[1]{\textcolor{temporal}{#1}}
\newcommand{\viewpointCol}[1]{\textcolor{viewpoint}{#1}}

\newcommand{\material}{\materialCol{material understanding}}
\newcommand{\Material}{\materialCol{Material understanding}}
\newcommand{\mechanics}{\mechanicsCol{mechanics}}
\newcommand{\Mechanics}{\mechanicsCol{Mechanics}}
\newcommand{\spatial}{\spatialCol{spatial reasoning}}
\newcommand{\Spatial}{\spatialCol{Spatial reasoning}}
\newcommand{\permanence}{\permanenceCol{permanence}}
\newcommand{\Permanence}{\permanenceCol{Permanence}}
\newcommand{\temporal}{\temporalCol{temporal reasoning}}
\newcommand{\Temporal}{\temporalCol{Temporal reasoning}}
\newcommand{\viewpoint}{\viewpointCol{viewpoint}}
\newcommand{\Viewpoint}{\viewpointCol{Viewpoint}}

\newcommand{\condenseparagraph}[1]{{\vspace*{0.1em}\noindent\textbf{#1}\quad}}

\crefname{section}{Sec.}{Secs.}
\Crefname{section}{Section}{Sections}
\Crefname{table}{Table}{Tables}
\crefname{table}{Tab.}{Tabs.}

\makeatletter
\DeclareRobustCommand\onedot{\futurelet\@let@token\@onedot}
\def\@onedot{\ifx\@let@token.\else.\null\fi\xspace}

\def\eg{\emph{e.g}\onedot} 
\def\ie{\emph{i.e}\onedot} 
\def\cf{\emph{cf}\onedot} 
\def\etc{\emph{etc}\onedot} \def\vs{\emph{vs}\onedot}
\def\wrt{w.r.t\onedot}

\definecolor{best}{HTML}{C6EFCE}    %
\definecolor{second}{HTML}{FFF2CC}  %

\definecolor{pgrade1}{RGB}{255,204,204}  %
\definecolor{pgrade2}{RGB}{255,221,204}  %
\definecolor{pgrade3}{RGB}{255,238,204}  %
\definecolor{pgrade4}{RGB}{255,255,204}  %
\definecolor{pgrade5}{RGB}{229,255,204}  %
\definecolor{pgrade6}{RGB}{204,255,204}  %
\definecolor{pgrade7}{RGB}{178,242,187}  %
\definecolor{pgrade8}{RGB}{153,230,170}  %
\definecolor{pgrade9}{RGB}{102,204,136}  %
\definecolor{pgrade10}{RGB}{51,153,102}  %

\definecolor{corr0}{RGB}{255,255,255}  %
\definecolor{corr1}{RGB}{229,245,233}
\definecolor{corr2}{RGB}{204,232,207}
\definecolor{corr3}{RGB}{168,221,181}
\definecolor{corr4}{RGB}{123,204,196}
\definecolor{corr5}{RGB}{78,179,211}
\definecolor{corr6}{RGB}{43,140,190}
\definecolor{corr7}{RGB}{29,116,155}
\definecolor{corr8}{RGB}{20,90,120}
\definecolor{corr9}{RGB}{0,68,89}      %

\def\paperID{*****} %
\def\confName{CVPR}
\def\confYear{2026}

\title{\method{}: Do Foundation Models Understand Newtonian Physics?} 

\author{
	Sebastian Cavada\textsuperscript{3,*} \, 
	Soumava Paul\textsuperscript{1} \,
	Tuan-Hung Vu\textsuperscript{1,2} \,
	Andrei Bursuc\textsuperscript{1,2} \,
	Raoul de Charette\textsuperscript{1,*} \\
	\\
	\textsuperscript{1}Inria, \quad\textsuperscript{2}Valeo.ai, \quad\textsuperscript{3}MBZUAI\\
	\textsuperscript{*}These authors contributed equally
}

\newcommand{\InsetY}{0.011}
\newcommand{\InsetLeftFrac}{0.44}  %
\newcommand{\InsetW}{0.15}
\pgfmathsetmacro{\InsetX}{\InsetLeftFrac + 0.5*\InsetW}
\newcommand{\InsetOpacity}{1.0}

\makeatletter
\apptocmd{\@maketitle}{
	\centering
    \vspace{-1.5em}\url{https://astra-vision.github.io/NewtPhys}\\[0.5em]
	\begin{minipage}{\textwidth}
		\centering
		\newcommand{\TeaserLeftW}{0.7\linewidth}
		\newcommand{\TeaserRightW}{0.2336\linewidth}
		\newcommand{\TeaserBigH}{8cm}
		
		\begin{minipage}[t]{\TeaserLeftW}
			\vspace{0pt} %
			\centering
			\begin{tikzpicture}[baseline=(big.north)] %
				\node[
				anchor=north west,
				inner sep=0,
				outer sep=0
				] (big) at (0,0) {%
					\includegraphics[width=\linewidth, trim=0 0 0 0, clip]{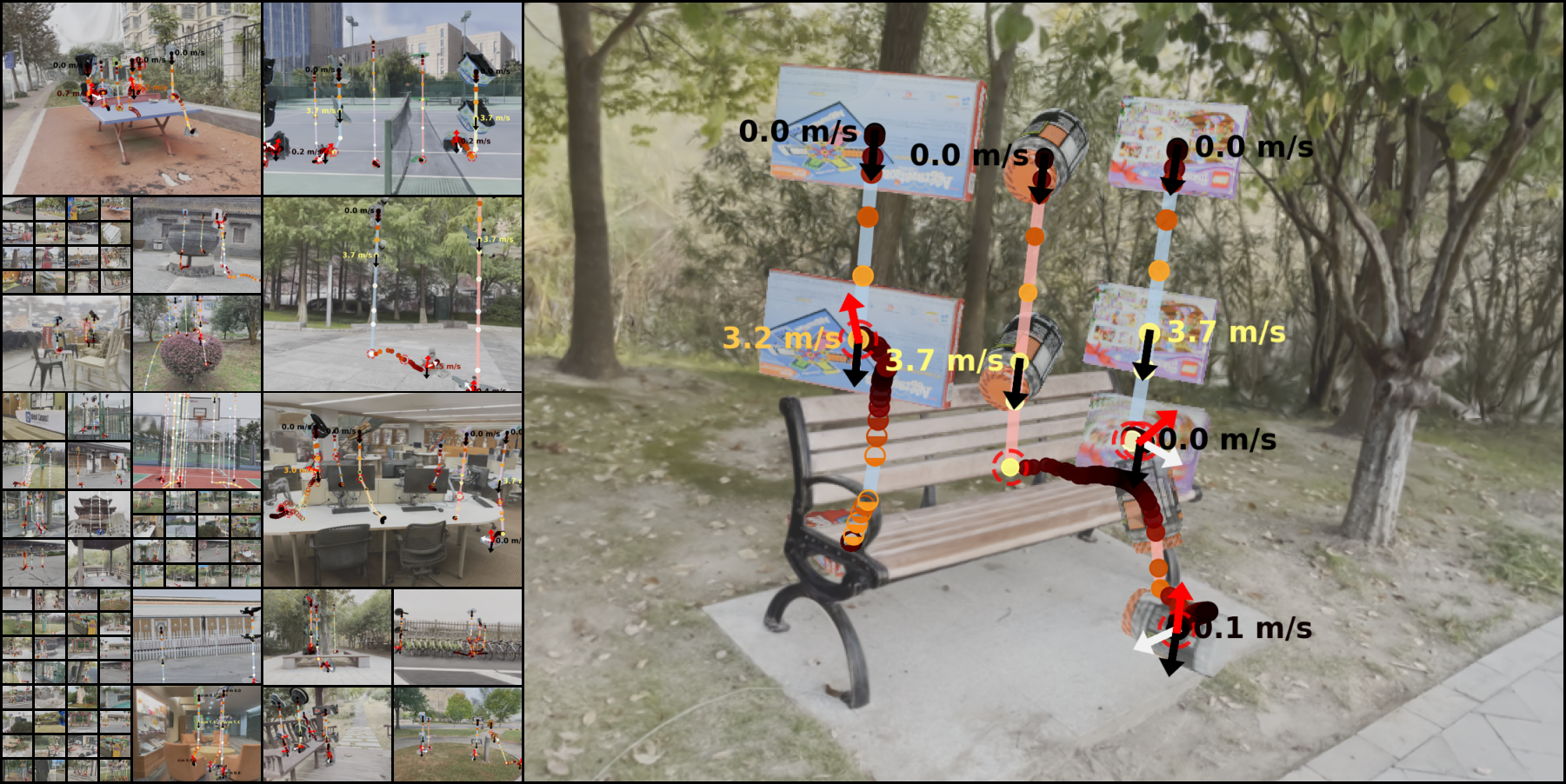}%
				};
				
				\pgfmathsetlengthmacro{\InsetYShift}{\InsetY*\TeaserBigH}
				
				\node[
				anchor=south,
				inner sep=0,
				outer sep=0,
				opacity=\InsetOpacity,
				xshift=-8mm,
				yshift=\InsetYShift
				] at ($(big.south west)!\InsetX!(big.south east)$) {%
					\includegraphics[width=\InsetW\linewidth]{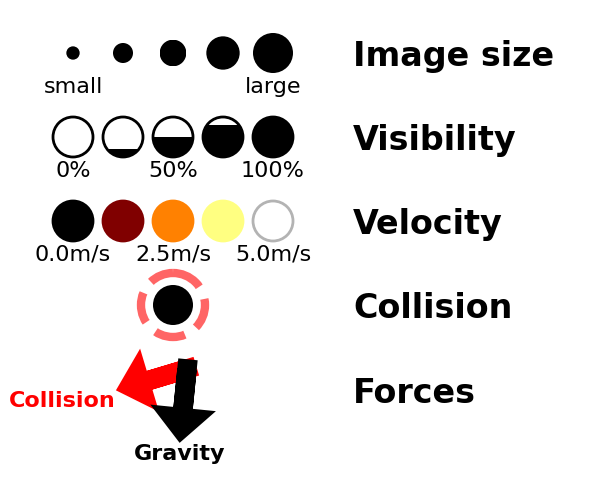}%
				};
			\end{tikzpicture}%
		\end{minipage}\hspace{0.5em}%
		\begin{minipage}[t]{\TeaserRightW}
			\vspace{0pt}
			\begin{flushright}
				\includegraphics[width=\linewidth]{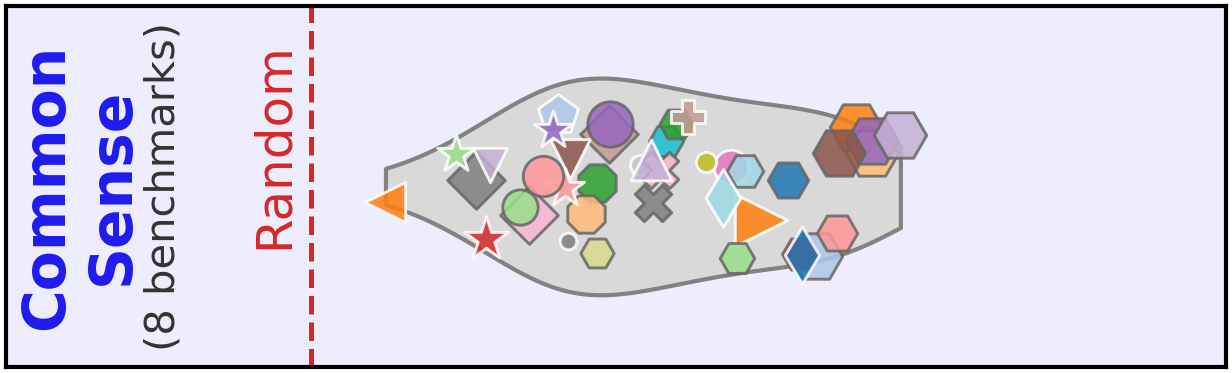}\\
				\includegraphics[width=\linewidth]{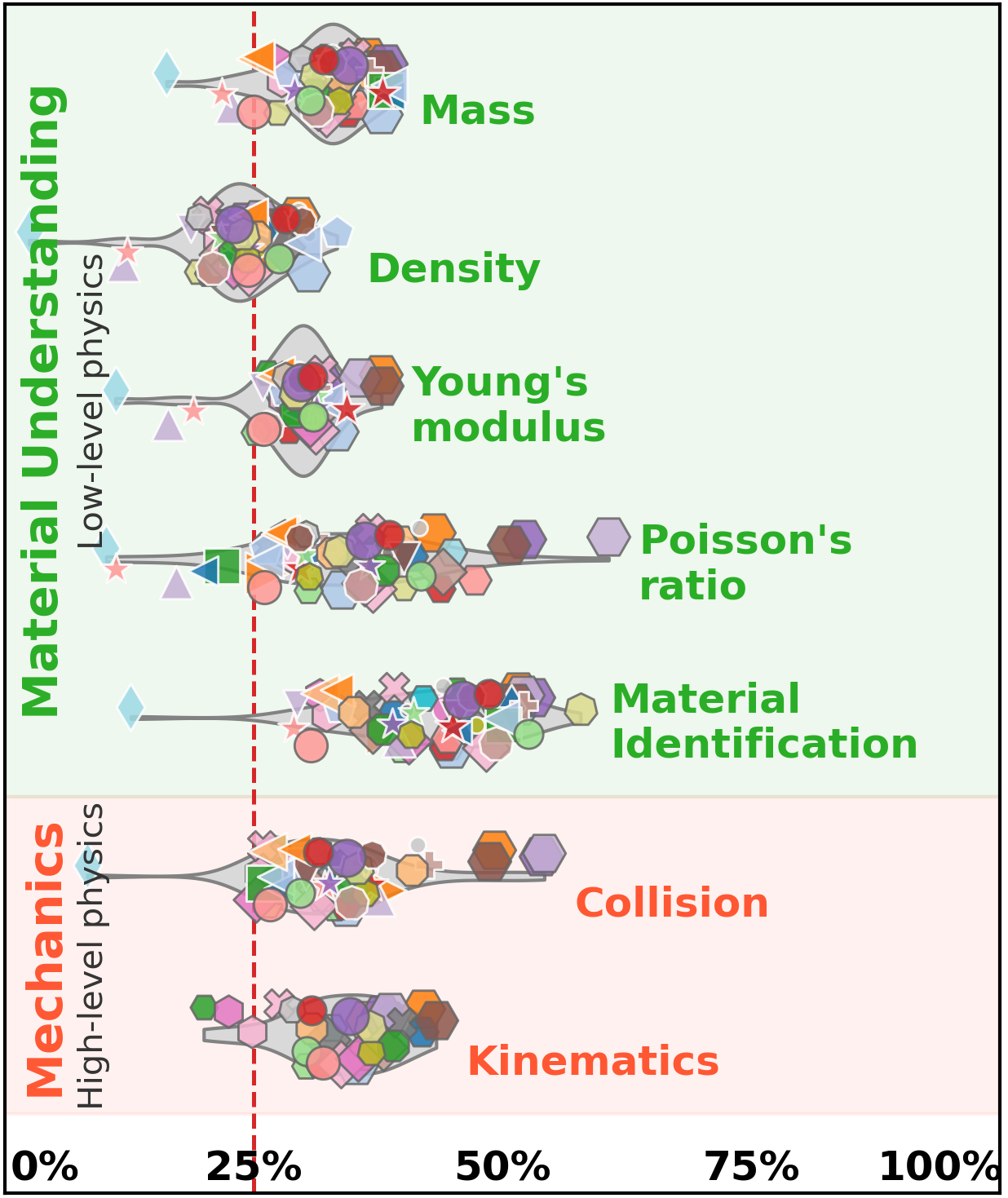}
			\end{flushright}
		\end{minipage}
        \vspace{-0.5em}
		\captionof{figure}{
			\textbf{\method{}} studies low-level physics understanding in vision-language models (VLMs) and vision foundation models (VFMs), revealing gaps in Newtonian reasoning. We introduce a physically annotated dataset spanning diverse scenes, objects (\textit{rigid and soft}) and dynamic interactions captured from static or moving cameras. It provides 2D and 3D annotations in metric units, including materials, kinematics, dynamics, camera parameters, and collision events. \textbf{Left}: We show exemplar scenes with selected overlays (real-world velocity, visibility, collisions, forces). \textbf{Right}: \method{} enables large-scale benchmarking of 56 VLMs{, including 54 open-weight models and 2 closed-source frontier models}, demonstrating that despite reasonable performance on commonsense tasks (top) they struggle to capture Newtonian physics.
		}
		\label{fig:teaser}
	\end{minipage}
	\vspace{1em}
}{}{}
\makeatother

\usepackage{titletoc}

\begin{document}
\captionsetup{hypcap=false}
\maketitle
\begin{abstract}
	Previous work has evaluated physics reasoning in foundation models using synthetic or semi-synthetic scenes and visual question-answering tasks.
	However, these benchmarks emphasize high-level events and lack the visual fidelity required to assess true low-level Newtonian understanding.
	We introduce \textbf{\method{}}, a 4D physically annotated dataset built from multiview images of real-world scenes with physics-grounded simulations.
	The dataset provides dense, fine-grained annotations across timesteps --- including 3D forces and {amodal per-pixel quantities covering physics, tracking, semantics and geometry} --- bridging the gap between simplistic synthetic setups and realistic visual complexity.
	Using \method{}, we systematically evaluate 56 VLMs, {including 54 open-weight models and 2 closed-source frontier models}, and 10 VFMs and reveal limitations in low-level physics reasoning.
	Beyond benchmarking, our dataset enables future research in physics-grounded vision and the development of next-generation physics-aware evaluations.
    Code and datasets are available at: \href{https://astra-vision.github.io/NewtPhys}{\texttt{https://astra-vision.github.io/NewtPhys}}.
\end{abstract}

\section{Introduction}

Humans develop intuitive Newtonian models early in life~\cite{baillargeon1994physical}, enabling prediction of motion, contact, and gravity, which allow them to navigate the physical world and solve complex visual tasks~\cite{mccloskey1983intuitive,carey2000origin}.
Building vision systems with comparable physical grounding remains a key goal, especially for embodied agents.

Recent Vision-Language Models (VLMs) and Vision Foundation Models (VFMs) excel at recognition and open-ended reasoning, raising the question of whether they truly understand the causal structure of physics or merely rely on correlations. However, common evaluations based on spatial awareness or 3D reasoning do not directly test physics awareness, and whether they understand low-level quantities such as forces, collisions, gravity, or deformation~\cite{zhan2024general,el2024probing} is yet unanswered. Existing benchmarks face a trade-off: realistic datasets rely on human judgments or high-level proxy labels and therefore do not provide physics labels, while benchmarks with explicit Newtonian annotations use simplified synthetic worlds~\cite{yi2020clevrer,riochet2021intphys,bear2021physion} or depict simplistic real-world dynamics in lab conditions~\cite{zhang2025morpheus}.

To bridge this gap, we introduce \method{}, a 4D benchmark for low and high-level Newtonian understanding in realistic settings. While physical annotation is impractical outside lab conditions, \method{} uses 3D Gaussian Splatting (3DGS)~\cite{kerbl20233d} to combine real-world scenes and objects as simulatable particles of a Newtonian simulator. Our benchmark not only captures the visual appearance as videos, but also pixel-aligned physical labels and events over time, including forces, deformation fields, material and instance maps, and scene flow, illustrated in~\cref{fig:teaser}.  We use \method{} to conduct a large-scale study on 56 Vision Language Models (VLMs) and 10 Vision Foundation Models (VFMs).

For VLMs, we create a physics-grounded Visual Question Answering (VQA) dataset, covering 141K question-answer pairs, across six physics understanding categories. 
Our extensive study shows that although advanced VLMs continue to grow in scale and capability alongside the evolution of LLMs, progress in physics understanding are surprisingly limited. {While frontier models continue to dominate the benchmark, large-scale open-source models are narrowing the {performance gap.}}
While these models can often identify materials, they struggle to estimate granular Newtonian properties like density or mass, suggesting that VLMs are not yet capturing the causal mechanics required for true physical world modeling.
Our study leads to two key observations. First, VLMs tend to ignore visual cues and rely on easier shortcuts derived from priors learned by LLMs. Second, most popular commonsense benchmarks correlate weakly with physics understanding, further hindering progress in physical reasoning.
Benefiting from \method{} pixel-wise physics annotation, we also benchmark 10 vision-only VFMs, introducing the task of \emph{Physics Probing} to evaluate whether Newtonian signals are accessible in the learned representations.
While stronger visual representations generally improve Physics Probing performance, our results suggest that VFMs are not inherently physics-grounded and may fail at reliably capturing physically meaningful signals.\\[-1em]

\noindent{}Our contributions are fourfold:
\begin{itemize}
	\item \method{}, a benchmark combining real 3DGS scenes with Newtonian simulation to produce rendered videos with dense pixel-aligned physical annotations.
	\item A point-based simulation pipeline supporting multi-object interactions and soft-body dynamics in realistic scenes.
	\item A large-scale dataset includes 141K VQA spanning six categories and 730K frames {with 11 ground-truth maps}, enabling statistically meaningful evaluation.
	\item Comprehensive evaluations of {66 VFMs/VLMs} revealing current limitations in low-level Newtonian physics understanding.
\end{itemize}

\section{Related works}
The idea that humans rely on an internal, approximate model of physics to understand and predict the visual world has long been studied in cognitive science under the umbrella of intuitive physics \cite{mccloskey1983intuitive, baillargeon1994physical, carey2000origin}. This perspective has strongly influenced machine learning and vision, where early and foundational work framed physical reasoning as inference over latent variables and object-centric world models \cite{tenenbaum2011grow, battaglia2013simulation, battaglia2018relational}. In parallel, others explored how visual representations can capture causal structure and physical regularities from data \cite{isola2015learning, zhu2015understanding}.

A substantial body of work aims to learn physical laws or latent dynamics from visual input, typically in simplified environments~\cite{wu2015galileo,wu2016physics}. 
These works demonstrate that structured physical representations can be learned, but they largely operate in toy 2D or simple 3D worlds with limited visual and physical complexity.
Some benchmarks study physical reasoning in richer scenes, including event prediction, counterfactual reasoning, and causal queries~\cite{yi2020clevrer, riochet2021intphys, bear2021physion}. 
However, physics annotations in these datasets is still mostly symbolic or event-level. 

\begin{figure*}[t!]
	\centering
    \newcommand{\samplemethod}[5]{%
        \begin{minipage}[m]{\linewidth}%
            \begin{tikzpicture}
                \node (img) [inner sep=0] {%
                    \adjustimage{clip,#4,width=\linewidth, height=0.9cm}{#5}%
                };
                \node[
                anchor=north east,            %
                fill=white,                   %
                fill opacity=0.65,             %
                text=black,                   %
                inner sep=1pt,                %
                font=\fontsize{3}{3}\selectfont\bfseries,          %
                align=center,
                text width=\linewidth 
                ] at (img.north east) {#1, #3};
            \end{tikzpicture}%
        \end{minipage}%
    }
    \begin{subfigure}{0.33\linewidth}%
        \resizebox{\linewidth}{!}{%
            \setlength{\tabcolsep}{0.2pt}%
            \begin{tabular}{p{0.25\linewidth}p{0.25\linewidth}p{0.25\linewidth}p{0.25\linewidth}}%
                \multicolumn{4}{c}{\scriptsize\textbf{High-level understanding}}\\
                \samplemethod{IntPhys}
                {riochet2021intphys}
                {2021}
                {trim=0 {\dimexpr0.2\height\relax} 0 0}
                {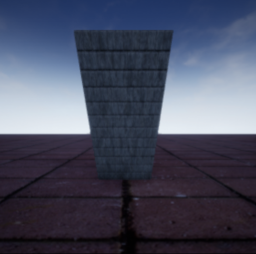}&
                \samplemethod{IntPhys2}
                {bordes2025intphys}
                {2025}
                {trim=0 {\dimexpr0.1\height\relax} 0 {\dimexpr0.1\height\relax}}
                {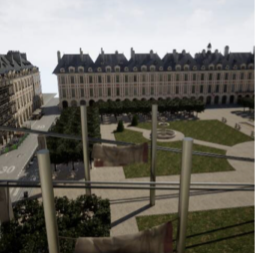}&
                \samplemethod{PhysBench}
                {chow2025physbench}
                {2025}
                {}
                {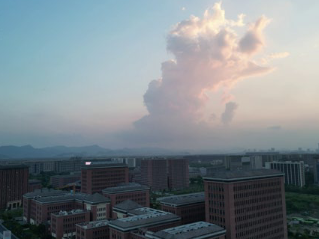}&
                \samplemethod{PhysicsIQ}
                {motamed2025generative}
                {2025}
                {trim={\dimexpr0.15\width\relax} 0 0 0}
                {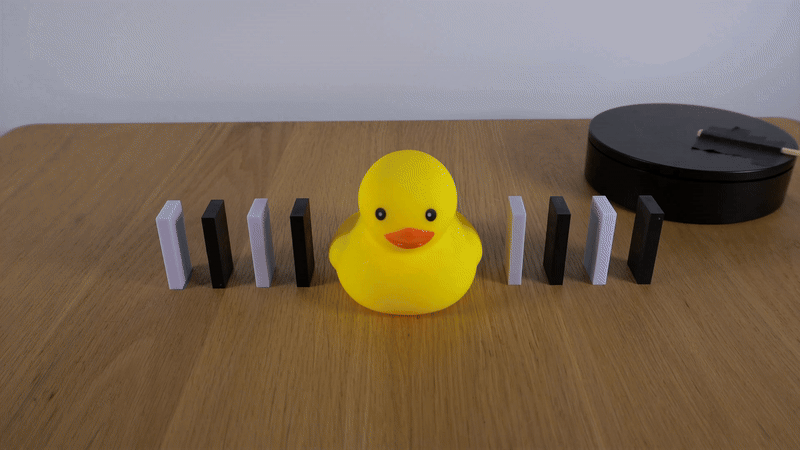}\\%
        \end{tabular}}
    \end{subfigure}\hfill%
    \begin{subfigure}{0.65\linewidth}%
        \resizebox{\linewidth}{!}{%
            \setlength{\tabcolsep}{0.2pt}
            \begin{tabular}{p{0.125\linewidth}p{0.125\linewidth}p{0.125\linewidth}p{0.125\linewidth}p{0.125\linewidth}p{0.125\linewidth}p{0.125\linewidth}p{0.125\linewidth}}
                \multicolumn{8}{c}{\scriptsize\textbf{Low-level understanding}}\\
                \samplemethod{Physics 101}
                {wu2016physics}
                {2016}
                {trim=0 {\dimexpr0.25\height\relax} 0 0}
                {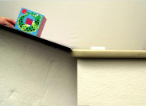}&%
                \samplemethod{Physion}
                {bear1physion}
                {2021}
                {trim=0 0 0 0}
                {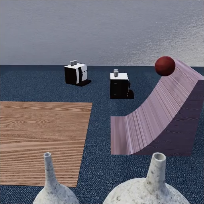}&%
                \samplemethod{CompPhy}
                {chen2022comphy}
                {2022}
                {trim={\dimexpr0.20\width\relax} 0 0 0}
                {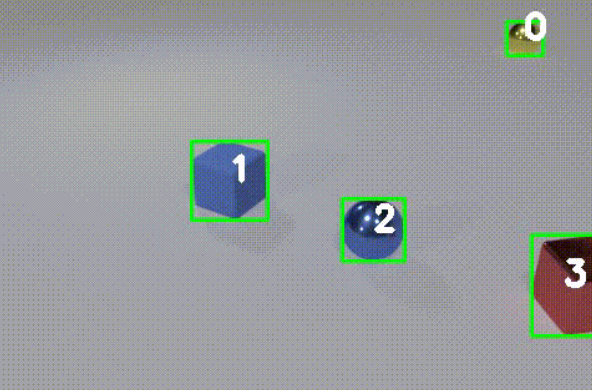}&%
                \samplemethod{CRIPPVQA}
                {patel2022cripp}
                {2022}
                {trim={\dimexpr0.15\width\relax} 0 0 0}
                {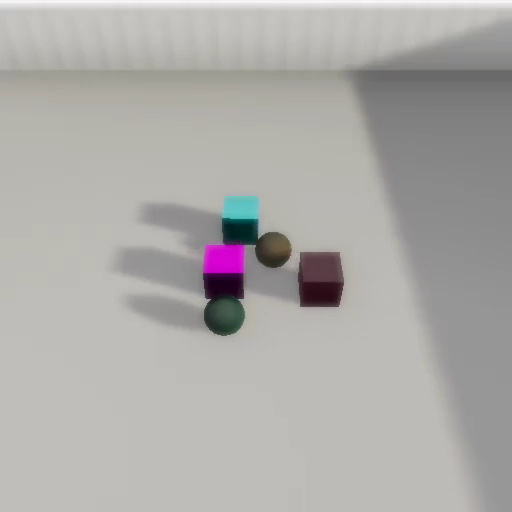}&%
                \samplemethod{ContPhy}
                {zheng2024contphy}
                {2024}
                {trim={\dimexpr0.30\width\relax} 0 0 0}
                {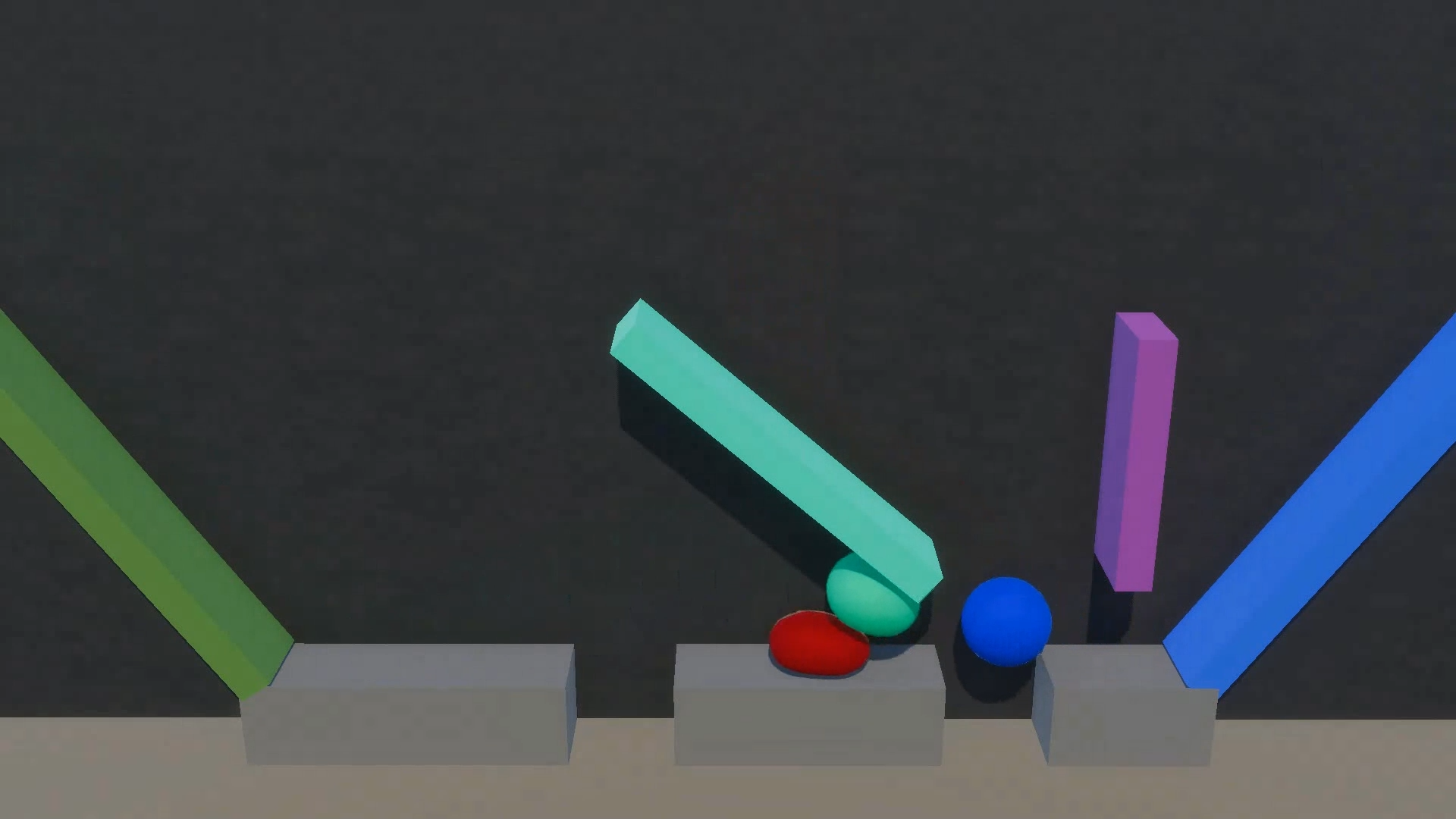}&%
                \samplemethod{PhysBench*}
                {chow2025physbench}
                {2025}
                {trim=0 {\dimexpr0.25\height\relax} 0 0 0}
                {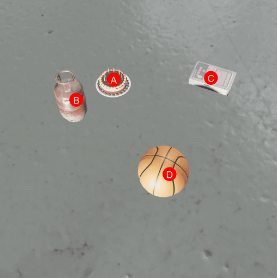}&
                \samplemethod{Physion++}
                {tung2023physion++}
                {2023}
                {trim={\dimexpr0.15\width\relax} 0 0 0}
                {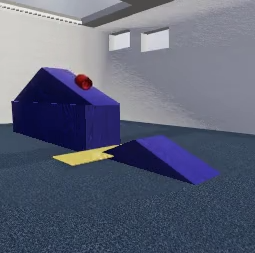}&%
                \samplemethod{PhyWorld}
                {phyworld}
                {2025}
                {trim=0 0 {\dimexpr0.15\width\relax} 0}
                {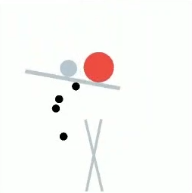}\\
        \end{tabular}}
    \end{subfigure}\\[0.5mm]
    \centerline{\rule{0.3\linewidth}{0.4pt}}\vspace{0.5mm}
    \begin{subfigure}{1\linewidth}
        \resizebox{\linewidth}{!}{%
            \setlength{\tabcolsep}{1pt}%
            \setlength{\baselineskip}{2pt}%
            \begin{tabular}{cccccc}%
                \includegraphics[width=0.16\linewidth,height=2cm,keepaspectratio]{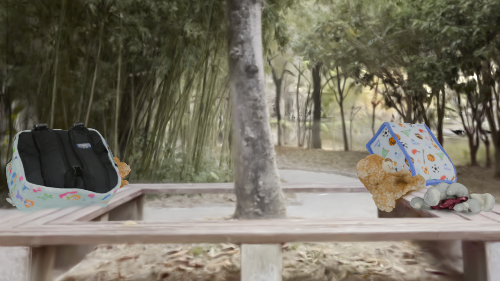}&%
                \includegraphics[width=0.16\linewidth,height=2cm,keepaspectratio]{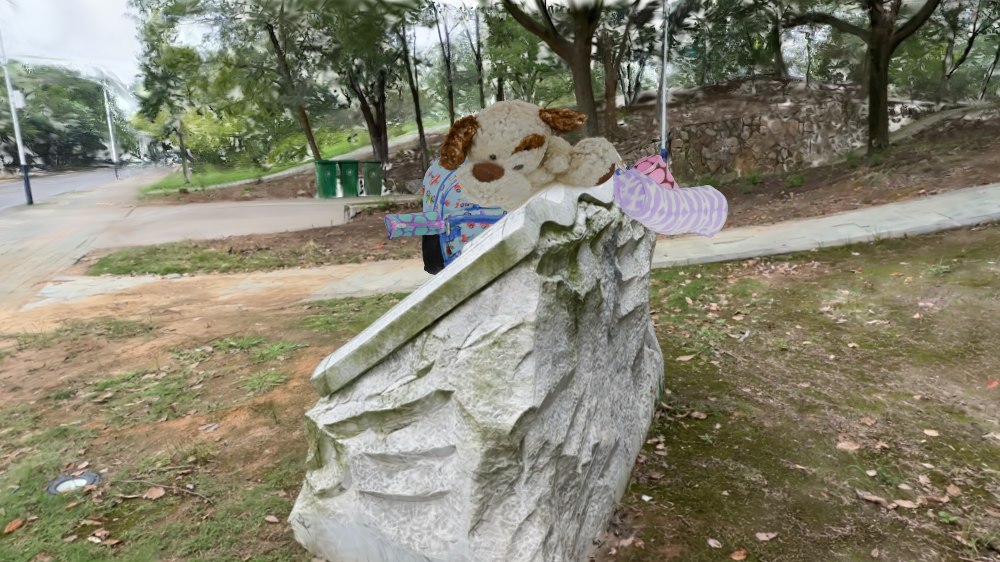}&%
                \includegraphics[width=0.16\linewidth,height=2cm,keepaspectratio]{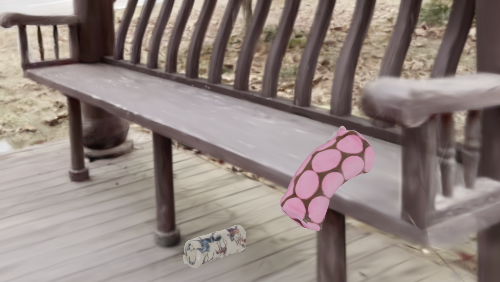}&%
                \includegraphics[width=0.16\linewidth,height=2cm,keepaspectratio]{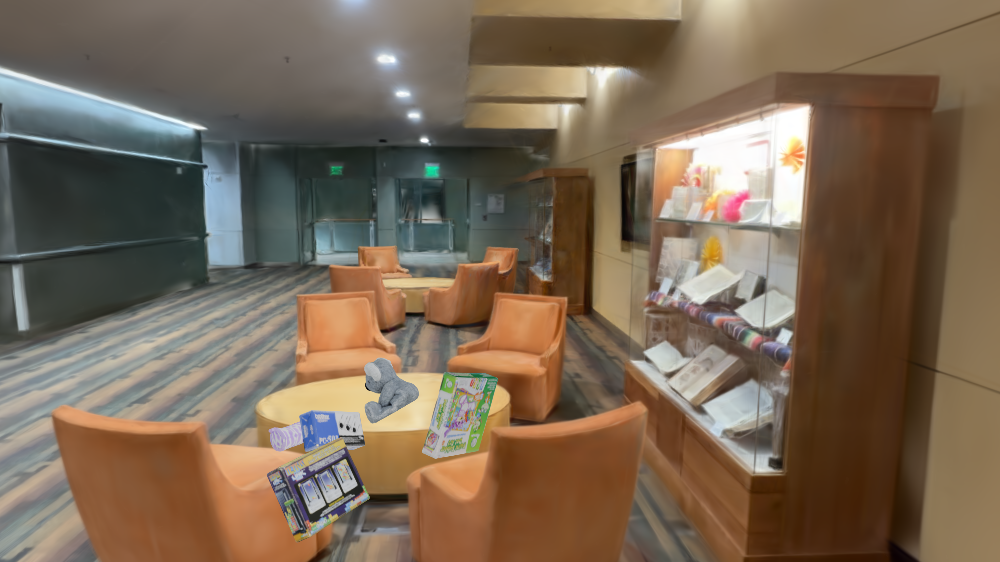}&%
                \includegraphics[width=0.16\linewidth,height=2cm,keepaspectratio]{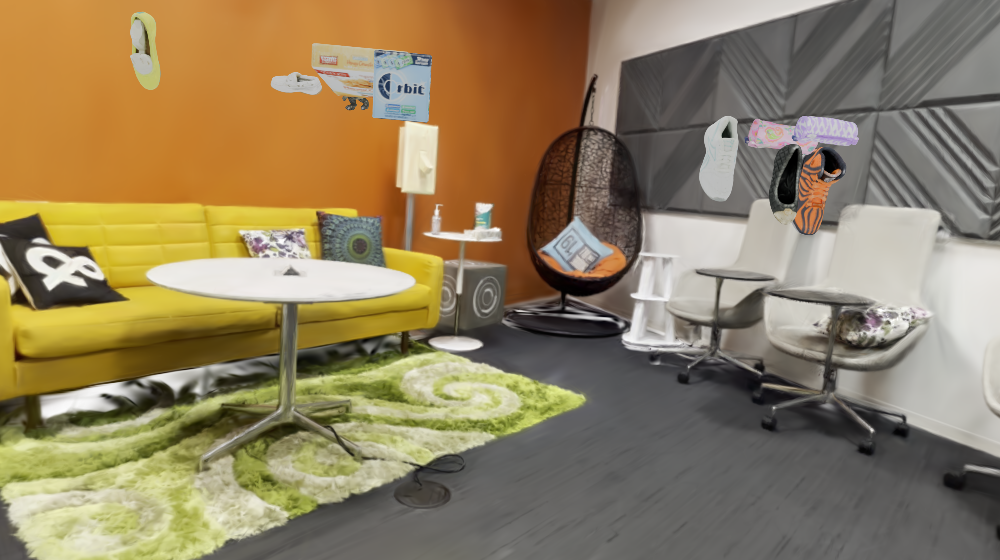}&%
                \includegraphics[width=0.16\linewidth,height=2cm,keepaspectratio]{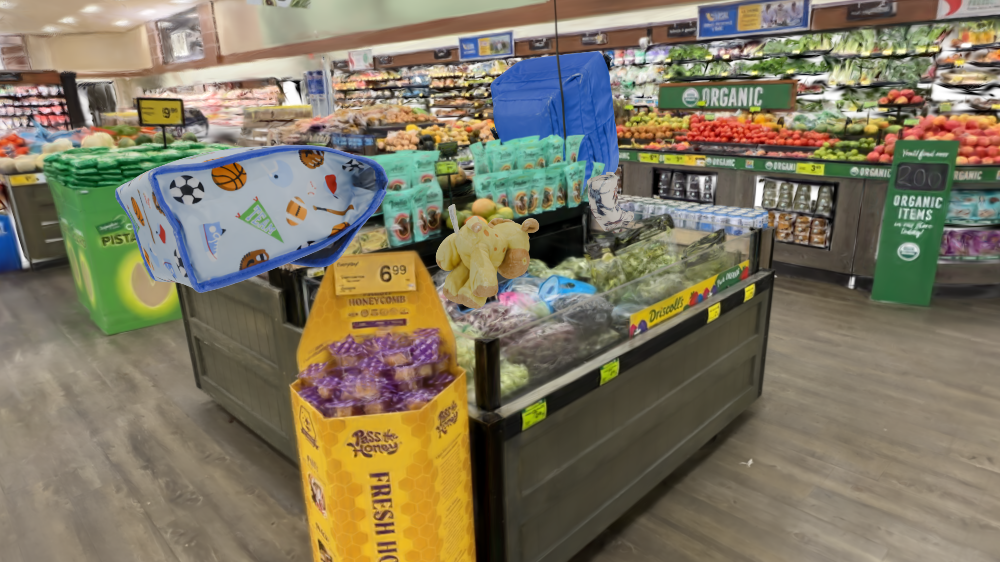}\\[-2mm]%
                \parbox[c]{0.16\linewidth}{\centering\fontsize{4.5}{5.5}\selectfont 5 objects colliding (crop)}&\parbox[c]{0.16\linewidth}{\centering\fontsize{4.5}{5.5}\selectfont 4 soft objects}&\parbox[c]{0.16\linewidth}{\centering\fontsize{4.5}{5.5}\selectfont Flexible pencil cases (crop)}&\parbox[c]{0.16\linewidth}{\centering\fontsize{4.5}{5.5}\selectfont 5 objects interacting}&\parbox[c]{0.16\linewidth}{\centering\fontsize{4.5}{5.5}\selectfont 10 objects in free fall}&\parbox[c]{0.16\linewidth}{\centering\fontsize{4.5}{5.5}\selectfont Multi-way interactions}\\[-1mm]
                \multicolumn{6}{c}{\scriptsize\textbf{\method{}} (high- and low-level understanding)}\\
            \end{tabular}%
        }
        \vspace{-1em}
    \end{subfigure}
    \caption{\textbf{Benchmarks for physical understanding.} {High-level physical understanding benchmarks (\eg, event ordering, general physics, frame reconstruction) are typically more realistic than low-level ones, which rely on toy simulators. In contrast, \method{} provides a realistic benchmark for both high- and low-level physical understanding in real-world scenes.}
    }
    \label{fig:qual_comparison}
\end{figure*}
		
Recent benchmarks move towards more realistic data and foundation model evaluation.
Physics-IQ evaluates generative video models based on physical plausibility judged by learned or human critics~\cite{motamed2025generative} while PhysBench evaluates VLMs on image–video–text questions about physical situations in real-world content \cite{chow2025physbench}.
NewtonGen and related datasets test qualitative or counterfactual physical reasoning \cite{Yuan_2025_NewtonGen}. 
While these benchmarks reveal important limitations of current models, they do not provide ground-truth Newtonian quantities and therefore cannot determine whether models represent forces, stresses, or contact patterns internally, rather than relying on high-level visual or statistical cues.
Some benchmarks target more diagnostic or application-oriented settings, such as PhysToolBench \cite{zhang2025phystoolbench}, PISA \cite{li2025pisa}, or Morpheus~\cite{zhang2025morpheus}. There also exist works on deformable or soft-body simulation environments~\cite{tung2023physion++}.

To our knowledge, {no benchmark provides dense}, pixel-aligned supervision of forces and deformations in visually realistic scenes. \method{} is designed to fill this gap: it provides per-pixel physical fields (forces, collisions, deformations, scene flow) aligned with rendered videos, enabling direct evaluation of whether vision models encode low-level Newtonian signals rather than only high-level physical plausibility.
{\cref{fig:qual_comparison} visualizes existing benchmarks\footnote{{Note PhysBench shows large variability in visual realism, from high-level (photographic) to low-level (toy-simulation) physics.}} (top) and {\method{} (bottom) which, in contrast, has realistic renderings and rich physical interactions for low- \textit{and} high- physics understanding.}}

\section{The \method{} benchmark}
\label{sec:bench}

\begin{figure*}[t]
	\centering
	\includegraphics[width=1.0\linewidth]{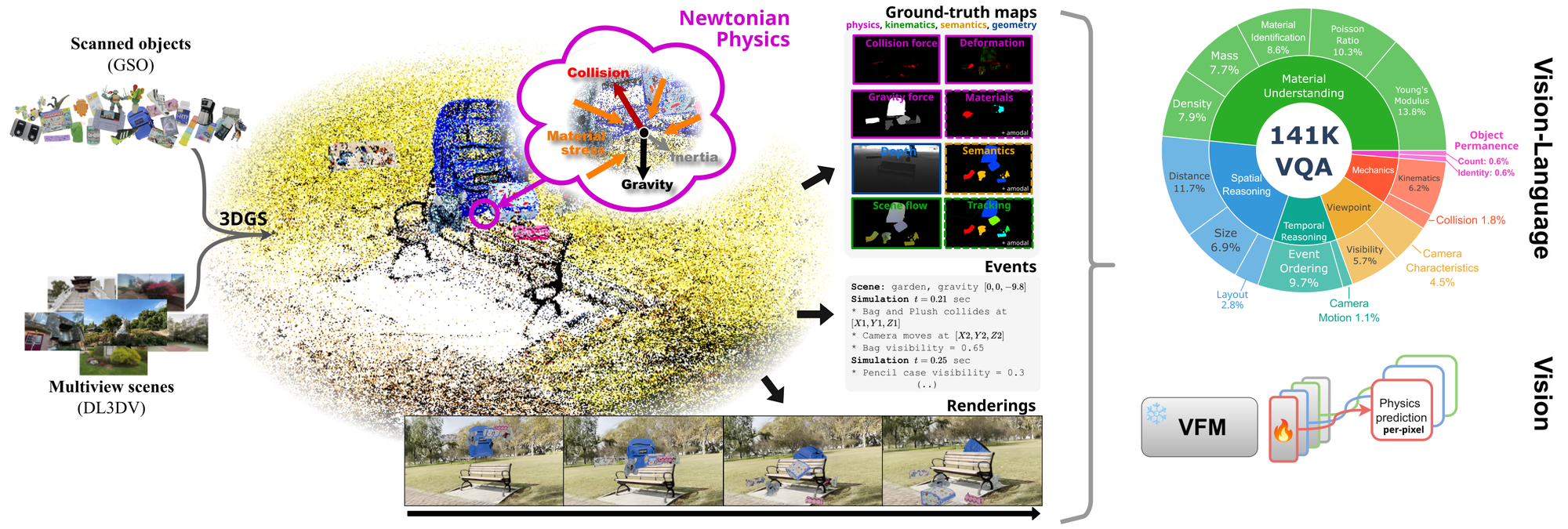}
	\caption{\textbf{Dataset construction.} We construct arbitrary scenarios from spawning up to ten objects (GSO~\cite{downs2022google}) into various scenes (DL3DV~\cite{ling2024dl3dv}). 
		The resulting 3DGS primitives serve as simulatable particles in Simplicits physical simulator~\cite{modi2024simplicits}, which we highly customize to exhaustively capture \textit{Newtonian forces in time and space}. Besides realistic renderings, the pipeline outputs {11 ground truth maps capturing pixel-level physics, kinematics, semantics and geometry (inc. 3 amodal not visualized); as well as frame/scene-level events capturing} collisions, forces, semantics, \etc. 
		Notably, it covers a wide range of material structures, such as rigid and deformable objects (notice how the blue bag compresses as it collides with the bench).
		\method{} encompasses 11k sequences totaling 730k frames. To evaluate Newtonian physics in VLMs we generate 141k VQA (top right) and use maps to assess pixel-level understanding of vision models (bottom right).
	}
	\label{fig:overview}
\end{figure*}

With \method{}, illustrated in~\cref{fig:overview}, we propose to augment real-world scenes as 3D Gaussian Splatting (3DGS) in a controllable fashion while recording Newtonian physics events (collisions, free fall, deformation, \etc) with force labels so as to study how vision models understand low-level physics.

\subsection{Dataset creation} 
\label{sec:bench_dataset}
Our pipeline uses 3DGS to augment real-world scenes~\cite{ling2024dl3dv} with scanned objects~\cite{downs2022google}, and processes the resulting representation as simulable particles in a Newtonian simulator for realistic dynamics.
Since existing mesh-free simulators are typically designed for single object and do not permit per point forces retrieval, we put significant effort to extend the Simplicits~\cite{modi2024simplicits} simulator to handle large scenes ($10^6$--$10^8$ particles) and enable the recording of individual forces acting at each position in space and time.
Such a pipeline allows us to script arbitrary scenarios (\eg, an object falling on a bench, a dozen of fluffy teddy bears colliding with a statue) covering a large range of dynamics, while producing 4D sequences realistic in both visual appearance, dynamics and camera motion.
Simulations also output a scenario description (physics properties, camera motion, \etc), world state with high-level events at each simulation step (\eg, kinematics, objects interaction, \etc.), as well as {ground-truth maps capturing \textit{per-pixel} physical phenomena (gravity, collision, deformation, materials) along with kinematics, semantics and geometry.}

\condenseparagraph{Scenario definition.} We construct the initial simulation state by combining real-world scene representations from DL3DV~\cite{ling2024dl3dv} with everyday objects from Google Scanned Objects (GSO)~\cite{downs2022google}, both represented as dense 3D Gaussian Splats (3DGS). Since splats serve as particles in our physical simulator, the 3DGS must be geometrically aligned, metrically scaled, and expressed in a shared canonical coordinate system with gravity along the negative $Z$ axis. Misalignment or scale inconsistencies lead to physically invalid interactions.

For DL3DV scenes, we recover camera poses using COLMAP~\cite{schoenberger2016sfm} and estimate metric scale via external priors, including known object dimensions or monocular depth predictions~\cite{depthanything3}, optionally refined for consistency. A canonical frame is enforced by aligning the reconstructed point cloud such that the ground plane lies at $Z=0$. To obtain dense and metrically consistent 3DGS, we align dense VGGT point clouds~\cite{wang2025vggt} to the metric COLMAP reconstruction using Kabsch–Umeyama alignment~\cite{umeyama2002least}, and initialize 3DGS optimization from the merged point cloud. For objects, we convert textured GSO meshes into metrically scaled 3DGS by rendering multi-view images in Blender and optimizing splats from these views.

Finally, scene and object splats are merged into a unified 3DGS representation. Objects are randomly placed within predefined regions of interest in each scene to generate diverse physically plausible interactions.

\condenseparagraph{Newtonian physics simulation.} Simulating physics directly on 3DGS is challenging due to their unstructured and sparse nature, which makes mesh-based simulators brittle and unstable under slight geometric noise. Instead, we adopt a mesh-free formulation based on Simplicits~\cite{modi2024simplicits}, which models deformable dynamics in a reduced deformation space and avoids explicit surface reconstruction.

While originally designed for single-object simulation, we extend this formulation to full scenes by treating 3DGS centers as particles and defining a joint reduced state over all objects. We dynamically allocate deformation handles based on object softness and restrict physical evaluation to strategically sampled cubature points near interaction regions, enabling stable simulation of scenes with $\sim10^7$ splats using $\sim10^4$ cubature points and $\sim10^2$ effective DoFs.

Scenes are treated as kinematic, while objects are deformable and equipped with learned skinning functions~\cite{modi2024simplicits}. Object material properties (Young’s modulus, Poisson ratio, density) are estimated from visual cues and object metadata, and combined with Monte Carlo volume estimation to derive masses in metric scale.

At each timestep, updated splat positions are rendered with the standard 3DGS rasterizer~\cite{kerbl20233d}. In addition to RGB frames, we extract dense physical annotations through a cubature-to-splat mapping, and store a structured simulation state capturing collisions, visibility, and camera motion, enabling automatic task generation. More details are in Appendix \cref{app:simulator}.

\subsection{Dataset details} 
\label{sec:bench_details}
We use $53$ different scenes and $109$ GSO objects both selected to maximize diversity of physics and appearances. For each object we train multiple skinning networks varying its physical properties within the VLM-queried ranges, amounting for a total of $333$ trainings. 
	To increase visual diversity we {randomize} camera trajectories though ensuring objects overlap with the camera frustum, for visible interactions. 
	{Each frame has 11 ground truth maps (inc. 3 amodal) capturing pixel-level physics (\texttt{collisions}, \texttt{gravity}, \texttt{materials}, and \texttt{deformation} which measures the stress -- compression or expansion -- of objects), kinematics (\texttt{scene flow}, \texttt{instance tracking}), semantics (\texttt{sem. segmentation}) and geometry (\texttt{depth}).} 
    Simulator details are in the Appendix \cref{app:simulator}.
	We generate 11k sequences at 25FPS, of various lengths but capped to 10 seconds, totaling 730K frames.
	
\begin{figure*}[t]
		\centering
		
		\begin{minipage}{1.0\linewidth}
			\centering
			\newcommand{\statsize}{0.19\linewidth}
			\includegraphics[width=\statsize]{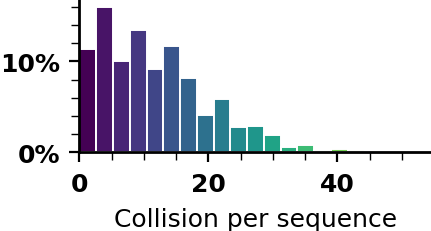}\hfill
			\includegraphics[width=\statsize]{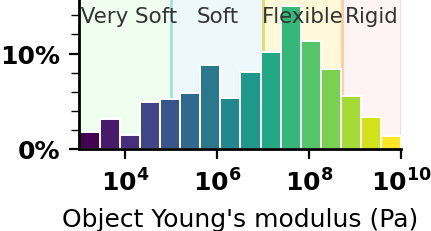}\hfill
			\includegraphics[width=\statsize]{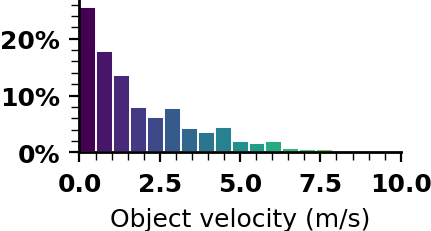}\hfill
			\includegraphics[width=\statsize]{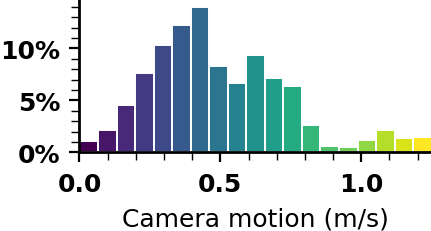}\hfill
			\begin{minipage}[b]{0.16\linewidth} 
				\centering
				\includegraphics[width=\linewidth]{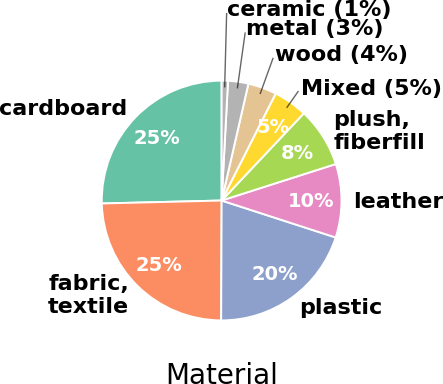}
			\end{minipage}
			\vspace{-0.3em}
		\end{minipage}
		\strut\hrulefill\strut %
		\vspace{-0.3em}
		\begin{minipage}{1.0\linewidth}
			\centering
			\vspace{2pt}
			
			\newcommand{\vqa}[7]{%
				\begin{minipage}[t]{0.24\linewidth}
					\centering
					\begin{tikzpicture}
						\node[anchor=north west, inner sep=0] (img) {\adjustimage{width=\linewidth, trim=0 0 0 0, clip}{#1}};
						\node[anchor=north west, fill=white, fill opacity=0.85, 
						draw=black, line width=0.2pt, rounded corners=0.5pt,
						text=black, inner sep=1.5pt, 
						font=\fontsize{6pt}{6.5pt}\selectfont\bfseries] at (img.north west) {#2};
					\end{tikzpicture}
					\begin{minipage}[t]{\linewidth}
						\vspace{-1em}
						\raggedright
						\fontsize{4.5pt}{4pt}\selectfont
						\parbox[c]{1.0\linewidth}{\fontsize{5.5pt}{5pt}\selectfont\centering \textbf{#3}}
						\parbox[t]{0.48\linewidth}{\textbf{A.} #4}\hfill
						\parbox[t]{0.48\linewidth}{\textbf{B.} #5}\par
						\parbox[t]{0.48\linewidth}{\textbf{C.} #6}\hfill
						\parbox[t]{0.48\linewidth}{\textbf{D.} #7}
					\end{minipage}
				\end{minipage}%
			}
			
			\vqa{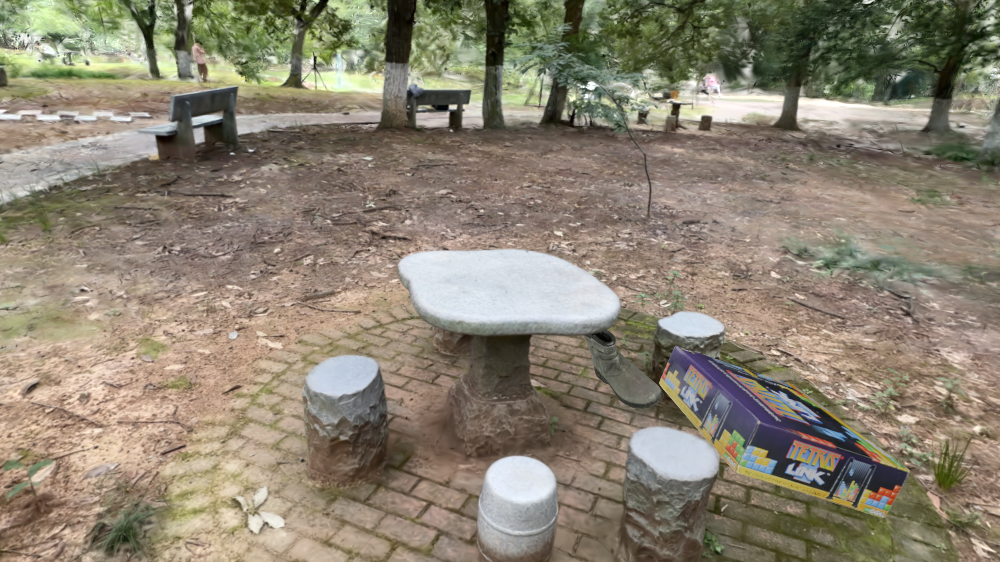}{\Material{}}{Which object has\\the biggest mass?}{\textbf{Tetris Link}}{TMNT figure}{Soccer shoe}{Ankle boot}\hfill
			\vqa{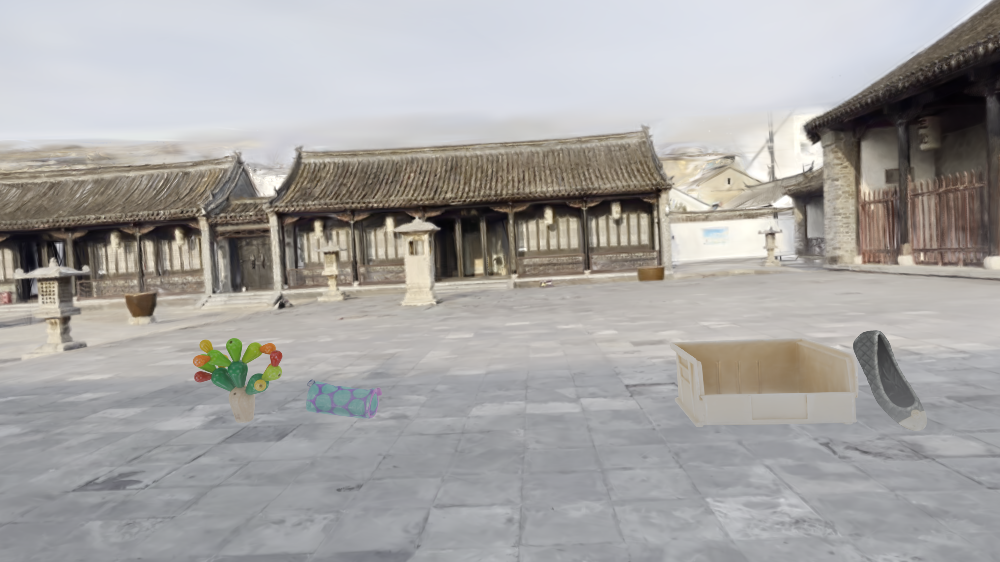}{\Mechanics{}}{What is the "dark flat shoe" colliding with?}{Pencil case}{\textbf{Shelf bin}}{Cactus toy}{Soccer shoe}\hfill
			\vqa{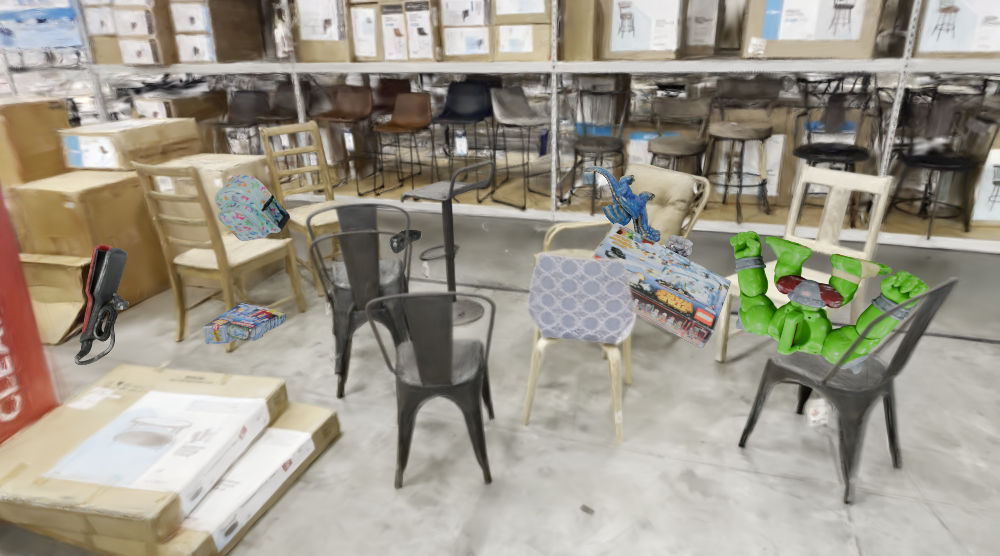}{\Material{}}{Which object has the highest Young's Modulus?}{Backpack}{LEGO set}{\textbf{Hair tool}}{Ogre toy}\hfill
			\vqa{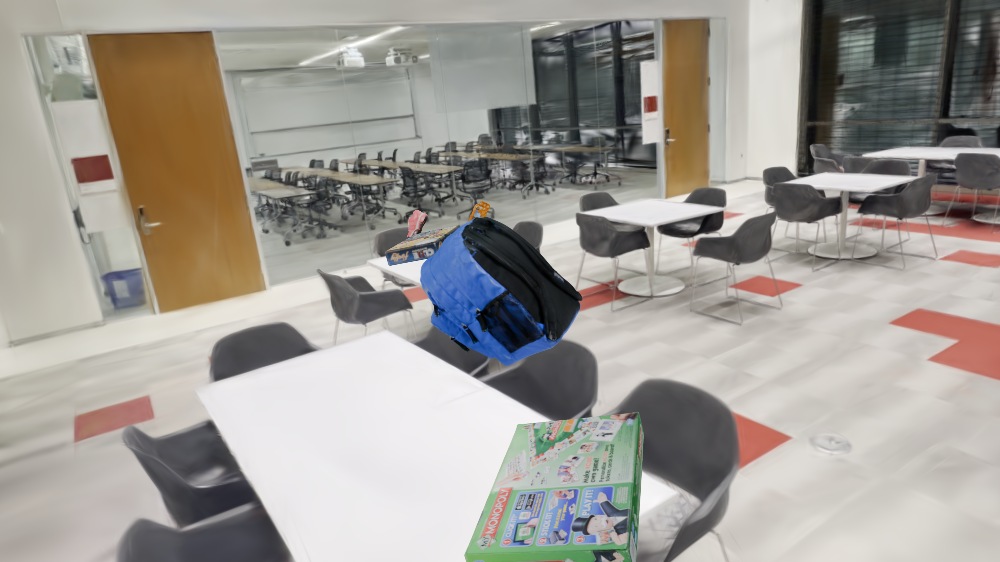}{\Spatial{}}{Where is "Monopoly" relative to "blue backpack"?}{Same depth}{Aligned}{To the left}{\textbf{Below}}
			
		\end{minipage}
		
		\caption{\textbf{Dataset statistics and VQA samples.} \textbf{Top}: Distribution of collisions, material properties, object velocity, camera motion, and material types. \textbf{Bottom}: Examples of VQA tasks including spatial reasoning, mechanics, and material understanding. We provide additional VQA examples in Appendix \cref{app:vqa_examples}.
		}
		\label{fig:combined_dataset_vqa}
	\end{figure*}

	Compared to prior physics datasets seen in \cref{fig:qual_comparison}, the statistics of our dataset in \cref{fig:combined_dataset_vqa} (top) exhibit more cluttered scenes from moving cameras, and $\approx$15 distinct collisions per sequence on average. The objects have varying softness and materials (right) and are simulated with a wide range of physical properties (middle), notably including highly deformable objects with low Young's Modulus.
	
	\subsection{Visual Question Answering}
	\label{sec:bench_vqa}
	
	To evaluate how vision models understand Newtonian physics, we design single and multi frame questions along six axes of study: \textbf{\Material{}} include questions on material intrinsics (density, Young modulus, Poisson ratio), objects masses and type of materials; \textbf{\Mechanics{}} include collision and interaction questions; \textbf{\Spatial{}} covers geometrical sensing of size, distances and general scene layout; \textbf{\Viewpoint{}} interrogates about visibility of objects, as well as camera-to-scene relation. For multi-frame sequences only, we also include: \textbf{\Temporal} which covers event ordering and camera motion and \textbf{\Permanence{}} which reflect the key ability to memorize invisible objects.
	
	In practice each question is implemented as a template function taking as input a frame or a sequence of frames, with the corresponding simulation state, and outputting a tailored question and four shuffled answers with a triplet of three wrong answers variably distant from the correct one. This process allows us to automatically generate a large number of questions for any simulation.
	We illustrate a few questions in \cref{fig:combined_dataset_vqa}.
	In total, \method{} includes 141K VQA pairs, comprising 84K multi-frame samples and 57K single-frame samples, whose statistics are shown in~\cref{fig:overview} (right).
    
    {For evaluation with smaller compute budgets, we also release a 15K subset, \method{}-15K, using a similar question balance and carefully selected to yield comparable VLM performance.}
	
	\section{Probing physics understanding}
	\label{sec:analysis}
	
	We study the physical understanding of 64 open-source models {and 2 closed-source models} {using \method{}};
	primarily focusing on VQA {since language is a natural proxy} for physical reasoning~\cite{Bisk2020piqa} 
	but also extending to the vision-only of pixel-wise physics prediction. We refer to Appendix~\cref{app:modelsspec} for a detailed listing but highlight that the chosen models cover a wide range of size (from 0.4B to 78B) and span over 27 {open-weights} families such as  LLaVA~\cite{liu2023visual}, InternVL~\cite{chen2024internvl}, PaliGemma~\cite{beyer2024paligemma}, Molmo~\cite{deitke2025molmo}, DeepSeek-VL~\cite{lu2024deepseek}, Cambrian~\cite{tong2024cambrian}, DINO~\cite{dino}, CLIP~\cite{clip}, etc. {For better positioning, we also include two frontier closed-source models, GTP~5.5~\cite{openai_gpt55} and Gemini~3.1~\cite{google_gemini31}}. {Unless stated otherwise, models are evaluated on the full NewtPhys, except closed-source models evaluated on NewtPhys-15K to reduce cost.}
    {Our study reveals that models struggle with physics understanding and that the field is poorly equipped to improve it.}
	
	\subsection{How do VLMs perform on fundamental physics?}
	\label{sec:analysis_perf}
	\begin{figure*}[t]
		\centering
		\begin{subfigure}{0.65\linewidth}
			\includegraphics[width=1.0\linewidth]{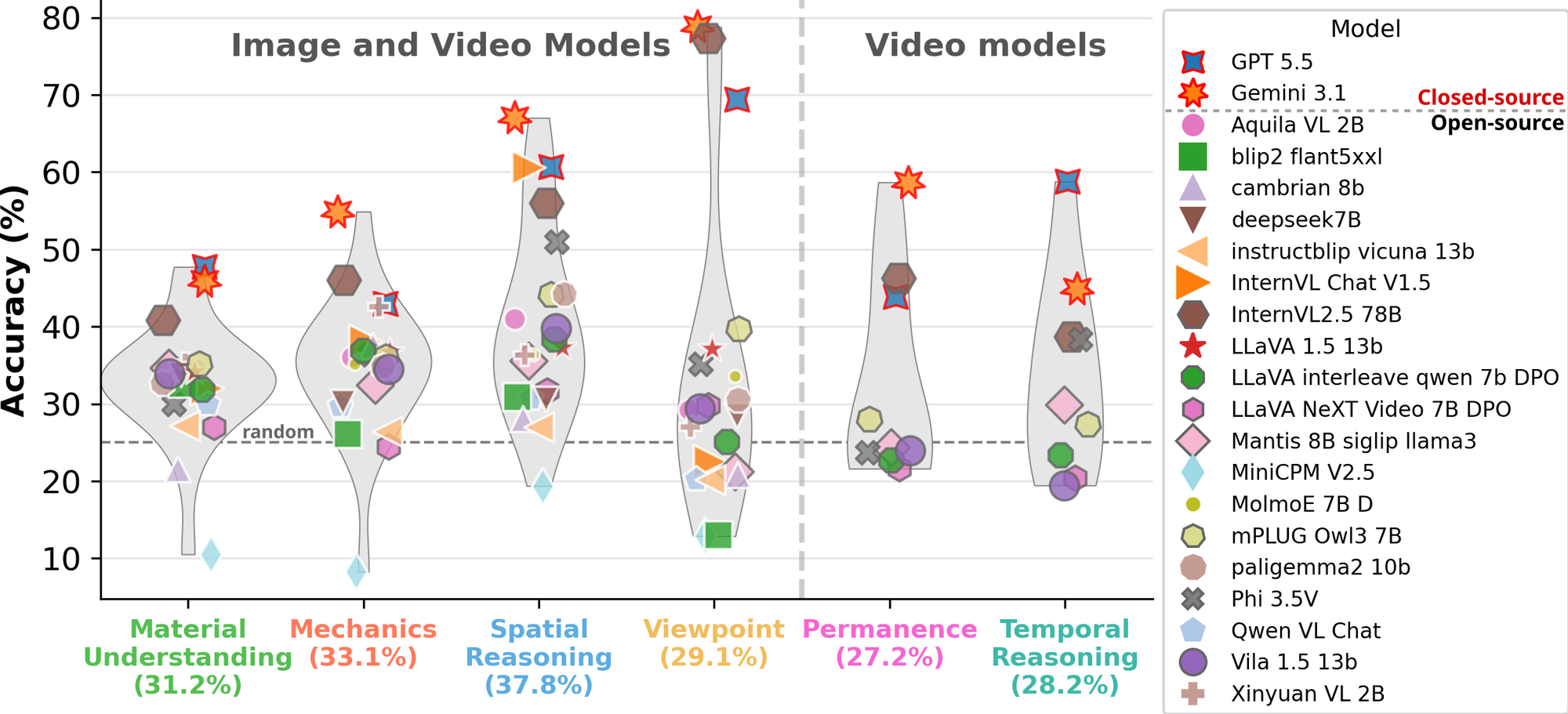}%
			\caption{Family's largest models} 
			\label{fig:perf}
		\end{subfigure}\hfill%
		\begin{subfigure}{0.33\linewidth}
			\includegraphics[width=1.0\linewidth]{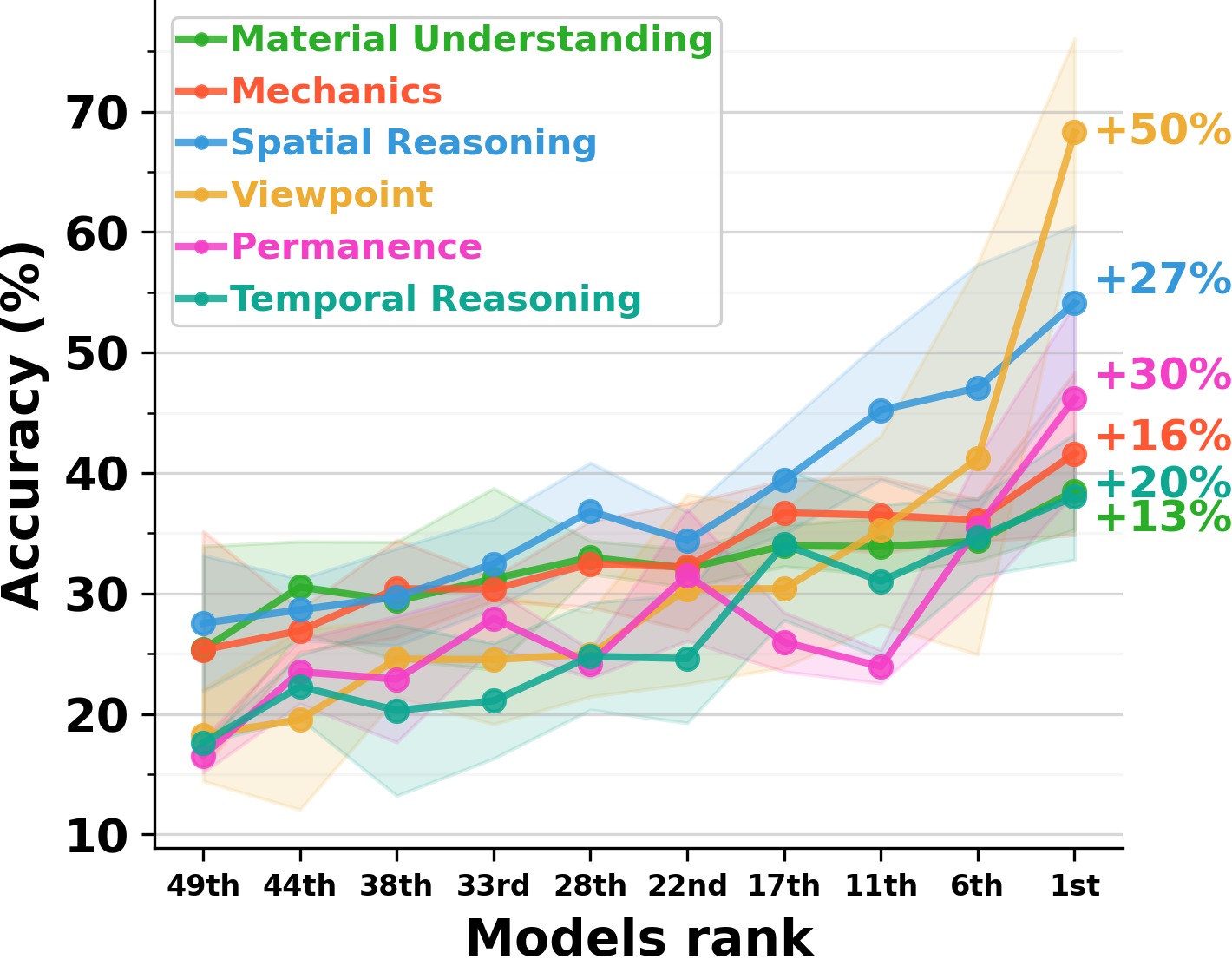}
			\caption{Performance per ranks}
			\label{fig:perf_per_rank}
		\end{subfigure}
		\caption{\textbf{Overall VQA performance.} \subref{fig:perf} {Performance of the largest per family models. The open-source InternVL2.5~78B narrows the gap with closed-source models, which dominate the board, while all models show similar trends. Parentheses indicate average open-source performance.}
            \subref{fig:perf_per_rank} Average performance {of all 54 open-source models} \wrt their overall rank, showing that some tasks are progressing more slowly.
			\colorbox{gray!15}{\textbf{Note on markers:}}\textit{Markers are consistent throughout the paper: shapes encodes model family, size scales with parameter count, colors encodes individual models. Details in Appendix~\cref{app:models}.}
		}
	\end{figure*}
	
	While models can solve high-level physics tasks like general questions answering~\cite{chow2025physbench} or video reconstruction~\cite{motamed2025generative}, it is not clear whether they rely solely on correlation patterns, or have understood the causal nature of physical principles. Therefore, we first aim at measuring \textit{to which extent} do VLMs understand physics principles, focusing in particular on \material{} and \mechanics{} which are underrepresented, if not omitted, in existing benchmarks. 
	We first report the performance of each family's largest model in~\cref{fig:perf}.
    {Overall, frontier models (GPT-5.5 and Gemini 3.1) dominate the board, although their trend is similar to other models.}
    {Notably, the best open-source model, InternVL2.5-78B, competes with frontier models, while most other models perform at around 25–35\% across tasks.}
    The low scores highlight the challenge of \method{} benchmark \wrt most benchmarks where objects of interest are typically large and visually in focus. 
	Overall, we note that video-only VQAs (\permanence{}, \temporal{}) perform low, as expected due to the added temporal dimension, although we highlight the near-random performance of \permanence{}, assessing that models still struggle at this core cognition task that requires a deep understanding of space and time. 
	Perhaps surprisingly, we note that \material{} and \mechanics{} perform only little lower than \spatial{} and even higher than \viewpoint{} although both of the latter are overwhelmingly seen in existing vision benchmarks. We conjecture that this results from the comparison of models with very different sizes in \cref{fig:perf} (from 2B to 78B).
	
	To verify this, in \cref{fig:perf_per_rank} we study how `equally performing' models behave by evaluating the 54 {open-weights} VLMs, binned by their overall rank on \method{}. 
	This highlights two key observations. 
	(i) Comparing worse-to-best models performance (\ie, rank 49th vs 1st) reveals a different story: with {only \materialCol{+13\%} and \mechanicsCol{+16\%} improvement} for material and mechanics, while other categories exhibit 20--50\%. This demonstrates that VLMs are poorly progressing in their physics understanding. 
	(ii) Looking at the best models ({ranks 11th to 1st}){, we note that their average per category performance is significantly better than those of the biggest per family models in}~\cref{fig:perf}, reaching around 45\% for \permanence{}
    (\vs 27.2\%)
    and 40\% for \material{}/\temporal{} 
    (\vs 31.2\%/28.2\%),
    suggesting that good models are not well spread across families and/or that largest models may not perform best. 
	Both of these observations also suggest that physics understanding might be under-looked by the existing models and benchmarks. 
	
	\begin{figure}[t]
		\centering
		\begin{subfigure}{1.0\linewidth}
			\includegraphics[width=\linewidth]{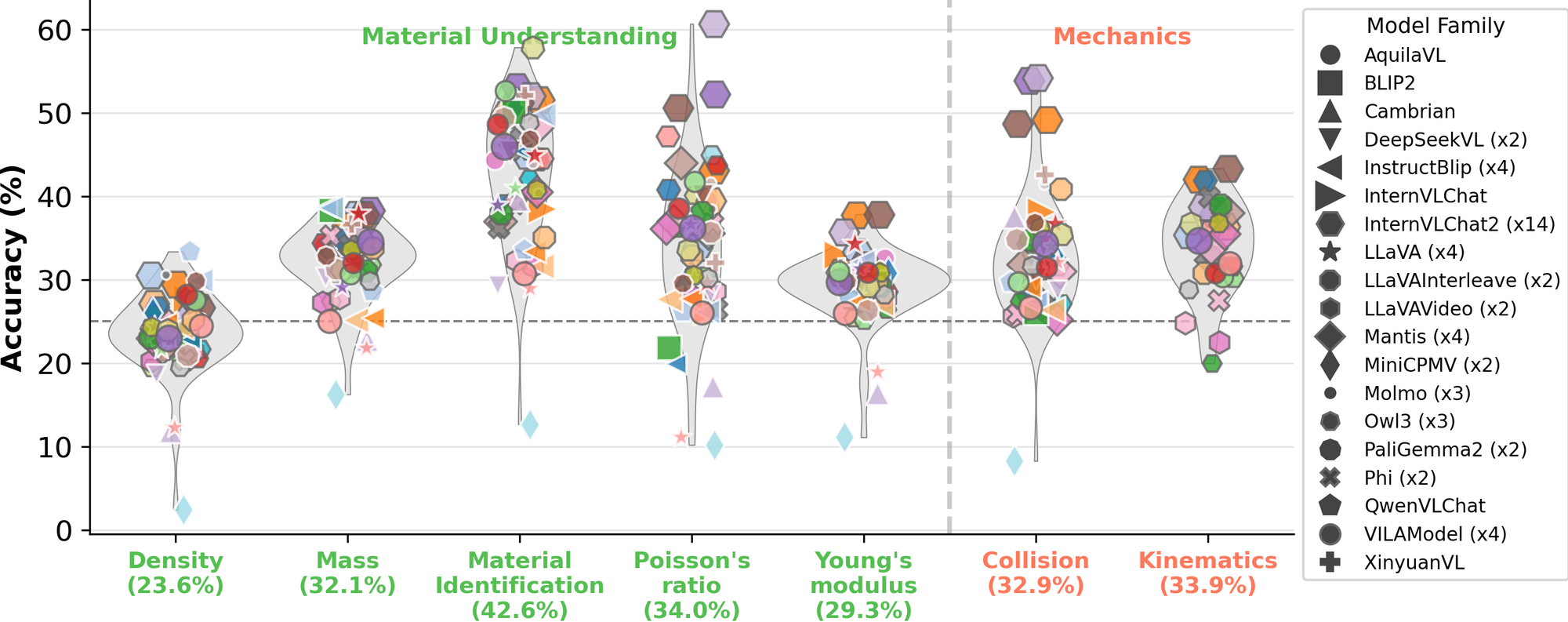}
			\caption{Physics understanding}
			\label{fig:perf_physics}
		\end{subfigure}\\%
		\begin{subfigure}{0.49\linewidth}
			\includegraphics[width=\linewidth]{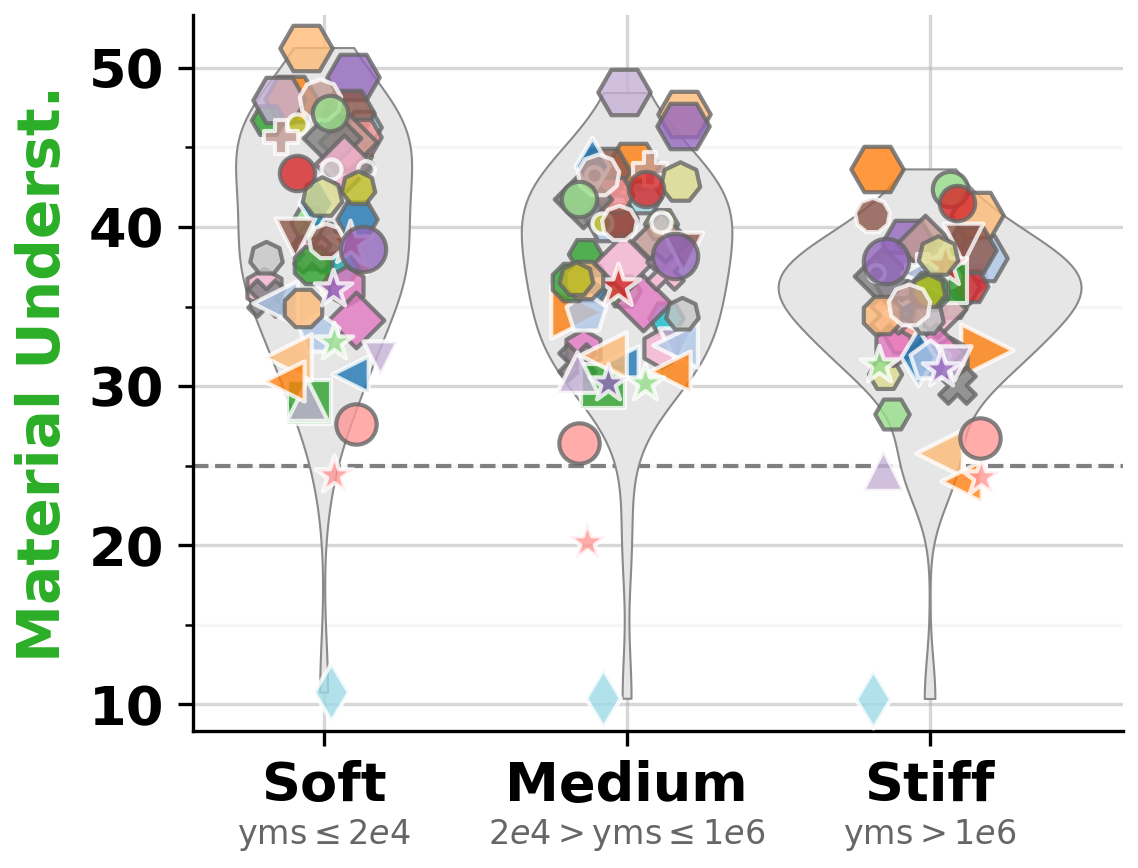}
			\caption{Variation per YMS}
			\label{fig:perf_per_yms}
		\end{subfigure}\hfill%
		\begin{subfigure}{0.49\linewidth}
			\includegraphics[width=\linewidth]{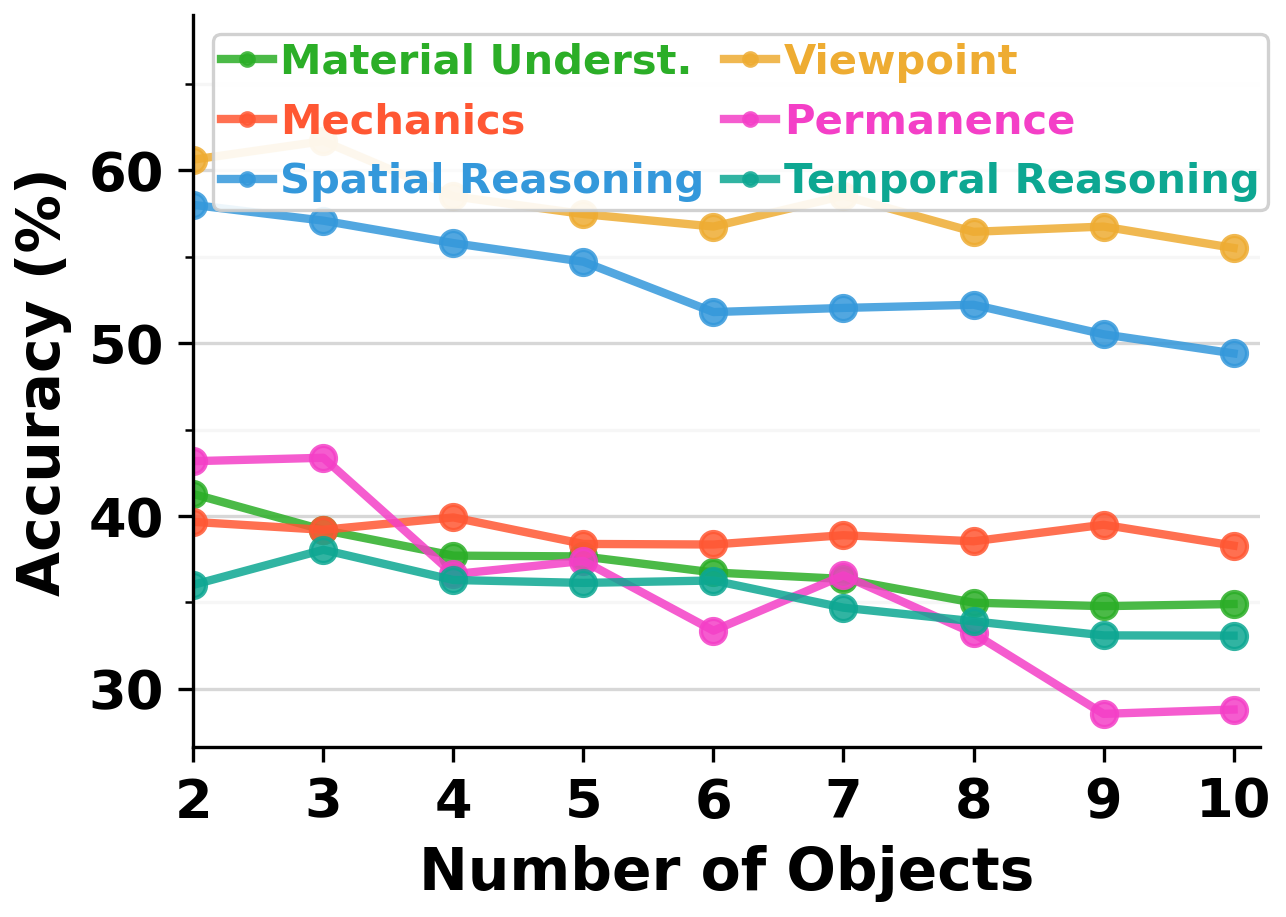}
			\caption{Objects variations}
			\label{fig:perf_per_object}
		\end{subfigure}%
		
		\caption{\textbf{Physics VQA and variations.} \subref{fig:perf_physics}~Subcategories performance reveals that material understanding performance is dominated by `material identification' (42.6\%) while properties estimation ranges much below (23.6\%--34.0\%). \subref{fig:perf_per_yms}~Varying object softness, we notice that VLMs perform better on soft objects. \subref{fig:perf_per_object}~Varying the number of objects shows that some VQAs are more stable than others.\vspace{-1em}
            }
	\label{fig:perf_per_subcat_and_rank}
\end{figure}

We provide detailed VQA physics performance in~\cref{fig:perf_physics} which highlights that \mechanics{} have stable performance unlike \material{}. Indeed we notice that models struggle at estimating \textit{density}, a relatively complex task that requires to properly estimate both the object's volume and material intrinsics, while they instead perform well on \textit{material identification} which we attribute to the nature of the task -- distinguishing materials (\eg, cotton, wood, \etc) -- being conceptually close to classification that is overwhelmingly present in benchmarks. Benefiting from attributes in \method{}, we also conduct analysis by varying the type of object's softness. Results in~\cref{fig:perf_per_yms} indicate that softer objects (\ie, low Young's modulus; YMS) lead to better accuracy \wrt scenes with stiff objects. 
This is partially explained by the exponential growth of the YMS scale, which makes the difference between ``stiff'' and ``super-stiff'' visually ambiguous unless a significant mechanical constraint is observed. Conversely, as soft objects deform easily upon contact, it provides a critical visual cue to estimate their material properties like density, Poisson's ratio, \etc. 
In \cref{fig:perf_per_object} we report performance for scenes with varying numbers of objects, showing no degradation for \mechanics{} and a minimal drop for \material{} and \temporal{}. 

\begin{figure}[t]
	\centering
	\begin{subfigure}{0.53\linewidth}
		\centering
		\includegraphics[width=1.0\linewidth, height=2.7cm, keepaspectratio]{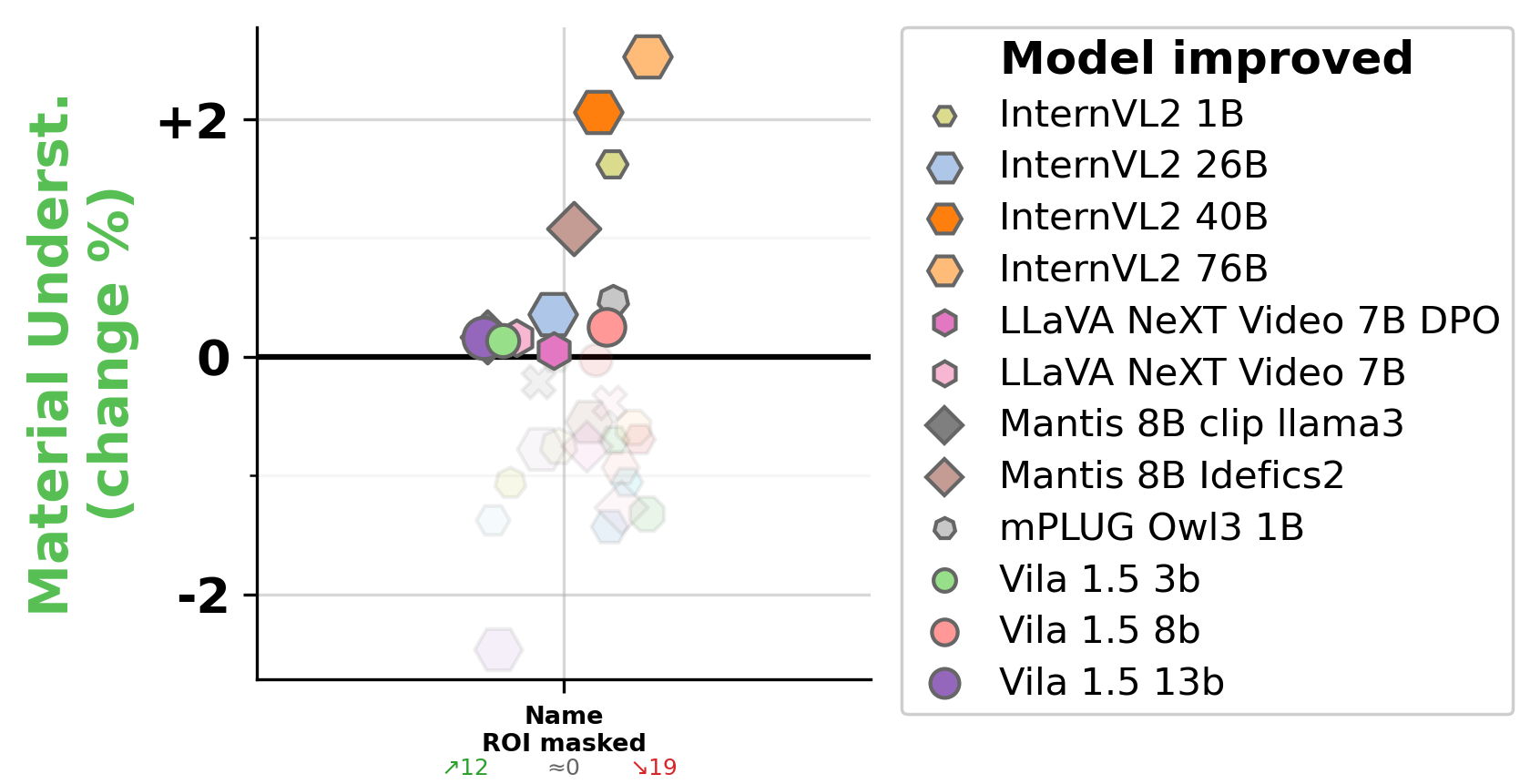}%
		\caption{ROI masked}
		\label{fig:bias_llm}
	\end{subfigure}\hfill%
	\begin{subfigure}{0.45\linewidth}
		\centering
		\includegraphics[width=1.0\linewidth, height=2.7cm, keepaspectratio]{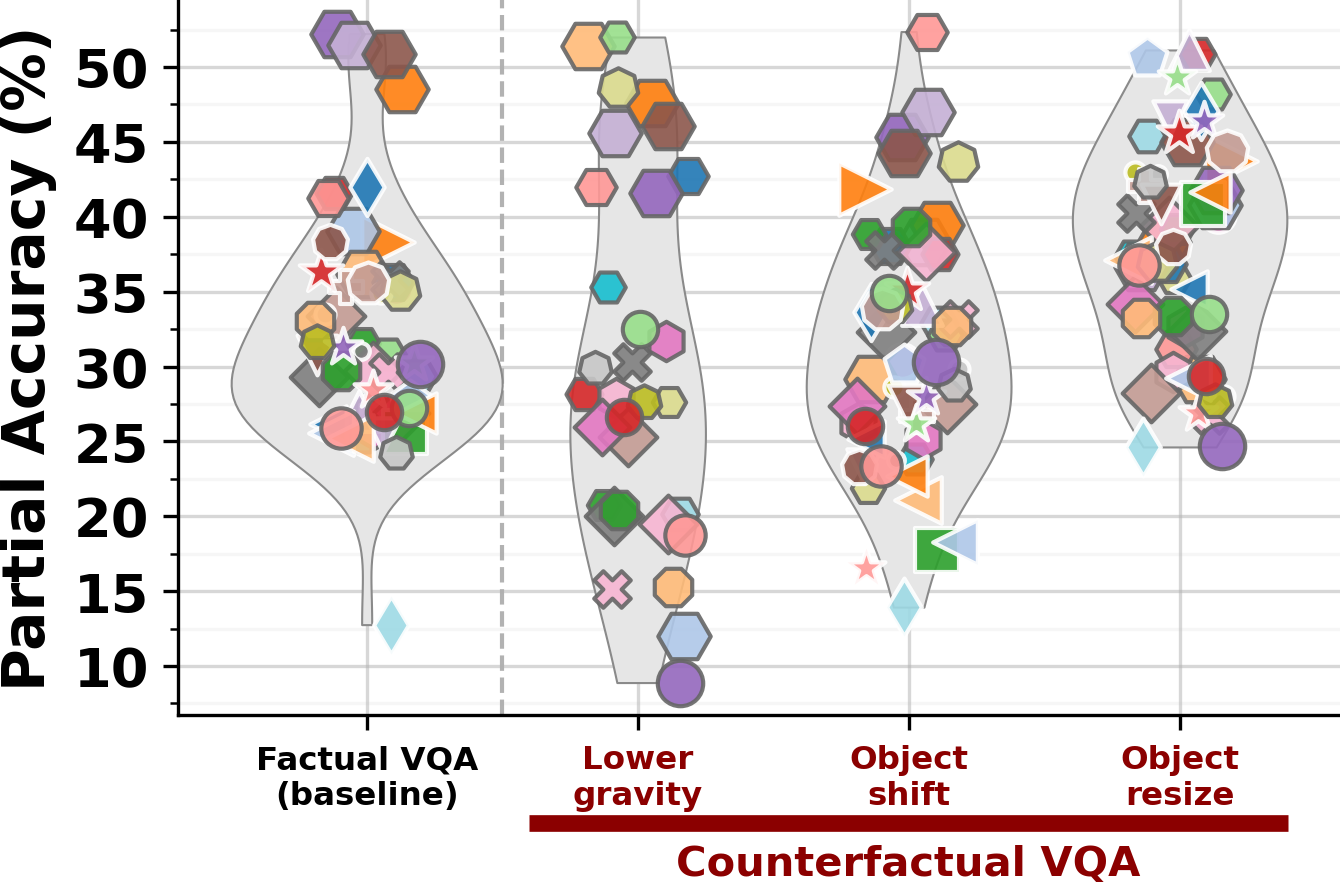}%
		\caption{Counterfactual VQAs}
		\label{fig:bias_counterfact}
	\end{subfigure}
	\caption{\textbf{Evaluation of LLM priors.} We design experiments to assess whether and how VLMs rely on LLM prior knowledge. In \subref{fig:bias_llm} we report VLM performance in the degenerated VQA scenario where the region of interest is masked out. 
		This sheds light on a handful of LLM-biased VLMs that benefit from \textit{not} accessing visual data.
		In \subref{fig:bias_counterfact}, various counterfactual scenarios are evaluated. Refer to the text for details.\vspace{-0.5em}
	}
	\label{fig:bias}
\end{figure}

\subsection{Do VLMs reason solely from LLMs knowledge?} 
\label{sec:analysis_llmbias}
A natural question is whether VLMs' ability to perform physical VQAs (at least above random chance) results from a true understanding of the visual data or rather from their prior LLM knowledge. Indeed, while \method{} varies object sizes and physical properties to mitigate such risk, they remain relatively close to their original values and questions such as asking the ``mass of a board game'' may find a logical answer without need of looking at the visual data.
To investigate this we design two experiments which aim at (a) assessing the impact of LLMs knowledge, (b) assessing the ability of VLMs to reason visually.
\textbf{(a)} To assess LLM bias we create a `ROI masked' variation of \method{} where regions of interest of questions are masked out (\ie, masking the object whose mass is being asked about). We then evaluate 31 Video VQA models on this `ROI masked' and report the per model change \wrt the original VQA in~\cref{fig:bias_llm}. We note that 12 out 31 models \textit{improve} performance when ROI is masked, showing that they \textit{can} rely solely on LLM knowledge and leakage from the question. Interestingly, most of these so-called ``LLM-biased VLMs'' are among the large models of our study ($\geq13$B params.) which possibly advocates that large models may overly rely on prior knowledge and \textit{answer without looking}. 
To further emphasize the importance of looking at visual data, we propose a second experiment \textbf{(b)} which consists of evaluating models on \textit{counterfactual questions}. Specifically, we follow \cref{sec:bench} to generate counterfactual scenes perfectly identical to the existing ones but either lowering the gravity, or randomly shifting/resizing an object of interest. Such scenes are then used to evaluate counter factual answers about unseen events. An example of such question is: \textit{Would A collide with B if A was shifted by 1 meter ahead?}. Since not all questions can be formulated counterfactually, in~\cref{fig:bias_counterfact} we report the partial accuracy of overlapping questions. It results that all 54 VLMs perform reasonably well on counterfactual although `lower gravity' exhibits a drop which may result from such scenario being OOD for both vision and language models which are unlikely to have seen such ten times lower gravity. Conversely, counterfactual VQA for varying objects' location or size is equivalent if not better than the factual VQA. A reasonable assumption is that, {similarly to CoT prompting shown to elicit reasoning in LLMs~\cite{chainofthoughts}}, counterfactual formulations (\textit{What if ..}) may encourage VLMs to reason rather than relying on prior knowledge. 

\subsection{Are we equipped for improved physics understanding?} 
\label{sec:analysis_commonsense}
\begin{figure}
	\centering
	\includegraphics[width=\linewidth]{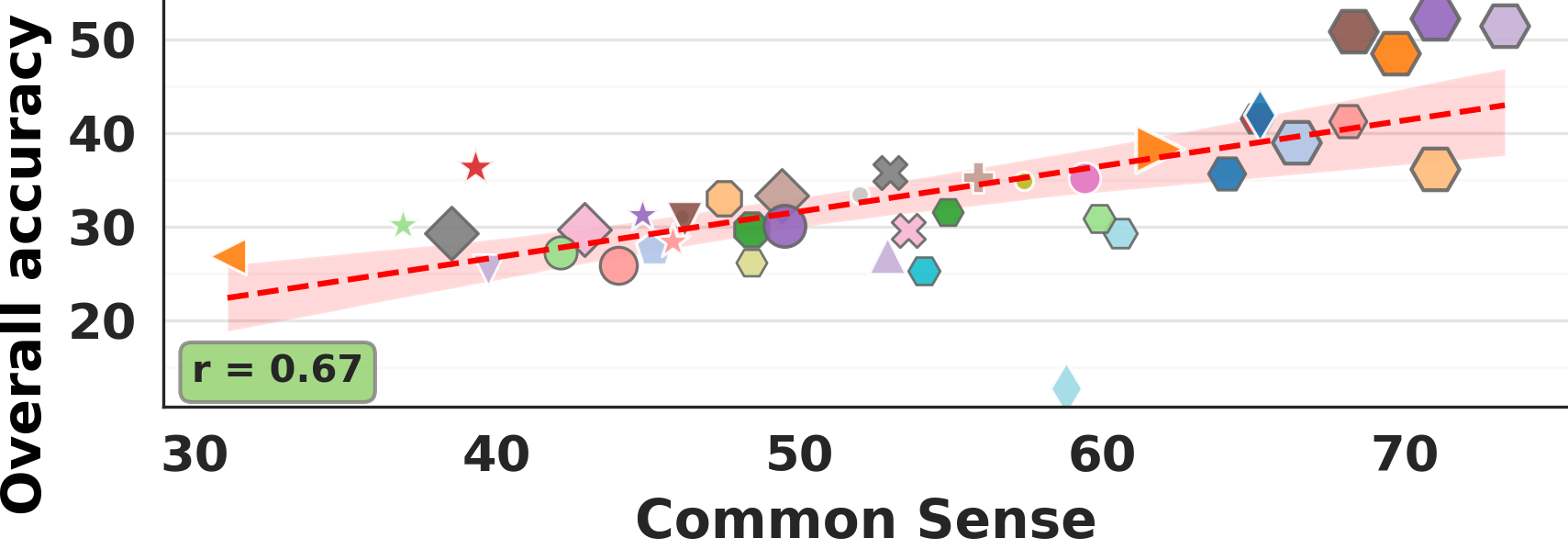}\\
	\begin{minipage}{1.0\linewidth}
		\begin{minipage}[m]{0.33\linewidth}
			\centering%
			\includegraphics[width=\linewidth]{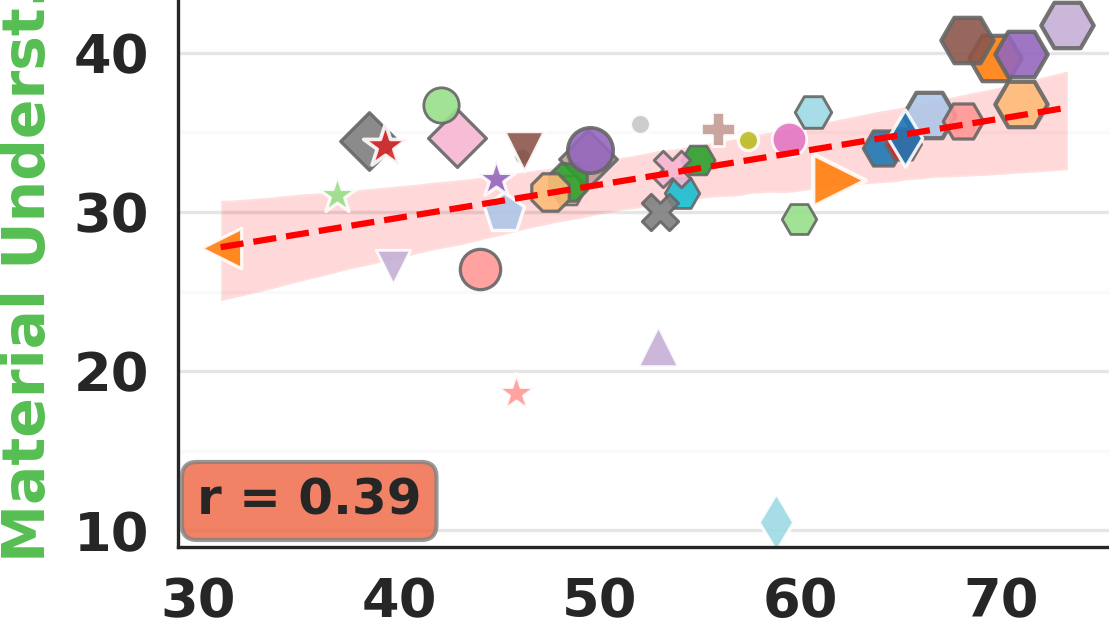}
		\end{minipage}\hfill%
		\begin{minipage}[m]{0.33\linewidth}
			\centering
			\includegraphics[width=\linewidth]{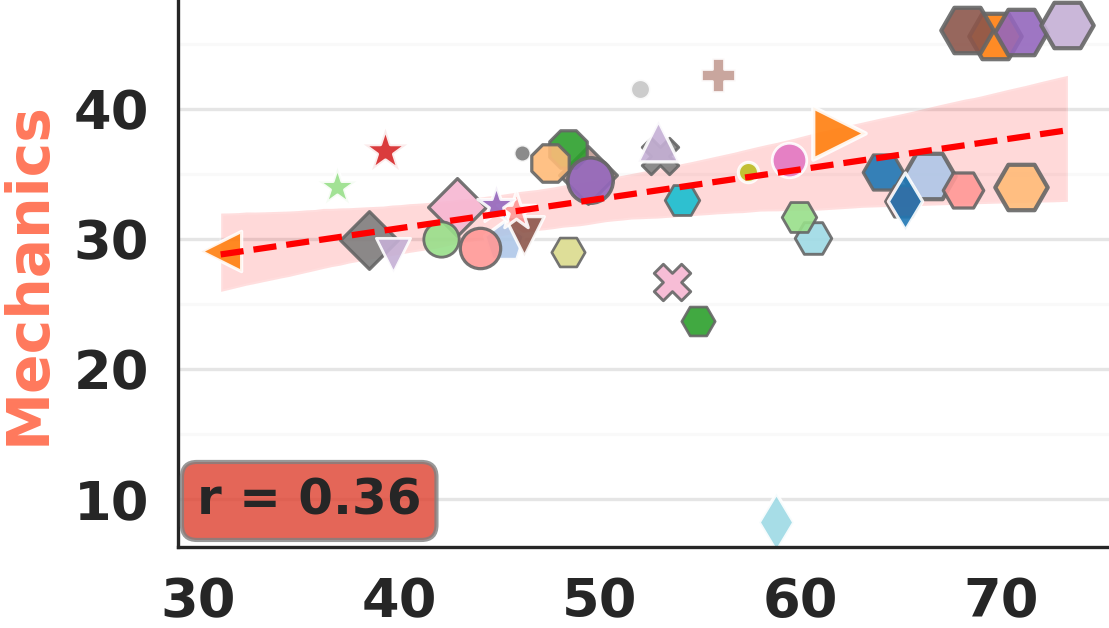}
		\end{minipage}\hfill%
		\begin{minipage}[m]{0.33\linewidth}
			\centering
			\includegraphics[width=\linewidth]{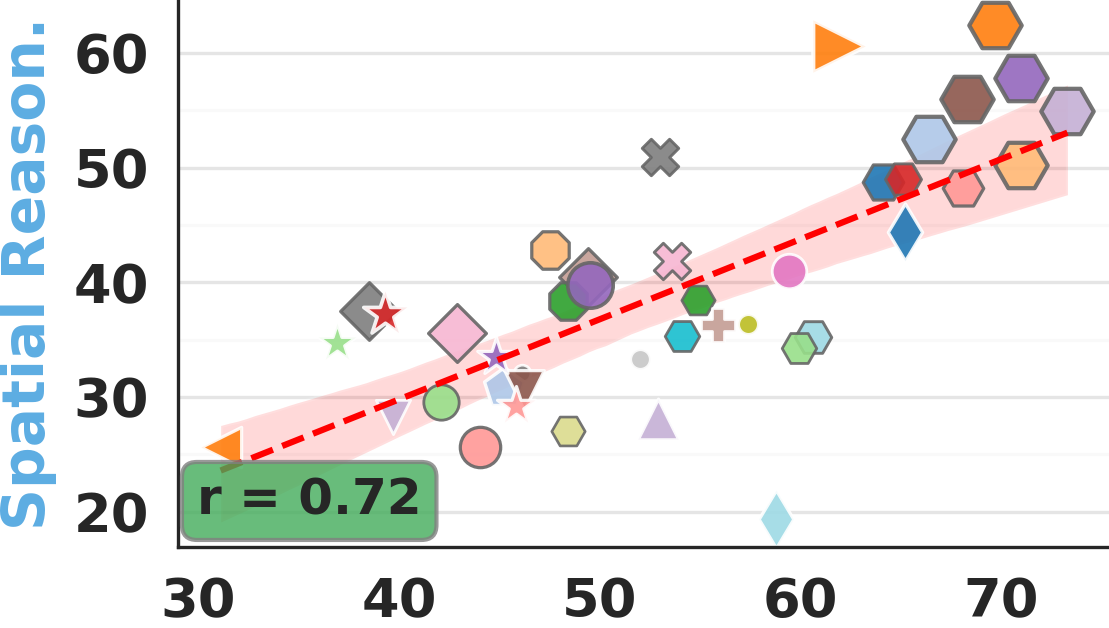}
		\end{minipage}\\
		\begin{minipage}[m]{0.33\linewidth}
			\centering
			\includegraphics[width=\linewidth]{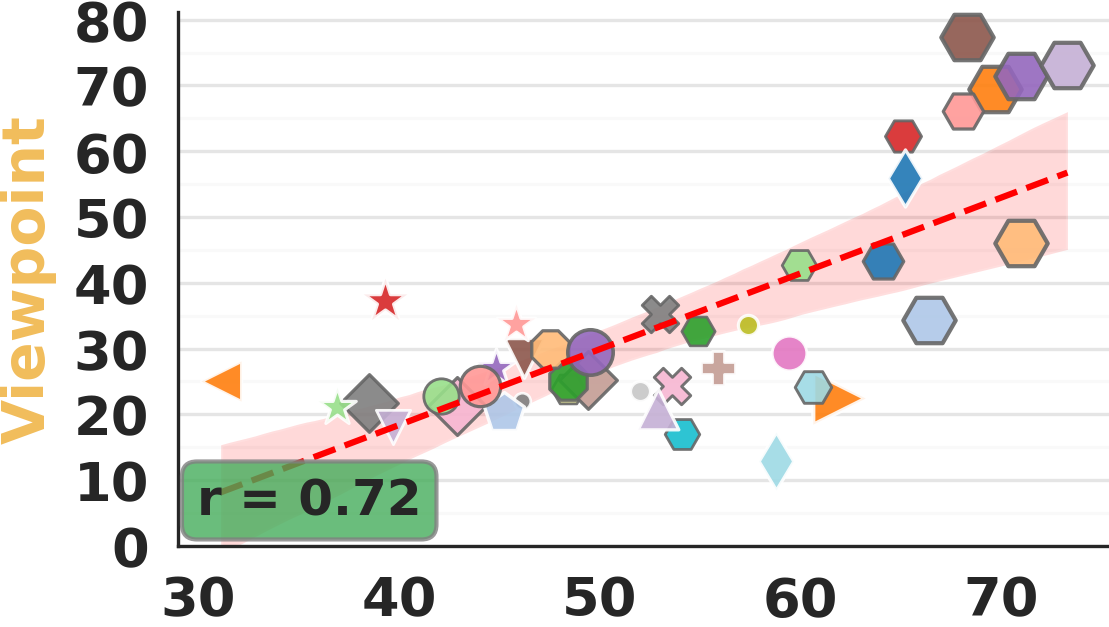}
		\end{minipage}\hfill%
		\begin{minipage}[m]{0.33\linewidth}
			\centering		\includegraphics[width=\linewidth]{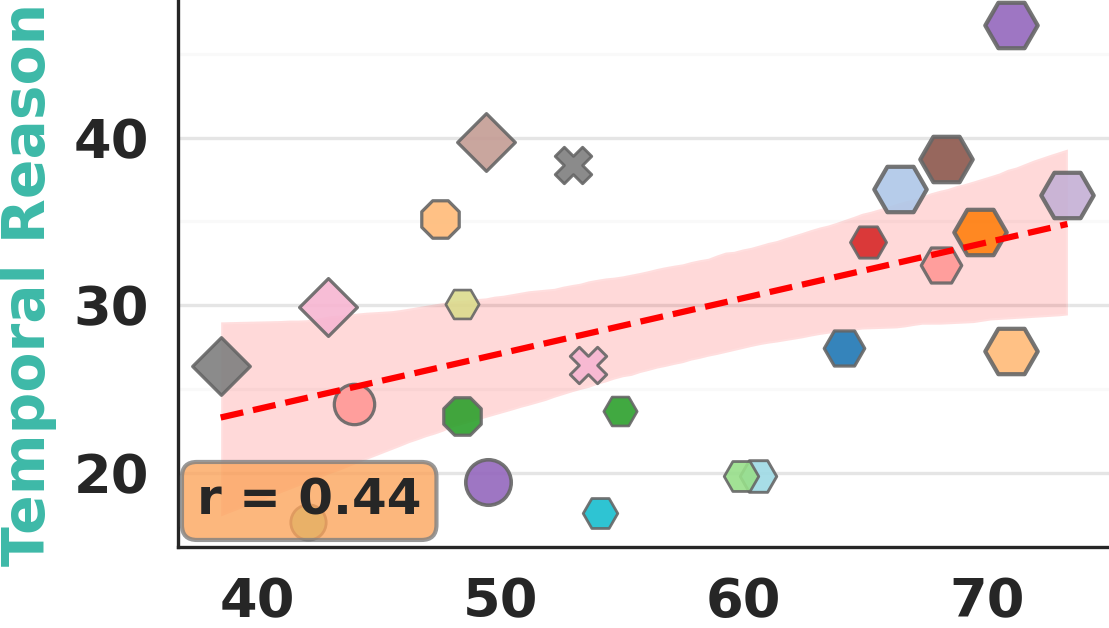}
		\end{minipage}\hfill%
		\begin{minipage}[m]{0.33\linewidth}
			\centering
			\includegraphics[width=\linewidth]{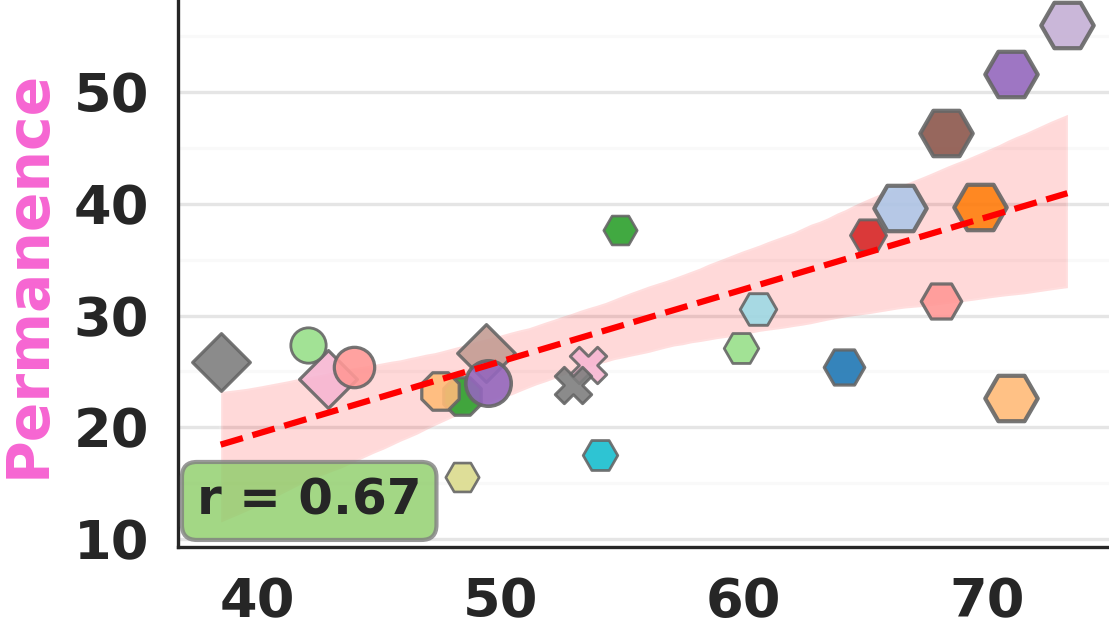}
		\end{minipage}%
	\end{minipage}
	
	\caption{\textbf{Common sense correlation.} Overall accuracy on \method{} correlates with the 8-benchmarks average performance, but per-category correlations exhibit significant discrepancies with weak correlation on physics (\material{}, \mechanics{}).}
	\label{fig:perf_commonsense_avg}
\end{figure}

Having shown that physics understanding is lagging~(\cref{sec:analysis_perf}) and may predominantly emerge from correlation rather than causality~(\cref{sec:analysis_llmbias}), we now question whether the computer vision community is equipped with the right tools to measure and improve physics understanding. 
We 
measure the correlation between the performance of 41 VLMs on \method{} and their average performance on eight common sense benchmarks~\cite{ai2d, hallusionbench,mmbenchv1_1,mmmu,mmstar,mmvet,mathvista,ocrbench}. 
\cref{fig:perf_commonsense_avg} (top) shows an overall strong correlation, with a Pearson Correlation Coefficient $r=0.67$, though correlation with individual categories (bottom) reveals discrepancies between low-level physics reasoning and others. As it appears, \material{} and \mechanics{} exhibit very low correlation ($r=0.39$ and $r=0.36$, respectively) compared to \spatial{} and \viewpoint{} ($r=0.72$ both). This reveals that current common sense's definition typically aligns with questions on visual attributes like object size, position, visibility, and camera characteristics but less with low-level physics like Young’s Modulus, density, \etc. 
Further, this suggests that while current VLMs grasp general world knowledge, they struggle with the nuance of low-level physical reasoning.

\begin{figure}
	\centering
	\includegraphics[width=1.0\linewidth, keepaspectratio]{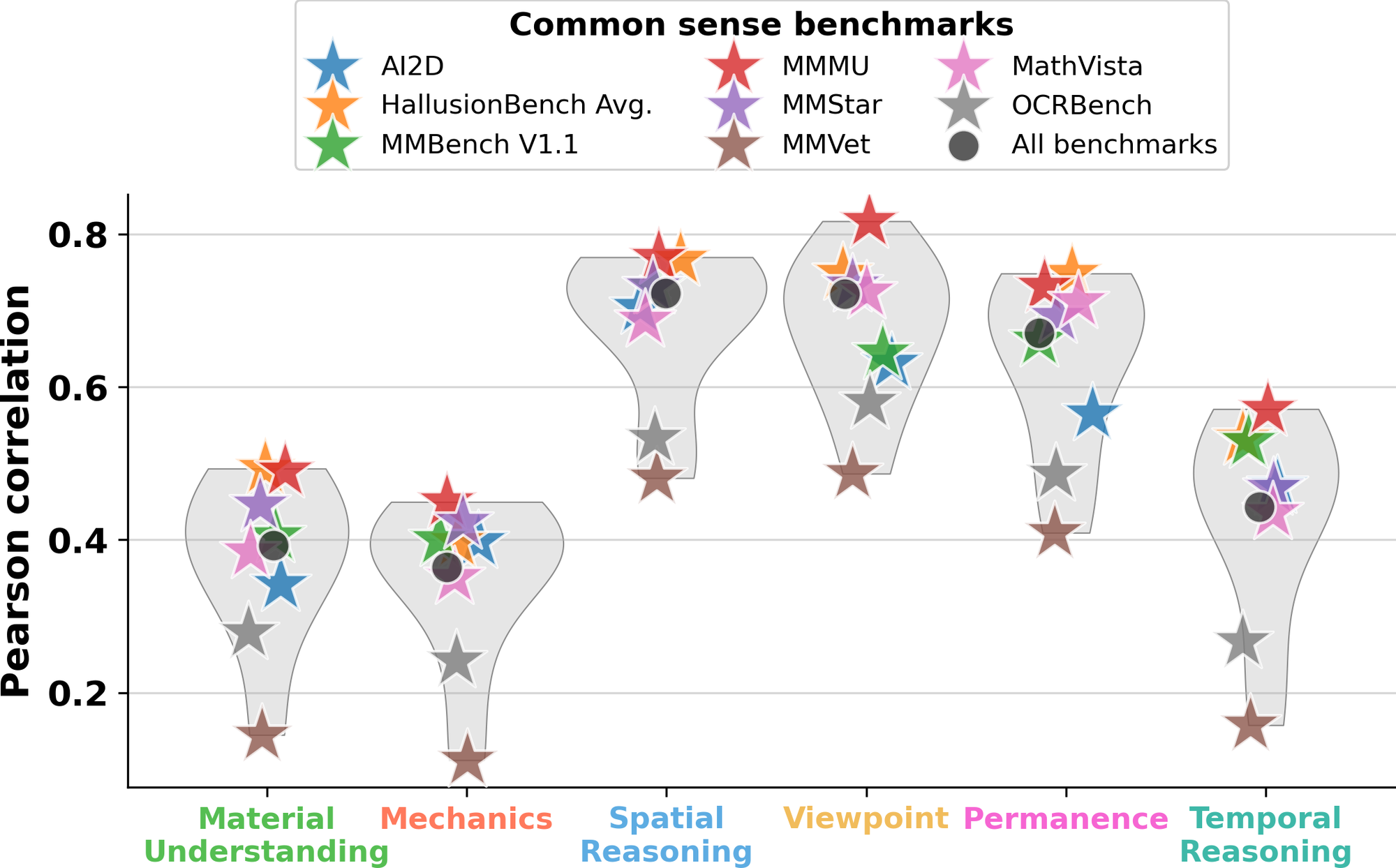}
	
	\caption{\textbf{Common sense correlation per benchmark.}Our study reveals a striking evidence that existing benchmarks weakly correlate with physics, with MMVet and OCRBench having a Pearson coefficient as low as $r=0.1$.}
\label{fig:perf_commonsense_benches}
\end{figure}

To clarify the picture, we study the per-benchmark correlation in~\cref{fig:perf_commonsense_benches}.
This reveals a striking observation: all benchmarks weakly correlate ($r<0.5$) with low-level physics. Moreover, MMVet~\cite{mmvet} and OCRBench~\cite{ocrbench} exhibit a surprisingly low correlation ($r\approx{}0.2$), which we conjecture originates from them being overly object-centric and mono-task, respectively. 
{We highlight that for image-only models some benchmarks even inversely correlate with physics; therefore favoring shortcuts in the learning rather than the true world model.}
On the other hand, MMMU~\cite{mmmu} and HallusionBench~\cite{hallusionbench}, which are largely multi domains and multi modals benchmarks, exhibit stronger correlation with physics ($r\approx{}0.45$), advocating for the Platonic hypothesis~\cite{platonic_hypothesis} where an increased number of domains and modalities lead to better convergence towards the true model of the world. 

From above observations, we conclude that current definitions of ``common sense'' in the literature are insufficient proxies to unlock true physical understanding. 

\subsection{Can models perform better on physics?}
\begin{figure}[t]
\centering
\begin{subfigure}{0.63\linewidth}
	\includegraphics[width=1.0\linewidth, height=2.5cm, keepaspectratio]{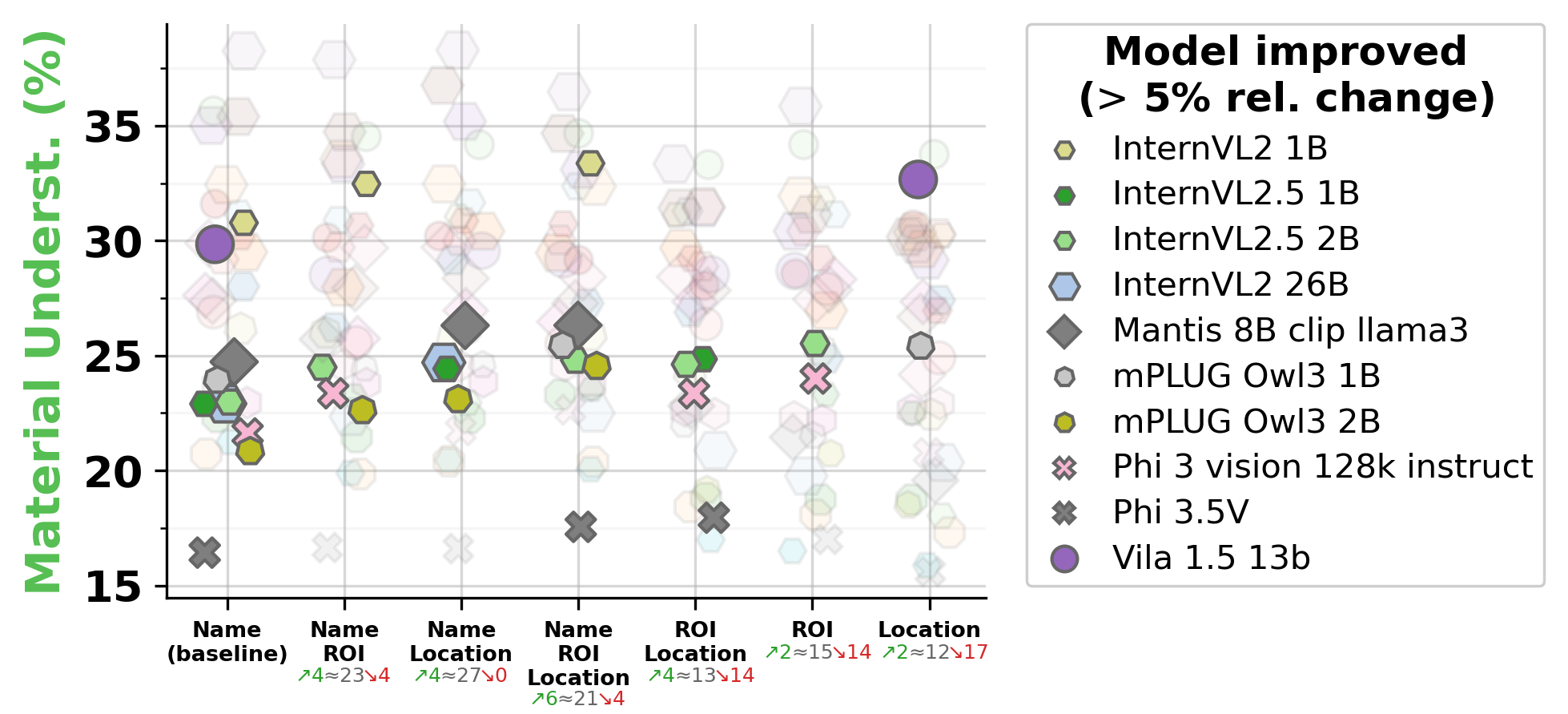}%
	\caption{Spatial cues}
	\label{fig:ablations_spatial}
\end{subfigure}\hfill%
\begin{subfigure}{0.35\linewidth}
	\includegraphics[width=1.0\linewidth, height=2.5cm, keepaspectratio]{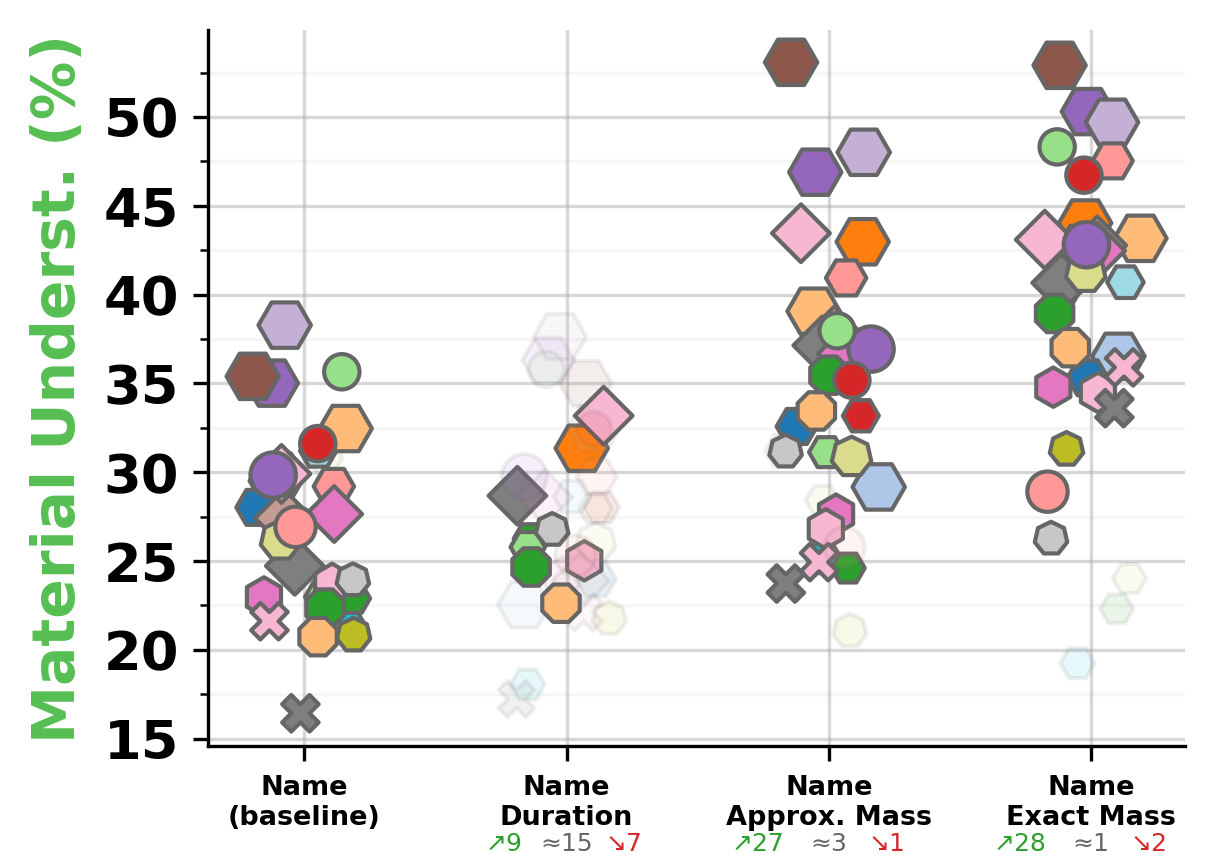}%
	\caption{Physical cues}
	\label{fig:ablations_physics}
\end{subfigure}
\caption{\textbf{Effects of spatial and physical cues.} For 31 video VLMs, we evaluate the effect of adding \subref{fig:ablations_spatial} spatial cues, which help some smaller models, or \subref{fig:ablations_physics} physical cues, which strongly improve performance.}
\label{fig:ablations}
\end{figure}

While finetuning is beyond our study, prior works~\cite{shtedritski2023does, sun2024alpha} demonstrated the benefits of adding cues to improve VQA as it encourages the VLMs to reason rather than complete.
Subsequently, we explore two strategies consisting into providing text/visual cues or changing our questions formulation. 

\subsubsection{Spatial and Physical cues.} On cues, objects' names already provide a strong cue, evidenced by results in~\cref{sec:analysis_llmbias}, although well-functioning VLMs still have to locate objects before estimating their physical properties. To ease this, we introduce spatial cues either in the visual form by circling the Region of Interest (\textit{ROI})~\cite{shtedritski2023does} or in the textual form by providing coarse \textit{Location} of objects in the image (\eg, `top-right`). This setup leads to seven variants, each with $\approx2\text{k}$ VQA pairs. While all 31 video VLMs are evaluated, for clarity, in~\cref{fig:ablations_spatial} we emphasize only models whose performance improved by at least five percent \wrt to the baseline VQA using only objects \textit{Name}. 
We note that only five families benefit from spatial cues, and models that improved are typically smaller ones (mainly $\leq\text{8B}$). 
This corroborates the observed ability of large VLMs to spatially locate objects in images~\cite{dorkenwald2024pin, xue2025point}. Specifically, mPlug 2B and InternVL2 1B get a large $+5$ points boost with all three combined cues. Besides localization, we also explore how physical cues affect physical reasoning, adding to each question hints about \textit{duration} or objects \textit{approximate} or \textit{exact mass}. 
Results in~\cref{fig:ablations_physics} highlight that almost all models are subsequently boosted. 
Of note, adding \textit{duration} (\eg, ``In this sequence of 3.5 seconds, ...'') improves a few of the large and best performing models, an interesting insight as duration is virtually free to provide. Interestingly, providing \textit{approximate mass} (\eg, ``... object of approximate 2kg mass ...'') brings a similar boost with the \textit{exact mass} (\eg, ``... object of 2.35kg ...'') though easier to provide. We however highlight that this is not sufficient to assess whether these physical cues help the model to reason visually or simply strengthen priors to the LLM.

\begin{figure}[t]
	\centering
	\includegraphics[width=1.0\linewidth, keepaspectratio]{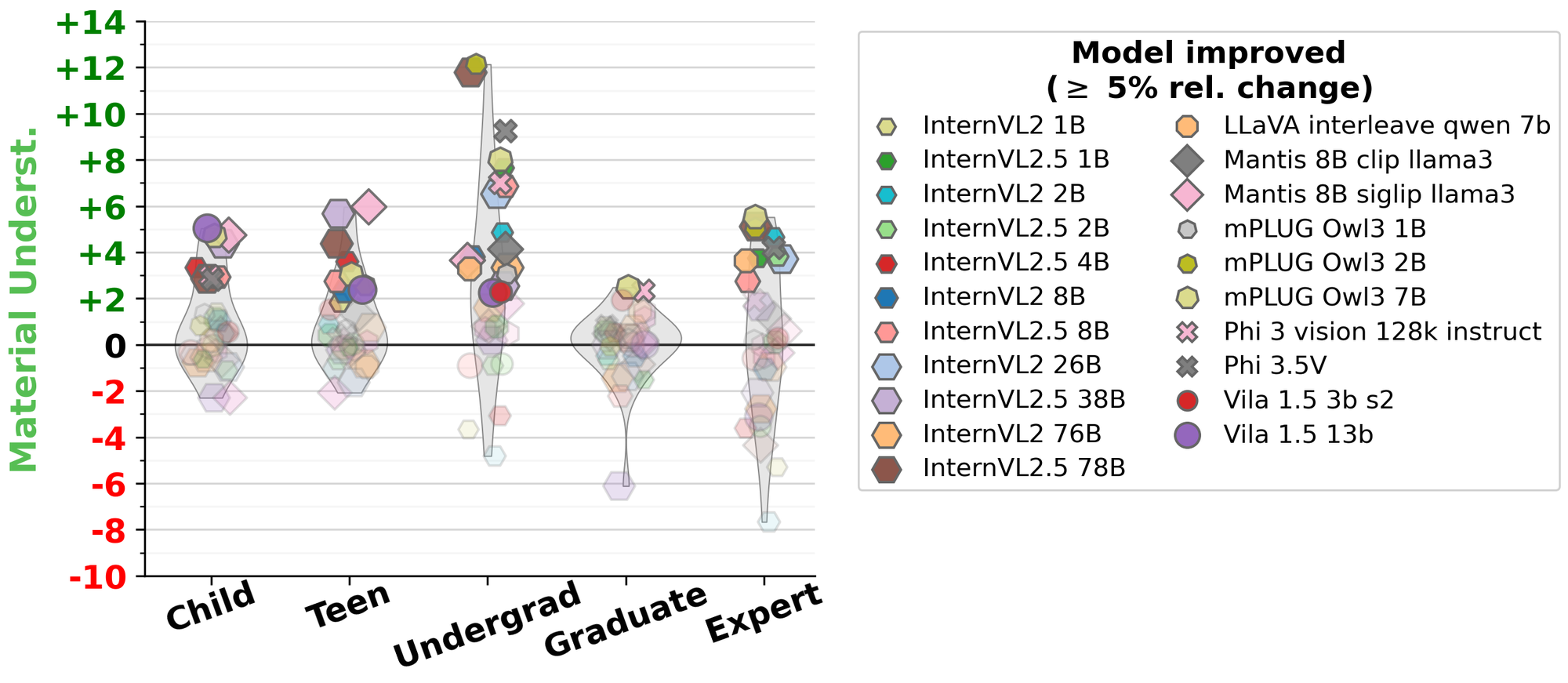}%
	\caption{\textbf{Experts to Novices prompting.} Inspired by physics education research we evaluate 31 video VLMs with questions reformulated with different level of expertise, ranging from child ($\approx$10yo) to expert (physicist) and report their performance \wrt the original \method{} VQA formulation. This showcases that \method{} aligns with \textit{graduate} prompting and that most VLMs perform better with \textit{undergrad} formulation.}
	\label{fig:ablations_levels}
\end{figure}

\subsubsection{Prompting level} One could argue that understanding physics principles differ from the ability to estimate quantities such as object’s Young’s modulus, for instance. Such an example are infants which have a sense of physical plausibility~\cite{baillargeon1994physical} but are unable to leverage complex physics metrics. 
This resonated with work in the field of physics education research which categorizes knowledge by experts and novices~\cite{chi1981categorization}. Similarly, VLMs could grasp physics principles but be limited in their expressivity. Inspired by this, we explore physical understanding across a spectrum of expertise ranging from novice (child) to expert (physicist). Practically, we select a subset of comparative VQAs which can be reformulated, and rewrite them with five levels (10 years old child, teen, undergrad, graduate, expert).
Details and exemplar questions are provided in Appendix~\cref{app:five_levels}. 

For all 31 video VLMs, \cref{fig:ablations_levels}~reports the performance change \wrt performance of similar questions in our benchmark. We observe a consistent performance peak at the \textit{Undergrad} level although further increasing questions complexity toward expert-like formulation significantly spreads models performance. We hypothesize that this is due to biases in the data used for training and instruction-tuning these models, which are largely crawled from sources such as undergraduate courses and Wikipedia, where most documents are written at a level comparable to undergraduate material. We also observe that \textit{Graduate} formulation aligns with \method{}, exhibiting the smallest change; a reasonable outcome given that original questions were formulated by computer vision scientists, not physicists experts. We highlight that our study provides a readily available tool for boosting performance on physics VQAs in VLMs. 

\subsection{Probing spatial understanding of physics in vision models?}
\label{sec:analysis_vision}
For applications such as robotics, it might be beneficial to have pixel-aligned physical predictions. Such task requires models that predict physical maps, much like ground truth provided in \method{} (\cf, \cref{sec:bench_details}). For that purpose, we evaluate Vision Foundation Models (VFMs) which are representation learning models trained with different forms of supervision on massive amounts of images to discriminate visual elements. Although not explicitly designed for physics, one may argue that their semantic understanding constitutes a step toward physical reasoning, suggesting a correlation between visual and physics representations. 

Specifically, we address \emph{Physics Probing} in VFMs by attaching small physics decoders to frozen visual encoders of pretrained VFMs, which estimate pixel-level physics maps (\eg, gravity, collision) and are trained on ground truth maps from \method{}.
The models take as inputs frame sequences and produce pixel-wise force prediction for the last frames. During training and evaluation, only predictions on the objects of interest are considered.
We focus our study on pixel-wise collision, gravity (magnitude and direction) and scene flow as those convey critical physical cues. 
For gravity prediction, we additionally construct a subset of out-of-distribution (OOD) videos in which we randomize the magnitude of gravitation $[0.98 m/s^2 - 20 m/s^2]$, making them different from the training videos, where Earth gravity of $9.81 m/s^2$ is always used.

\begin{table}
\centering
\resizebox{1.0\linewidth}{!}{%
\setlength{\tabcolsep}{1pt}
\begin{tabular}{c|l|l|c|cc|cc|c}
	\toprule
	\textbf{Supervision type}&\textbf{Model} & \textbf{Objective} &
	\textbf{Collision} &
	\multicolumn{2}{c|}{\textbf{Gravity}} &
	\multicolumn{2}{c|}{\textbf{Gravity-OOD}} &
	\textbf{Scene Flow} \\
	&& & F1 $\uparrow$
	& mAE$\downarrow$ & magE$\downarrow$
	& mAE$\downarrow$ & magE$\downarrow$
	& AEE$\downarrow$ \\[0.4em]
	\midrule
	\multicolumn{9}{c}{\cellcolor{gray!20}\textbf{Vision}}\\ \multirow{3}{*}{Fully-supervised}&DeiT III  & Classification & \cellcolor{pgrade4}48.47 & \cellcolor{pgrade6}19.34 & \cellcolor{pgrade5}21.44 & \cellcolor{pgrade6}15.20 & \cellcolor{pgrade3}35.43 & \cellcolor{pgrade4}1.29 \\
	&SAM       & Segmentation   & \cellcolor{pgrade8}54.80 & \cellcolor{pgrade5}20.73 & \cellcolor{pgrade7}17.04 & \cellcolor{pgrade5}16.28 & \cellcolor{pgrade3}33.98 & \cellcolor{pgrade10}0.94 \\
	&MiDaS     & Depth          & \cellcolor{pgrade9}54.95 & \cellcolor{pgrade9}12.12 & \cellcolor{pgrade8}15.06 & \cellcolor{pgrade9}8.23  & \cellcolor{pgrade4}33.80 & \cellcolor{pgrade8}0.95 \\
	\cline{1-9}
	\multirow{3}{*}{Self-supervised}&MAE       & SSL            & \cellcolor{pgrade1}28.61 & \cellcolor{pgrade1}45.69 & \cellcolor{pgrade1}31.79 & \cellcolor{pgrade2}42.91 & \cellcolor{pgrade1}42.50 & \cellcolor{pgrade3}1.29 \\
	&DINO      & SSL            & \cellcolor{pgrade10}56.54 & \cellcolor{pgrade8}13.79 & \cellcolor{pgrade9}14.92 & \cellcolor{pgrade10}9.30 & \cellcolor{pgrade6}33.33 & \cellcolor{pgrade9}0.94 \\
	&DINOv2    & SSL            & \cellcolor{pgrade9}56.52 & \cellcolor{pgrade7}14.95 & \cellcolor{pgrade10}14.76 & \cellcolor{pgrade8}11.02 & \cellcolor{pgrade5}33.81 & \cellcolor{pgrade7}0.95 \\
	\cline{1-9}
	\multirow{1}{*}{Agglomerative}&AM-Radio  & Distillation   & \cellcolor{pgrade10}56.96 & \cellcolor{pgrade7}13.90 & \cellcolor{pgrade10}13.95 & \cellcolor{pgrade9}8.25  & \cellcolor{pgrade5}33.84 & \cellcolor{pgrade7}0.97\\
	
	\multicolumn{9}{c}{\cellcolor{gray!20}\textbf{Vision, Language}}\\ 
	\multirow{2}{*}{Vision-Language}&CLIP      & alignment            & \cellcolor{pgrade7}53.85 & \cellcolor{pgrade10}11.30 & \cellcolor{pgrade10}14.65 & \cellcolor{pgrade10}7.62 & \cellcolor{pgrade5}34.35 & \cellcolor{pgrade6}0.98 \\
	&SigLIP    & alignment            & \cellcolor{pgrade3}40.91 & \cellcolor{pgrade2}41.87 & \cellcolor{pgrade3}27.52 & \cellcolor{pgrade3}40.19 & \cellcolor{pgrade2}38.54 & \cellcolor{pgrade5}1.27 \\
	\cline{1-9}
	\multirow{1}{*}{Reconstruction}&StableDiffusion & Generation & \cellcolor{pgrade5}50.39 & \cellcolor{pgrade4}21.50 & \cellcolor{pgrade5}21.22 & \cellcolor{pgrade5}15.64 & \cellcolor{pgrade5}33.81 & \cellcolor{pgrade2}1.30 \\[0.7em]
	\bottomrule
\end{tabular}
}
\setlength{\fboxsep}{0.2pt}
\newcommand{\letterbox}[2]{\colorbox{#1}{\raisebox{0pt}[0.7em][0.2em]{\makebox[0.7em][c]{#2}}}}
\caption{\textbf{Physics Probing} results for 10 VFMs, grouped by supervision type. Performance is encoded as \letterbox{pgrade1}{w}\letterbox{pgrade2}{o}\letterbox{pgrade3}{r}\letterbox{pgrade4}{s}\letterbox{pgrade5}{e}\letterbox{pgrade6}{-}\letterbox{pgrade7}{b}\letterbox{pgrade8}{e}\letterbox{pgrade9}{s}\letterbox{pgrade10}{t}.\vspace{-1.5em}
}
\label{tbl:physics_probing}
\end{table}

For experiments, we look at a set of ten prominent VFMs, all except Stable Diffusion~\cite{stablediff} use a transformer architecture. Models, training and metrics are detailed in Appendix~\cref{app:modelsspec}. 
We report results in~\cref{tbl:physics_probing}, grouping VFMs according to their supervision types. 
Overall, we note that \textit{self-supervised} models tend to perform better than other types, exhibiting that some sense of physics ability emerges.
In the \textit{fully-supervised} group, MiDaS~\cite{midas} which is trained with pixel-level depth, a somehow physical task, outperforms other models like DeiT~\cite{deit} or SAM~\cite{SAM} despite similar architectures.
Among self-supervised methods, DINO \cite{dino} stands out, demonstrating the effectiveness of its training strategy and achieving a large margin over gravity MAE.
For VLMs, the image encoder of CLIP \cite{clip} appears to outperform that of SigLIP \cite{siglip}, which may stem from the training data, although this is highly speculative.
The agglomerative model (AM-Radio~\cite{amradio}) performs well overall but at a cost of extremely slow inference.
Lastly, the generative model Stable Diffusion does not exhibit strong performance in physics probing, remaining inferior to the top models in the other groups.
Scene flow results appear to be correlated with physics probing performance, consolidating the intuition that motion prediction is a viable proxy for approximating physical reasoning.

On Gravity-OOD results, we observe angular prediction performance (mAE) comparable to Gravity, as expected since direction does not change, but we note a severe degradation in magnitude (magE) showing that all models have severely to Earth gravity which is their sole observation point.

Empirical results indicate that stronger visual representations tend to yield better performance in physics probing. However, the current performance remains far from being useful for downstream applications and we encourage future work to explore alternative supervision strategies, together with suitable datasets, to learn of visual representations that are truly physics-grounded.

\section{Discussion}
\label{sec:discussion}

\method{} provides dense, physically grounded annotations, but several aspects leave room for extension. Our rendering pipeline prioritizes physical consistency over full photometric realism and does not model complex lighting effects such as rich shadows or advanced specularity. The simulator also inherits modeling assumptions (\eg, contact, material parameterization, and actuation design) that can introduce a domain gap relative to real-world dynamics. Our evaluation focuses on open-source foundation models under standard inference settings for reproducibility and controlled comparison, though future studies may broaden the scope to proprietary systems, stronger test-time reasoning, and alternative prompting or tool use. Finally, controlled human studies remain an important direction for establishing reference points for low-level Newtonian reasoning.

Beyond evaluation, \method{} is designed as an extensible framework. The simulation and annotation pipeline enables researchers to construct new tasks, filter scenes by physical events, and probe models across levels of abstraction using pixel-aligned supervision over time. We see \method{} not merely as a benchmark, but as a foundation for investigating how physical structure is represented, reasoned about, and ultimately integrated into emerging visual world models.

\paragraph{Acknowledgments.}
This work was conducted at Inria. It was supported by the European Union’s Horizon Europe research and innovation programme under grant agreement number 101214398 (ELLIOT).

\clearpage
\appendix

\setcounter{tocdepth}{2} 

\section*{Appendix}

{The appendix details the Newtonian physics simulator (\cref{app:simulator}), \method{} dataset statistics (\cref{app:dataset}), as well as the taxonomy, questions, and VQA creation details (\cref{app:vqa}). 
It further includes specification of the experts-to-notice questions (\cref{app:five_levels}) and list all model specifications (\cref{app:modelsspec}).}

\vspace{0.5cm}
\hrule
\vspace{0.3em}
\startcontents[appendices]

\titlecontents{section}[2.5em]{\vspace{0.15cm}\bfseries}{\contentslabel{2.0em}}{}{\titlerule*[0.5pc]{.}\contentspage}

\titlecontents{subsection}[4.5em]{\normalfont}{\contentslabel{2.0em}}{}{\titlerule*[0.5pc]{.}\contentspage}

\printcontents[appendices]{}{1}{}
\hrule
\vspace{0.5cm}

\section{Detailed Newtonian physics simulation}
\label{app:simulator}

In the following, we provide additional details about the physical simulator, mentioned in main paper Sec.~3.1, which constitutes the backbone of our benchmark.

The key complexity of simulating 3DGS $G = \{(\mu_i,\sigma_i, c_i, \alpha_i)\}^N_{i=1}$ is that they encode sparse and unstructured information, as opposed to meshes which are structured 3D data, typically preferred for simulation. We highlight that while prior works~\cite{guedon2024sugar,guedon2025milo} demonstrated the ability to optimize meshes with 3DGS, our experiments show that they hardly generalize across scenes and objects, and that even a slightly noisy mesh can produce highly unrealistic simulation with mesh-dependent simulators. 

We instead rely on Simplicits~\cite{modi2024simplicits}, an extension of the Nvidia Kaolin~\cite{kaolinlib}, which enables mesh-free simulator for time-varying elastodynamics by learning reduced deformation space of complex shapes. We careful extend the latter to allow handling large scenes ($\approx{}10^7$ particles) with up to 50 objects.

\newcommand{\state}{\ensuremath{\textbf{z}}}
\newcommand{\stateOrderOne}{\ensuremath{\bar{\textbf{z}}}}
\newcommand{\massTensor}{\ensuremath{\textbf{M}}}
\newcommand{\EPot}{\ensuremath{E_\text{pot}}}
\newcommand{\skinningFct}{\ensuremath{\phi}}

\subsection{Solver adaptation.}

Let us consider a single deformable object with particles $x \in \mathcal{X}$. Rather than simulating all particles, Simplicits models a simulation state as a time-varying vector $\state_t$ having much reduced Degrees of Freedom (DoFs) \wrt $\mathcal{X}$. The next state is estimated using a Newton-based solver with the following optimization objective:
\begin{equation}
	\state_{t+1} = \arg \min_\state \frac{1}{2} \| \state - \stateOrderOne_t \|_\massTensor^2 + h^2 E_{\text{pot}}(\state),
	\label{eq:simp_solver}
\end{equation}
where $||\cdot||^2_\massTensor$ is the squared norm weighted by the mass matrix $\massTensor$, $h$ the simulation timestep, $\stateOrderOne_t$ is the first order predictor for \state{}, and \EPot{} is the potential energy of the system. We note that collision and forces constraints are also added to \cref{eq:simp_solver}.
The mapping $\skinningFct(Z)\mapsto{}X$ is learned with a small skinning network $\skinningFct(\cdot)$ as-yet undefined, given a fixed number of deformable handles and physical properties. Having no mesh, the mass and potential energy \EPot{} are approximated by \textit{cubature points} randomly sampled in $X$, \ie, $Q \in X$. Intuitively, cubature points serve as locations for solving the physical forces of the system.

To simulate a full scene we consider only the centers of our 3DGS particles $X=\{\mu_i\}^N_{i=1}$ as our physical world, and apply crucial adjustments.
Instead of individual object, we define our simulation state as the union of all objects in the scene such that $\state_\text{sim} = \{\state_\text{scene}, \state_{\text{obj}_1}, \dots, \state_{\text{obj}_N}\}$. 
By keeping track of the mapping between particles and objects, the complete 3DGS state can then be updated after each simulation step by applying the relevant $\skinningFct(\cdot)$ mapping for each object: $G = \{\skinningFct_\text{scene}(\cdot), \skinningFct_{\text{obj}_1}(\cdot), \dots, \skinningFct_{\text{obj}_N}(\cdot)\}$.
Further, since highly deformable objects intuitively require more DoFs, rather than using a fixed set of handles, we logarithmically vary the number as a function of each object softness. This reduces the rank of $\state_\text{sim}$ and significantly lowers the complexity of solving~\cref{eq:simp_solver}. Another source of computational cost is the large number of cubature points needed for accurate simulation. 
Using only a few cubature points leads to penetration/collision of objects due to lack of physics evaluation basis, while large numbers bring intractable computational costs for our large scenes. Instead, we employ a simple strategy that primarily samples cubature points near the object spawning positions where fine-grained physics interaction is likely to occur. Lastly, to compensate for sparse evaluation basis, each cubature point is modeled as a small sphere rather than a point location.

\paragraph{Assessing correctness of the simulation.}
{By design, simulation requires 3DGS primitives to lie on object surfaces. For objects, because the GSO dataset provides all-around views, the resulting 3DGS reconstructions are highly accurate; leading to only 2.48\% error in dimensions upon manual verifications on 20 objects.}
{Instead for the scene reconstructions, the combination of VGGT and COLMAP enables high-density as well as accurate 3DGS reconstruction. Furthermore, our simulations account for sparsity by treating primitives as small spheres (r=3mm), thereby theoretically ensuring tight 6 mm-precision collisions. These choices, together, ensure primitives on the surface, enabling our large-scene, automated simulation pipeline.}

We highlight that, without noticeable losses in quality, our adaptation drastically lowers the complexity of each simulation which typically have $10^7$ 3DGS to a simulation state integrated over ${\approx}10^4$ cubature points with only ${\approx}10^2$ DoFs. Further, the object-to-particle mapping allows retrieving per-point forces which is crucial for our need of physically-annotated dataset. 

\subsection{Estimation of objects' physical properties.} While the scene itself is kinematic (\ie, non-deformable) the objects need tailored skinning functions for deformation. We follow~\cite{modi2024simplicits}, first densifying 3DGS models to compensate for 3DGS only capturing visible shell, and then learning skinning functions from small individual networks (10 MLP layers) given the densified 3DGS centers (\ie, rest pose) and physics properties. The training objective minimizes the elastic energy \wrt the rest pose while enforcing orthogonality of the skinning weights. 
We highlight that GSO~\cite{downs2022google} does \textit{not} provide physical quantities\footnote{Note that while GSO mentions the release of physical labels, the absence of the latter was confirmed via email by the Google GSO team.}. Therefore, for each object, we estimate the possible Young modulus, Poisson ratio, and density by querying GPT-5 with 4 object views, while also querying for names of the visible materials. Object volumes are then computed with a greedy Monte Carlo from our metric-scaled 3DGS, which allows accurate estimation of their masses. 
{All properties are manually verified.}

\begin{figure}[t]
    \centering
    \vspace{-0.6em}

    \begin{subfigure}{0.48\linewidth}
        \centering
        \includegraphics[width=\linewidth]{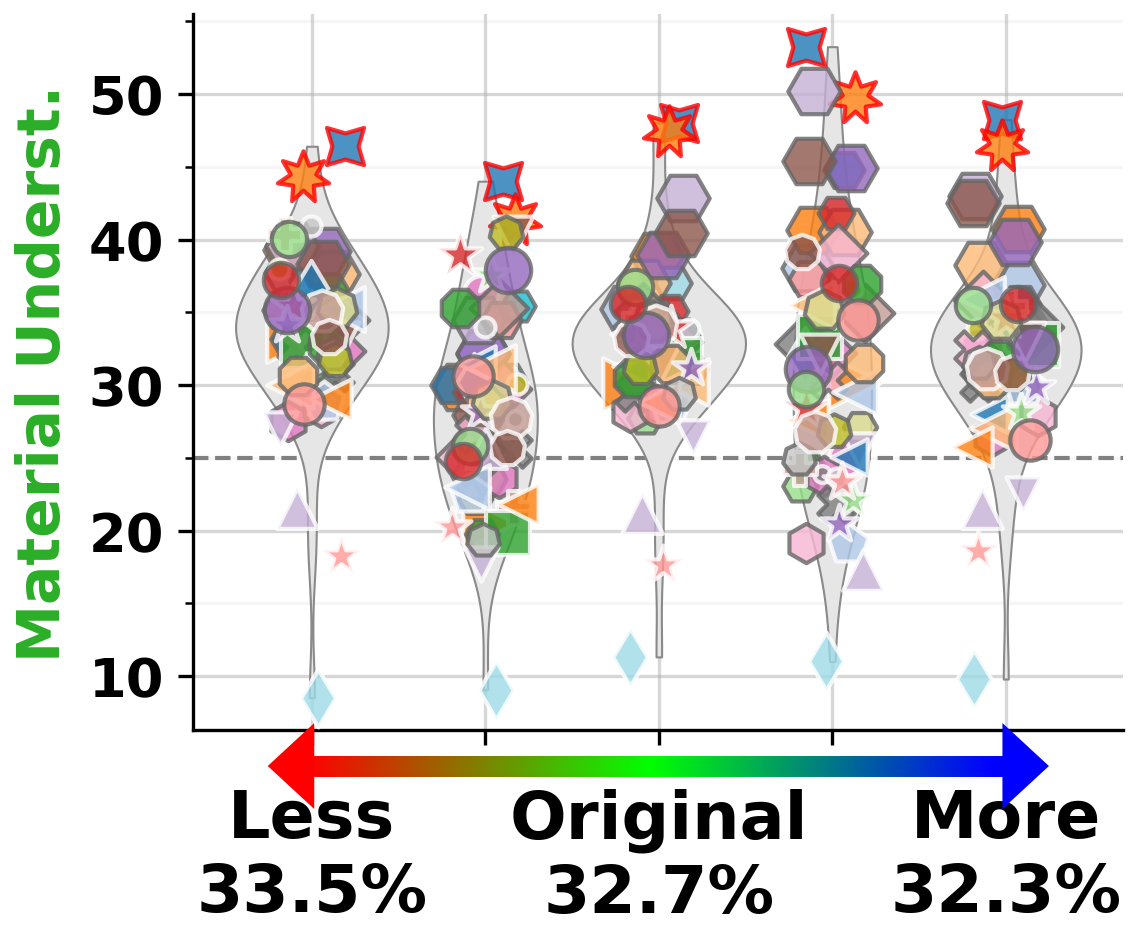}
        \caption{\textbf{YMS shift}}
        \label{fig:yms_shift}
    \end{subfigure}
    \hfill
    \begin{subfigure}{0.48\linewidth}
        \centering
        \includegraphics[width=\linewidth]{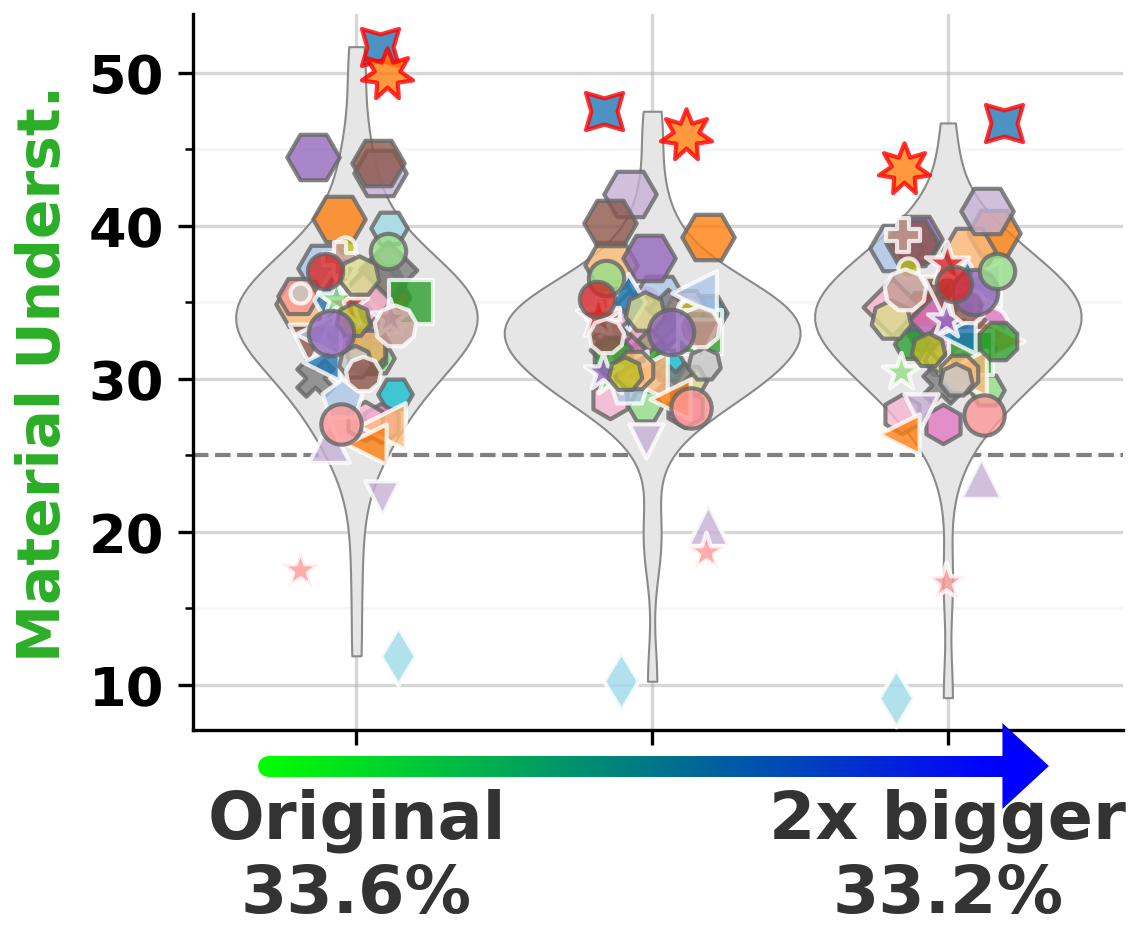}
        \caption{\textbf{Object scale}}
        \label{fig:scale_shift}
    \end{subfigure}

    \caption{\textbf{Robustness to intrinsic perturbations.}
    We perturb \subref{fig:yms_shift} Young's Modulus and \subref{fig:scale_shift} object scale and observe that VQA accuracy remains largely stable, indicating that our original intrinsics are coherent.}
    \label{fig:intrinsics_shift}
    \vspace{-0.8em}
\end{figure}

\paragraph{Assessing properties correctness.}
{In order to assess the correctness of properties, we manually verified vendor properties by scraping the web for 20 GSO objects. This further check shows that our estimated properties yield an average error of 14.58\% for mass (which accounts for both dimension and density), and 36.09\% for volume -- so that our properties are reliably close. The volume error can be explained by the cubic nature of the metric, and the Monte-Carlo volume estimation approximating discrete points as small spheres, which biases volumes of small objects. Furthermore, to assess the impact of our intrinsics, we perturb the Young's Modulus (YMS) in~\cref{fig:yms_shift} and scale in~\cref{fig:scale_shift}, of all models on subsets of \method{}. For both our 'original' intrinsics, we observe that the modification leads to coherent VQA accuracy that does not change significantly under perturbation. }

\subsection{Rendering}
At each time step, the Simplicits routine resolves all physics constraint with a Newton solver while saving only the resulting particles states. 

\paragraph{RGB rendering.} 
To render the simulation, we update 3DGS positions and orientations from the current simulation state $\state_t$, using the objects skinning functions, and simply render with the vanilla 3DGS rasterizer~\cite{kerbl20233d}. 

\paragraph{Ground-truth maps.} 
With some engineering efforts, {we capture individual forces} (material stress, gravity, collision, \etc) along with other kinematics, semantics and geometric labels, by modifying the Newton solver routine in Simplicits. 

Rendering per-pixel forces from camera perspective is complex. We do so by duplicating the 3DGS renderers, obtaining one renderer per force to extract (and one for RGB). We then use each \textit{force renderer} as a proxy to store the per-point force value of each Gaussian mapped to spherical harmonics (\ie, ultimately to RGB) after careful binarization of the Gaussians' opacity\footnote{We highlight that binarization is required. Failure to do so, would lead to integration of multiple force values along a single ray, making per pixel invalid, ergo, erroneous physical maps.}. 
Importantly, forces renderers do \textit{not} use the skinning functions, as the latter are only valid in the spatial domain. Instead, we use a nearest-neighbour cubature to 3DGS mapping. Subsequently, each 3DGS particle is rendered as having the force value of the closest cubature point. 
Given that our objects are relatively small in size and uniformly covered by cubature points, the resulting approximation is found to be negligible.
A similar process is followed to render kinematics, semantics, and geometry maps.

\paragraph{Events recording.} 
Alongside RGB and physical maps, events like collisions, camera motion, object visibility, \etc are capture at each rendering steps, and encoded as a JSON file which later allows, along with the physical maps, for automatic scripting of Visual Question Answering (VQA).

\paragraph{Ground truth consistency.}
{Since labels are rendered directly from the simulator state used to generate the observations as described above, \textit{they are always perfectly aligned with the observed dynamics}. Thus, the labels remain consistent with the simulated scene, independently of the simulator's intrinsic physical accuracy.}

\section{Additional dataset details}
\label{app:dataset}

\begin{figure}[h!]
	\newcommand{\TeaserLeftW}{1.0\linewidth}
	\newcommand{\TeaserRightW}{0.27\linewidth}
	\newcommand{\TeaserBigH}{7cm}
	
	\renewcommand{\InsetY}{0.011}
	\renewcommand{\InsetLeftFrac}{0.44}  %
	\renewcommand{\InsetW}{0.15}
	\pgfmathsetmacro{\InsetX}{\InsetLeftFrac + 0.5*\InsetW} %
	\renewcommand{\InsetOpacity}{1.0}
	
	\noindent
	\begin{minipage}[t]{\TeaserLeftW}
		\vspace{0pt} %
		\centering
		\begin{tikzpicture}[baseline=(big.north)] %
			\node[
			anchor=north west,
			inner sep=0,
			outer sep=0
			] (big) at (0,0) {%
				\includegraphics[width=\linewidth]{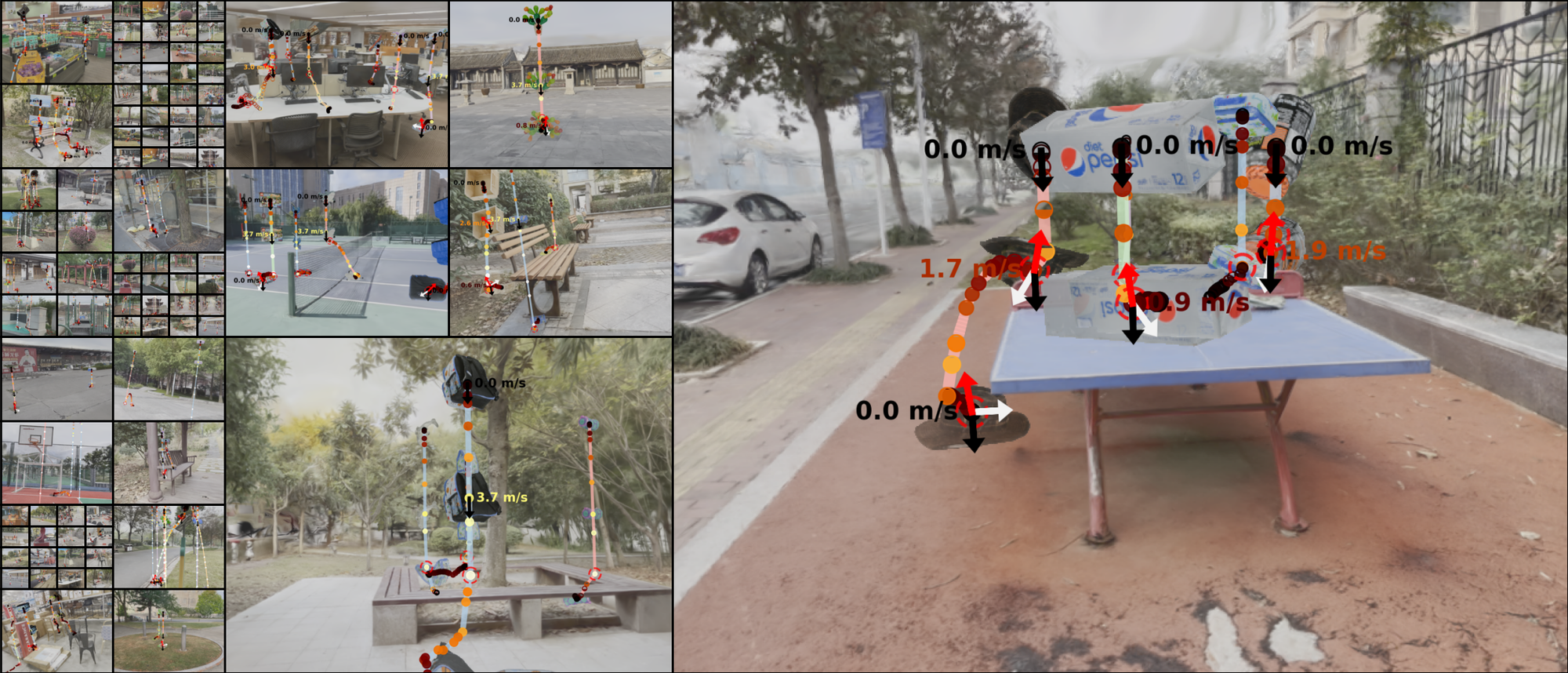}%
			};
			
			\coordinate (bm) at ($(big.south west)!0.51!(big.south east)$); %
			\coordinate (tm) at ($(big.north west)!0.51!(big.north east)$);
			
			\node[
			anchor=south,
			inner sep=0,
			outer sep=0,
			opacity=\InsetOpacity
			] at ($(bm)!\InsetY!(tm)$) {%
				\includegraphics[width=\InsetW\linewidth]{figures/viz/legend.png}%
			};
		\end{tikzpicture}
	\end{minipage}%
	
	\vspace{-0.6em}
	\captionsetup{type=figure}
	\captionof{figure}{
		\textbf{\method{} additional sequences}. We display sequences from our \method{} benchmark, though rendered from a stationary viewpoint for visualization purpose. Dynamics are highlighted as overlaid (cf. inset legend).
	}
	\label{fig:dataset_more_viz}
\end{figure}

\cref{fig:dataset_more_viz} shows additional exemplar sequences with physical annotations. For better forces visualization, we highlight that renderings are taken from stationary cameras and faded gray. \textit{We refer to the website for full quality illustration of simulation sequences.}

\begin{figure}[h!]
	\centering
	
	\newcommand{\rowgap}{\vspace{2pt}}
	
	\newlength{\statXh}
	\setlength{\statXh}{0.18\linewidth} %
	
	\newcommand{\statimg}[2]{%
		\begin{minipage}[b]{#1}
			\centering
			\includegraphics[width=\linewidth,height=\statXh,keepaspectratio]{#2}
		\end{minipage}%
	}
	
	\newcommand{\statimgV}[2]{%
		\begin{minipage}[b]{#1}
			\centering
			\includegraphics[width=\linewidth,height=1.5\statXh,keepaspectratio]{#2}
		\end{minipage}%
	}
	
	\begin{minipage}[m]{0.75\linewidth}
		\centering
		
		{\scriptsize\textbf{Sequences characteristics}}\\
		\statimg{0.33\linewidth}{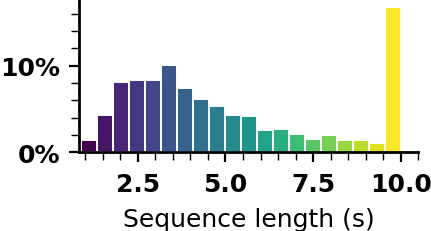}\hfill
		\statimg{0.33\linewidth}{figures/viz/v4_dl3dv_random/stats/stats_collisions_events.png}\hfill
		\statimg{0.33\linewidth}{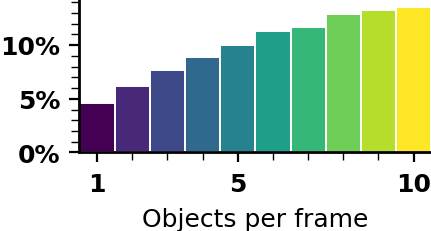}\\

		{\scriptsize\textbf{Object's physical properties}}\\
		\statimg{0.33\linewidth}{figures/viz/v4_dl3dv_random/stats/stats_yms.png}\hfill
		\statimg{0.33\linewidth}{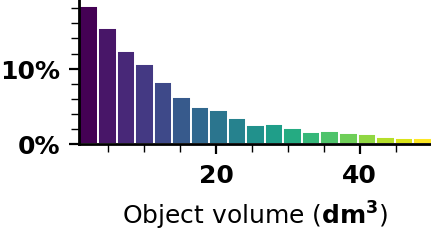}\hfill
		\statimg{0.33\linewidth}{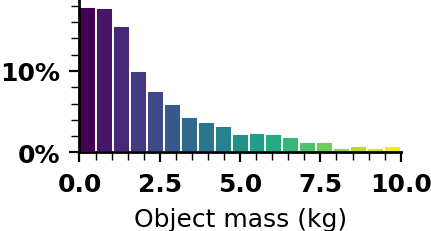}

		{\scriptsize\textbf{Motion and Visibility}}\\
		\statimg{0.33\linewidth}{figures/viz/v4_dl3dv_random/stats/stats_motion.png}\hfill
		\statimg{0.33\linewidth}{figures/viz/v4_dl3dv_random/stats/stats_cam_motion.png}\hfill
		\statimg{0.33\linewidth}{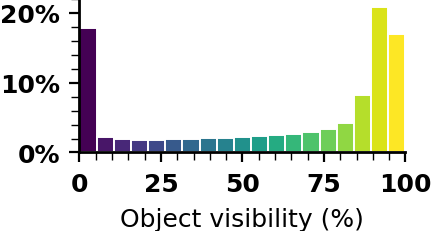}
		
	\end{minipage}\hfill
	\begin{minipage}[m]{0.25\linewidth}
		\centering
		
		\statimg{1.0\linewidth}{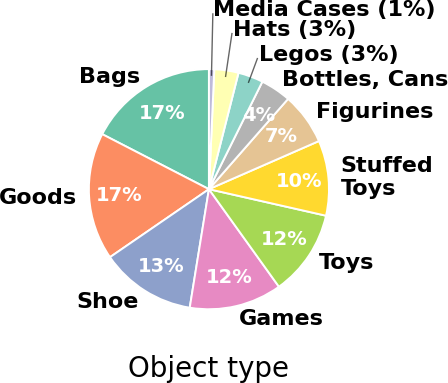}
		
		\vspace{0.3cm}
		
		\statimg{1.0\linewidth}{figures/viz/v4_dl3dv_random/stats/stats_material_pie.png}
		
	\end{minipage}
	
	\caption{\textbf{Detailed datasets statistics.} We report here additional dataset statistics, highlighting the variability of \method{} dataset along various axes of study.}
	\label{fig:dataset_stats_details}
\end{figure}

Further, we report additional statistics in \cref{fig:dataset_stats_details}. They highlight the high diversity and variability of scenes and dynamics included in \method{} benchmark.

\section{Details on Visual Question Answering (VQA)}
\label{app:vqa}
In the following, we detail the taxonomy used for \method{} VQA (\cref{app:question_taxonomy}), along with exemplar VQA with visuals (\cref{app:vqa_examples}), and details on our automation process (\cref{app:details_vqa_creation}).

\clearpage 
\onecolumn
\subsection{Taxonomy}
\label{app:question_taxonomy}

We design multiple-choice questions (MCQ) having 4 answers, grouped into five high-level categories. Each question is instantiated from a template by replacing placeholders (\eg, \texttt{<OBJECT>}, \texttt{<OBJECT\_1>}, \texttt{<OBJECT\_2>}) with scene-specific object names. We also define task splits: \textit{single}-frame (answerable from one frame) and \textit{multi}-frame (requires temporal evidence across a sequence). We try to use different scenes for each question to showcase the variety of the dataset.

\newcolumntype{C}[1]{>{\centering\arraybackslash}p{#1}}
\newcolumntype{L}[1]{>{\raggedright\arraybackslash}p{#1}}
\definecolor{catmechanics}{HTML}{FF5733}
\definecolor{catspatialreasoning}{HTML}{3498DB}
\definecolor{catpermanence}{HTML}{F43FC7}
\definecolor{cattemporal}{HTML}{0DA792}
\definecolor{catviewpoint}{HTML}{EEAC32}
\definecolor{catmaterialunderstanding}{HTML}{2BAE27}
\definecolor{catvisualpercetion}{HTML}{9B59B6}

\newcommand{\taxonomyCat}[2]{\noindent\makebox[\columnwidth][l]{%
		\colorbox{#1!18}{%
			\parbox{\columnwidth}{\rule{0pt}{1.25em}\bfseries{}#2}%
		}%
}%
}

\small
\renewcommand{\arraystretch}{1.12}
\setlength{\tabcolsep}{2pt}

\begin{longtable}{| C{0.05\textwidth} | C{0.10\textwidth} | L{0.85\textwidth} |}
	\caption{SimpleVQA question IDs grouped by category and sub-category.} \\
	\hline
	\rowcolor[gray]{0.9}
	\rotatebox[origin=c]{90}{\textbf{Category}} & \rotatebox[origin=c]{90}{\textbf{\shortstack[c]{Sub-\\Category}}} & \textbf{Question ID and Prompt} \\ \hline
	\endfirsthead
	\hline
	\rowcolor[gray]{0.9}
	\rotatebox[origin=c]{90}{\textbf{Category}} & \rotatebox[origin=c]{90}{\textbf{\shortstack[c]{Sub-\\Category}}} & \textbf{Question ID and Prompt} \\ \hline
	\endhead
	\hline
	\multicolumn{3}{r}{\textit{Continued on next page}} \\
	\endfoot
	\hline
	\endlastfoot
	
	\multirow[t]{7}{*}{\vspace{7pt}\raisebox{0pt}{\rotatebox[origin=c]{90}{\colorbox{catspatialreasoning!18}{\hspace{2pt}\rule{0pt}{1.25em}\bfseries Spatial Reasoning\hspace{2pt}}\hspace{92pt}}}} & \multirow[t]{4}{*}{\vspace{7pt}\raisebox{0pt}{\rotatebox[origin=c]{90}{\colorbox{catspatialreasoning!10}{\hspace{2pt}\rule{0pt}{1.25em}\bfseries Distance\hspace{2pt}}\hspace{48pt}}}} & \colorbox{catspatialreasoning!6}{\parbox{\dimexpr\linewidth-2\fboxsep\relax}{\raggedright\arraybackslash \footnotesize\ttfamily\bfseries F\_\allowbreak{}DISTANCE\_\allowbreak{}OBJECT\_\allowbreak{}OBJECT}}\par \textbf{Single Image:} Based on the image, what is the real-world distance between <OBJECT\_1> and the <OBJECT\_2>?\par \textbf{Multi Image:} Considering all frames, what is the real-world distance between <OBJECT\_1> and the <OBJECT\_2> in the last frame? \\
	\cline{3-3}
	&  & \colorbox{catspatialreasoning!6}{\parbox{\dimexpr\linewidth-2\fboxsep\relax}{\raggedright\arraybackslash \footnotesize\ttfamily\bfseries F\_\allowbreak{}CLOSEST\_\allowbreak{}OBJECT\_\allowbreak{}OBJECT}}\par \textbf{Single Image:} Based on the image, which object is closest to the <OBJECT> in real-world distance?\par \textbf{Multi Image:} Considering all frames, which object is closest to the <OBJECT> in real-world distance in the last frame? \\
	\cline{3-3}
	&  & \colorbox{catspatialreasoning!6}{\parbox{\dimexpr\linewidth-2\fboxsep\relax}{\raggedright\arraybackslash \footnotesize\ttfamily\bfseries F\_\allowbreak{}DISTANCE\_\allowbreak{}OBJECT\_\allowbreak{}CAMERA\_\allowbreak{}DISTANCE}}\par \textbf{Single Image:} Based on the image, what is the real-world distance between the <OBJECT> and the camera?\par \textbf{Multi Image:} Considering all frames, what is the real-world distance between the <OBJECT> and the camera in the last frame? \\
	\cline{3-3}
	&  & \colorbox{catspatialreasoning!6}{\parbox{\dimexpr\linewidth-2\fboxsep\relax}{\raggedright\arraybackslash \footnotesize\ttfamily\bfseries F\_\allowbreak{}CLOSEST\_\allowbreak{}OBJECT\_\allowbreak{}CAMERA}}\par \textbf{Single Image:} Based on the image, which object is the closest to the camera in real-world distance?\par \textbf{Multi Image:} Considering all frames, which object is the closest to the camera in real-world distance in the last frame? \\
	\cline{2-3}
	& \multirow[t]{2}{*}{\vspace{7pt}\raisebox{0pt}{\rotatebox[origin=c]{90}{\colorbox{catspatialreasoning!10}{\hspace{2pt}\rule{0pt}{1.25em}\bfseries Size\hspace{2pt}}\hspace{24pt}}}} & \colorbox{catspatialreasoning!6}{\parbox{\dimexpr\linewidth-2\fboxsep\relax}{\raggedright\arraybackslash \footnotesize\ttfamily\bfseries F\_\allowbreak{}SIZE\_\allowbreak{}OBJECT}}\par \textbf{Single Image:} Based on the image, what are the real-world dimensions of the <OBJECT>?\par \textbf{Multi Image:} Considering all frames, what are the real-world dimensions of the <OBJECT> in the last frame? \\
	\cline{3-3}
	&  & \colorbox{catspatialreasoning!6}{\parbox{\dimexpr\linewidth-2\fboxsep\relax}{\raggedright\arraybackslash \footnotesize\ttfamily\bfseries F\_\allowbreak{}SIZE\_\allowbreak{}OBJECT\_\allowbreak{}BIGGER}}\par \textbf{Single Image:} Based on the image, which single object has the biggest real-world volume?\par \textbf{Multi Image:} Considering all frames, which single object, visible in the last frame, has the biggest real-world volume? \\
	\cline{2-3}
	& \multirow[t]{1}{*}{\vspace{7pt}\raisebox{0pt}{\rotatebox[origin=c]{90}{\colorbox{catspatialreasoning!10}{\hspace{2pt}\rule{0pt}{1.25em}\bfseries Layout\hspace{2pt}}\hspace{36pt}}}} & \colorbox{catspatialreasoning!6}{\parbox{\dimexpr\linewidth-2\fboxsep\relax}{\raggedright\arraybackslash \footnotesize\ttfamily\bfseries F\_\allowbreak{}LAYOUT\_\allowbreak{}POSITION\_\allowbreak{}OBJECT\_\allowbreak{}OBJECT}}\par \textbf{Single Image:} From the camera’s perspective, where is the <OBJECT\_1> relative to the <OBJECT\_2> in the image?\par \textbf{Multi Image:} Considering all frames, from the camera’s perspective, where is the <OBJECT\_1> relative to the <OBJECT\_2> in the last frame? \\
	\hline
	\multirow[t]{7}{*}{\vspace{7pt}\raisebox{0pt}{\rotatebox[origin=c]{90}{\colorbox{catmechanics!18}{\hspace{2pt}\rule{0pt}{1.25em}\bfseries Mechanics\hspace{2pt}}\hspace{54pt}}}} & \multirow[t]{4}{*}{\vspace{7pt}\raisebox{0pt}{\rotatebox[origin=c]{90}{\colorbox{catmechanics!10}{\hspace{2pt}\rule{0pt}{1.25em}\bfseries Kinematics\hspace{2pt}}\hspace{60pt}}}} & \colorbox{catmechanics!6}{\parbox{\dimexpr\linewidth-2\fboxsep\relax}{\raggedright\arraybackslash \footnotesize\ttfamily\bfseries F\_\allowbreak{}KINEMATICS\_\allowbreak{}SPEED\_\allowbreak{}OBJECT}}\par \textbf{Multi Image:} Considering all frames, what is the real-world speed of the <OBJECT> at the time of the last frame? \\
	\cline{3-3}
	&  & \colorbox{catmechanics!6}{\parbox{\dimexpr\linewidth-2\fboxsep\relax}{\raggedright\arraybackslash \footnotesize\ttfamily\bfseries F\_\allowbreak{}KINEMATICS\_\allowbreak{}ACCEL\_\allowbreak{}OBJECT}}\par \textbf{Multi Image:} Considering all frames, what is the real-world magnitude of acceleration of the <OBJECT> at the time of the last frame? \\
	\cline{3-3}
	&  & \colorbox{catmechanics!6}{\parbox{\dimexpr\linewidth-2\fboxsep\relax}{\raggedright\arraybackslash \footnotesize\ttfamily\bfseries F\_\allowbreak{}KINEMATICS\_\allowbreak{}DISTANCE\_\allowbreak{}TRAVELED\_\allowbreak{}INTERVAL}}\par \textbf{Multi Image:} Considering all frames, what is the real-world displacement of the centroid of the <OBJECT> from the first to the last frame? \\
	\cline{3-3}
	&  & \colorbox{catmechanics!6}{\parbox{\dimexpr\linewidth-2\fboxsep\relax}{\raggedright\arraybackslash \footnotesize\ttfamily\bfseries F\_\allowbreak{}KINEMATICS\_\allowbreak{}SYSTEM\_\allowbreak{}STABILITY}}\par \textbf{Multi Image:} Analyzing the motion trend across the sequence, which statement best describes the system's state at the final frame? \\
	\cline{2-3}
	& \multirow[t]{3}{*}{\vspace{7pt}\raisebox{0pt}{\rotatebox[origin=c]{90}{\colorbox{catmechanics!10}{\hspace{2pt}\rule{0pt}{1.25em}\bfseries Collision\hspace{2pt}}\hspace{54pt}}}} & \colorbox{catmechanics!6}{\parbox{\dimexpr\linewidth-2\fboxsep\relax}{\raggedright\arraybackslash \footnotesize\ttfamily\bfseries F\_\allowbreak{}COLLISION\_\allowbreak{}OBJECT\_\allowbreak{}OBJECT\_\allowbreak{}FRAME\_\allowbreak{}SINGLE}}\par \textbf{Single Image:} Based on the image, which visible object is the <OBJECT> colliding with? \\
	\cline{3-3}
	&  & \colorbox{catmechanics!6}{\parbox{\dimexpr\linewidth-2\fboxsep\relax}{\raggedright\arraybackslash \footnotesize\ttfamily\bfseries F\_\allowbreak{}COLLISION\_\allowbreak{}OBJECT\_\allowbreak{}OBJECT\_\allowbreak{}FRAME\_\allowbreak{}MULTI}}\par \textbf{Multi Image:} In which frame is the <OBJECT> most likely colliding with another object? \\
	\cline{3-3}
	&  & \colorbox{catmechanics!6}{\parbox{\dimexpr\linewidth-2\fboxsep\relax}{\raggedright\arraybackslash \footnotesize\ttfamily\bfseries F\_\allowbreak{}COLLISION\_\allowbreak{}OBJECT\_\allowbreak{}SCENE\_\allowbreak{}FRAME\_\allowbreak{}MULTI}}\par \textbf{Multi Image:} In which frame is the <OBJECT> most likely colliding with the static scene? \\
	\hline
	\multirow[t]{21}{*}{\vspace{7pt}\raisebox{0pt}{\rotatebox[origin=c]{90}{\colorbox{catmaterialunderstanding!18}{\hspace{2pt}\rule{0pt}{1.25em}\bfseries Material Understanding\hspace{2pt}}\hspace{126pt}}}} & \multirow[t]{3}{*}{\vspace{7pt}\raisebox{0pt}{\rotatebox[origin=c]{90}{\colorbox{catmaterialunderstanding!10}{\hspace{2pt}\rule{0pt}{1.25em}\bfseries Mass\hspace{2pt}}\hspace{30pt}}}} & \colorbox{catmaterialunderstanding!6}{\parbox{\dimexpr\linewidth-2\fboxsep\relax}{\raggedright\arraybackslash \footnotesize\ttfamily\bfseries F\_\allowbreak{}MASS\_\allowbreak{}OBJECT}}\par \textbf{Single Image:} Based on the image, what is the mass of the <OBJECT>?\par \textbf{Multi Image:} Considering all frames, what is the mass of the <OBJECT>? \\
	\cline{3-3}
	&  & \colorbox{catmaterialunderstanding!6}{\parbox{\dimexpr\linewidth-2\fboxsep\relax}{\raggedright\arraybackslash \footnotesize\ttfamily\bfseries F\_\allowbreak{}MASS\_\allowbreak{}HEAVIEST\_\allowbreak{}OBJECT}}\par \textbf{Single Image:} Based on the image, which visible single object has the greatest mass?\par \textbf{Multi Image:} Considering all frames, which single object, visible in the last frame, has the greatest mass? \\
	\cline{3-3}
	&  & \colorbox{catmaterialunderstanding!6}{\parbox{\dimexpr\linewidth-2\fboxsep\relax}{\raggedright\arraybackslash \footnotesize\ttfamily\bfseries F\_\allowbreak{}MASS\_\allowbreak{}LIGHTEST\_\allowbreak{}OBJECT}}\par \textbf{Single Image:} Based on the image, which visible single object has the least mass?\par \textbf{Multi Image:} Considering all frames, which single object, visible in the last frame, has the least mass? \\
	\cline{2-3}
	& \multirow[t]{2}{*}{\vspace{7pt}\raisebox{0pt}{\rotatebox[origin=c]{90}{\colorbox{catmaterialunderstanding!10}{\hspace{2pt}\rule{0pt}{1.25em}\bfseries Density\hspace{2pt}}\hspace{42pt}}}} & \colorbox{catmaterialunderstanding!6}{\parbox{\dimexpr\linewidth-2\fboxsep\relax}{\raggedright\arraybackslash \footnotesize\ttfamily\bfseries F\_\allowbreak{}PHYSICS\_\allowbreak{}PROPERTY\_\allowbreak{}DENSITY\_\allowbreak{}OBJECT}}\par \textbf{Single Image:} Based on the image, what is the estimated mean density of the <OBJECT>?\par \textbf{Multi Image:} Considering all frames, what is the estimated mean density of the <OBJECT>? \\
	\cline{3-3}
	&  & \colorbox{catmaterialunderstanding!6}{\parbox{\dimexpr\linewidth-2\fboxsep\relax}{\raggedright\arraybackslash \footnotesize\ttfamily\bfseries F\_\allowbreak{}PHYSICS\_\allowbreak{}PROPERTY\_\allowbreak{}DENSITY\_\allowbreak{}OBJECT\_\allowbreak{}RELATIVE}}\par \textbf{Single Image:} Based on the image, which visible object has the highest effective density?\par \textbf{Multi Image:} Considering all frames, which object, visible in the last frame, has the highest effective density? \\
	\cline{2-3}
	& \multirow[t]{6}{*}{\vspace{7pt}\raisebox{0pt}{\rotatebox[origin=c]{90}{\colorbox{catmaterialunderstanding!10}{\hspace{2pt}\rule{0pt}{1.25em}\bfseries \shortstack[c]{Young\\Modulus}\hspace{2pt}}\hspace{60pt}}}} & \colorbox{catmaterialunderstanding!6}{\parbox{\dimexpr\linewidth-2\fboxsep\relax}{\raggedright\arraybackslash \footnotesize\ttfamily\bfseries F\_\allowbreak{}PHYSICS\_\allowbreak{}PROPERTY\_\allowbreak{}YOUNG\_\allowbreak{}MODULUS\_\allowbreak{}OBJECT\_\allowbreak{}METRIC\_\allowbreak{}PREFIX}}\par \textbf{Single Image:} Based on the image, what is the Young's modulus of the <OBJECT>?\par \textbf{Multi Image:} Considering all frames, what is the Young's modulus of the <OBJECT>? \\
	\cline{3-3}
	&  & \colorbox{catmaterialunderstanding!6}{\parbox{\dimexpr\linewidth-2\fboxsep\relax}{\raggedright\arraybackslash \footnotesize\ttfamily\bfseries F\_\allowbreak{}PHYSICS\_\allowbreak{}PROPERTY\_\allowbreak{}YOUNG\_\allowbreak{}MODULUS\_\allowbreak{}OBJECT\_\allowbreak{}SIMILAR}}\par \textbf{Single Image:} Based on the image, which visible object has a Young's Modulus most similar to that of the <OBJECT>?\par \textbf{Multi Image:} Considering all frames, which object, visible in the last frame, has a Young's Modulus most similar to that of the <OBJECT>? \\
	\cline{3-3}
	&  & \colorbox{catmaterialunderstanding!6}{\parbox{\dimexpr\linewidth-2\fboxsep\relax}{\raggedright\arraybackslash \footnotesize\ttfamily\bfseries F\_\allowbreak{}PHYSICS\_\allowbreak{}PROPERTY\_\allowbreak{}YOUNG\_\allowbreak{}MODULUS\_\allowbreak{}OBJECT\_\allowbreak{}SIMILAR\_\allowbreak{}NON\_\allowbreak{}TECHNICAL}}\par \textbf{Single Image:} Based on the image, which visible object has a softness most similar to that of the <OBJECT>?\par \textbf{Multi Image:} Considering all frames, which object, visible in the last frame, has a softness most similar to that of the <OBJECT>? \\
	\cline{3-3}
	&  & \colorbox{catmaterialunderstanding!6}{\parbox{\dimexpr\linewidth-2\fboxsep\relax}{\raggedright\arraybackslash \footnotesize\ttfamily\bfseries F\_\allowbreak{}PHYSICS\_\allowbreak{}PROPERTY\_\allowbreak{}YOUNG\_\allowbreak{}MODULUS\_\allowbreak{}HIGHEST}}\par \textbf{Single Image:} Based on the image, which visible object exhibits the highest Young's Modulus?\par \textbf{Multi Image:} Considering all frames, which object, visible in the last frame, exhibits the highest Young's Modulus? \\
	\cline{3-3}
	&  & \colorbox{catmaterialunderstanding!6}{\parbox{\dimexpr\linewidth-2\fboxsep\relax}{\raggedright\arraybackslash \footnotesize\ttfamily\bfseries F\_\allowbreak{}PHYSICS\_\allowbreak{}PROPERTY\_\allowbreak{}YOUNG\_\allowbreak{}MODULUS\_\allowbreak{}HIGHEST\_\allowbreak{}NON\_\allowbreak{}TECHNICAL}}\par \textbf{Single Image:} Based on the image, which visible object is the stiffest?\par \textbf{Multi Image:} Considering all frames, which object, visible in the last frame, is the stiffest? \\
	\cline{3-3}
	&  & \colorbox{catmaterialunderstanding!6}{\parbox{\dimexpr\linewidth-2\fboxsep\relax}{\raggedright\arraybackslash \footnotesize\ttfamily\bfseries F\_\allowbreak{}PHYSICS\_\allowbreak{}PROPERTY\_\allowbreak{}YOUNG\_\allowbreak{}MODULUS\_\allowbreak{}OBJECT\_\allowbreak{}HIGH\_\allowbreak{}LEVEL}}\par \textbf{Single Image:} Based on the image, which attribute best describes the <OBJECT> in terms of deformability?\par \textbf{Multi Image:} Considering all frames, which attribute best describes the <OBJECT>, visible in the last frame, in terms of deformability? \\
	\cline{2-3}
	& \multirow[t]{6}{*}{\vspace{7pt}\raisebox{0pt}{\rotatebox[origin=c]{90}{\colorbox{catmaterialunderstanding!10}{\hspace{2pt}\rule{0pt}{1.25em}\bfseries \shortstack[c]{Poisson\\Ratio}\hspace{2pt}}\hspace{50pt}}}} & \colorbox{catmaterialunderstanding!6}{\parbox{\dimexpr\linewidth-2\fboxsep\relax}{\raggedright\arraybackslash \footnotesize\ttfamily\bfseries F\_\allowbreak{}PHYSICS\_\allowbreak{}PROPERTY\_\allowbreak{}POISSON\_\allowbreak{}RATIO\_\allowbreak{}OBJECT}}\par \textbf{Single Image:} Based on the image, what is the Poisson ratio of the <OBJECT>?\par \textbf{Multi Image:} Considering all frames, what is the Poisson ratio of the <OBJECT>? \\
	\cline{3-3}
	&  & \colorbox{catmaterialunderstanding!6}{\parbox{\dimexpr\linewidth-2\fboxsep\relax}{\raggedright\arraybackslash \footnotesize\ttfamily\bfseries F\_\allowbreak{}PHYSICS\_\allowbreak{}PROPERTY\_\allowbreak{}POISSON\_\allowbreak{}RATIO\_\allowbreak{}OBJECT\_\allowbreak{}SIMILAR}}\par \textbf{Single Image:} Based on the image, which visible object has a Poisson ratio most similar to that of the <OBJECT>?\par \textbf{Multi Image:} Considering all frames, which visible object has a Poisson ratio most similar to that of the <OBJECT> visible in the last frame? \\
	\cline{3-3}
	&  & \colorbox{catmaterialunderstanding!6}{\parbox{\dimexpr\linewidth-2\fboxsep\relax}{\raggedright\arraybackslash \footnotesize\ttfamily\bfseries F\_\allowbreak{}PHYSICS\_\allowbreak{}PROPERTY\_\allowbreak{}POISSON\_\allowbreak{}RATIO\_\allowbreak{}OBJECT\_\allowbreak{}SIMILAR\_\allowbreak{}NON\_\allowbreak{}TECHNICAL}}\par \textbf{Single Image:} Based on the image, which visible object acts most like the <OBJECT> in terms of how it bulges sideways when squeezed?\par \textbf{Multi Image:} Considering all frames, which visible object acts most like the <OBJECT>, visible in the last frame, in terms of how it bulges sideways when squeezed? \\
	\cline{3-3}
	&  & \colorbox{catmaterialunderstanding!6}{\parbox{\dimexpr\linewidth-2\fboxsep\relax}{\raggedright\arraybackslash \footnotesize\ttfamily\bfseries F\_\allowbreak{}PHYSICS\_\allowbreak{}PROPERTY\_\allowbreak{}POISSON\_\allowbreak{}RATIO\_\allowbreak{}HIGHEST}}\par \textbf{Single Image:} Based on the image, which visible object exhibits the largest Poisson ratio?\par \textbf{Multi Image:} Considering all frames, which object, visible in the last frame, exhibits the largest Poisson ratio? \\
	\cline{3-3}
	&  & \colorbox{catmaterialunderstanding!6}{\parbox{\dimexpr\linewidth-2\fboxsep\relax}{\raggedright\arraybackslash \footnotesize\ttfamily\bfseries F\_\allowbreak{}PHYSICS\_\allowbreak{}PROPERTY\_\allowbreak{}POISSON\_\allowbreak{}RATIO\_\allowbreak{}HIGHEST\_\allowbreak{}NON\_\allowbreak{}TECHNICAL}}\par \textbf{Single Image:} Based on the image, which visible object bulges out the most when you press on it?\par \textbf{Multi Image:} Considering all frames, which object, visible in the last frame, bulges out the most when you press on it? \\
	\cline{3-3}
	&  & \colorbox{catmaterialunderstanding!6}{\parbox{\dimexpr\linewidth-2\fboxsep\relax}{\raggedright\arraybackslash \footnotesize\ttfamily\bfseries F\_\allowbreak{}PHYSICS\_\allowbreak{}PROPERTY\_\allowbreak{}POISSON\_\allowbreak{}HIGH\_\allowbreak{}LEVEL}}\par \textbf{Single Image:} Based on the image, if the <OBJECT> were compressed vertically, how would its horizontal dimensions change?\par \textbf{Multi Image:} Considering all frames, if the <OBJECT>, visible in the last frame, were compressed vertically, how would its horizontal dimensions change? \\
	\cline{2-3}
	& \multirow[t]{4}{*}{\vspace{7pt}\raisebox{0pt}{\rotatebox[origin=c]{90}{\colorbox{catmaterialunderstanding!10}{\hspace{2pt}\rule{0pt}{1.25em}\bfseries \shortstack[c]{Material\\Identification}\hspace{2pt}}\hspace{84pt}}}} & \colorbox{catmaterialunderstanding!6}{\parbox{\dimexpr\linewidth-2\fboxsep\relax}{\raggedright\arraybackslash \footnotesize\ttfamily\bfseries F\_\allowbreak{}MATERIAL\_\allowbreak{}IDENTIFICATION\_\allowbreak{}SIMILAR\_\allowbreak{}OBJECT}}\par \textbf{Single Image:} Based on the image, which visible object is made of a material most similar to that of the <OBJECT>?\par \textbf{Multi Image:} Considering all frames, which visible object is made of a material most similar to that of the <OBJECT>? \\
	\cline{3-3}
	&  & \colorbox{catmaterialunderstanding!6}{\parbox{\dimexpr\linewidth-2\fboxsep\relax}{\raggedright\arraybackslash \footnotesize\ttfamily\bfseries F\_\allowbreak{}MATERIAL\_\allowbreak{}IDENTIFICATION\_\allowbreak{}OBJECT\_\allowbreak{}LEVEL\_\allowbreak{}1}}\par \textbf{Single Image:} Based on the image, what material is the <OBJECT> made of?\par \textbf{Multi Image:} Considering all frames, what material is the <OBJECT> made of? \\
	\cline{3-3}
	&  & \colorbox{catmaterialunderstanding!6}{\parbox{\dimexpr\linewidth-2\fboxsep\relax}{\raggedright\arraybackslash \footnotesize\ttfamily\bfseries F\_\allowbreak{}MATERIAL\_\allowbreak{}IDENTIFICATION\_\allowbreak{}OBJECT\_\allowbreak{}LEVEL\_\allowbreak{}2}}\par \textbf{Single Image:} Based on the image, what material is the <OBJECT> made of?\par \textbf{Multi Image:} Considering all frames, what material is the <OBJECT> made of? \\
	\cline{3-3}
	&  & \colorbox{catmaterialunderstanding!6}{\parbox{\dimexpr\linewidth-2\fboxsep\relax}{\raggedright\arraybackslash \footnotesize\ttfamily\bfseries F\_\allowbreak{}MATERIAL\_\allowbreak{}IDENTIFICATION\_\allowbreak{}OBJECT\_\allowbreak{}LEVEL\_\allowbreak{}3}}\par \textbf{Single Image:} Based on the image, what material is the <OBJECT> made of?\par \textbf{Multi Image:} Considering all frames, what material is the <OBJECT> made of? \\
	\hline
	\multirow[t]{5}{*}{\vspace{7pt}\raisebox{0pt}{\rotatebox[origin=c]{90}{\colorbox{catviewpoint!18}{\hspace{2pt}\rule{0pt}{1.25em}\bfseries View Point\hspace{2pt}}\hspace{60pt}}}} & \multirow[t]{3}{*}{\vspace{7pt}\raisebox{0pt}{\rotatebox[origin=c]{90}{\colorbox{catviewpoint!10}{\hspace{2pt}\rule{0pt}{1.25em}\bfseries Visibility\hspace{2pt}}\hspace{60pt}}}} & \colorbox{catviewpoint!6}{\parbox{\dimexpr\linewidth-2\fboxsep\relax}{\raggedright\arraybackslash \footnotesize\ttfamily\bfseries F\_\allowbreak{}VISIBILITY\_\allowbreak{}OBJECT}}\par \textbf{Single Image:} Based on the image, which of these objects is visible?\par \textbf{Multi Image:} Considering all frames, which of these objects is visible in the last frame? \\
	\cline{3-3}
	&  & \colorbox{catviewpoint!6}{\parbox{\dimexpr\linewidth-2\fboxsep\relax}{\raggedright\arraybackslash \footnotesize\ttfamily\bfseries F\_\allowbreak{}OCCLUSION\_\allowbreak{}PERCENTAGE\_\allowbreak{}OBJECT}}\par \textbf{Single Image:} Based on the image, how much of the <OBJECT> is occluded?\par \textbf{Multi Image:} Considering all frames, how much of the <OBJECT> is occluded in the last frame? \\
	\cline{3-3}
	&  & \colorbox{catviewpoint!6}{\parbox{\dimexpr\linewidth-2\fboxsep\relax}{\raggedright\arraybackslash \footnotesize\ttfamily\bfseries F\_\allowbreak{}VISIBILITY\_\allowbreak{}OBJECT\_\allowbreak{}COUNT}}\par \textbf{Single Image:} Based on the image, how many objects are visible?\par \textbf{Multi Image:} Considering all frames, how many objects are visible in the last frame? \\
	\cline{2-3}
	& \multirow[t]{2}{*}{\vspace{7pt}\raisebox{0pt}{\rotatebox[origin=c]{90}{\colorbox{catviewpoint!10}{\hspace{2pt}\rule{0pt}{1.25em}\bfseries \shortstack[c]{Camera\\Characteristics}\hspace{2pt}}\hspace{90pt}}}} & \colorbox{catviewpoint!6}{\parbox{\dimexpr\linewidth-2\fboxsep\relax}{\raggedright\arraybackslash \footnotesize\ttfamily\bfseries F\_\allowbreak{}VIEWPOINT\_\allowbreak{}CAMERA\_\allowbreak{}ANGLE}}\par \textbf{Single Image:} From the camera’s perspective, what is the viewing direction relative to the horizon?\par \textbf{Multi Image:} Considering all frames, from the camera’s perspective, what is the viewing direction relative to the horizon in the last frame? \\
	\cline{3-3}
	&  & \colorbox{catviewpoint!6}{\parbox{\dimexpr\linewidth-2\fboxsep\relax}{\raggedright\arraybackslash \footnotesize\ttfamily\bfseries F\_\allowbreak{}FOCAL\_\allowbreak{}LENGTH\_\allowbreak{}CLASS}}\par \textbf{Single Image:} Based on the image, which focal‑length class best matches the perspective observed?\par \textbf{Multi Image:} Considering all frames, which focal‑length class best matches the perspective observed? \\
	\hline
    \clearpage
	\multirow[t]{8}{*}{\vspace{7pt}\raisebox{0pt}{\rotatebox[origin=c]{90}{\colorbox{cattemporal!18}{\hspace{2pt}\rule{0pt}{1.25em}\bfseries Temporal\hspace{2pt}}\hspace{48pt}}}} & \multirow[t]{6}{*}{\vspace{7pt}\raisebox{0pt}{\rotatebox[origin=c]{90}{\colorbox{cattemporal!10}{\hspace{2pt}\rule{0pt}{1.25em}\bfseries \shortstack[c]{Event\\Ordering}\hspace{2pt}}\hspace{48pt}}}} & \colorbox{cattemporal!6}{\parbox{\dimexpr\linewidth-2\fboxsep\relax}{\raggedright\arraybackslash \footnotesize\ttfamily\bfseries F\_\allowbreak{}TEMPORAL\_\allowbreak{}SEQUENCE\_\allowbreak{}IMAGES}}\par \textbf{Multi Image:} Given the four unordered frames (A, B, C, D) of the same scene, what is the correct temporal ordering of the events? \\
	\cline{3-3}
	&  & \colorbox{cattemporal!6}{\parbox{\dimexpr\linewidth-2\fboxsep\relax}{\raggedright\arraybackslash \footnotesize\ttfamily\bfseries F\_\allowbreak{}TEMPORAL\_\allowbreak{}PREDICTION\_\allowbreak{}NEXT\_\allowbreak{}IMAGE\_\allowbreak{}GRANULARITY\_\allowbreak{}1}}\par \textbf{Multi Image:} Which individual frame (A, B, C, or D) is most likely to display an event that occurred after the provided frame sequence? \\
	\cline{3-3}
	&  & \colorbox{cattemporal!6}{\parbox{\dimexpr\linewidth-2\fboxsep\relax}{\raggedright\arraybackslash \footnotesize\ttfamily\bfseries F\_\allowbreak{}TEMPORAL\_\allowbreak{}PREDICTION\_\allowbreak{}NEXT\_\allowbreak{}IMAGE\_\allowbreak{}GRANULARITY\_\allowbreak{}2}}\par \textbf{Multi Image:} Which individual frame (A, B, C, or D) is most likely to display an event that occurred after the provided frame sequence? \\
	\cline{3-3}
	&  & \colorbox{cattemporal!6}{\parbox{\dimexpr\linewidth-2\fboxsep\relax}{\raggedright\arraybackslash \footnotesize\ttfamily\bfseries F\_\allowbreak{}TEMPORAL\_\allowbreak{}PREDICTION\_\allowbreak{}NEXT\_\allowbreak{}IMAGE\_\allowbreak{}GRANULARITY\_\allowbreak{}5}}\par \textbf{Multi Image:} Which individual frame (A, B, C, or D) is most likely to display an event that occurred after the provided frame sequence? \\
	\cline{3-3}
	&  & \colorbox{cattemporal!6}{\parbox{\dimexpr\linewidth-2\fboxsep\relax}{\raggedright\arraybackslash \footnotesize\ttfamily\bfseries F\_\allowbreak{}TEMPORAL\_\allowbreak{}PREDICTION\_\allowbreak{}PREVIOUS\_\allowbreak{}IMAGE}}\par \textbf{Multi Image:} Which individual frame (A, B, C, or D) is most likely to display an event that occurred before the provided frame sequence? \\
	\cline{3-3}
	&  & \colorbox{cattemporal!6}{\parbox{\dimexpr\linewidth-2\fboxsep\relax}{\raggedright\arraybackslash \footnotesize\ttfamily\bfseries F\_\allowbreak{}TEMPORAL\_\allowbreak{}PREDICTION\_\allowbreak{}MISSING\_\allowbreak{}IMAGE}}\par \textbf{Multi Image:} Which individual frame (A, B, C, or D) is most likely to display an event that occurred during the time span of the provided frame sequence? \\
	\cline{2-3}
	& \multirow[t]{2}{*}{\vspace{7pt}\raisebox{0pt}{\rotatebox[origin=c]{90}{\colorbox{cattemporal!10}{\hspace{2pt}\rule{0pt}{1.25em}\bfseries \shortstack[c]{Camera\\Motion}\hspace{2pt}}\hspace{46pt}}}} & \colorbox{cattemporal!6}{\parbox{\dimexpr\linewidth-2\fboxsep\relax}{\raggedright\arraybackslash \footnotesize\ttfamily\bfseries F\_\allowbreak{}CAMERA\_\allowbreak{}MOTION\_\allowbreak{}DIRECTION}}\par \textbf{Multi Image:} Across the frame sequence, what is the predominant direction of the camera’s motion? \\
	\cline{3-3}
	&  & \colorbox{cattemporal!6}{\parbox{\dimexpr\linewidth-2\fboxsep\relax}{\raggedright\arraybackslash \footnotesize\ttfamily\bfseries F\_\allowbreak{}CAMERA\_\allowbreak{}ZOOM\_\allowbreak{}BEHAVIOR}}\par \textbf{Multi Image:} Across the frame sequence, how does the camera’s zoom level change? \\
	\hline
	\multirow[t]{4}{*}{\vspace{7pt}\raisebox{0pt}{\rotatebox[origin=c]{90}{\colorbox{catpermanence!18}{\hspace{2pt}\rule{0pt}{1.25em}\bfseries Persistence\hspace{2pt}}\hspace{66pt}}}} & \multirow[t]{2}{*}{\vspace{7pt}\raisebox{0pt}{\rotatebox[origin=c]{90}{\colorbox{catpermanence!10}{\hspace{2pt}\rule{0pt}{1.25em}\bfseries Identity\hspace{2pt}}\hspace{48pt}}}} & \colorbox{catpermanence!6}{\parbox{\dimexpr\linewidth-2\fboxsep\relax}{\raggedright\arraybackslash \footnotesize\ttfamily\bfseries F\_\allowbreak{}PERSISTENCE\_\allowbreak{}OBJECT\_\allowbreak{}PRESENT}}\par \textbf{Multi Image:} Considering all frames, which object was visible but disappeared in the last frame? \\
	\cline{3-3}
	&  & \colorbox{catpermanence!6}{\parbox{\dimexpr\linewidth-2\fboxsep\relax}{\raggedright\arraybackslash \footnotesize\ttfamily\bfseries F\_\allowbreak{}PERSISTENCE\_\allowbreak{}OBJECT\_\allowbreak{}DISAPPEAR}}\par \textbf{Multi Image:} Considering all frames, which visible object disappeared and does not reappear in the last frame? \\
	\cline{2-3}
	& \multirow[t]{2}{*}{\vspace{7pt}\raisebox{0pt}{\rotatebox[origin=c]{90}{\colorbox{catpermanence!10}{\hspace{2pt}\rule{0pt}{1.25em}\bfseries Counting\hspace{2pt}}\hspace{48pt}}}} & \colorbox{catpermanence!6}{\parbox{\dimexpr\linewidth-2\fboxsep\relax}{\raggedright\arraybackslash \footnotesize\ttfamily\bfseries F\_\allowbreak{}PERSISTENCE\_\allowbreak{}OBJECT\_\allowbreak{}TOTAL\_\allowbreak{}COUNT}}\par \textbf{Multi Image:} Considering all frames, how many objects are present at the time of the last frame, including those currently hidden or out of frame? \\
	\cline{3-3}
	&  & \colorbox{catpermanence!6}{\parbox{\dimexpr\linewidth-2\fboxsep\relax}{\raggedright\arraybackslash \footnotesize\ttfamily\bfseries F\_\allowbreak{}PERSISTENCE\_\allowbreak{}OBJECT\_\allowbreak{}TOTAL\_\allowbreak{}COUNT\_\allowbreak{}HIDDEN}}\par \textbf{Multi Image:} Considering all frames, how many objects are present at the time of the last frame, but not visible? \\
	\hline
\end{longtable}
\twocolumn

\onecolumn
\subsection{VQA examples}
\label{app:vqa_examples}

Below, we present examples of the actual VQA questions used to evaluate the 54 VLMs. For each question\_id, we show one representative single-image question and one representative multi-image question. In total, the benchmark contains 85 distinct questions, spanning both single-image and multi-image settings. Some questions are defined only for single-image inputs or only for multi-image inputs; this is intentional and reflects the original design of the VQA benchmark.

\newcolumntype{M}[1]{>{\centering\arraybackslash}m{#1}}

\newcommand{\opts}[4]{%
	\newline \vspace{12pt}
	\noindent
	\begin{tabular}{@{}p{0.9\linewidth}}
		A) #1 \\ B) #2 \\
		C) #3 \\ D) #4
	\end{tabular}
}

\newcommand{\vqaImgNode}[2]{%
	\begin{tikzpicture}
		\node[inner sep=0, outer sep=0, draw=black, line width=0.4pt] (img) 
		{\includegraphics[width=0.46\linewidth, height=1.6cm, keepaspectratio]{figures/supp/questions_paper/#1}};
		\node[anchor=north west, fill=white, fill opacity=0.9, text=black, 
		inner sep=1.5pt, font=\tiny\bfseries, xshift=0.4pt, yshift=-0.4pt] 
		at (img.north west) {#2};
	\end{tikzpicture}%
}

\newcommand{\stripFour}[4]{%
	\begingroup
	\setlength{\tabcolsep}{1pt}
	\renewcommand{\arraystretch}{0} 
	\centering
	\begin{tabular}{cc}
		\vqaImgNode{#1}{A} & \vqaImgNode{#2}{B} \\ \noalign{\vskip 2pt}
		\vqaImgNode{#3}{C} & \vqaImgNode{#4}{D}
	\end{tabular}
	\endgroup
}

\newcommand{\stripEight}[8]{%
	\begin{tikzpicture}
		\def\dx{0.3cm} %
		\def\dy{0.3cm} %
		\def\opa{0.7}  %
		
		\node[inner sep=0, outer sep=0, draw=black, line width=0.4pt, opacity=\opa] 
		at (7*\dx,7*\dy) {\includegraphics[width=0.46\linewidth, height=1.6cm, keepaspectratio]{figures/supp/questions_paper/#8}};
		
		\node[inner sep=0, outer sep=0, draw=black, line width=0.4pt, opacity=\opa] 
		at (6*\dx,6*\dy) {\includegraphics[width=0.46\linewidth, height=1.6cm, keepaspectratio]{figures/supp/questions_paper/#7}};
		
		\node[inner sep=0, outer sep=0, draw=black, line width=0.4pt, opacity=\opa] 
		at (5*\dx,5*\dy) {\includegraphics[width=0.46\linewidth, height=1.6cm, keepaspectratio]{figures/supp/questions_paper/#6}};
		
		\node[inner sep=0, outer sep=0, draw=black, line width=0.4pt, opacity=\opa] 
		at (4*\dx,4*\dy) {\includegraphics[width=0.46\linewidth, height=1.6cm, keepaspectratio]{figures/supp/questions_paper/#5}};
		
		\node[inner sep=0, outer sep=0, draw=black, line width=0.4pt, opacity=\opa] 
		at (3*\dx,3*\dy) {\includegraphics[width=0.46\linewidth, height=1.6cm, keepaspectratio]{figures/supp/questions_paper/#4}};
		
		\node[inner sep=0, outer sep=0, draw=black, line width=0.4pt, opacity=\opa] 
		at (2*\dx,2*\dy) {\includegraphics[width=0.46\linewidth, height=1.6cm, keepaspectratio]{figures/supp/questions_paper/#3}};
		
		\node[inner sep=0, outer sep=0, draw=black, line width=0.4pt, opacity=\opa] 
		at (1*\dx,1*\dy) {\includegraphics[width=0.46\linewidth, height=1.6cm, keepaspectratio]{figures/supp/questions_paper/#2}};
		
		\node[inner sep=0, outer sep=0, draw=black, line width=0.4pt, opacity=\opa] 
		at (0*\dx,0*\dy) {\includegraphics[width=0.46\linewidth, height=1.6cm, keepaspectratio]{figures/supp/questions_paper/#1}};
	\end{tikzpicture}%
}
\newcommand{\insertImageIfGiven}[1]{%
	\if\relax\detokenize{#1}\relax
	\else
	\includegraphics[width=0.9\linewidth, keepaspectratio]{figures/supp/questions_paper/#1}
	\fi
}

\newcommand{\vqaSplitEntry}[7]{%
	\multicolumn{4}{|@{}c@{}|}{%
		\begin{tabular}{ M{0.05\textwidth} | M{0.08\textwidth} | @{}c@{} }
			\rotatebox{90}{\parbox{5cm}{\centering \tiny \textbf{#1}}} &
			\rotatebox{90}{\textbf{#2} \scriptsize (#3)} &
			\begin{tabular}{ M{0.38\textwidth} | m{0.41\textwidth} }
				\vspace{4pt} \centering \insertImageIfGiven{#4} \vspace{4pt} & 
				\vspace{4pt} \raggedright\arraybackslash \textbf{Image:} #6 \vspace{4pt} \\ 
				\hline
				\vspace{4pt} \centering #5 \vspace{4pt} & 
				\vspace{4pt} \raggedright\arraybackslash \textbf{Image sequences:} #7 \vspace{4pt} \\
			\end{tabular}
		\end{tabular}%
	} \\ \hline
}

\setlength{\tabcolsep}{4pt}
\renewcommand{\arraystretch}{1.3}

\begingroup
{
	\scriptsize
	\begin{longtable}{| M{0.05\textwidth} | M{0.08\textwidth} | M{0.38\textwidth} | m{0.41\textwidth} |}
		\caption{Dataset Samples} \\
		\hline
		\textbf{ID} & \textbf{Type} & \textbf{Visual Input} & \textbf{Task \& Logic} \\ 
		\hline
		\endfirsthead
		\hline
		\textbf{ID} & \textbf{Type} & \textbf{Visual Input} & \textbf{Task \& Logic} \\ 
		\hline
		\endhead
		
		\vqaSplitEntry{F\_CLOSEST\_OBJECT\_OBJECT}
		{Spatial Reasoning}
		{Distance}
		{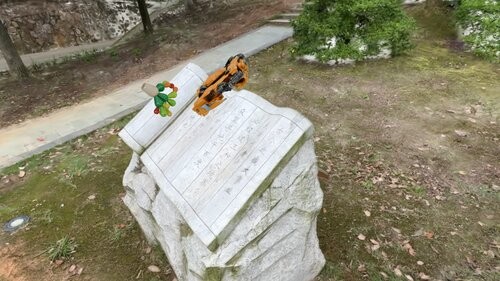}
		{\stripEight{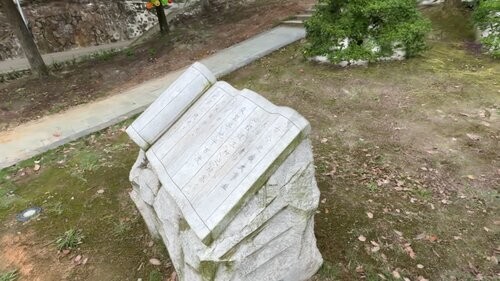}{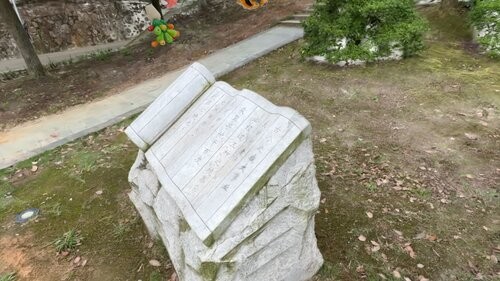}{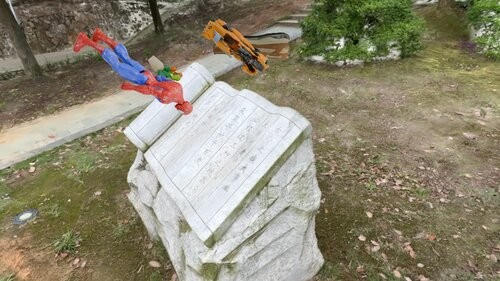}{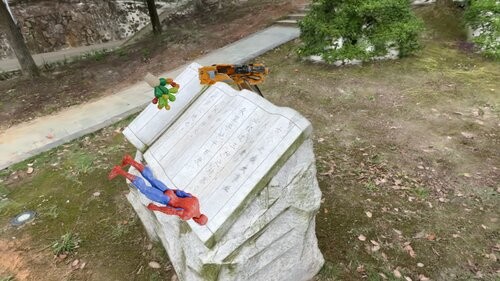}
			{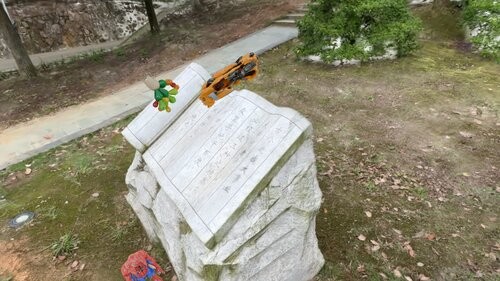}{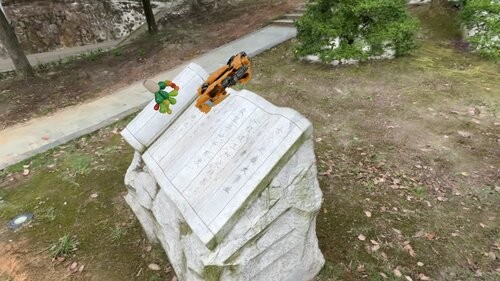}{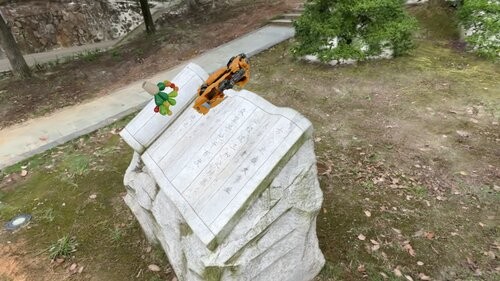}{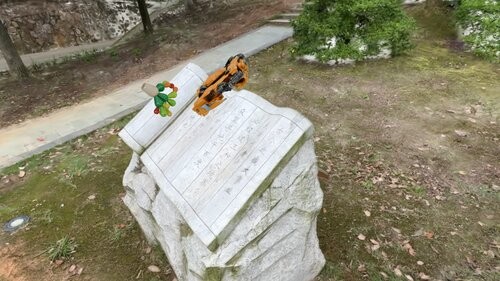}}
		{Based on the image, which object is closest to the "orange Transformers action figure" in real-world distance? \opts{Teenage Mutant Ninja Turtles action figurine}{white soccer shoe}{\textbf{colored cactus balancing toy}}{small blue painted bucket}}
		{Considering all frames, which object is closest to the "orange Transformers action figure" in real-world distance in the last frame? \opts{Teenage Mutant Ninja Turtles action figurine}{white soccer shoe}{\textbf{colored cactus balancing toy}}{small blue painted bucket}}
		
		\vqaSplitEntry{F\_DISTANCE\_\allowbreak OBJECT\_CAMERA\_DISTANCE}
		{Spatial Reasoning}
		{Distance}
		{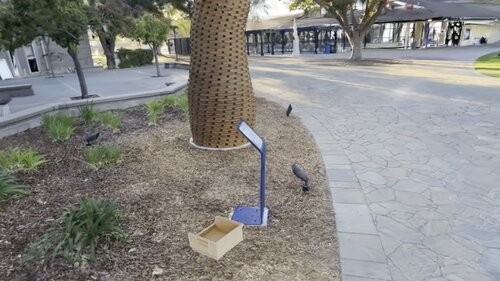}
		{\stripEight{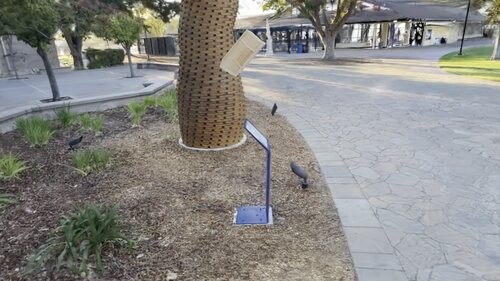}{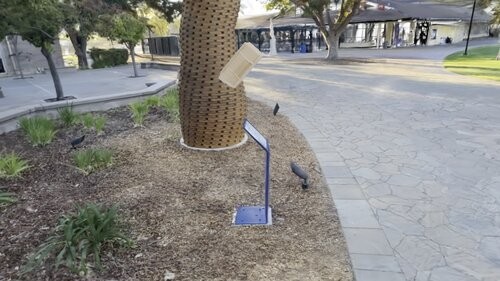}{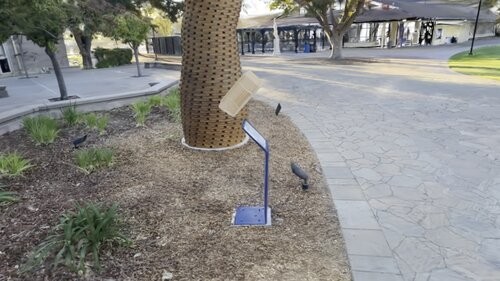}{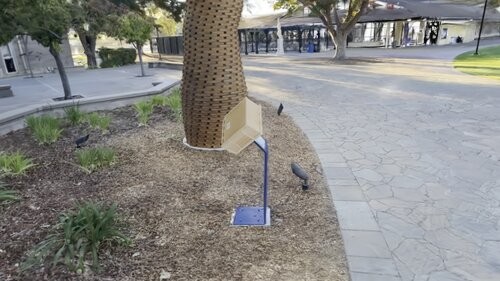}
			{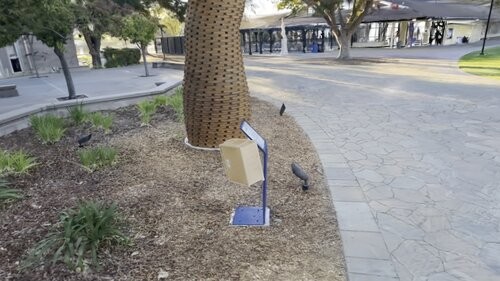}{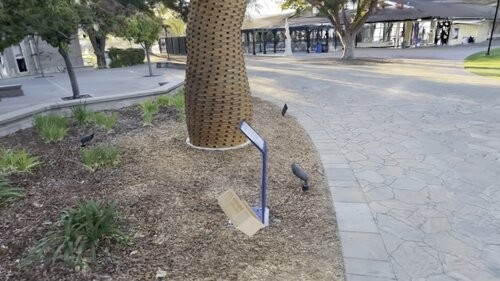}{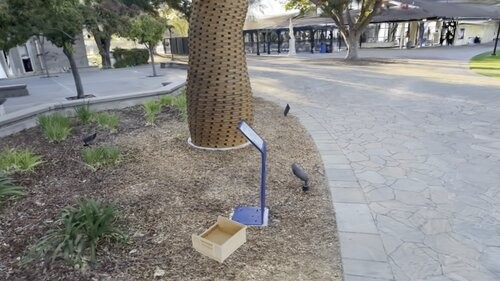}{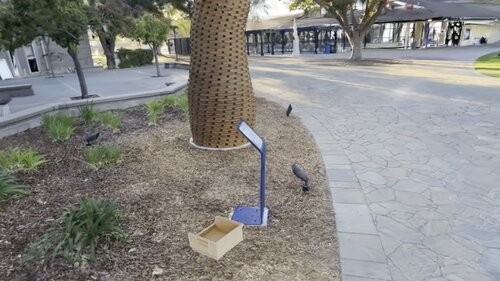}}
		{Based on the image, what is the real-world distance between the "shelf bin" and the camera? \opts{1.32 meters}{2.05 meters}{\textbf{2.79 meters}}{0.58 meters}}
		{Considering all frames, what is the real-world distance between the "shelf bin" and the camera in the last frame? \opts{1.32 meters}{2.05 meters}{\textbf{2.79 meters}}{0.58 meters}}
		
		\vqaSplitEntry{F\_CLOSEST\_OBJECT\_CAMERA}
		{Spatial Reasoning}
		{Distance}
		{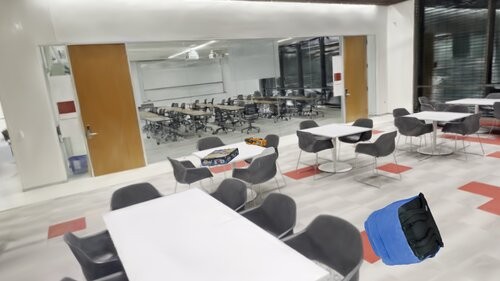}
		{\stripEight{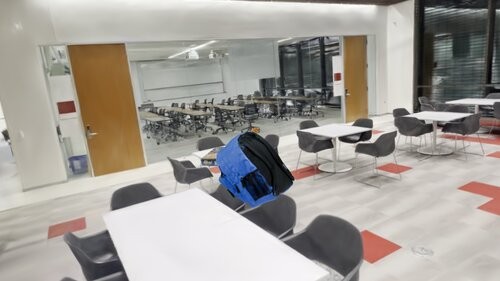}{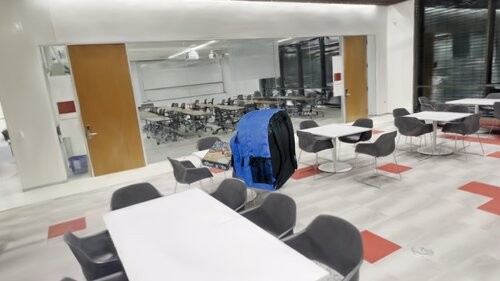}{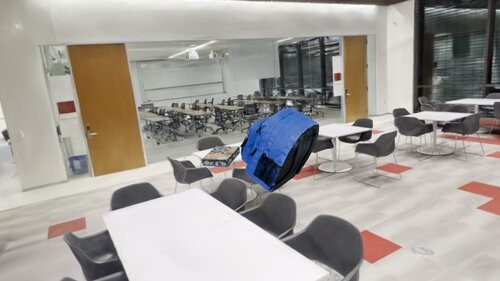}{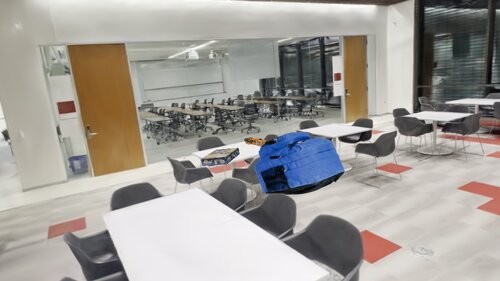}
			{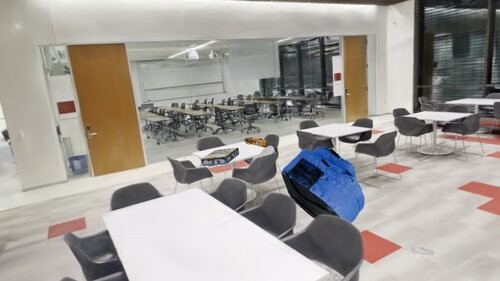}{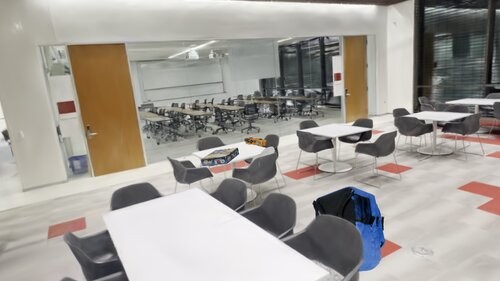}{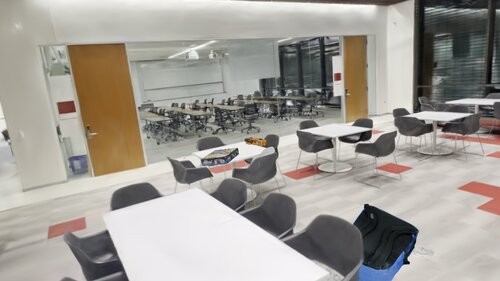}{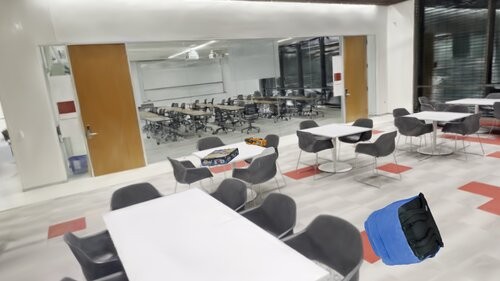}}
		{Based on the image, which object is the closest to the camera in real-world distance? \opts{white soccer shoe}{CLUE board game box}{\textbf{blue backpack}}{Teenage Mutant Ninja Turtles action figurine}}
		{Considering all frames, which object is the closest to the camera in real-world distance in the last frame? \opts{white soccer shoe}{CLUE board game box}{\textbf{blue backpack}}{Teenage Mutant Ninja Turtles action figurine}}
		
		\vqaSplitEntry{F\_SIZE\_OBJECT}
		{Spatial Reasoning}
		{Size}
		{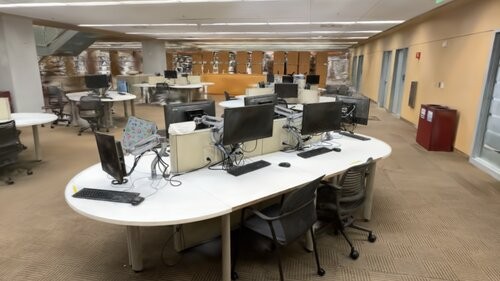}
		{\stripEight{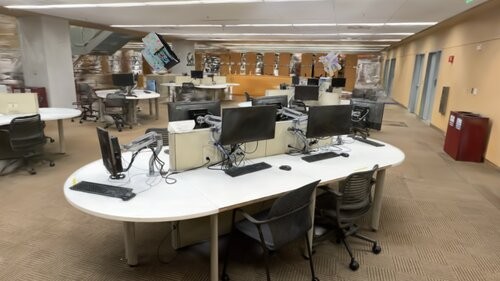}{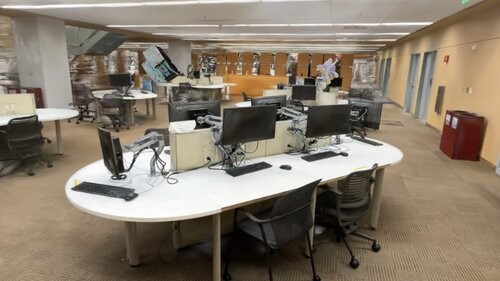}{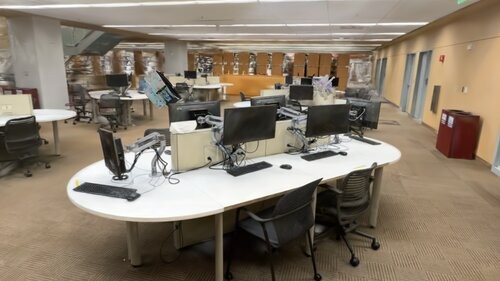}{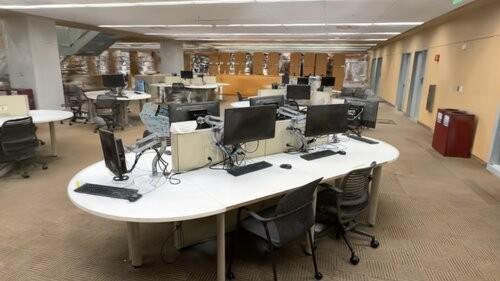}
			{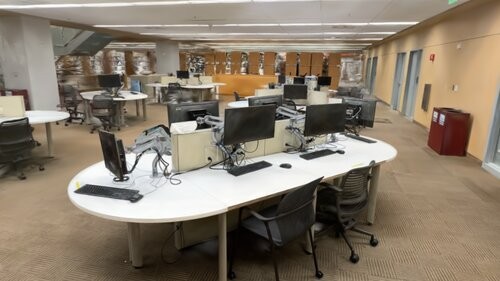}{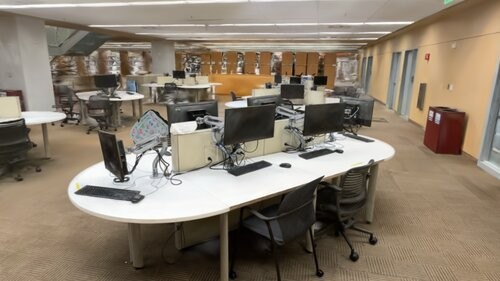}{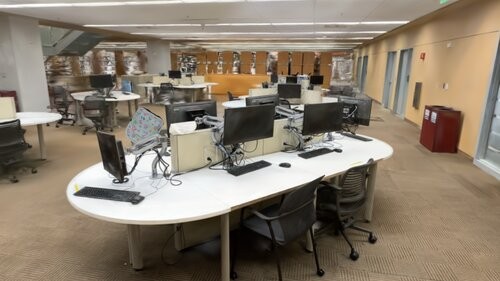}{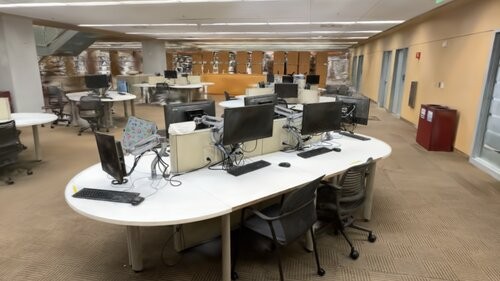}}
		{Based on the image, what are the real-world dimensions of the "turquoise mermaid-design backpack"? \opts{0.62m x 0.53m x 0.43m}{\textbf{0.41m x 0.35m x 0.28m}}{0.74m x 0.63m x 0.51m}{0.83m x 0.71m x 0.57m}}
		{Considering all frames, what are the real-world dimensions of the "turquoise mermaid-design backpack" in the last frame? \opts{0.62m x 0.53m x 0.43m}{\textbf{0.41m x 0.35m x 0.28m}}{0.74m x 0.63m x 0.51m}{0.83m x 0.71m x 0.57m}}
		
		\vqaSplitEntry{F\_SIZE\_OBJECT\_BIGGER}
		{Spatial Reasoning}
		{Size}
		{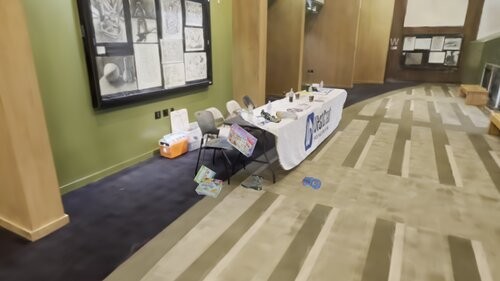}
		{\stripEight{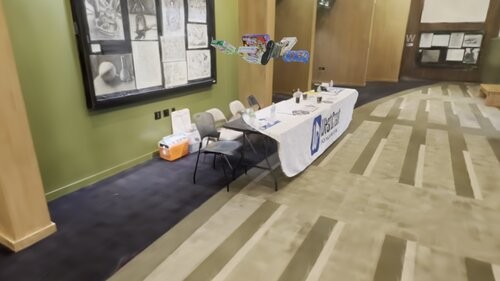}{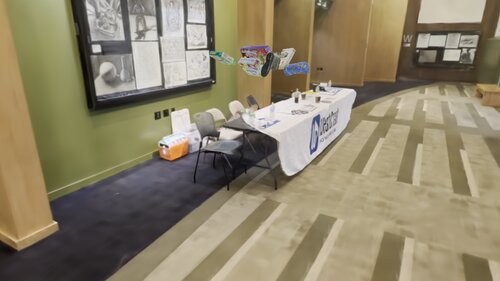}{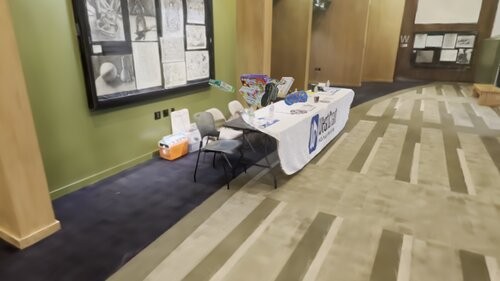}{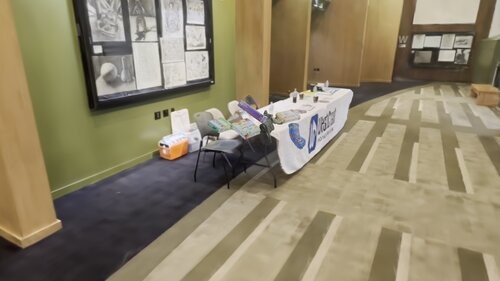}
			{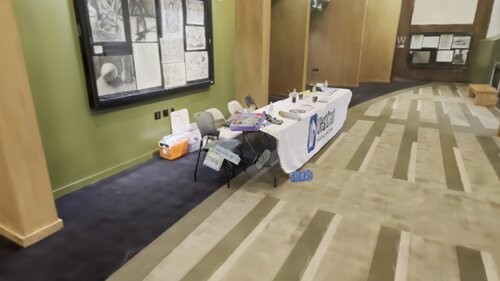}{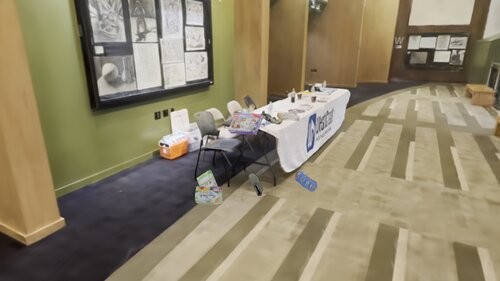}{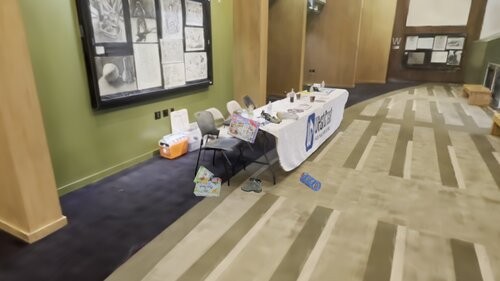}{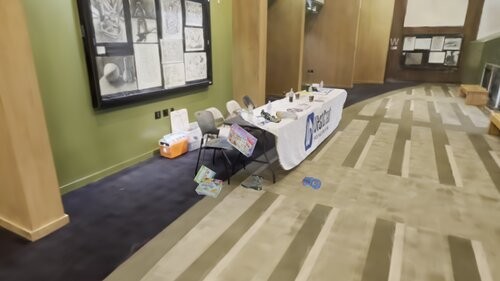}}
		{Based on the image, which single object has the biggest real-world volume? \opts{Teenage Mutant Ninja Turtles action figurine}{\textbf{228-piece purple LEGO Friends boxed set}}{retail box of fruit snacks (10 pouches)}{white soccer shoe}}
		{Considering all frames, which single object, visible in the last frame, has the biggest real-world volume? \opts{Teenage Mutant Ninja Turtles action figurine}{\textbf{228-piece purple LEGO Friends boxed set}}{retail box of fruit snacks (10 pouches)}{white soccer shoe}}
		
		\vqaSplitEntry{F\_LAYOUT\_\allowbreak POSITION\_OBJECT\_OBJECT}
		{Spatial Reasoning}
		{Layout}
		{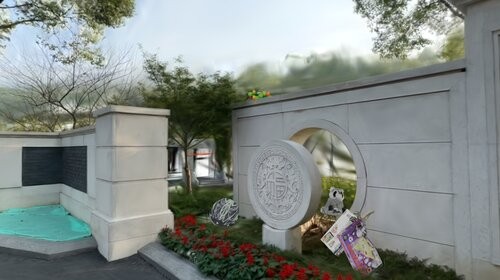}
		{\stripEight{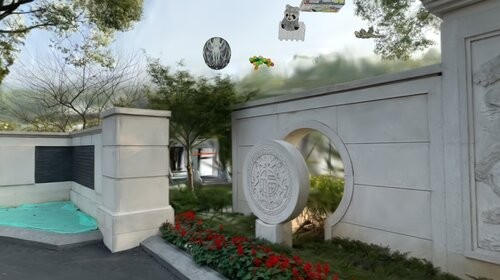}{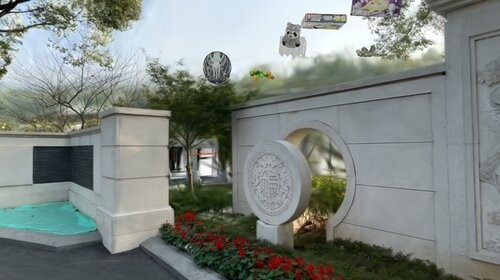}{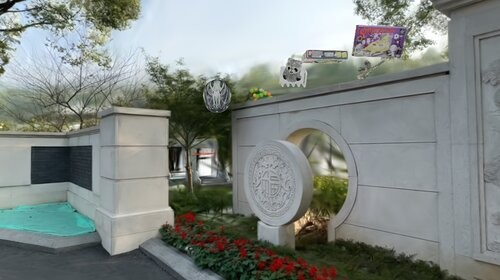}{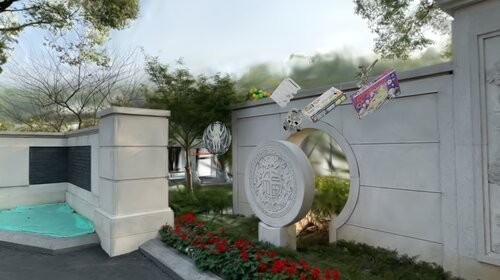}
			{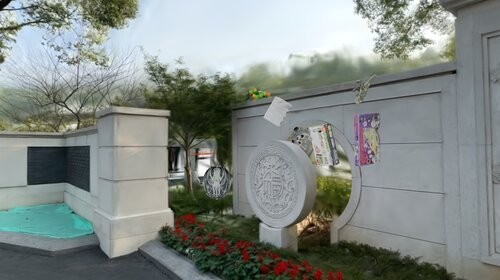}{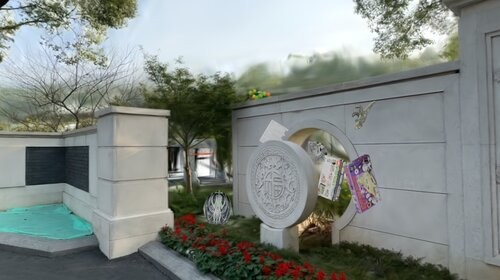}{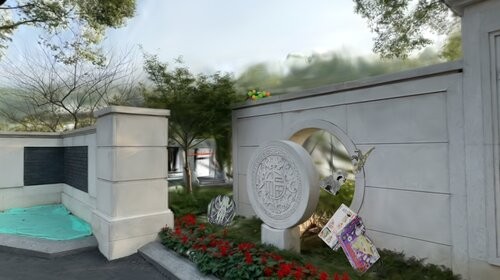}{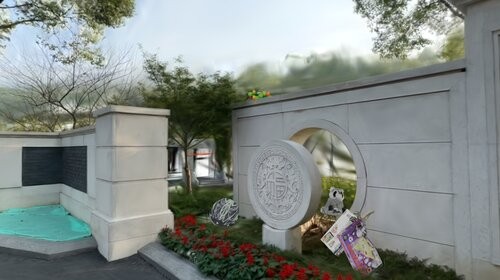}}
		{From the camera's perspective, where is the "motorcycle helmet" relative to the "gray raccoon toy" in the image? \opts{to the right}{\textbf{to the left}}{in front}{horizontally aligned}}
		{Considering all frames, from the camera's perspective, where is the "motorcycle helmet" relative to the "gray raccoon toy" in the last frame? \opts{to the right}{\textbf{to the left}}{in front}{horizontally aligned}}
		
		\vqaSplitEntry{F\_KINEMATICS\_SPEED\_OBJECT}
		{Mechanics}
		{Kinematics}
		{}
		{\stripEight{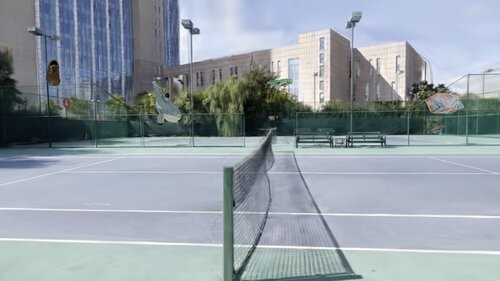}{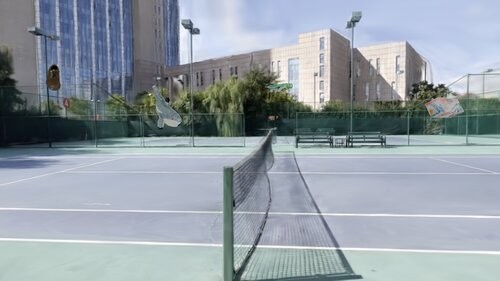}{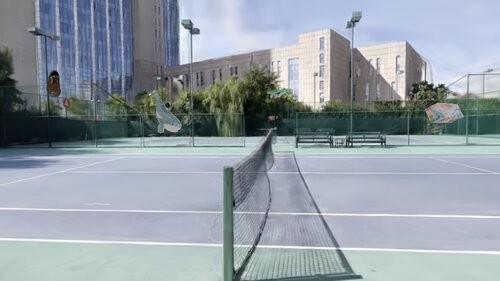}{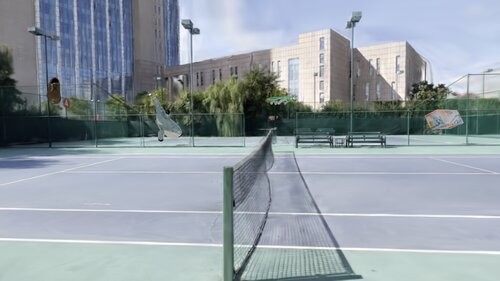}
			{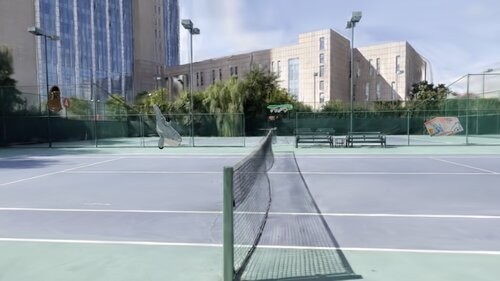}{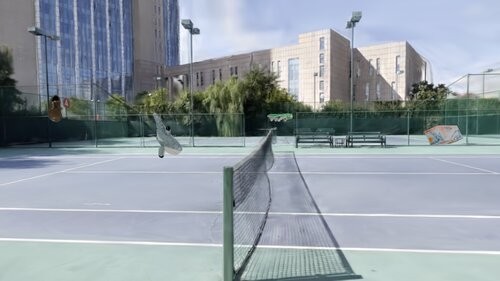}{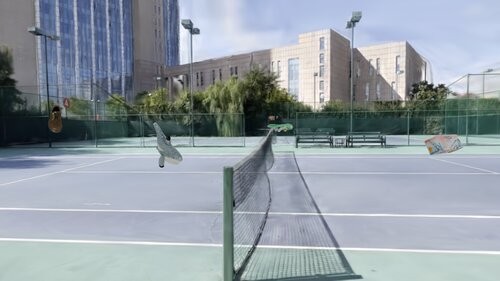}{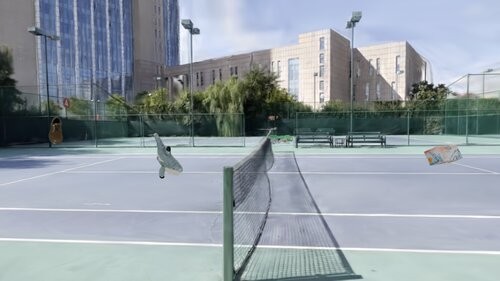}}
		{\textbf{No question provided/possible}}
		{Considering all frames, what is the real-world speed of the "12-pack Fresca soda carton" at the time of the last frame? \opts{7.71 m/s}{1.95 m/s}{\textbf{3.87 m/s}}{5.79 m/s}}
		
		\vqaSplitEntry{F\_KINEMATICS\_ACCEL\_OBJECT}
		{Mechanics}
		{Kinematics}
		{}
		{\stripEight{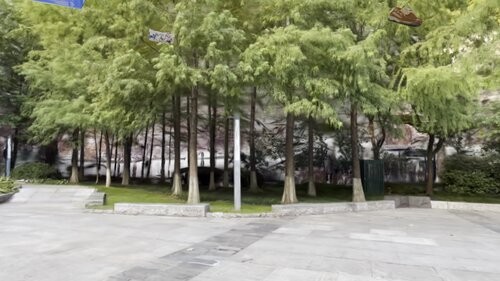}{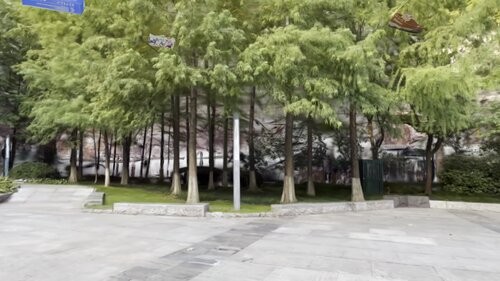}{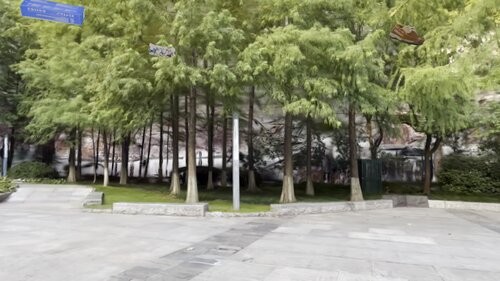}{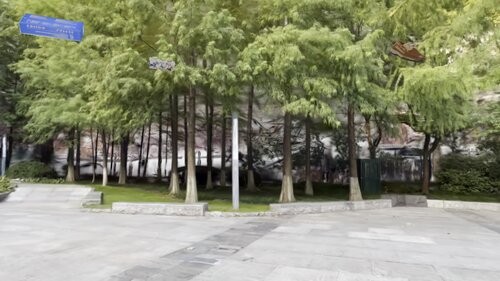}
			{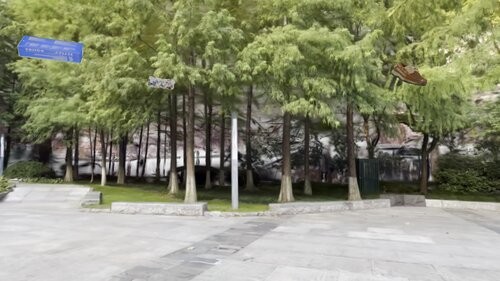}{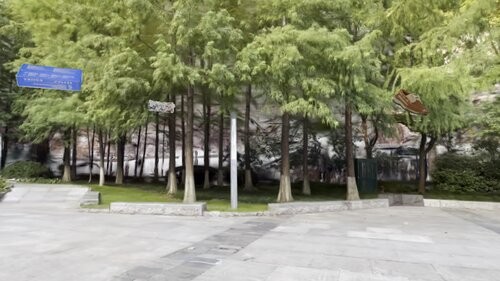}{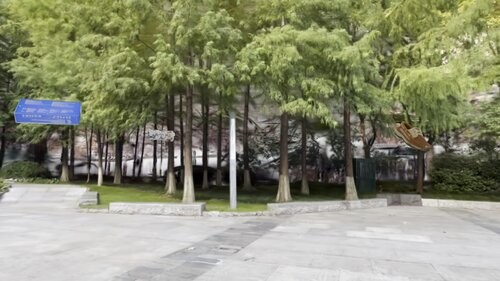}{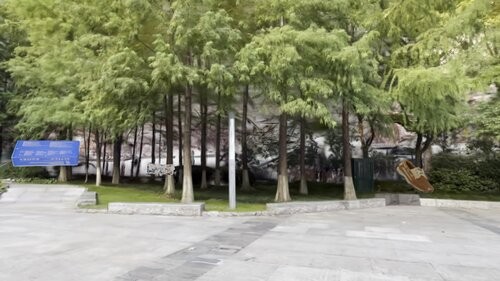}}
		{\textbf{No question provided/possible}}
		{Considering all frames, what is the real-world magnitude of acceleration of the "brown boat shoe" at the time of the last frame? \opts{\textbf{9.79 m/s\textasciicircum{}2}}{5.05 m/s\textasciicircum{}2}{0.32 m/s\textasciicircum{}2}{14.53 m/s\textasciicircum{}2}}
		
		\vqaSplitEntry{F\_KINEMATICS\_\allowbreak DISTANCE\_TRAVELED\_INTERVAL}
		{Mechanics}
		{Kinematics}
		{}
		{\stripEight{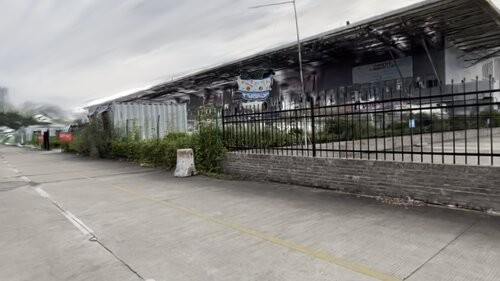}{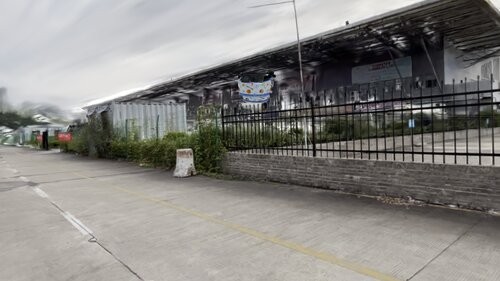}{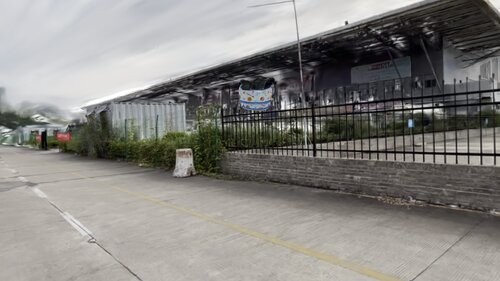}{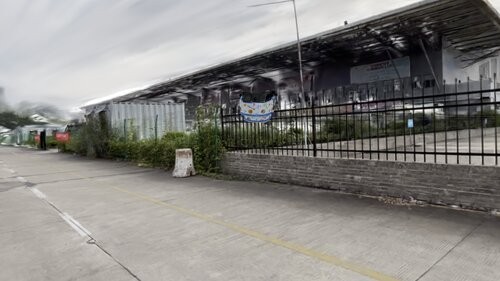}
			{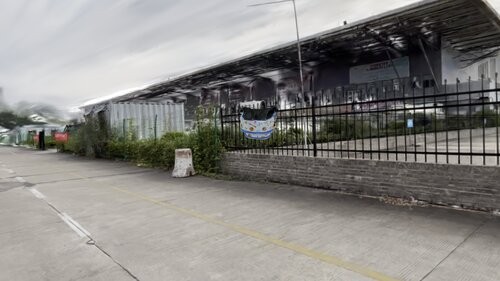}{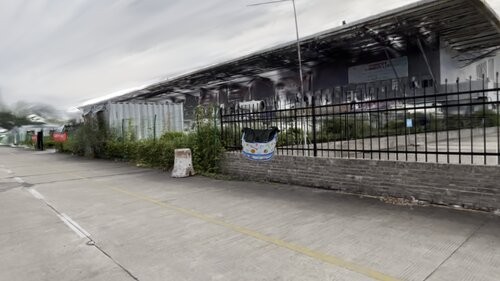}{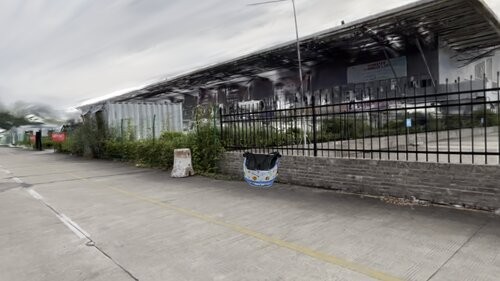}{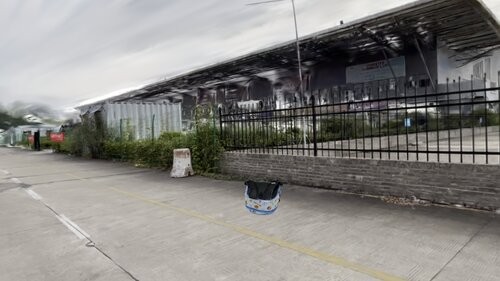}}
		{\textbf{No question provided/possible}}
		{Considering all frames, what is the real-world displacement of the centroid of the "blue sport-design kid backpack" from the first to the last frame? \opts{2.7 meters}{0.5 meters}{3.7 meters}{\textbf{1.6 meters}}}
		
		\vqaSplitEntry{F\_KINEMATICS\_SYSTEM\_STABILITY}
		{Mechanics}
		{Kinematics}
		{}
		{\stripEight{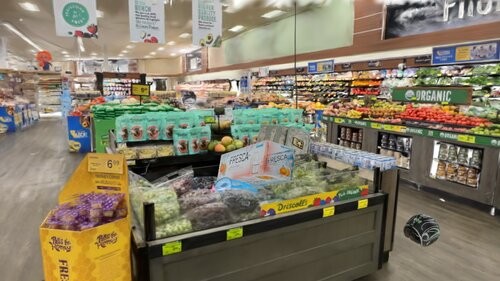}{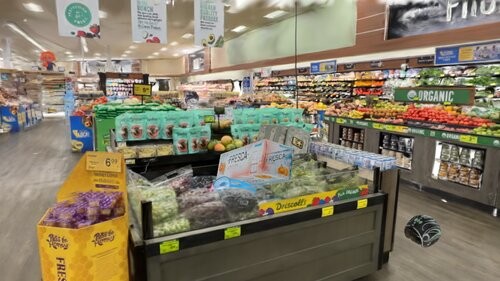}{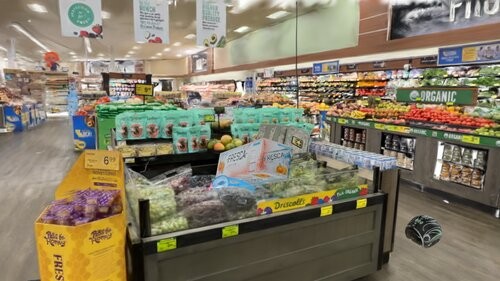}{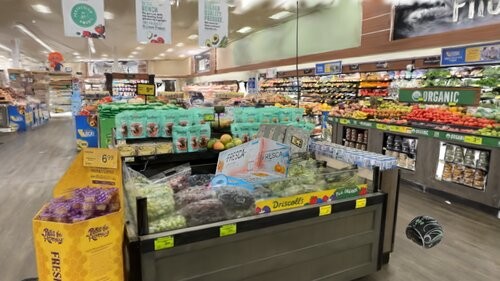}
			{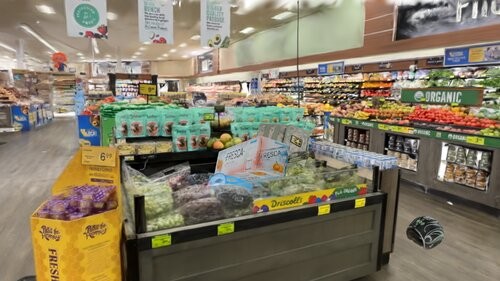}{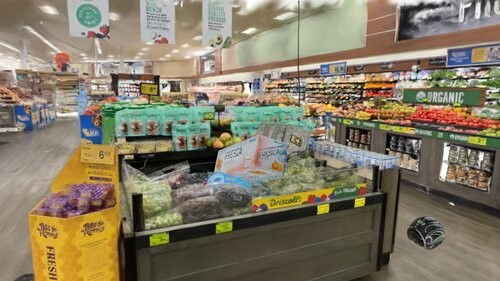}{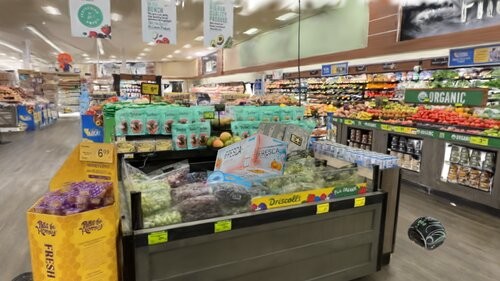}{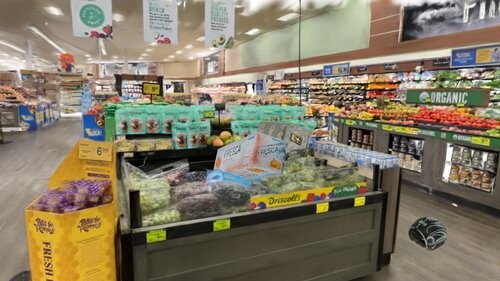}}
		{\textbf{No question provided/possible}}
		{Analyzing the motion trend across the sequence, which statement best describes the system's state at the final frame? \opts{Cyclic: All objects returned to their initial position}{\textbf{Stable: The system has stopped}}{Unstable: The system is currently moving}{Invisible: All objects have moved out of the frame entirely}}
		
		\vqaSplitEntry{F\_COLLISION\_OBJECT\_\allowbreak OBJECT\_FRAME\_SINGLE}
		{Mechanics}
		{Collision}
		{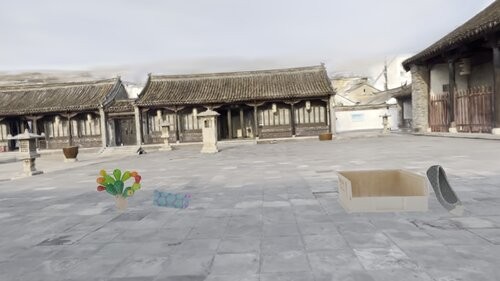}
		{}
		{Based on the image, which visible object is the "dark ballet flat shoe" colliding with? \opts{\textbf{shelf bin}}{white soccer shoe}{colored cactus balancing toy}{empty purple and green pencil case}}
		{\textbf{No question provided/possible}}
		
		\vqaSplitEntry{F\_COLLISION\_OBJECT\_\allowbreak OBJECT\_FRAME\_MULTI}
		{Mechanics}
		{Collision}
		{}
		{\stripFour{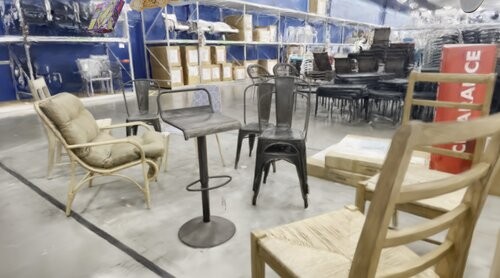}{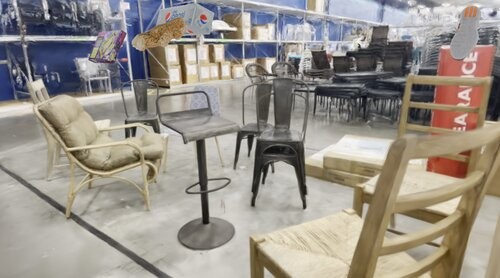}{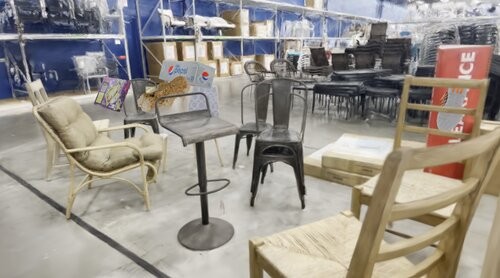}{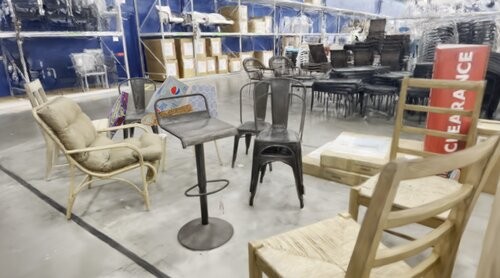}}
		{\textbf{No question provided/possible}}
		{In which frame is the "12-pack carton of diet pepsi cans" most likely colliding with another object? \opts{A}{B}{C}{\textbf{D}}}
		
		\vqaSplitEntry{F\_COLLISION\_OBJECT\_\allowbreak SCENE\_FRAME\_MULTI}
		{Mechanics}
		{Collision}
		{}
		{\stripFour{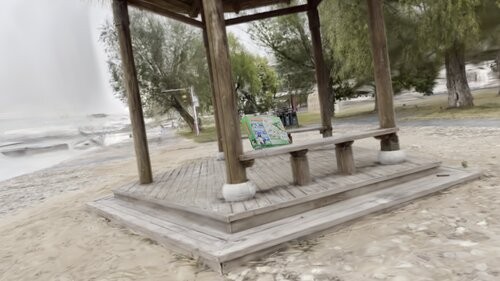}{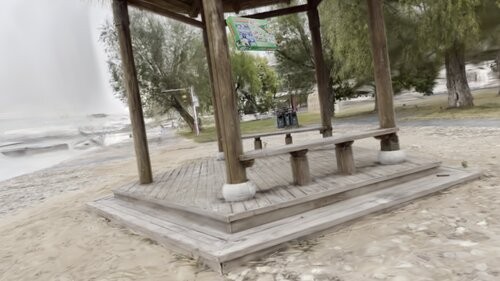}{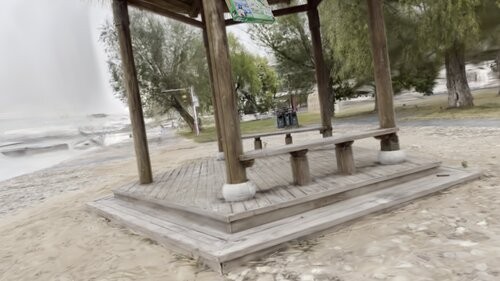}{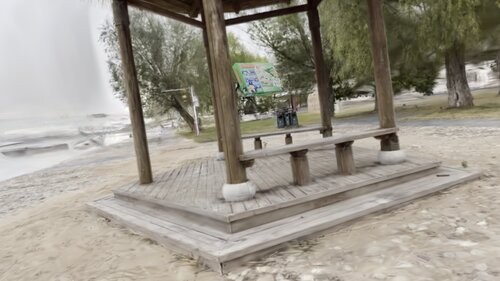}}
		{\textbf{No question provided/possible}}
		{In which frame is the "boxed My Monopoly board game" most likely colliding with the static scene? \opts{\textbf{A}}{B}{C}{D}}
		
		\vqaSplitEntry{F\_MASS\_OBJECT}
		{Material Understanding}
		{Mass}
		{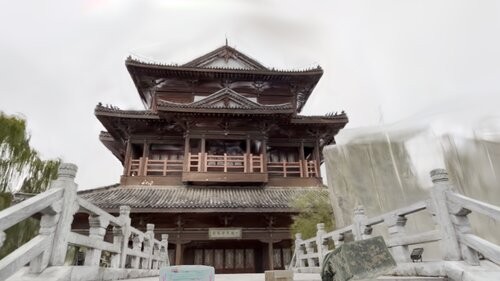}
		{\stripEight{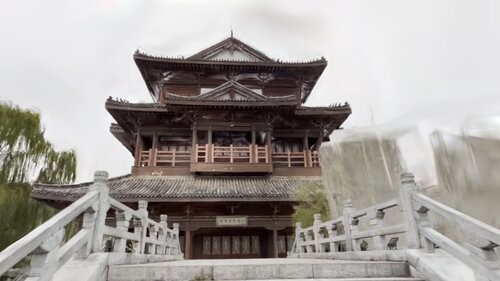}{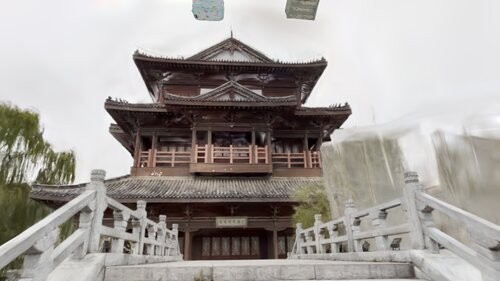}{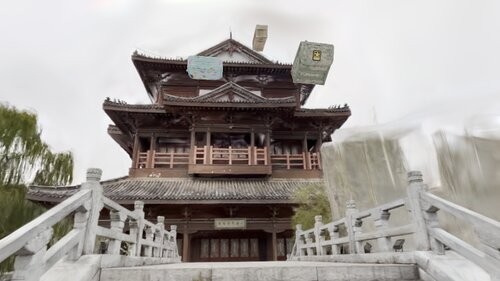}{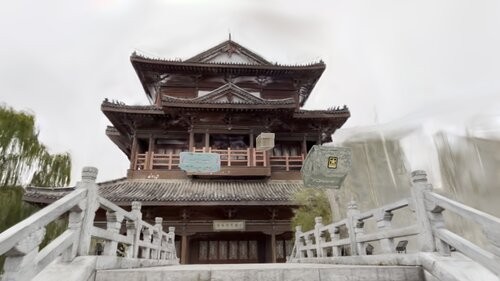}
			{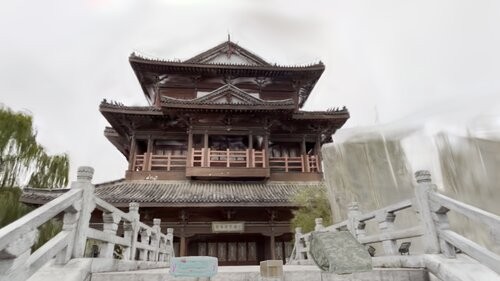}{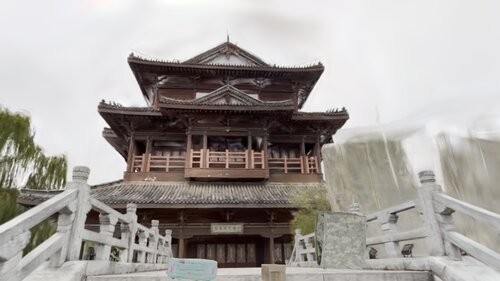}{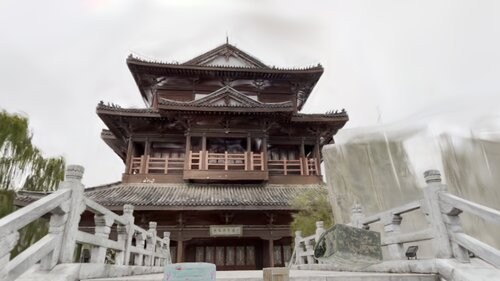}{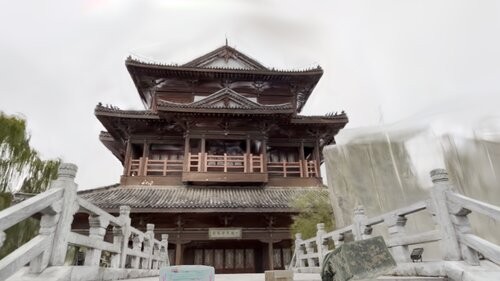}}
		{Based on the image, what is the mass of the "army-design lunch bag"? \opts{2.00 kgs}{\textbf{3.12 kgs}}{4.24 kgs}{0.89 kgs}}
		{Considering all frames, what is the mass of the "army-design lunch bag"? \opts{2.00 kgs}{\textbf{3.12 kgs}}{4.24 kgs}{0.89 kgs}}
		
		\vqaSplitEntry{F\_MASS\_HEAVIEST\_OBJECT}
		{Material Understanding}
		{Mass}
		{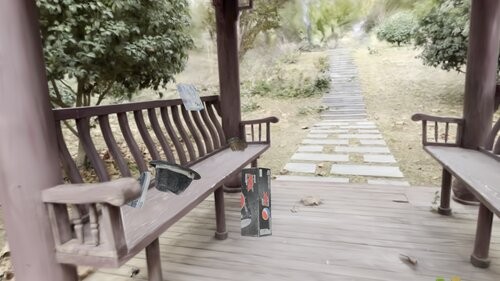}
		{\stripEight{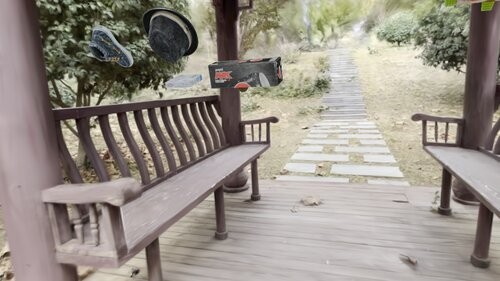}{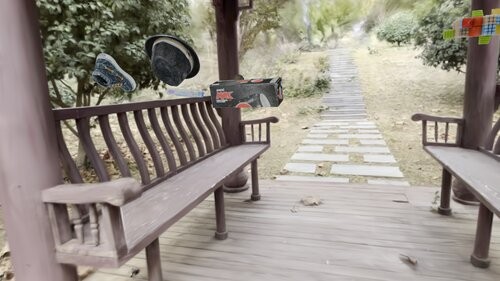}{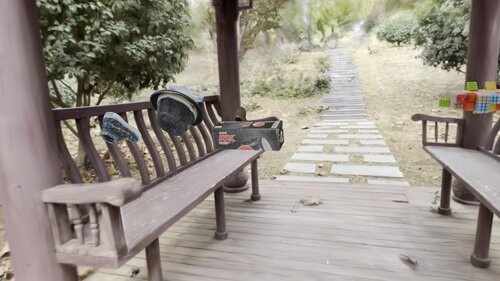}{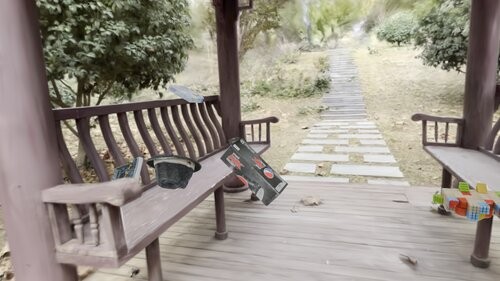}
			{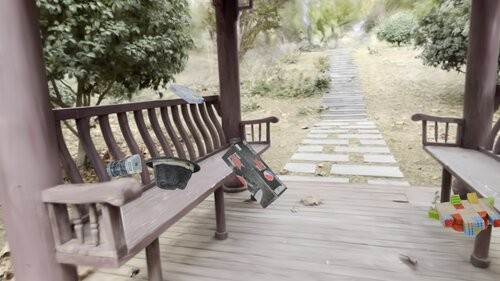}{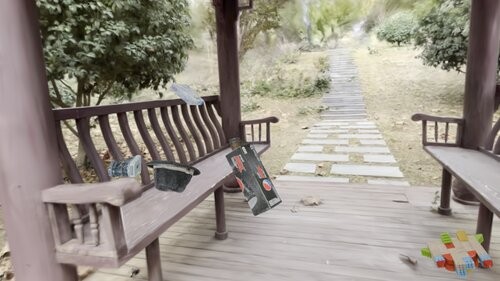}{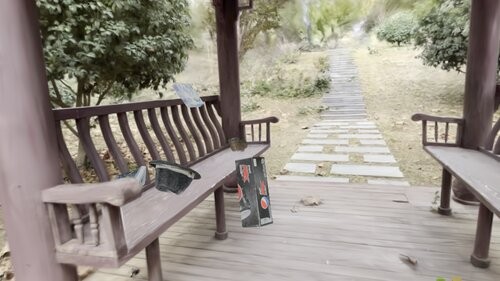}{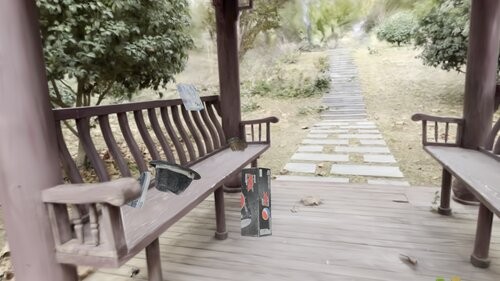}}
		{Based on the image, which visible single object has the greatest mass? \opts{boxed Frozen board game}{Teenage Mutant Ninja Turtles action figurine}{woven straw fedora hat}{\textbf{12-pack carton of Pepsi MAX cans}}}
		{Considering all frames, which single object, visible in the last frame, has the greatest mass? \opts{boxed Frozen board game}{Teenage Mutant Ninja Turtles action figurine}{woven straw fedora hat}{\textbf{12-pack carton of Pepsi MAX cans}}}
		
		\vqaSplitEntry{F\_MASS\_LIGHTEST\_OBJECT}
		{Material Understanding}
		{Mass}
		{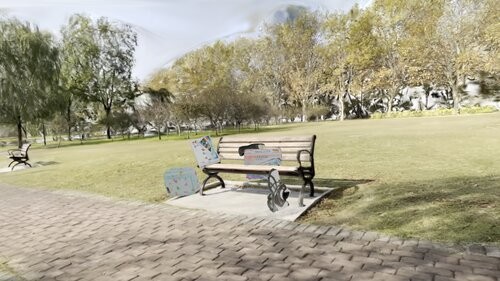}
		{\stripEight{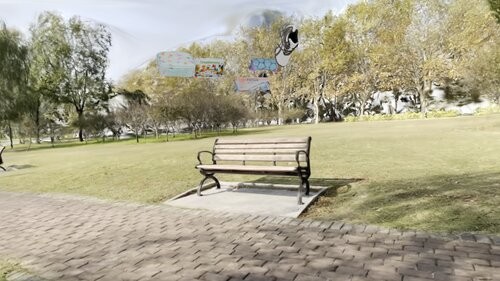}{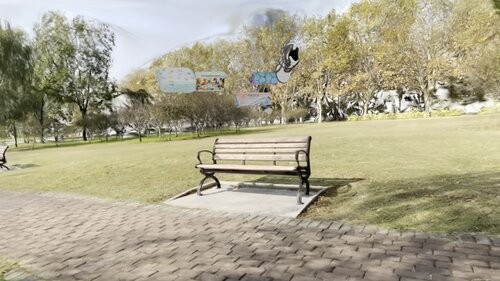}{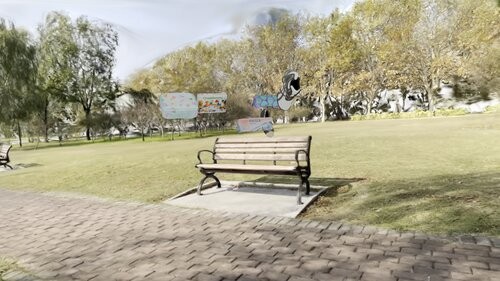}{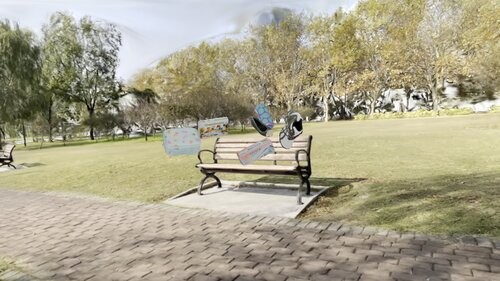}
			{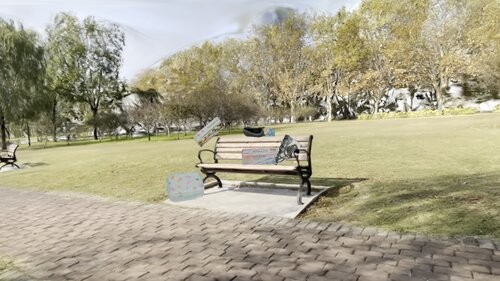}{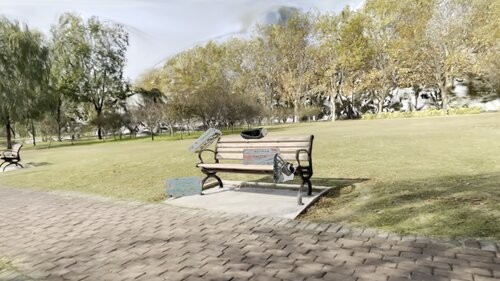}{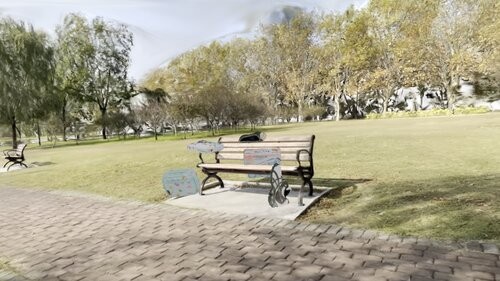}{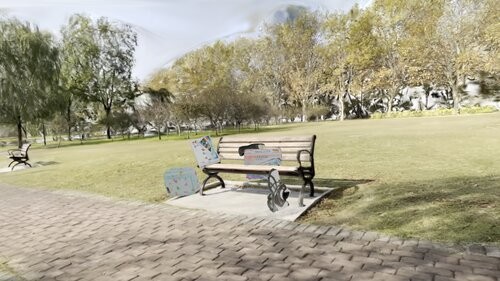}}
		{Based on the image, which visible single object has the least mass? \opts{12-pack Fresca soda carton}{turquoise insulated lunch bag}{boxed white Creationary board game}{\textbf{black and white ASICS golf shoe}}}
		{Considering all frames, which single object, visible in the last frame, has the least mass? \opts{12-pack Fresca soda carton}{turquoise insulated lunch bag}{boxed white Creationary board game}{\textbf{black and white ASICS golf shoe}}}
		
		\vqaSplitEntry{F\_PHYSICS\_\allowbreak PROPERTY\_DENSITY\_OBJECT}
		{Material Understanding}
		{Density}
		{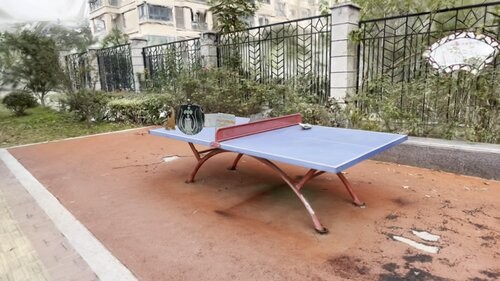}
		{\stripEight{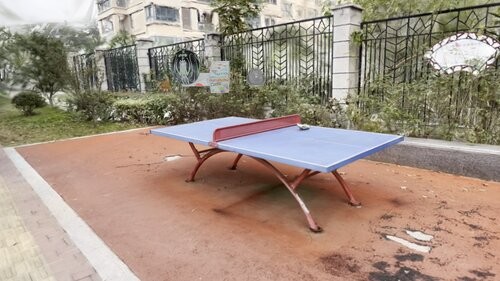}{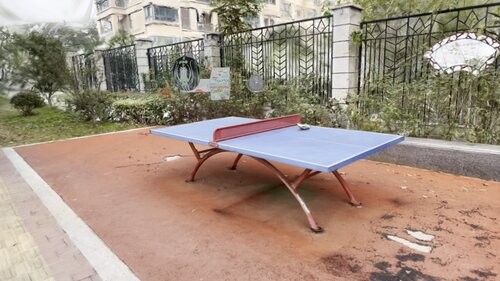}{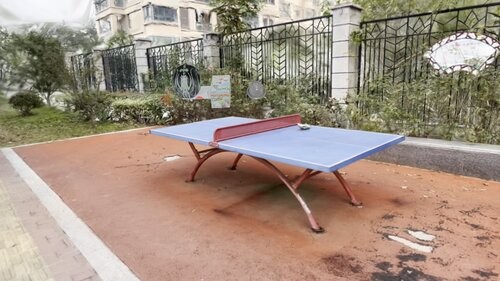}{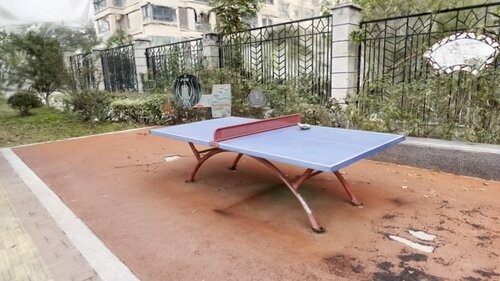}
			{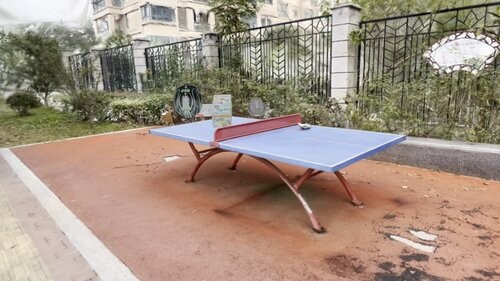}{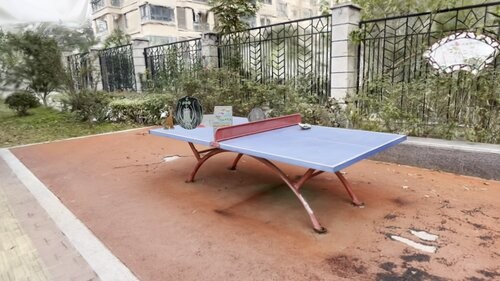}{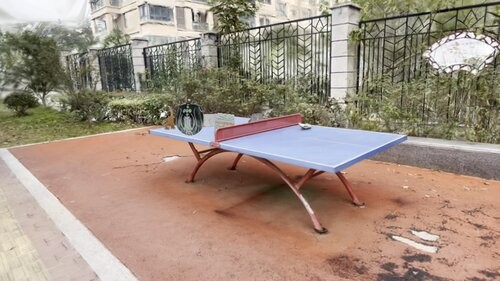}{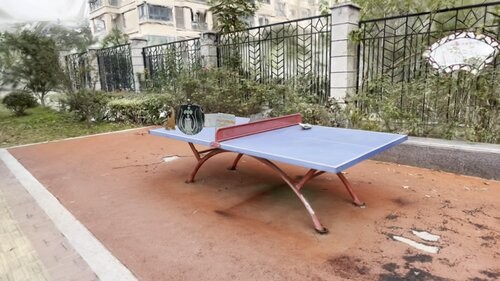}}
		{Based on the image, what is the estimated mean density of the "motorcycle helmet"? \opts{\textbf{110.0 kg/m\textasciicircum{}3}}{19.4 kg/m\textasciicircum{}3}{79.8 kg/m\textasciicircum{}3}{49.6 kg/m\textasciicircum{}3}}
		{Considering all frames, what is the estimated mean density of the "motorcycle helmet"? \opts{\textbf{110.0 kg/m\textasciicircum{}3}}{19.4 kg/m\textasciicircum{}3}{79.8 kg/m\textasciicircum{}3}{49.6 kg/m\textasciicircum{}3}}
		
		\vqaSplitEntry{F\_PHYSICS\_PROPERTY\_\allowbreak DENSITY\_OBJECT\_RELATIVE}
		{Material Understanding}
		{Density}
		{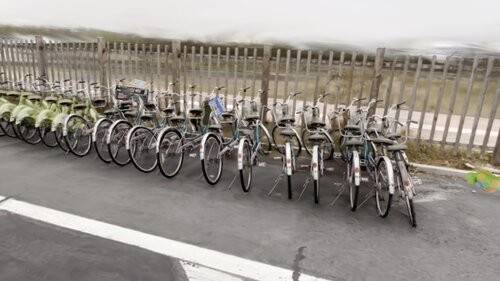}
		{\stripEight{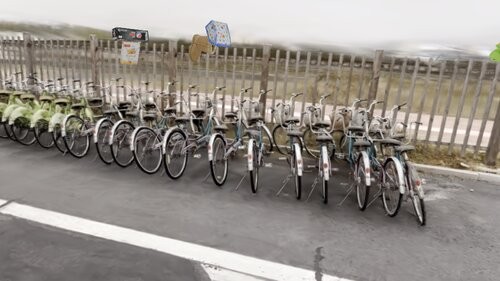}{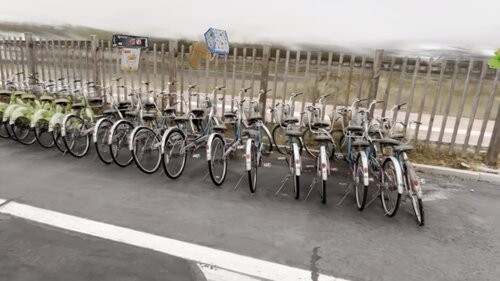}{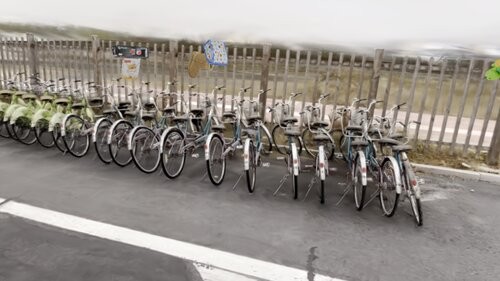}{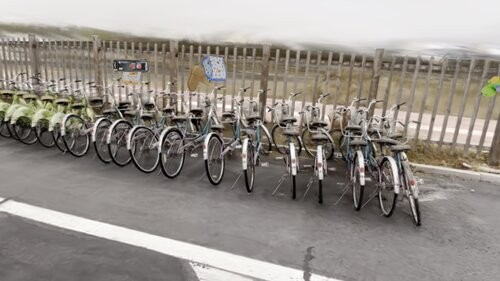}
			{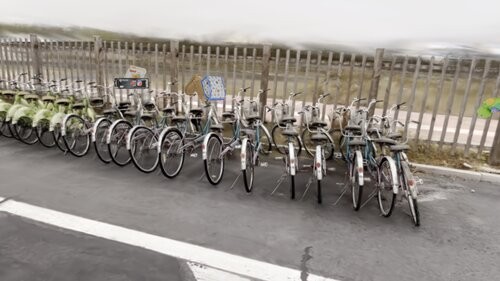}{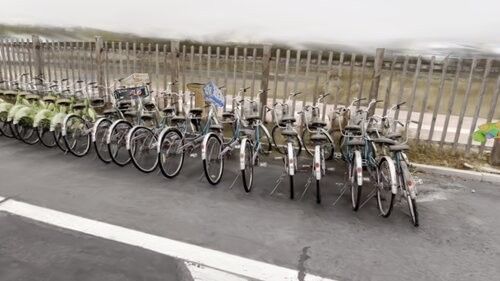}{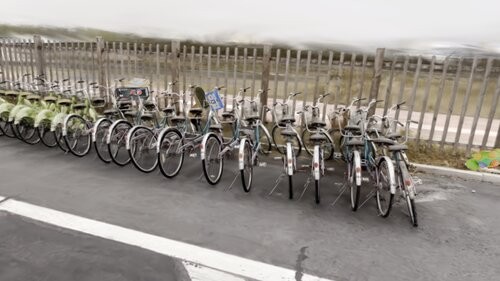}{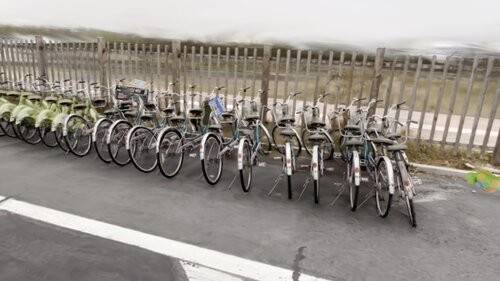}}
		{Based on the image, which visible object has the highest effective density? \opts{Teenage Mutant Ninja Turtles action figurine}{\textbf{12-pack carton of Pepsi MAX cans}}{white soccer shoe}{roll-along turtle toy}}
		{Considering all frames, which object, visible in the last frame, has the highest effective density? \opts{Teenage Mutant Ninja Turtles action figurine}{\textbf{12-pack carton of Pepsi MAX cans}}{white soccer shoe}{roll-along turtle toy}}
		
		\vqaSplitEntry{F\_PHYSICS\_PROPERTY\_YOUNG\_\allowbreak MODULUS\_OBJECT\_METRIC\_PREFIX}
		{Material Understanding}
		{Young Modulus}
		{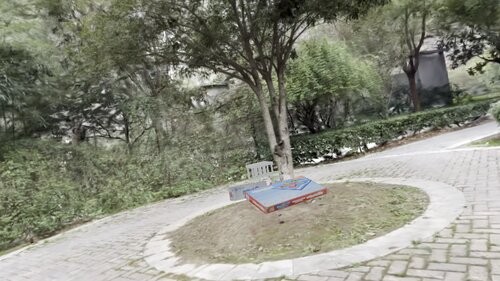}
		{\stripEight{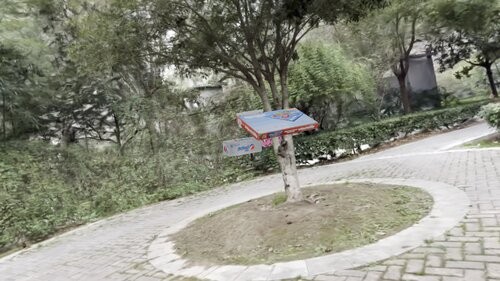}{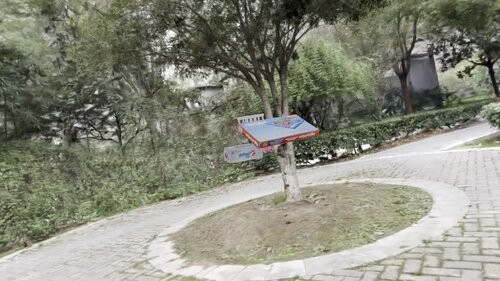}{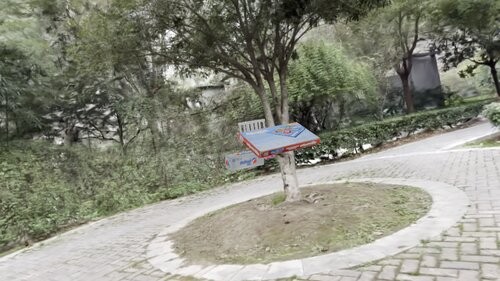}{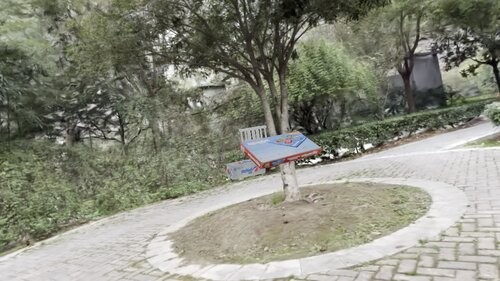}
			{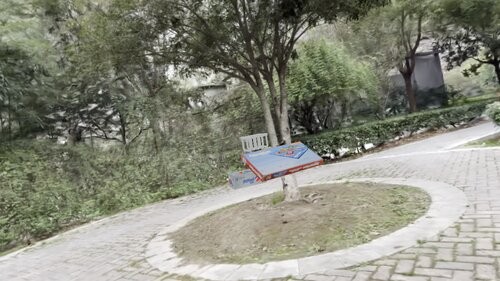}{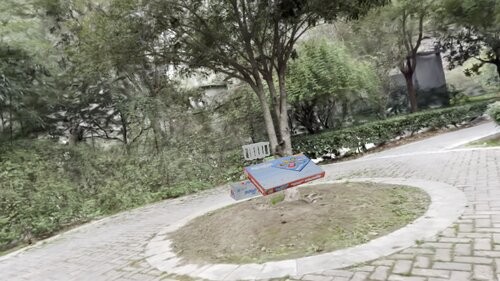}{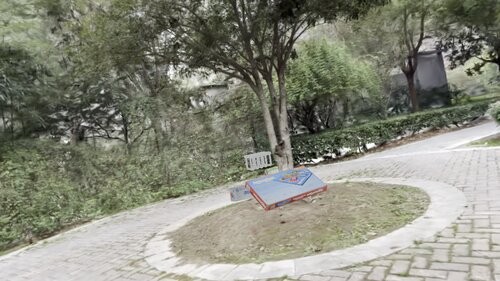}{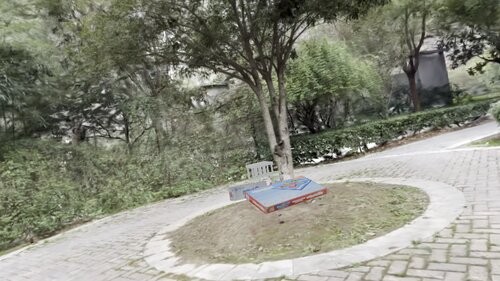}}
		{Based on the image, what is the Young's modulus of the "boxed Aggravation board game"? \opts{0,032 MPa}{3,15 MPa}{0,315 MPa}{\textbf{315,00 MPa}}}
		{Considering all frames, what is the Young's modulus of the "boxed Aggravation board game"? \opts{0,032 MPa}{3,15 MPa}{0,315 MPa}{\textbf{315,00 MPa}}}
		
		\vqaSplitEntry{F\_PHYSICS\_PROPERTY\_\allowbreak YOUNG\_MODULUS\_OBJECT\_SIMILAR}
		{Material Understanding}
		{Young Modulus}
		{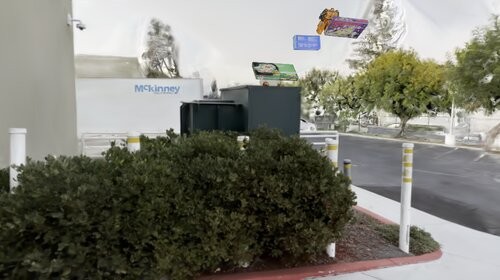}
		{\stripEight{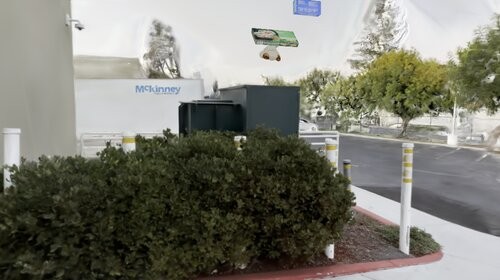}{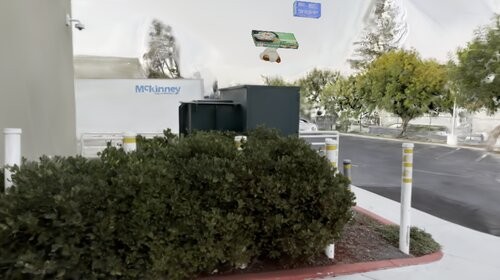}{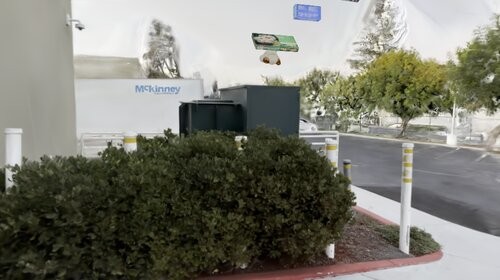}{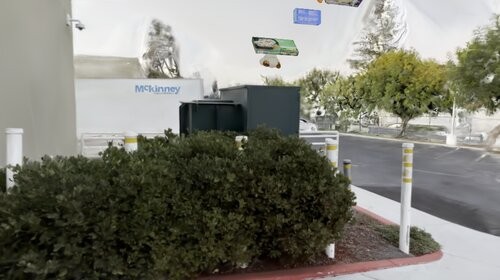}
			{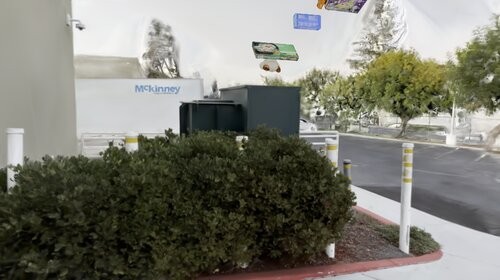}{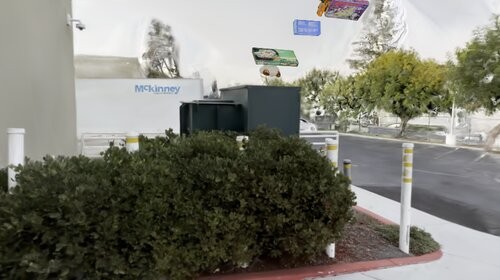}{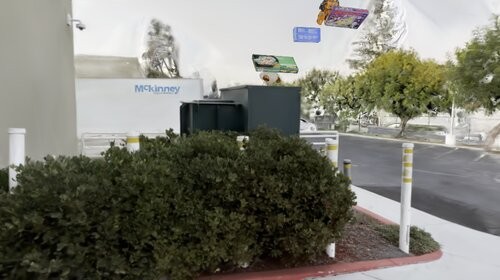}{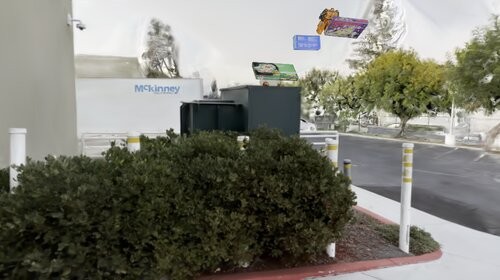}}
		{Based on the image, which visible object has a Young's Modulus most similar to that of the "boxed Clue board game"? \opts{\textbf{boxed purple Operation board game}}{Teenage Mutant Ninja Turtles action figurine}{orange Transformers action figure}{box of Samsung C406S cartridge}}
		{Considering all frames, which object, visible in the last frame, has a Young's Modulus most similar to that of the "boxed Clue board game"? \opts{\textbf{boxed purple Operation board game}}{Teenage Mutant Ninja Turtles action figurine}{orange Transformers action figure}{box of Samsung C406S cartridge}}
		
		\vqaSplitEntry{F\_PHYSICS\_PROPERTY\_YOUNG\_\allowbreak MODULUS\_OBJECT\_SIMILAR\_NON\_TECHNICAL}
		{Material Understanding}
		{Young Modulus}
		{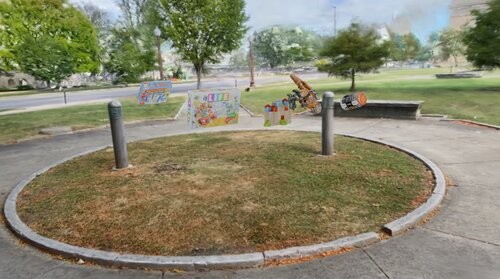}
		{\stripEight{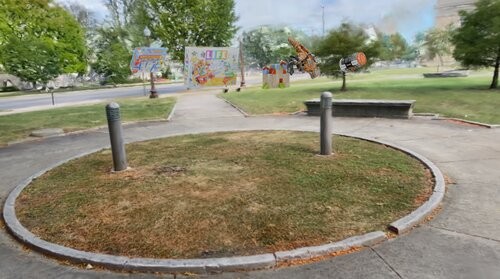}{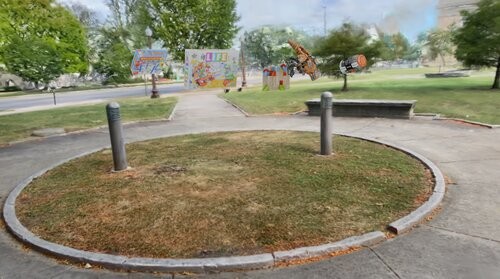}{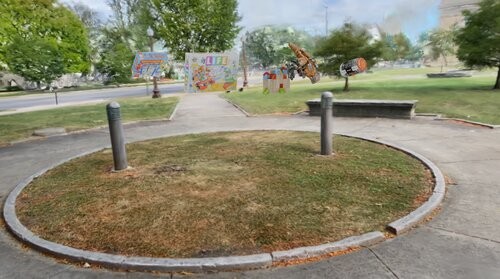}{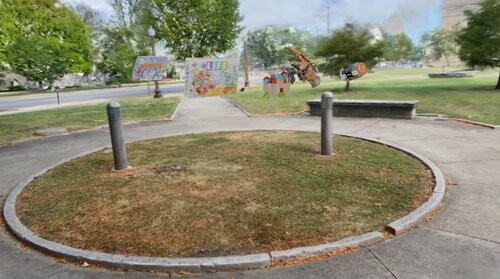}
			{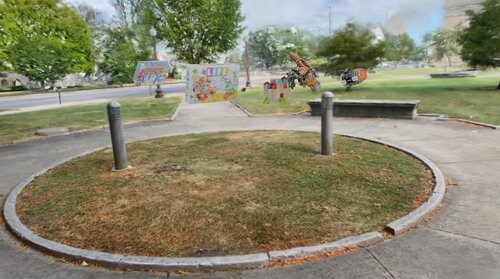}{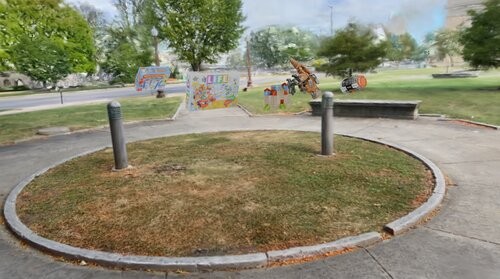}{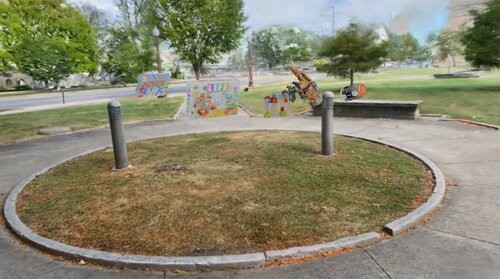}{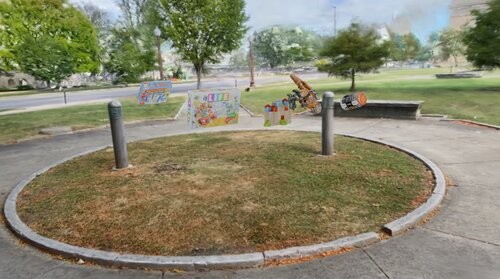}}
		{Based on the image, which visible object has a softness most similar to that of the "jar of Twinlab protein powder"? \opts{painted toy castle made of stacked blocks}{boxed LIFE board game}{\textbf{futuristic dinosaur robot figure}}{boxed Connect 4 Launchers game}}
		{Considering all frames, which object, visible in the last frame, has a softness most similar to that of the "jar of Twinlab protein powder"? \opts{painted toy castle made of stacked blocks}{boxed LIFE board game}{\textbf{futuristic dinosaur robot figure}}{boxed Connect 4 Launchers game}}
		
		\vqaSplitEntry{F\_PHYSICS\_PROPERTY\_\allowbreak YOUNG\_MODULUS\_HIGHEST}
		{Material Understanding}
		{Young Modulus}
		{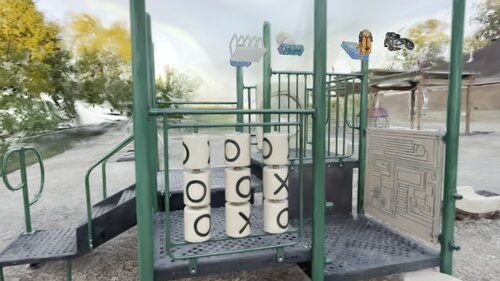}
		{\stripEight{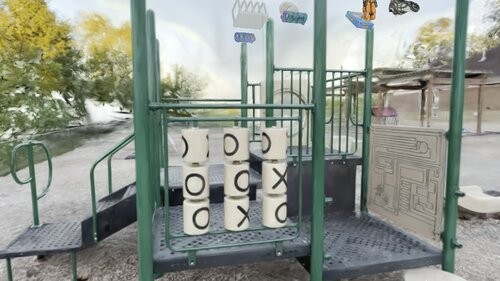}{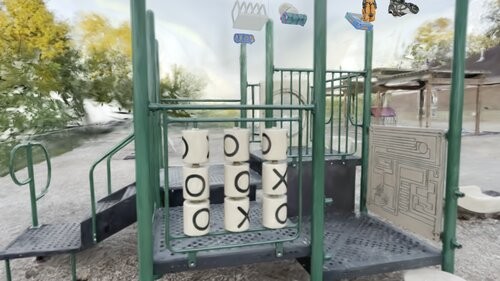}{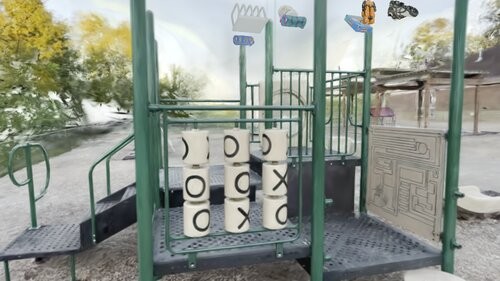}{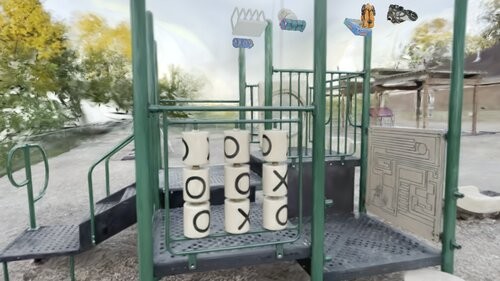}
			{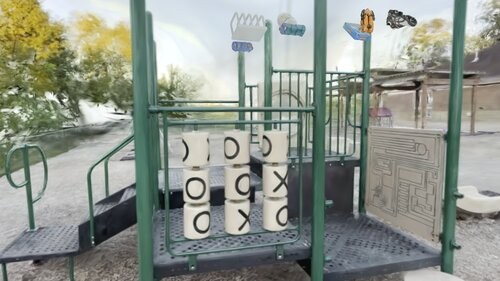}{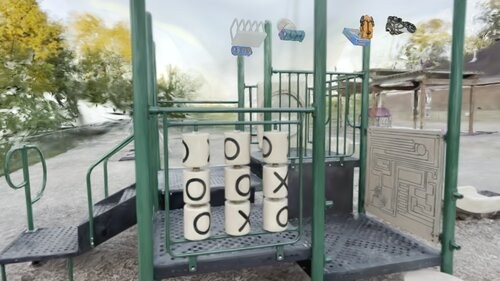}{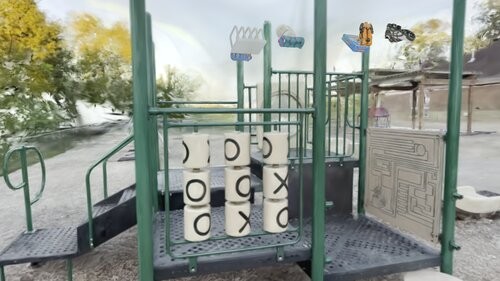}{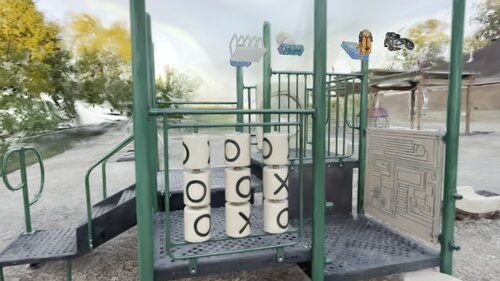}}
		{Based on the image, which visible object exhibits the highest Young's Modulus? \opts{boxed Balderdash board game}{\textbf{light gray file organizer rack}}{black and white ASICS golf shoe}{Teenage Mutant Ninja Turtles action figurine}}
		{Considering all frames, which object, visible in the last frame, exhibits the highest Young's Modulus? \opts{boxed Balderdash board game}{\textbf{light gray file organizer rack}}{black and white ASICS golf shoe}{Teenage Mutant Ninja Turtles action figurine}}
		
		\vqaSplitEntry{F\_PHYSICS\_PROPERTY\_YOUNG\_\allowbreak MODULUS\_HIGHEST\_NON\_TECHNICAL}
		{Material Understanding}
		{Young Modulus}
		{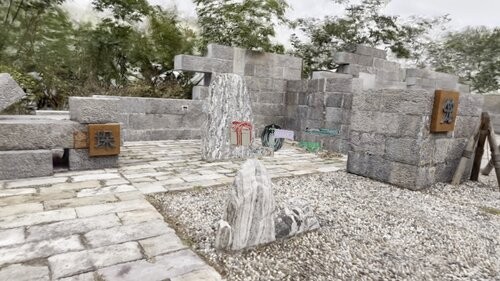}
		{\stripEight{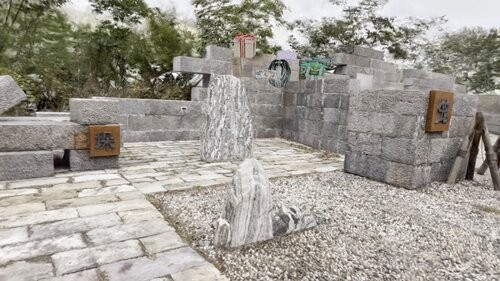}{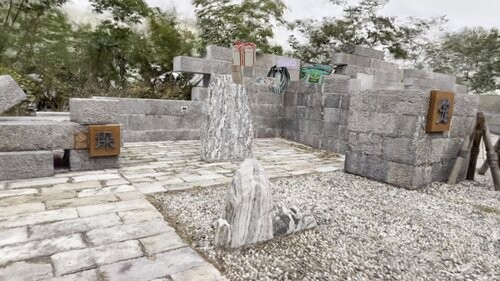}{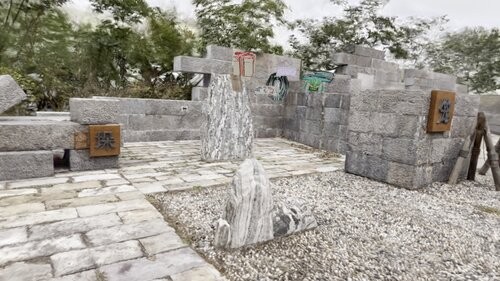}{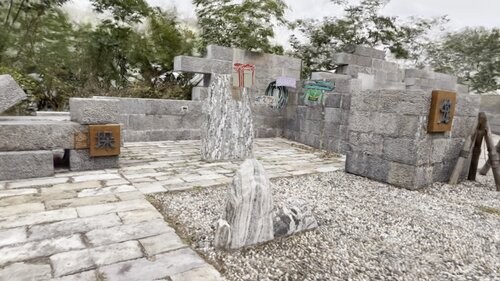}
			{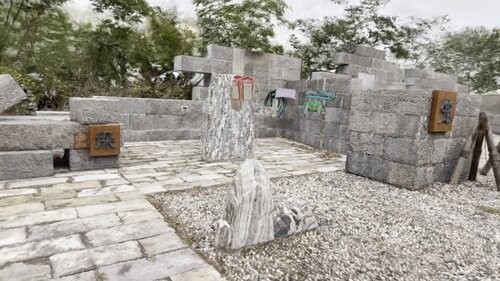}{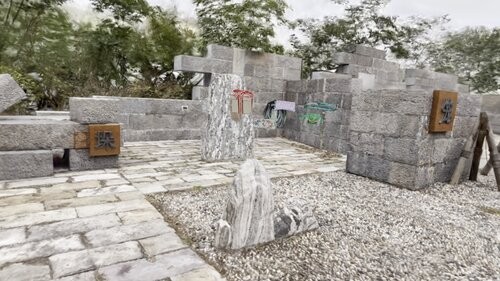}{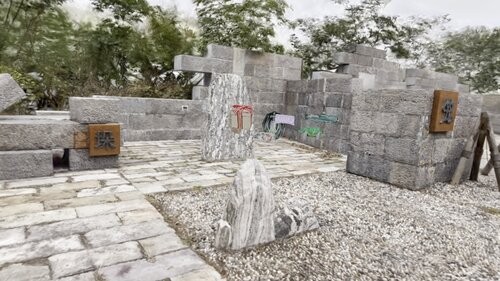}{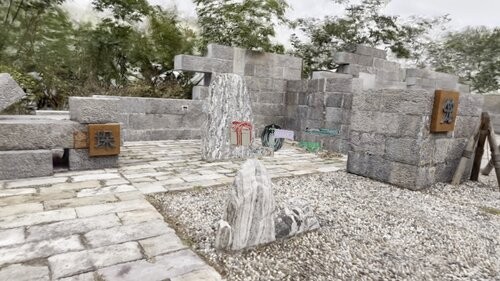}}
		{Based on the image, which visible object is the stiffest? \opts{gift box}{\textbf{motorcycle helmet}}{Teenage Mutant Ninja Turtles action figurine}{white soccer shoe}}
		{Considering all frames, which object, visible in the last frame, is the stiffest? \opts{gift box}{\textbf{motorcycle helmet}}{Teenage Mutant Ninja Turtles action figurine}{white soccer shoe}}
		
		\vqaSplitEntry{F\_PHYSICS\_PROPERTY\_YOUNG\_\allowbreak MODULUS\_OBJECT\_HIGH\_LEVEL}
		{Material Understanding}
		{Young Modulus}
		{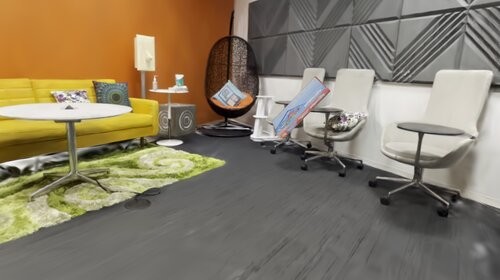}
		{\stripEight{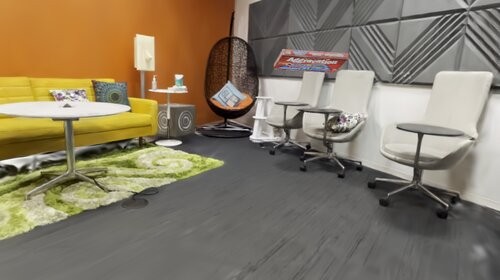}{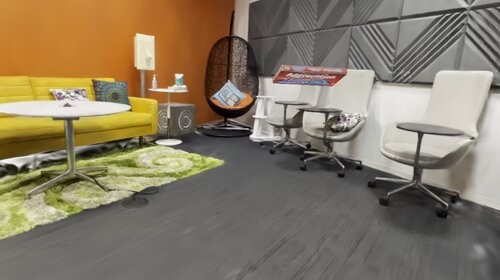}{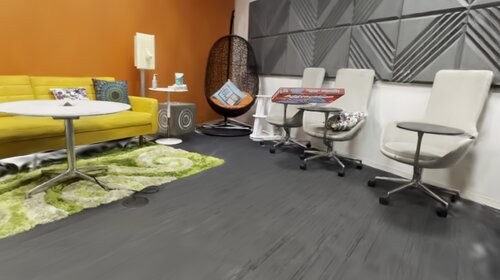}{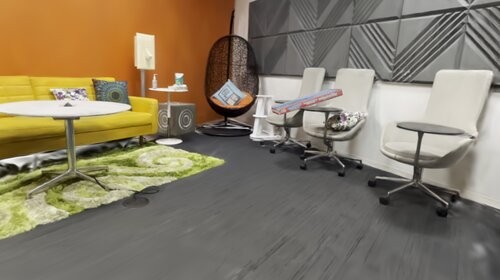}
			{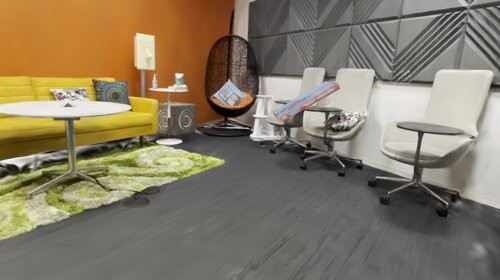}{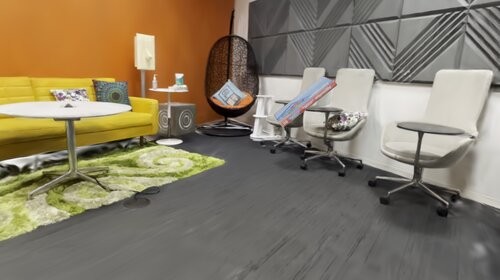}{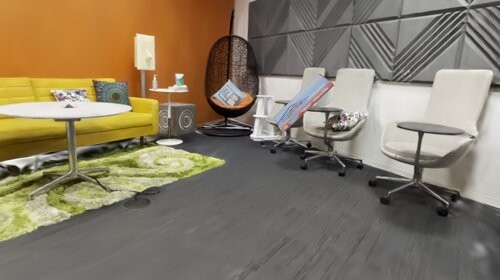}{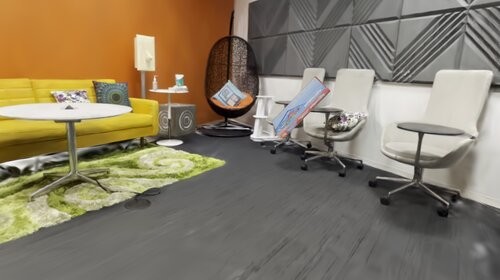}}
		{Based on the image, which attribute best describes the "boxed Aggravation board game" in terms of deformability? \opts{\textbf{Rigid (Holds shape perfectly)}}{Flexible (Bendable but tough)}{Soft (Deformable like stiff foam)}{Very Soft (No resistance, like a plush toy)}}
		{Considering all frames, which attribute best describes the "boxed Aggravation board game", visible in the last frame, in terms of deformability? \opts{\textbf{Rigid (Holds shape perfectly)}}{Flexible (Bendable but tough)}{Soft (Deformable like stiff foam)}{Very Soft (No resistance, like a plush toy)}}
		
		\vqaSplitEntry{F\_PHYSICS\_PROPERTY\_\allowbreak POISSON\_RATIO\_OBJECT}
		{Material Understanding}
		{Poisson Ratio}
		{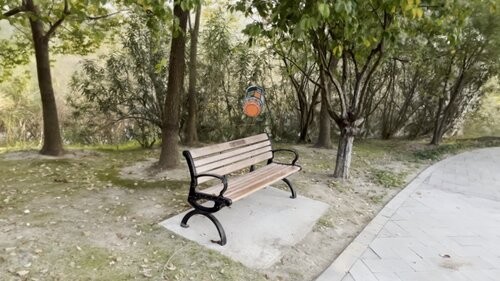}
		{\stripEight{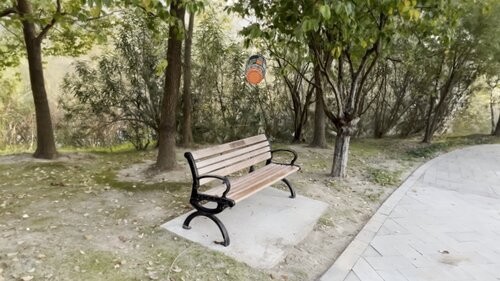}{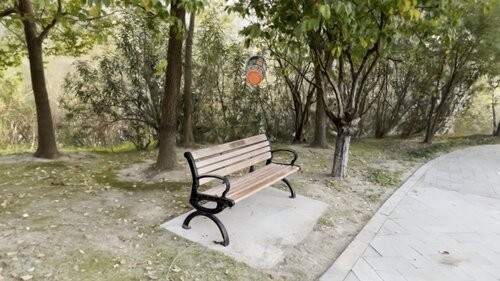}{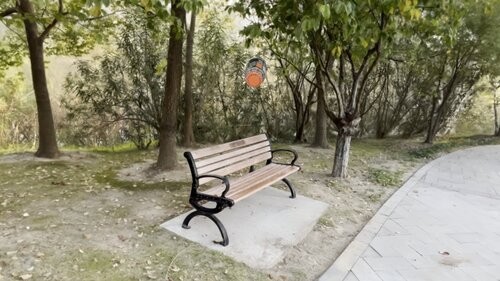}{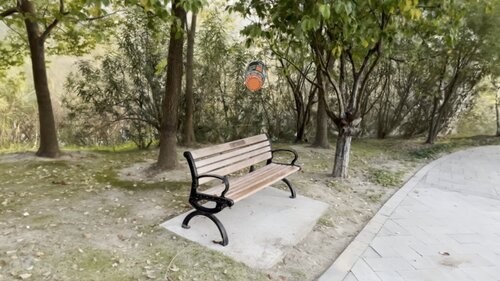}
			{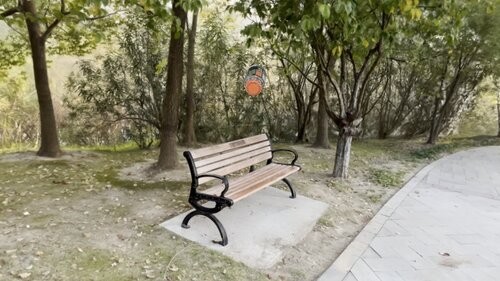}{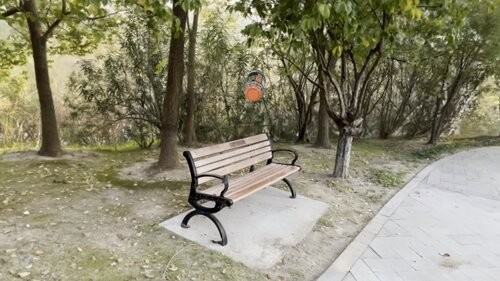}{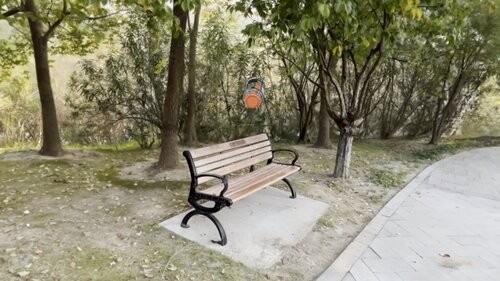}{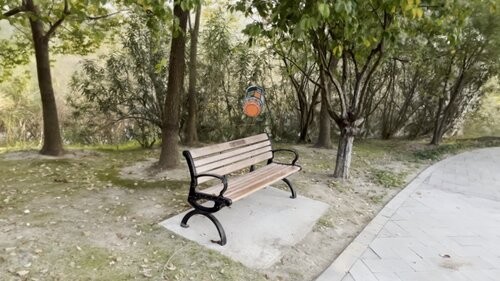}}
		{Based on the image, what is the Poisson ratio of the "jar of Twinlab protein powder"? \opts{0.10}{0.20}{0.40}{\textbf{0.30}}}
		{Considering all frames, what is the Poisson ratio of the "jar of Twinlab protein powder"? \opts{0.10}{0.20}{0.40}{\textbf{0.30}}}
		
		\vqaSplitEntry{F\_PHYSICS\_PROPERTY\_\allowbreak POISSON\_RATIO\_OBJECT\_SIMILAR}
		{Material Understanding}
		{Poisson Ratio}
		{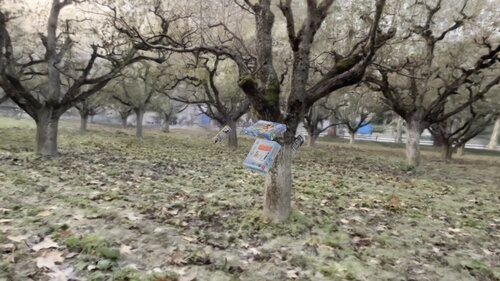}
		{\stripEight{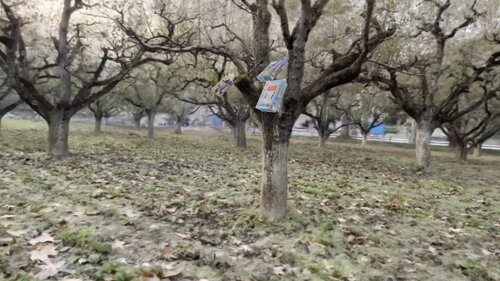}{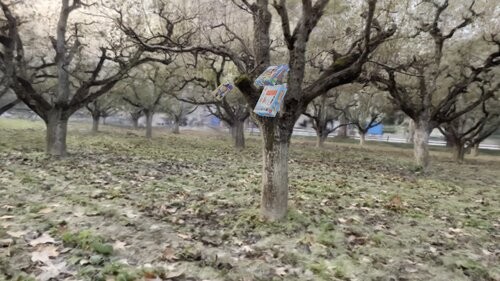}{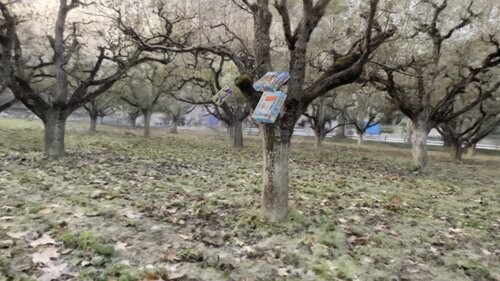}{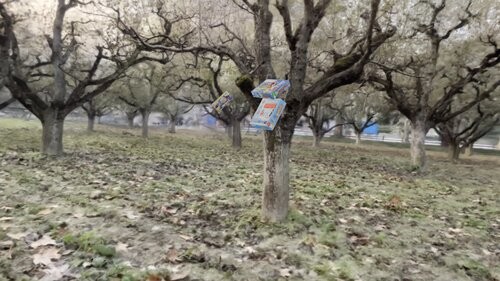}
			{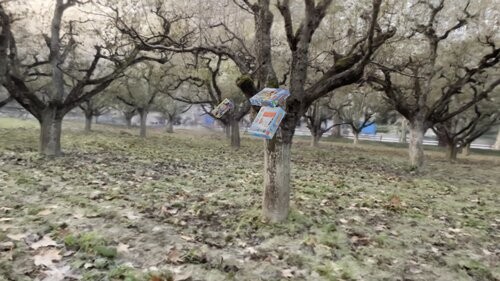}{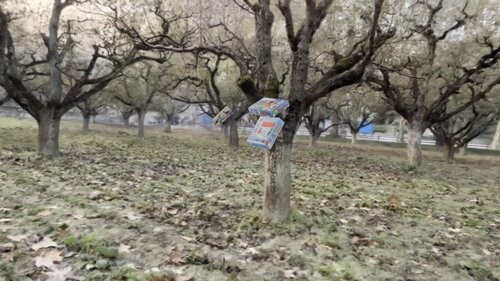}{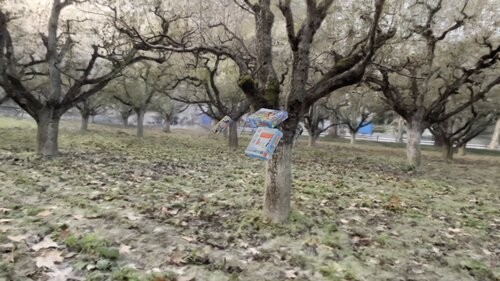}{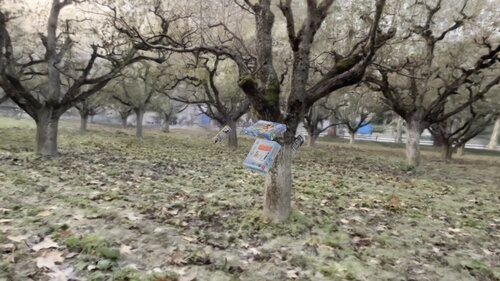}}
		{Based on the image, which visible object has a Poisson ratio most similar to that of the "boxed SLIDERS board game"? \opts{white soccer shoe}{\textbf{boxed Connect 4 Launchers game}}{small blue painted bucket}{Teenage Mutant Ninja Turtles action figurine}}
		{Considering all frames, which visible object has a Poisson ratio most similar to that of the "boxed SLIDERS board game" visible in the last frame? \opts{white soccer shoe}{\textbf{boxed Connect 4 Launchers game}}{small blue painted bucket}{Teenage Mutant Ninja Turtles action figurine}}
		
		\vqaSplitEntry{F\_PHYSICS\_PROPERTY\_POISSON\_\allowbreak RATIO\_OBJECT\_SIMILAR\_NON\_TECHNICAL}
		{Material Understanding}
		{Poisson Ratio}
		{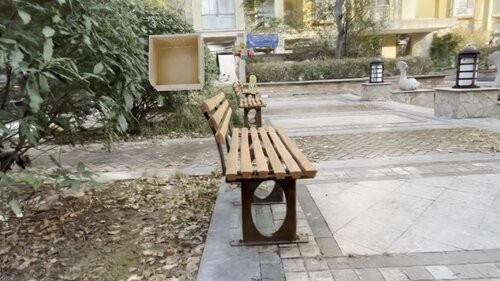}
		{\stripEight{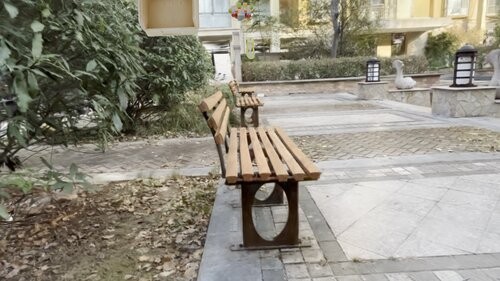}{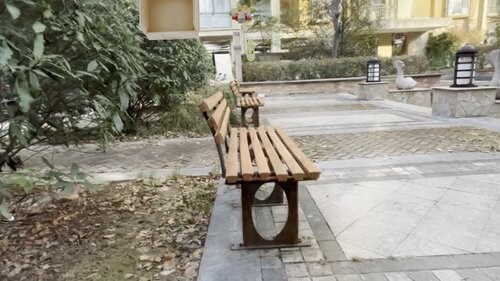}{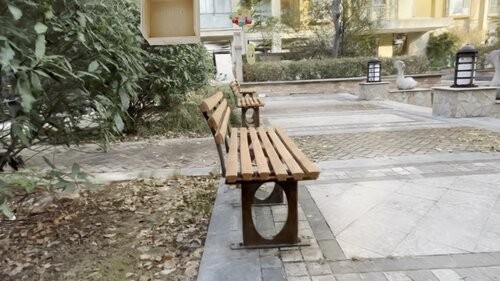}{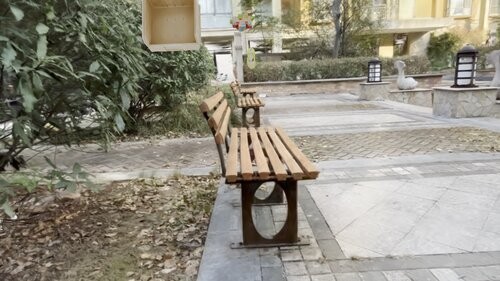}
			{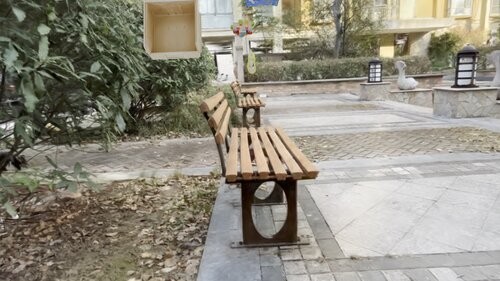}{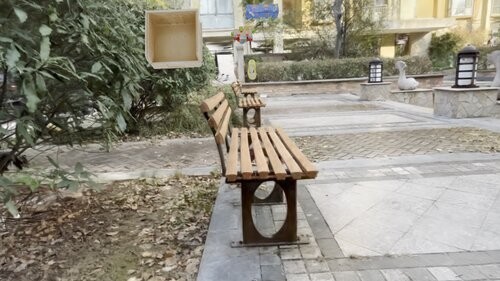}{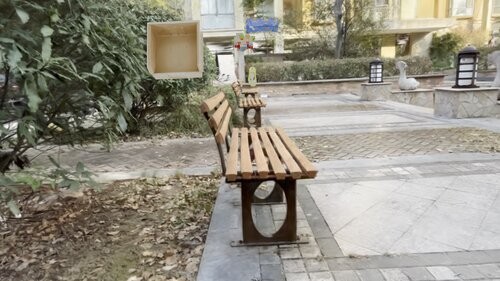}{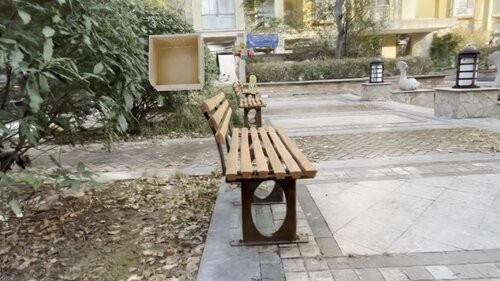}}
		{Based on the image, which visible object acts most like the "shelf bin" in terms of how it bulges sideways when squeezed? \opts{Teenage Mutant Ninja Turtles action figurine}{white soccer shoe}{small blue painted bucket}{\textbf{empty blue pencil case}}}
		{Considering all frames, which visible object acts most like the "shelf bin", visible in the last frame, in terms of how it bulges sideways when squeezed? \opts{Teenage Mutant Ninja Turtles action figurine}{white soccer shoe}{small blue painted bucket}{\textbf{empty blue pencil case}}}
		
		\vqaSplitEntry{F\_PHYSICS\_PROPERTY\_\allowbreak POISSON\_RATIO\_HIGHEST}
		{Material Understanding}
		{Poisson Ratio}
		{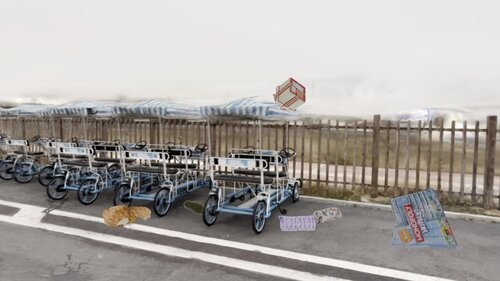}
		{\stripEight{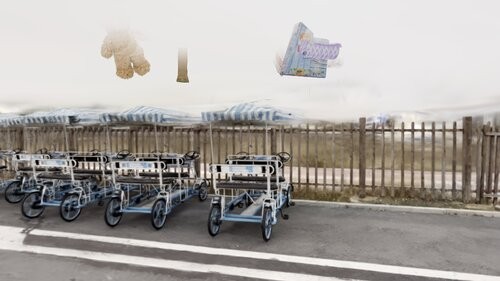}{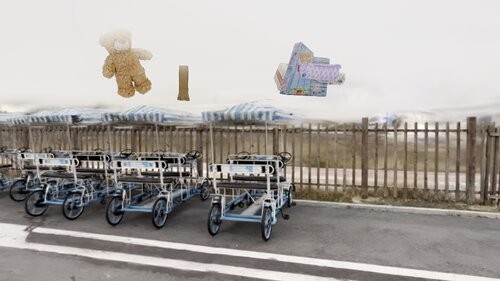}{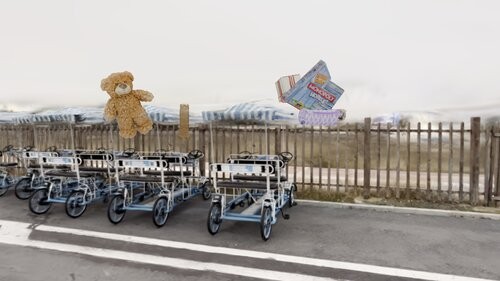}{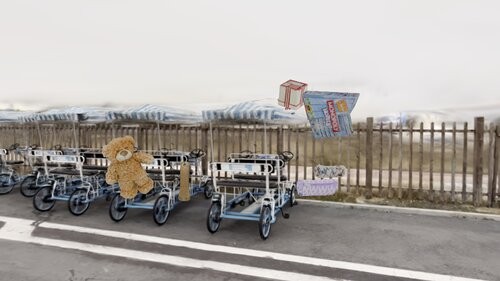}
			{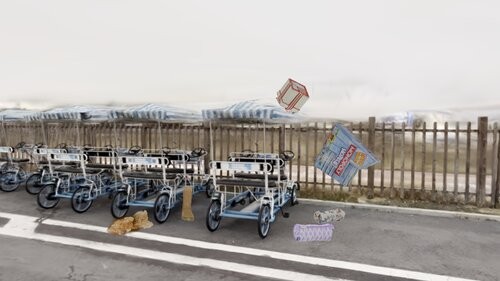}{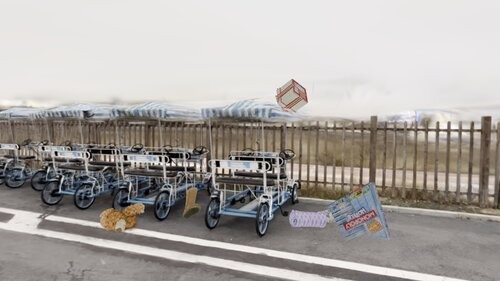}{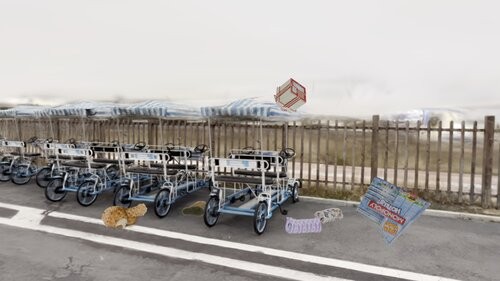}{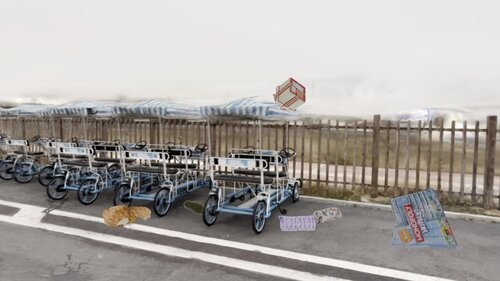}}
		{Based on the image, which visible object exhibits the largest Poisson ratio? \opts{\textbf{empty horse-design pencil case}}{gift box}{brown teddy bear toy}{boxed Monopoly Hotels board game}}
		{Considering all frames, which object, visible in the last frame, exhibits the largest Poisson ratio? \opts{\textbf{empty horse-design pencil case}}{gift box}{brown teddy bear toy}{boxed Monopoly Hotels board game}}
		
		\vqaSplitEntry{F\_PHYSICS\_PROPERTY\_POISSON\_\allowbreak RATIO\_HIGHEST\_NON\_TECHNICAL}
		{Material Understanding}
		{Poisson Ratio}
		{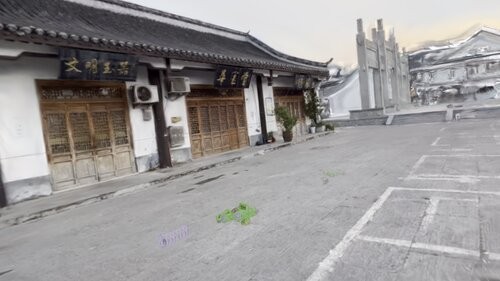}
		{\stripEight{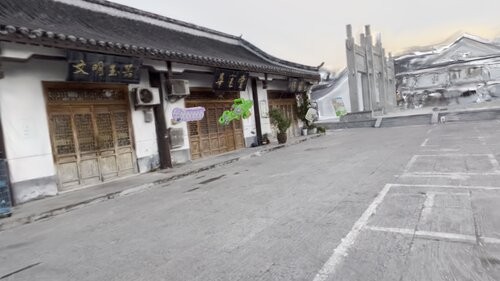}{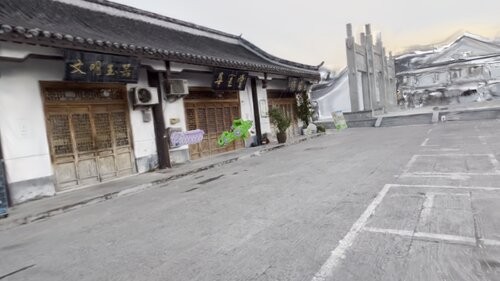}{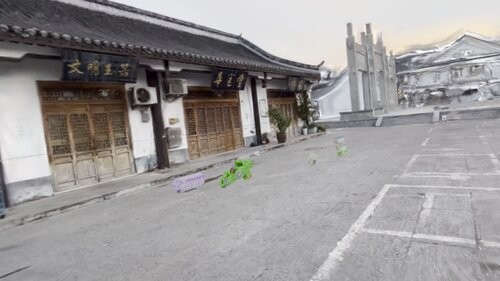}{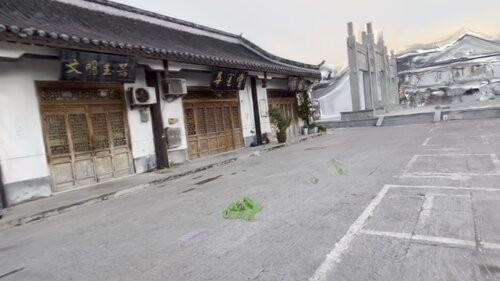}
			{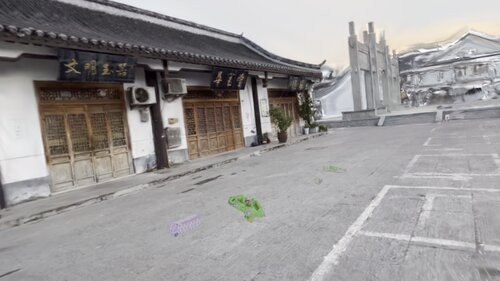}{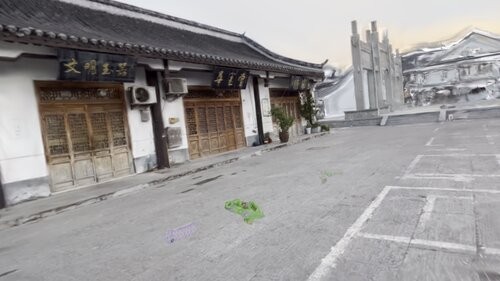}{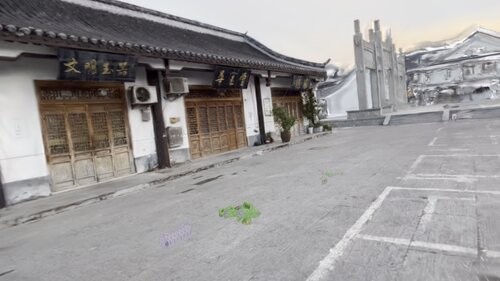}{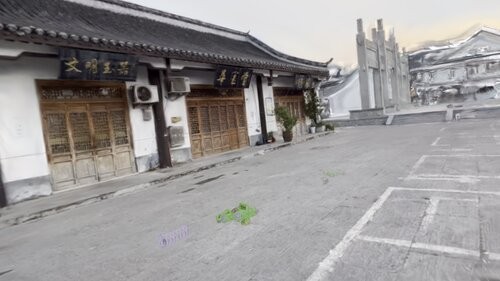}}
		{Based on the image, which visible object bulges out the most when you press on it? \opts{white-striped purple pencil case}{white soccer shoe}{Teenage Mutant Ninja Turtles action figurine}{\textbf{green ogre toy figurine}}}
		{Considering all frames, which object, visible in the last frame, bulges out the most when you press on it? \opts{white-striped purple pencil case}{white soccer shoe}{Teenage Mutant Ninja Turtles action figurine}{\textbf{green ogre toy figurine}}}
		
		\vqaSplitEntry{F\_PHYSICS\_PROPERTY\_\allowbreak POISSON\_HIGH\_LEVEL}
		{Material Understanding}
		{Poisson Ratio}
		{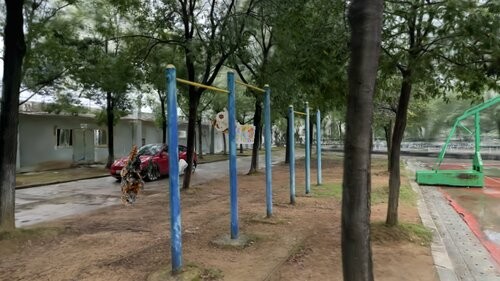}
		{\stripEight{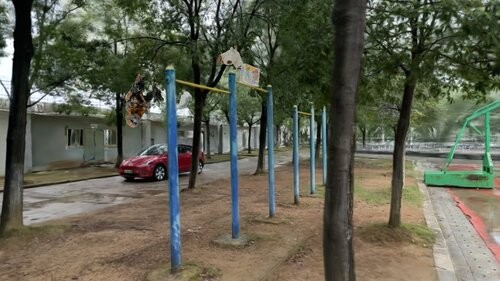}{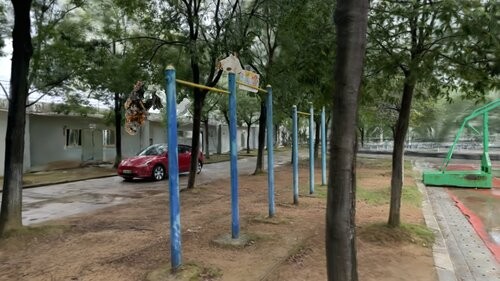}{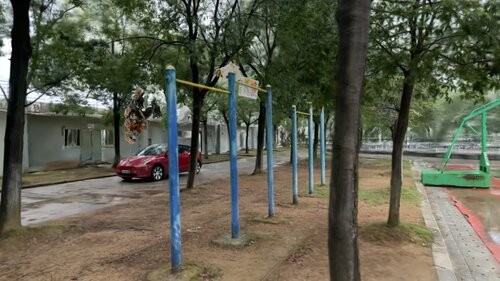}{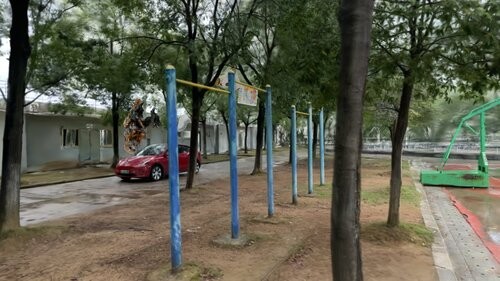}
			{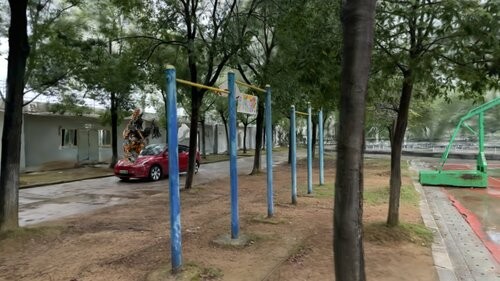}{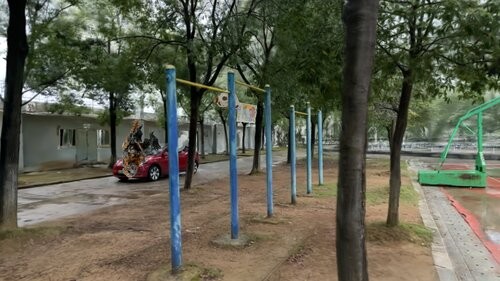}{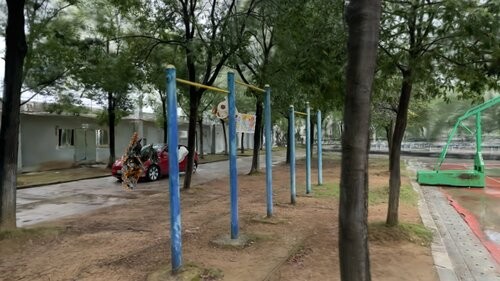}{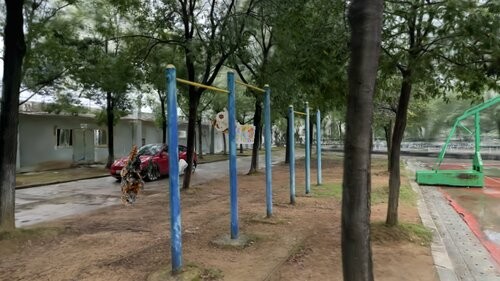}}
		{Based on the image, if the "boxed LIFE board game" were compressed vertically, how would its horizontal dimensions change? \opts{\textbf{It would expand sideways a moderate amount}}{It would barely change width}{It would contract inwards}{It would bulge out significantly}}
		{Considering all frames, if the "boxed LIFE board game", visible in the last frame, were compressed vertically, how would its horizontal dimensions change? \opts{\textbf{It would expand sideways a moderate amount}}{It would barely change width}{It would contract inwards}{It would bulge out significantly}}
		
		\vqaSplitEntry{F\_MATERIAL\_\allowbreak IDENTIFICATION\_SIMILAR\_OBJECT}
		{Material Understanding}
		{Material Identification}
		{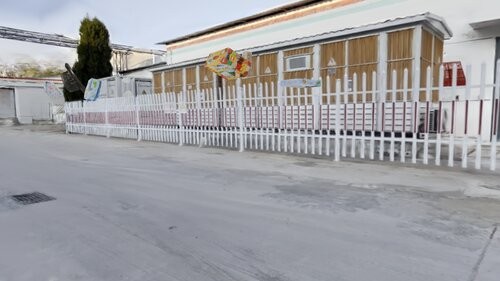}
		{\stripEight{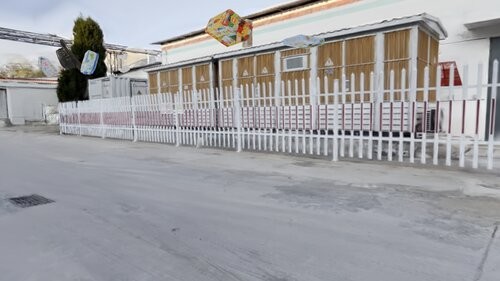}{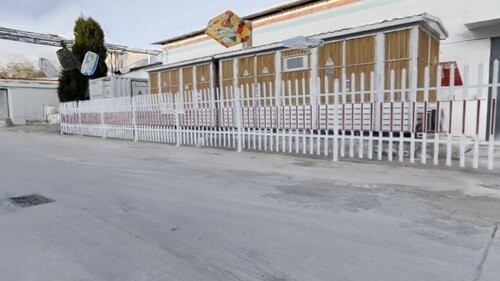}{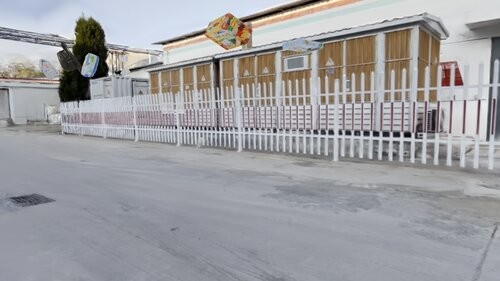}{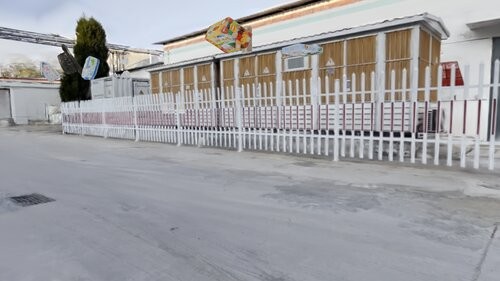}
			{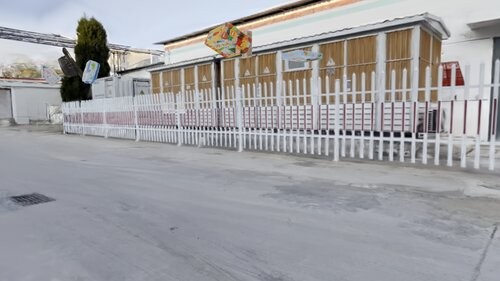}{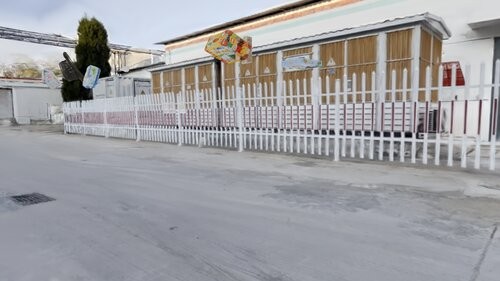}{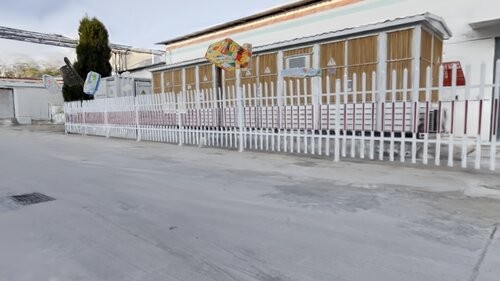}{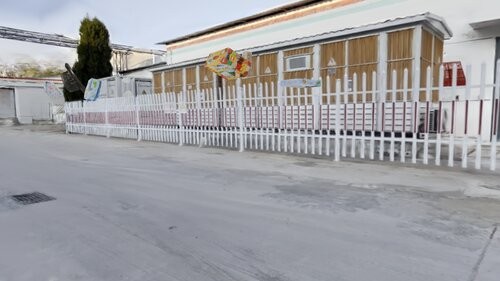}}
		{Based on the image, which visible object is made of a material most similar to that of the "Make-a-Match memory game box"? \opts{Teenage Mutant Ninja Turtles action figurine}{outback hat}{blue insulated lunch box with dinosaurs}{\textbf{1000-piece yellow LEGO brick assortment boxed set}}}
		{Considering all frames, which visible object is made of a material most similar to that of the "Make-a-Match memory game box"? \opts{Teenage Mutant Ninja Turtles action figurine}{outback hat}{blue insulated lunch box with dinosaurs}{\textbf{1000-piece yellow LEGO brick assortment boxed set}}}
		
		\vqaSplitEntry{F\_MATERIAL\_IDENTIFICATION\_\allowbreak OBJECT\_LEVEL\_1}
		{Material Understanding}
		{Material Identification}
		{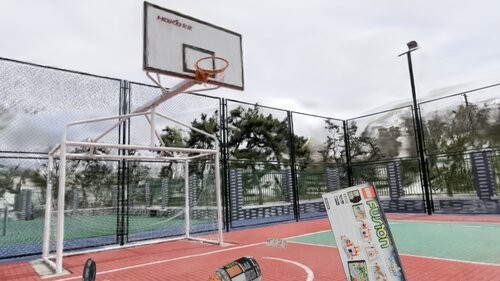}
		{\stripEight{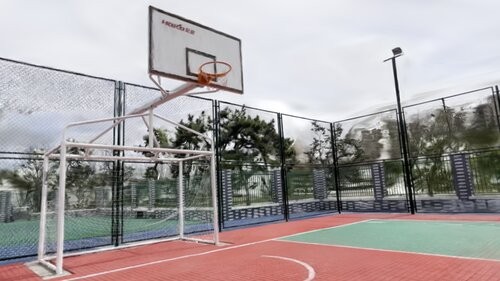}{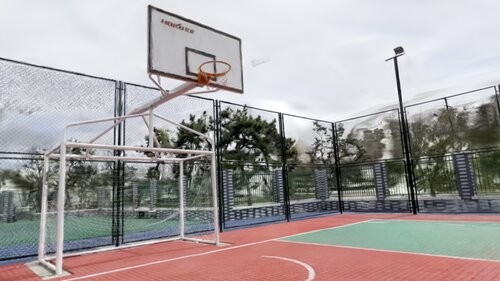}{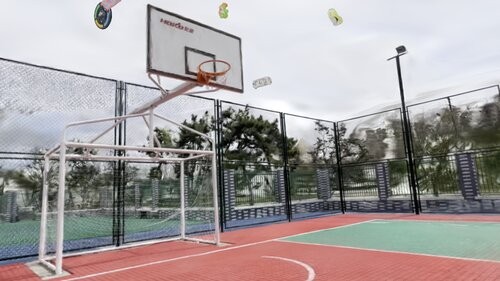}{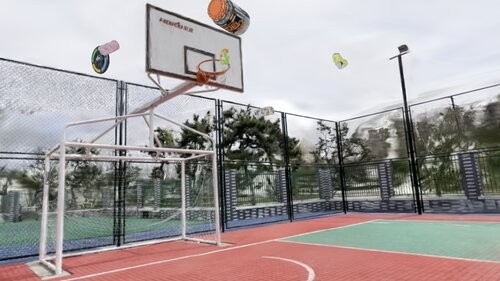}
			{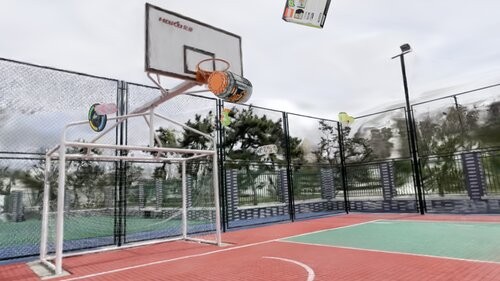}{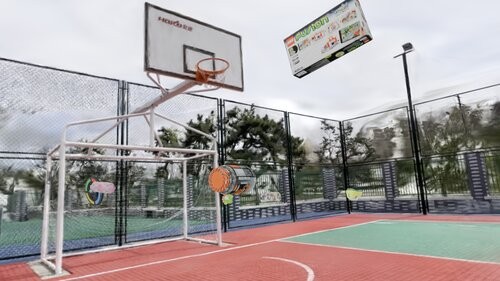}{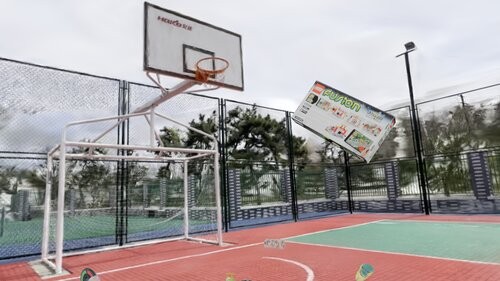}{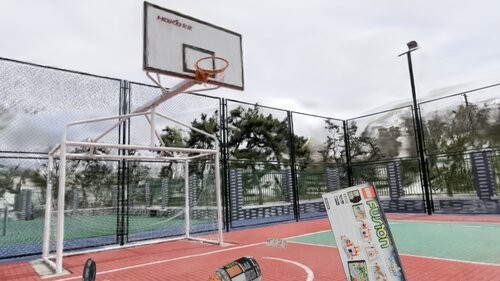}}
		{Based on the image, what material is the "jar of Twinlab protein powder" made of? \opts{\textbf{synthetic polymers}}{vegetal and cellulosic}{skins, textiles and fibers}{inorganic rigid}}
		{Considering all frames, what material is the "jar of Twinlab protein powder" made of? \opts{\textbf{synthetic polymers}}{vegetal and cellulosic}{skins, textiles and fibers}{inorganic rigid}}
		
		\vqaSplitEntry{F\_MATERIAL\_IDENTIFICATION\_\allowbreak OBJECT\_LEVEL\_2}
		{Material Understanding}
		{Material Identification}
		{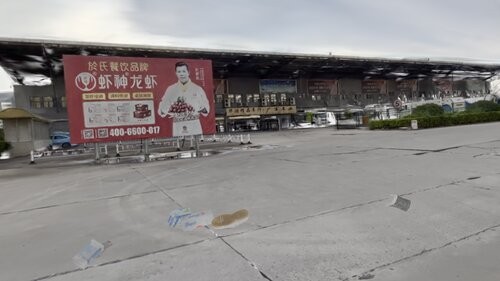}
		{\stripEight{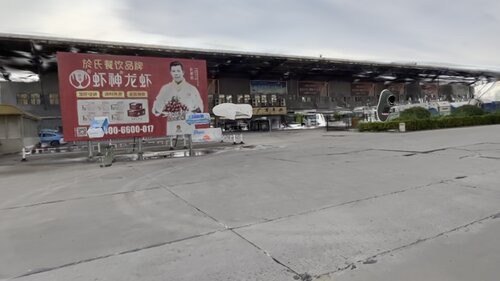}{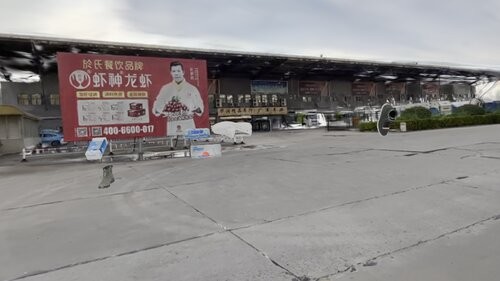}{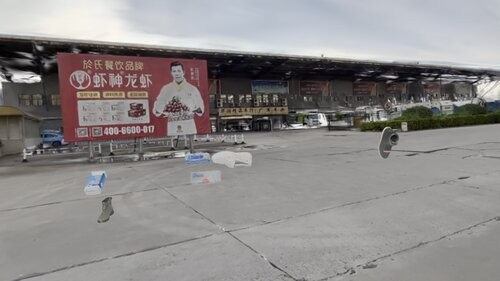}{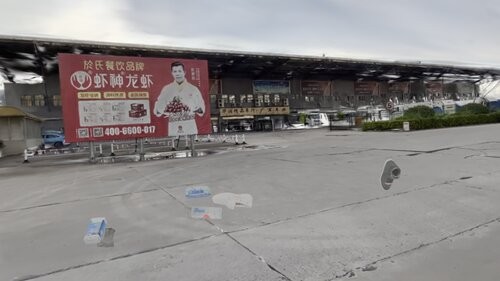}
			{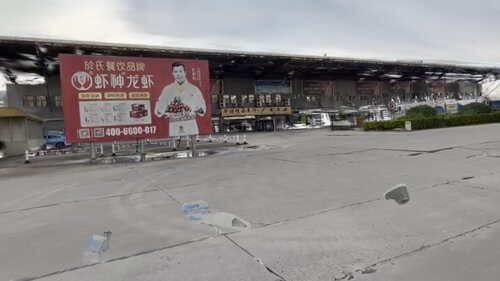}{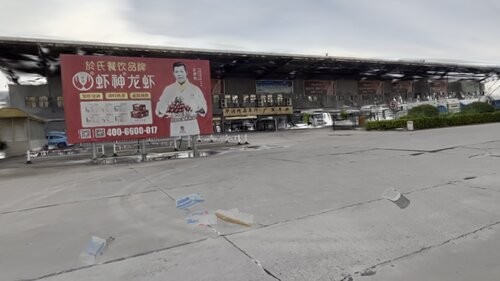}{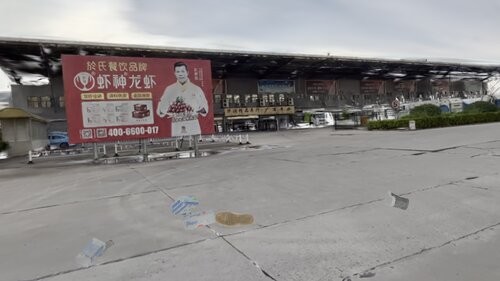}{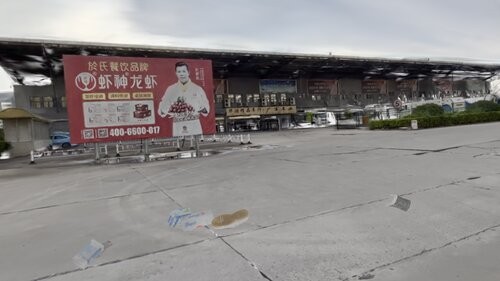}}
		{Based on the image, what material is the "12-pack carton of diet pepsi cans" made of? \opts{\textbf{processed cellulose}}{ligneous matter}{woven textiles}{cellular synthetics}}
		{Considering all frames, what material is the "12-pack carton of diet pepsi cans" made of? \opts{\textbf{processed cellulose}}{ligneous matter}{woven textiles}{cellular synthetics}}
		
		\vqaSplitEntry{F\_MATERIAL\_IDENTIFICATION\_\allowbreak OBJECT\_LEVEL\_3}
		{Material Understanding}
		{Material Identification}
		{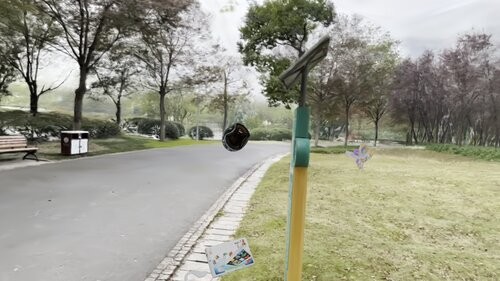}
		{\stripEight{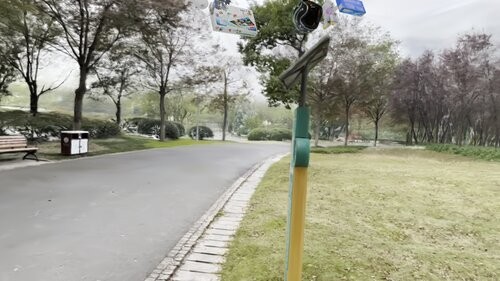}{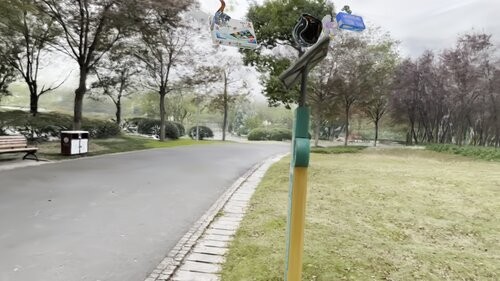}{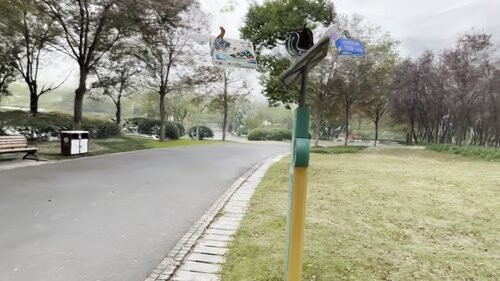}{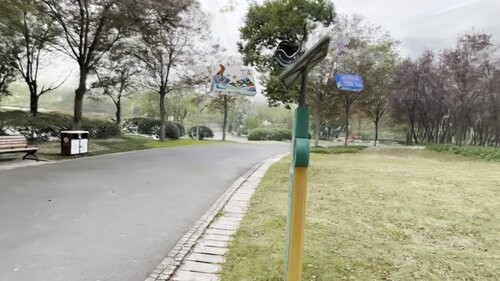}
			{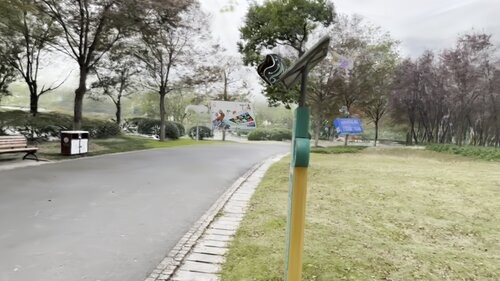}{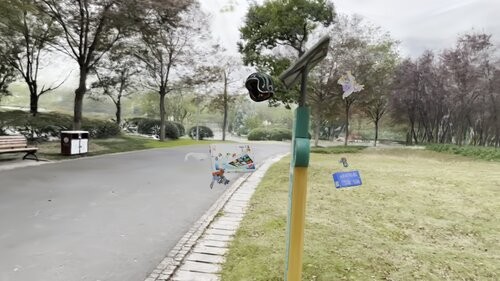}{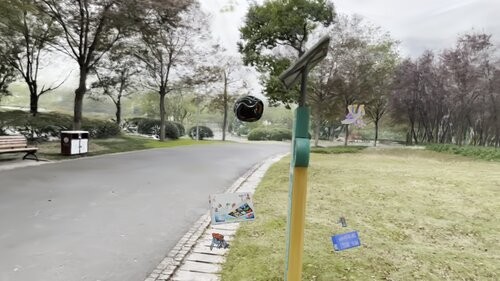}{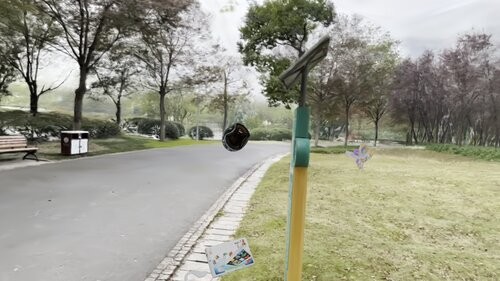}}
		{Based on the image, what material is the "Princess Celestia pony figurine toy" made of? \opts{metal}{\textbf{plastic}}{paper/cardboard}{ceramic}}
		{Considering all frames, what material is the "Princess Celestia pony figurine toy" made of? \opts{metal}{\textbf{plastic}}{paper/cardboard}{ceramic}}
		
		\vqaSplitEntry{F\_VISIBILITY\_OBJECT}
		{View Point}
		{Visibility}
		{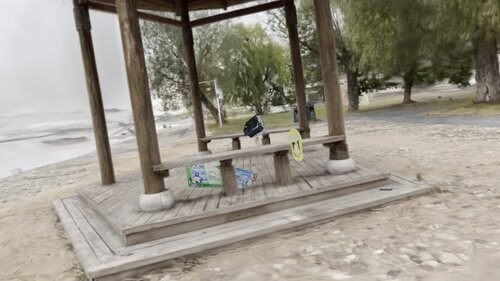}
		{\stripEight{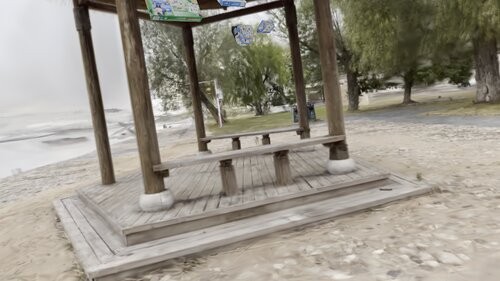}{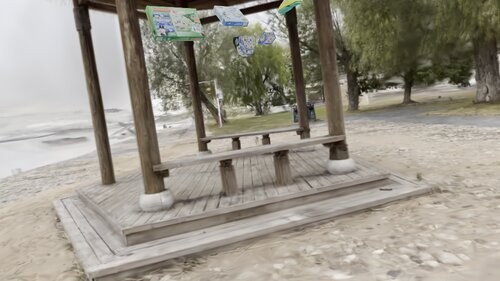}{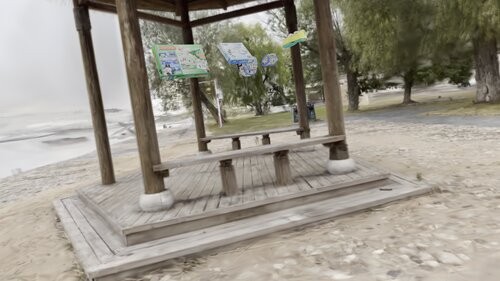}{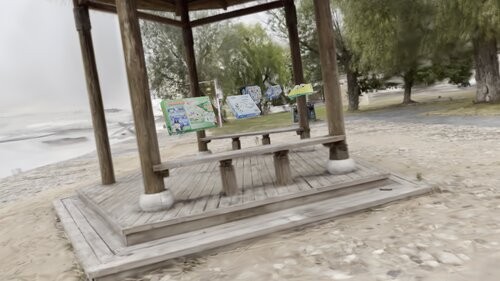}
			{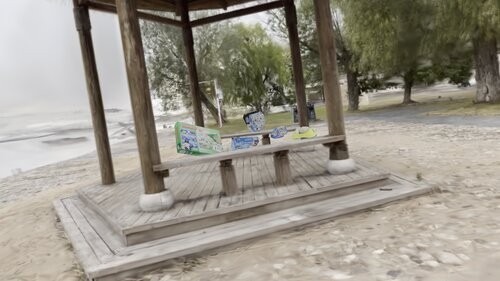}{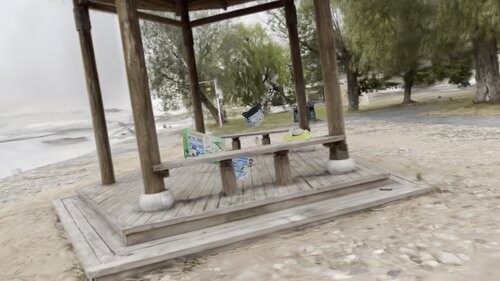}{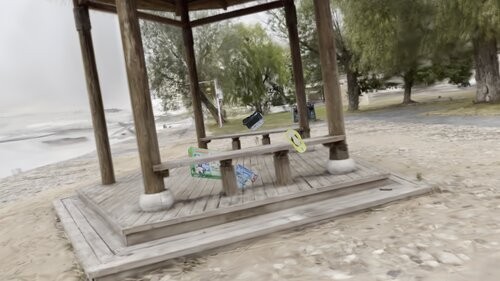}{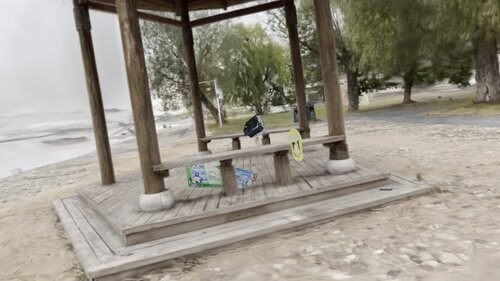}}
		{Based on the image, which of these objects is visible? \opts{turquoise mermaid-design backpack}{boxed Clue board game}{\textbf{boxed My Monopoly board game}}{white spoon}}
		{Considering all frames, which of these objects is visible in the last frame? \opts{turquoise mermaid-design backpack}{boxed Clue board game}{\textbf{boxed My Monopoly board game}}{white spoon}}
		
		\vqaSplitEntry{F\_OCCLUSION\_PERCENTAGE\_OBJECT}
		{View Point}
		{Visibility}
		{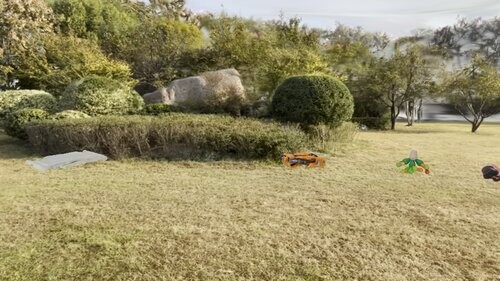}
		{\stripEight{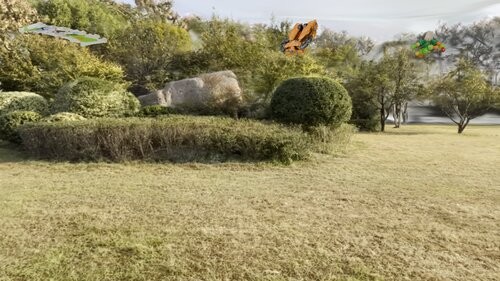}{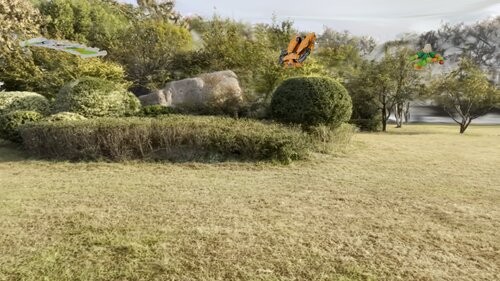}{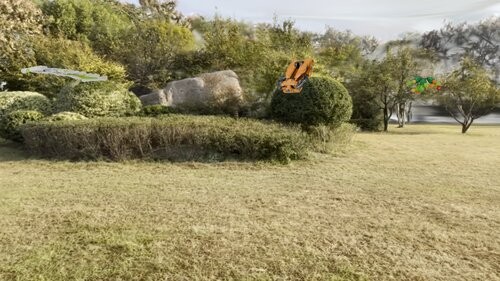}{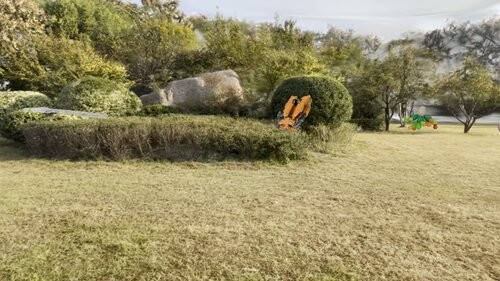}
			{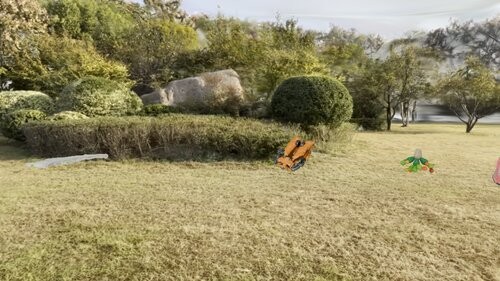}{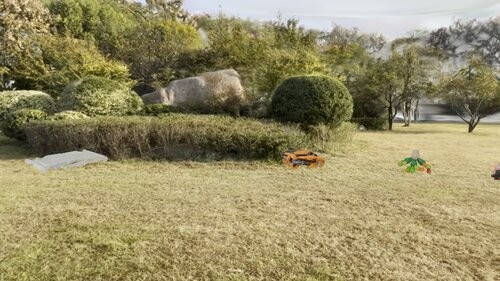}{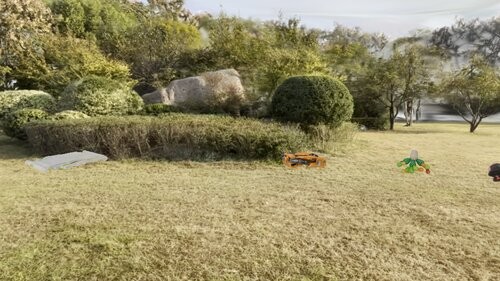}{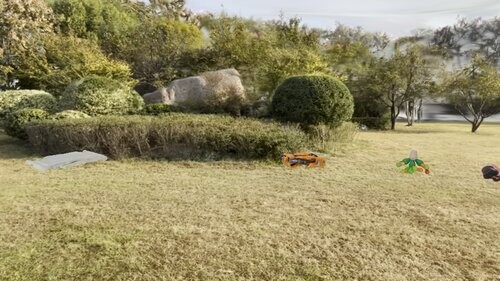}}
		{Based on the image, how much of the "orange Transformers action figure" is occluded? \opts{\textbf{Fully Visible (>95\% visible)}}{Partially Occluded (25-65\% visible)}{Slightly Occluded (65-95\% visible)}{Severely Occluded (0-25\% visible)}}
		{Considering all frames, how much of the "orange Transformers action figure" is occluded in the last frame? \opts{\textbf{Fully Visible (>95\% visible)}}{Partially Occluded (25-65\% visible)}{Slightly Occluded (65-95\% visible)}{Severely Occluded (0-25\% visible)}}
		
		\vqaSplitEntry{F\_VISIBILITY\_OBJECT\_COUNT}
		{View Point}
		{Visibility}
		{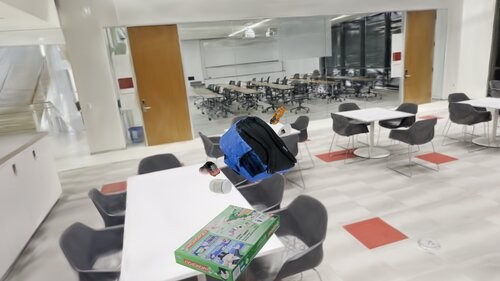}
		{\stripEight{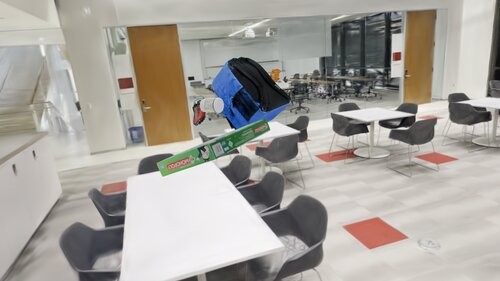}{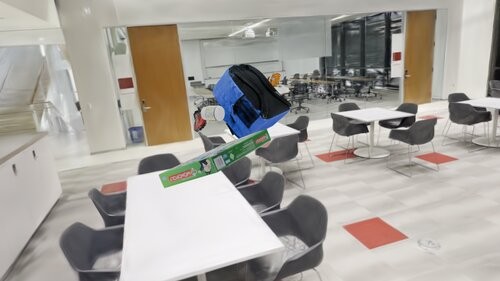}{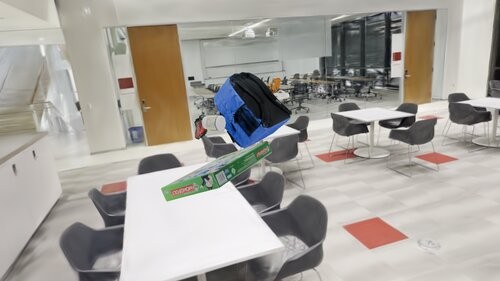}{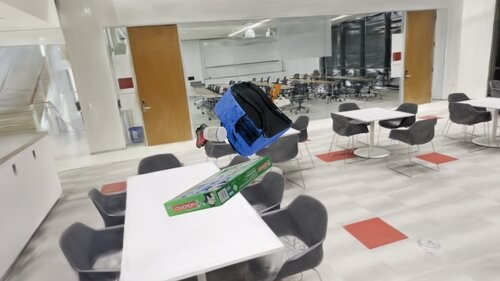}
			{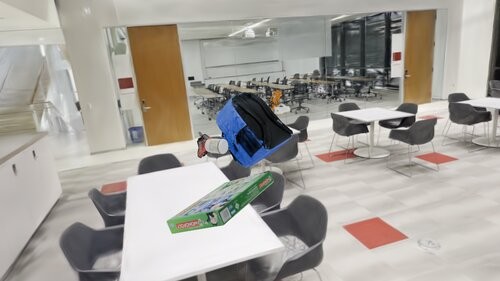}{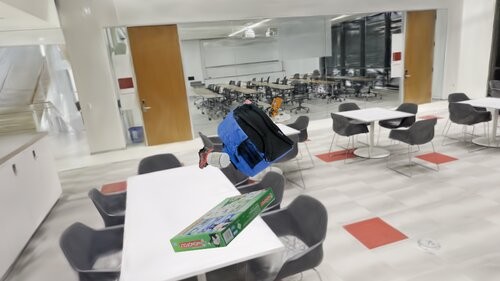}{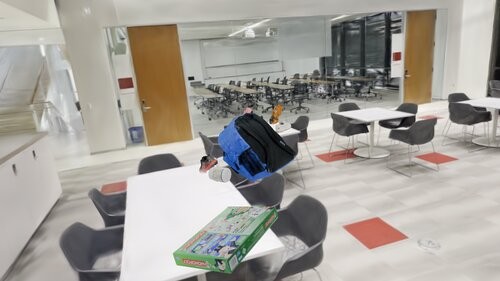}{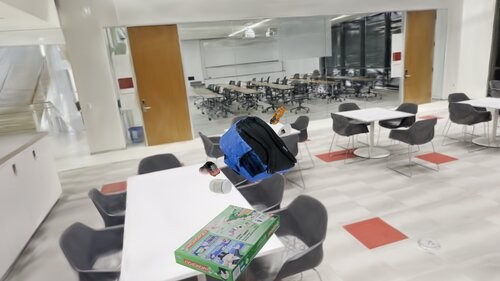}}
		{Based on the image, how many objects are visible? \opts{7}{4}{5}{\textbf{6}}}
		{Considering all frames, how many objects are visible in the last frame? \opts{7}{4}{5}{\textbf{6}}}
		
		\vqaSplitEntry{F\_VIEWPOINT\_CAMERA\_ANGLE}
		{View Point}
		{Camera Characteristics}
		{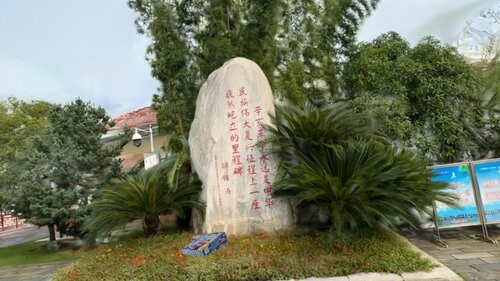}
		{\stripEight{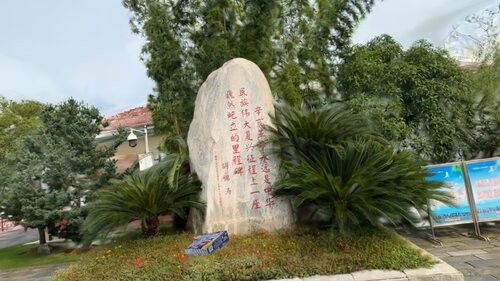}{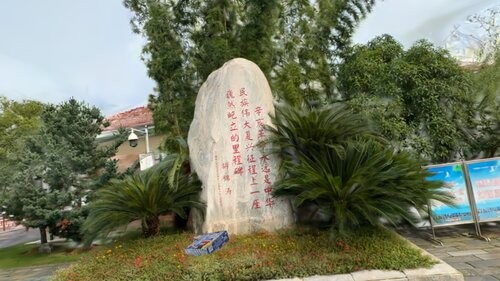}{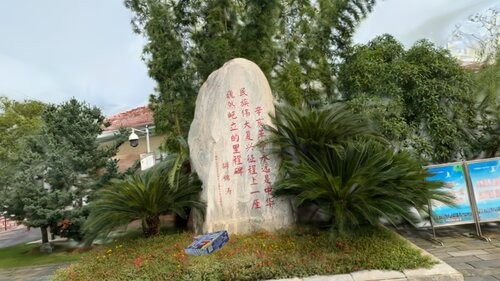}{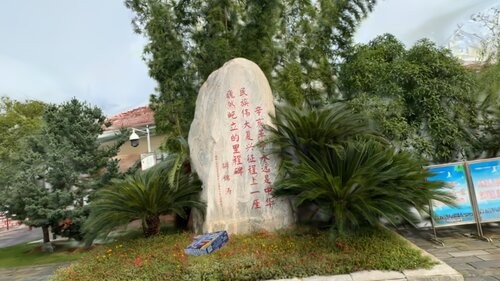}
			{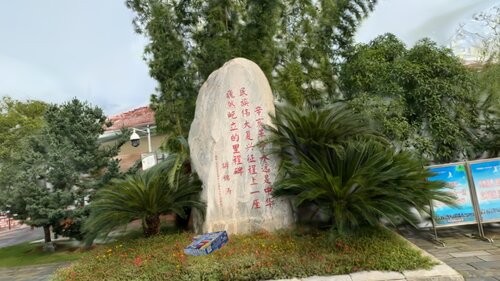}{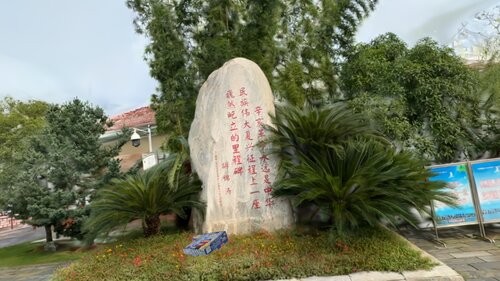}{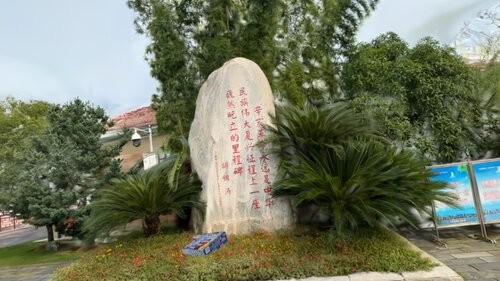}{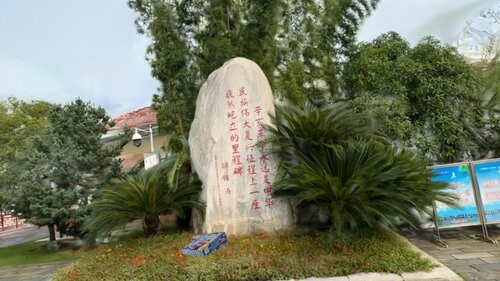}}
		{From the camera's perspective, what is the viewing direction relative to the horizon? \opts{\textbf{eye level (-15 to 15 degrees)}}{high angle (-60 to -15 degrees)}{bird's-eye (<=-60 degrees)}{low angle (15 to 60 degrees)}}
		{Considering all frames, from the camera's perspective, what is the viewing direction relative to the horizon in the last frame? \opts{\textbf{eye level (-15 to 15 degrees)}}{high angle (-60 to -15 degrees)}{bird's-eye (<=-60 degrees)}{low angle (15 to 60 degrees)}}
		
		\vqaSplitEntry{F\_FOCAL\_LENGTH\_CLASS}
		{View Point}
		{Camera Characteristics}
		{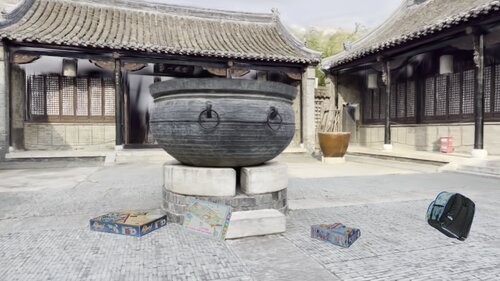}
		{\stripEight{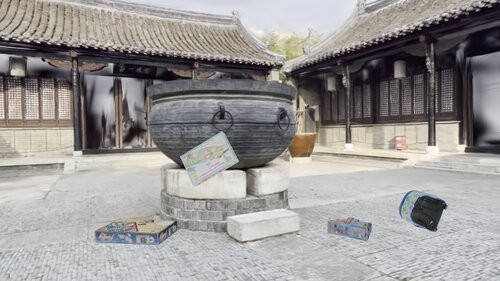}{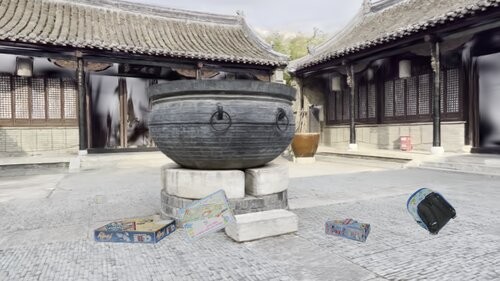}{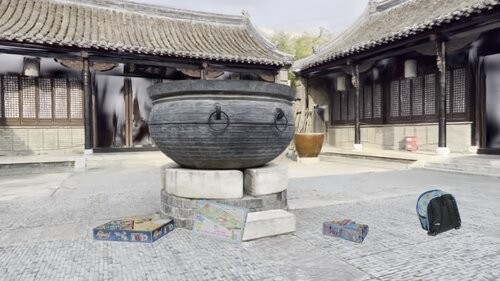}{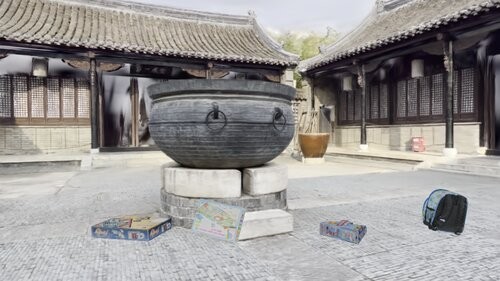}
			{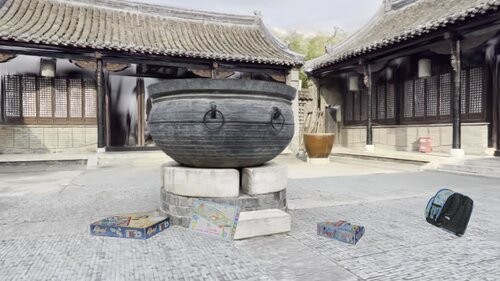}{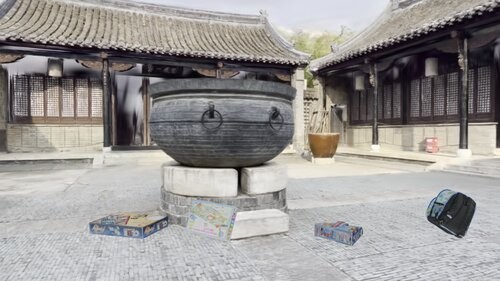}{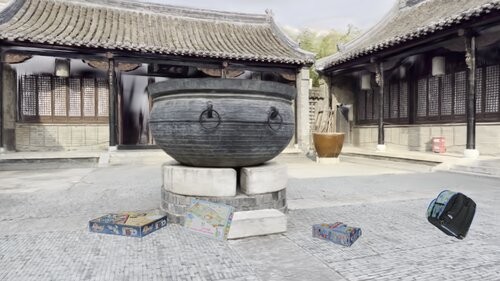}{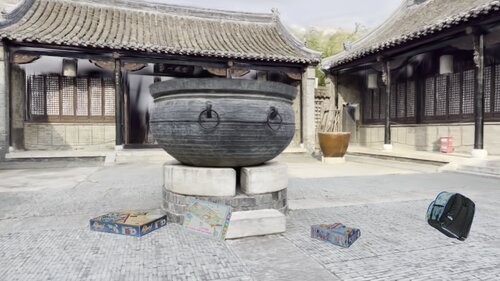}}
		{Based on the image, which focal-length class best matches the perspective observed? \opts{normal (20-60)}{\textbf{wide (60-100)}}{ultra-wide (>=100)}{telephoto (<20)}}
		{Considering all frames, which focal-length class best matches the perspective observed? \opts{normal (20-60)}{\textbf{wide (60-100)}}{ultra-wide (>=100)}{telephoto (<20)}}
		
		\vqaSplitEntry{F\_TEMPORAL\_SEQUENCE\_IMAGES}
		{Temporal}
		{Event Ordering}
		{}
		{\stripFour{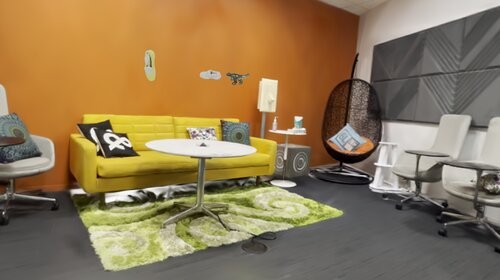}{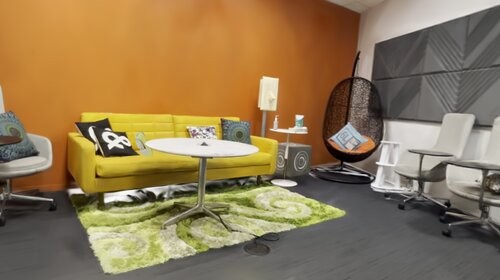}{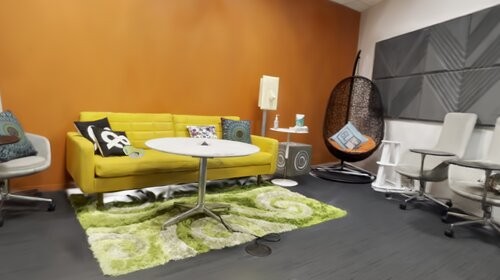}{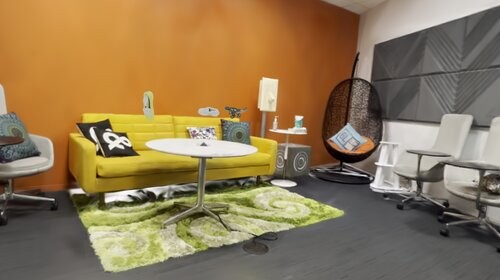}}
		{\textbf{No question provided/possible}}
		{Given the four unordered frames (A, B, C, D) of the same scene, what is the correct temporal ordering of the events? \opts{A-C-B-D}{\textbf{A-D-B-C}}{C-D-B-A}{C-A-B-D}}
		
		\vqaSplitEntry{F\_TEMPORAL\_PREDICTION\_\allowbreak NEXT\_IMAGE\_GRANULARITY\_1}
		{Temporal}
		{Event Ordering}
		{}
		{\stripEight{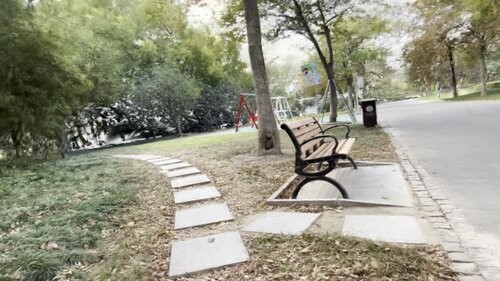}{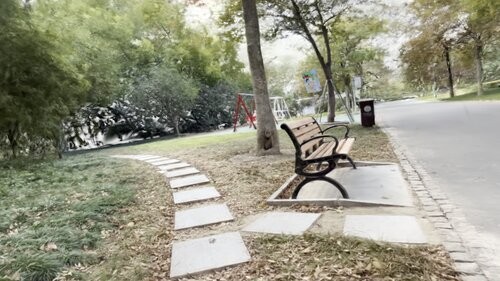}{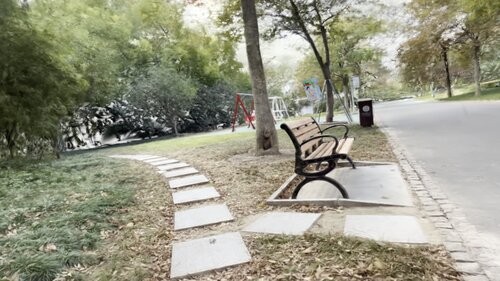}{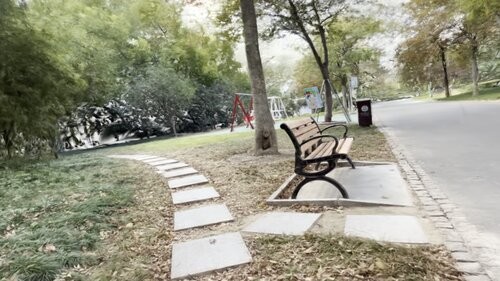}
			{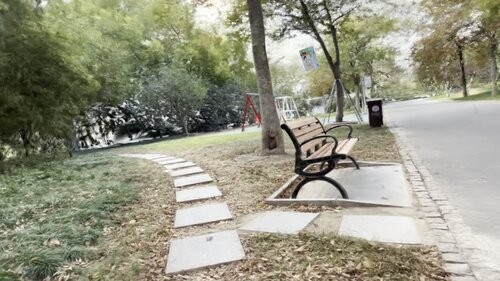}{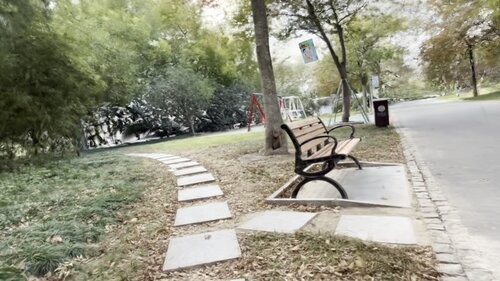}{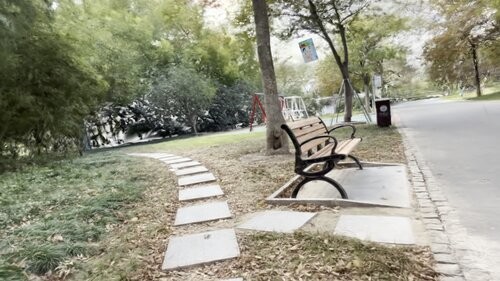}{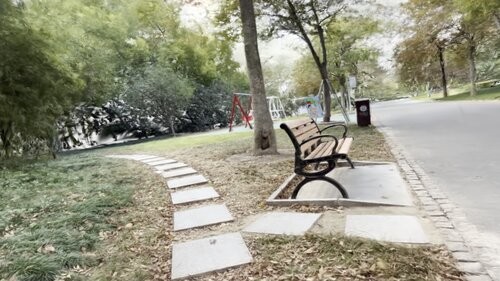}}
		{\textbf{No question provided/possible}}
		{Which individual frame (t5, t6, t7, or t8) is most likely to display an event that occurred after the provided frame sequence (t1, t2, t3, t4)? \opts{t5}{t6}{t7}{\textbf{t8}}}
		
		\vqaSplitEntry{F\_TEMPORAL\_PREDICTION\_\allowbreak NEXT\_IMAGE\_GRANULARITY\_2}
		{Temporal}
		{Event Ordering}
		{}
		{\stripEight{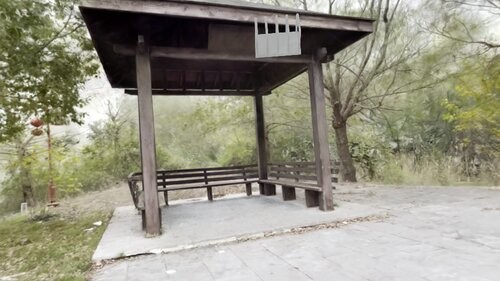}{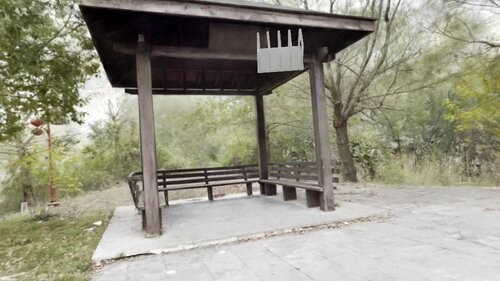}{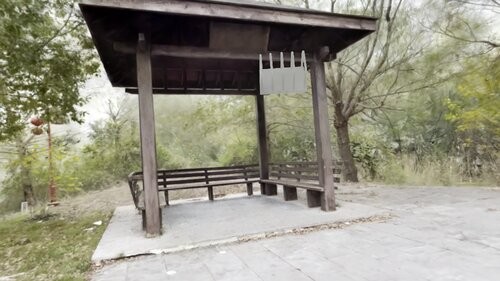}{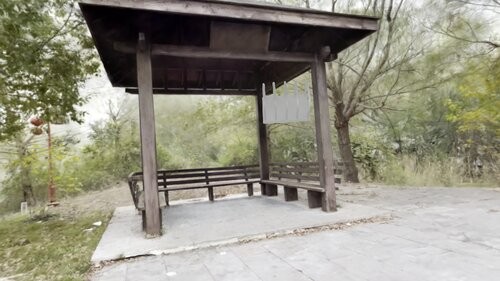}
			{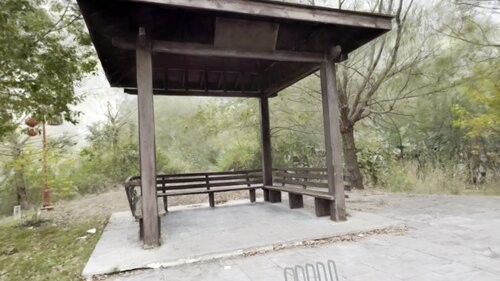}{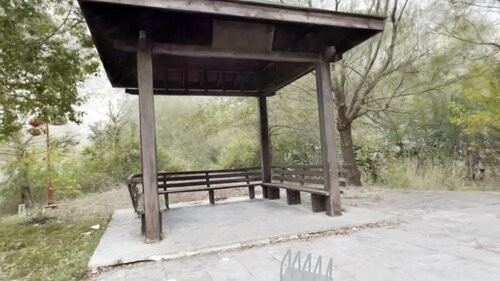}{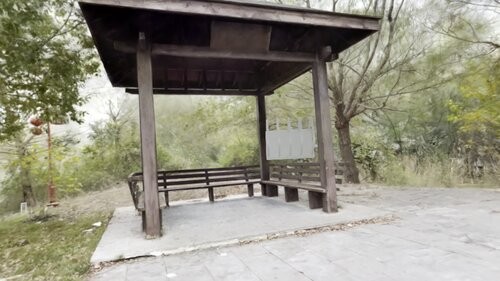}{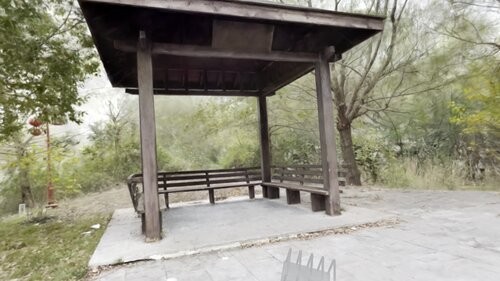}}
		{\textbf{No question provided/possible}}
		{Which individual frame (t5, t6, t7, or t8) is most likely to display an event that occurred after the provided frame sequence (t1, t2, t3, t4)? \opts{t5}{t6}{\textbf{t7}}{t8}}
		
		\vqaSplitEntry{F\_TEMPORAL\_PREDICTION\_\allowbreak NEXT\_IMAGE\_GRANULARITY\_5}
		{Temporal}
		{Event Ordering}
		{}
		{\stripEight{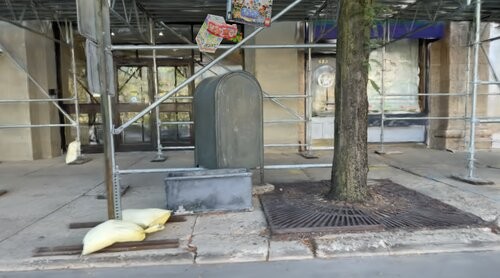}{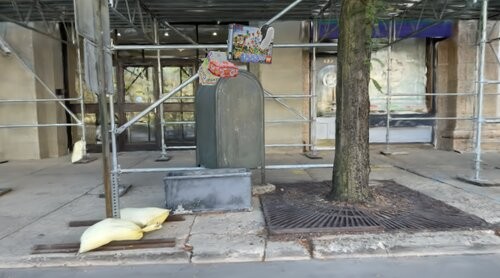}{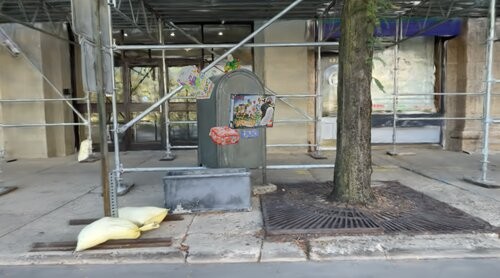}{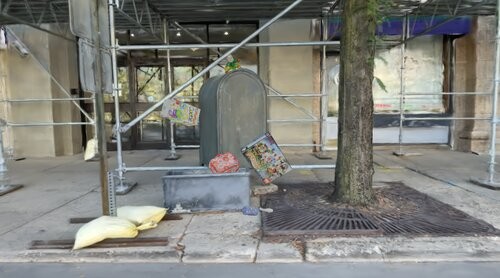}
			{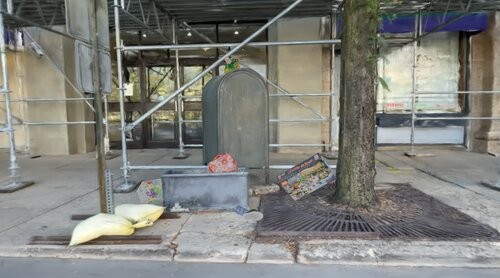}{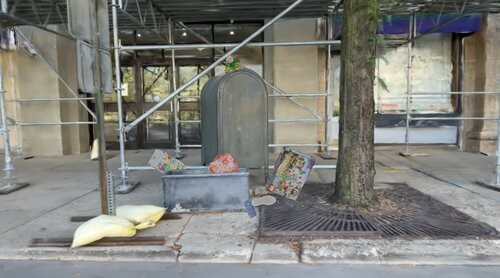}{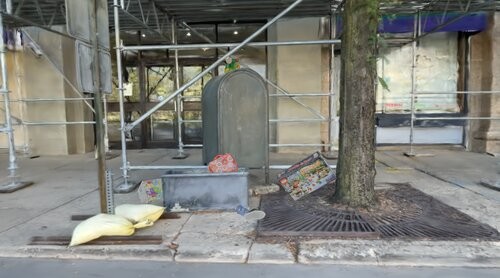}{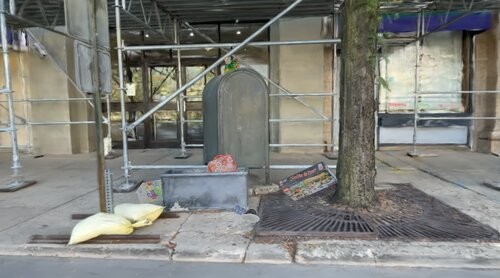}}
		{\textbf{No question provided/possible}}
		{Which individual frame (t5, t6, t7, or t8) is most likely to display an event that occurred after the provided frame sequence (t1, t2, t3, t4)? \opts{t5}{\textbf{t6}}{t7}{t8}}
		
		\vqaSplitEntry{F\_TEMPORAL\_\allowbreak PREDICTION\_PREVIOUS\_IMAGE}
		{Temporal}
		{Event Ordering}
		{}
		{\stripEight{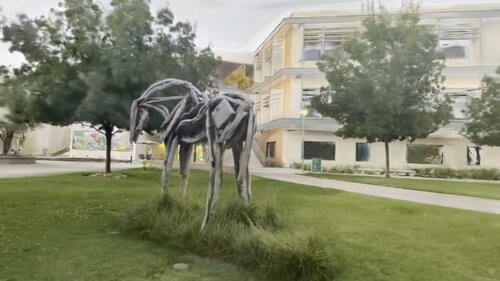}{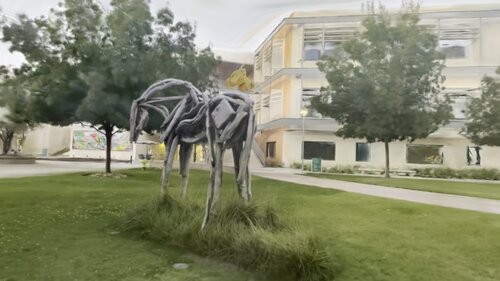}{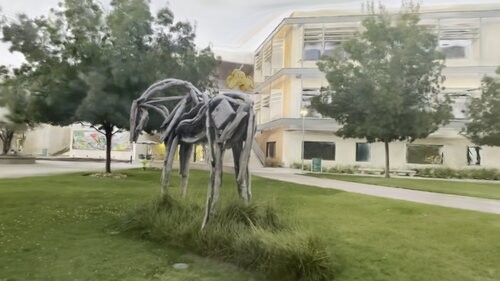}{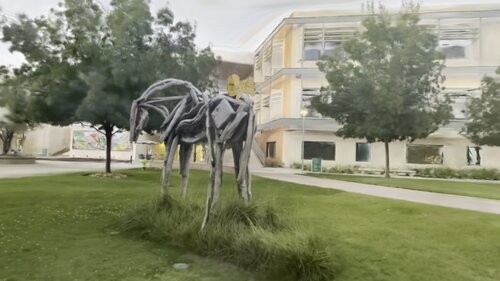}
			{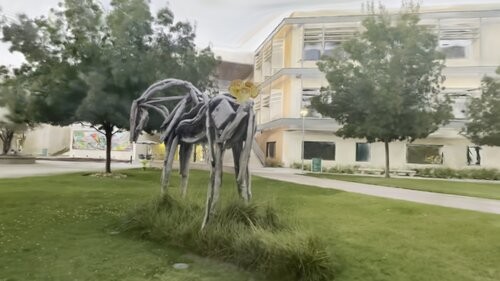}{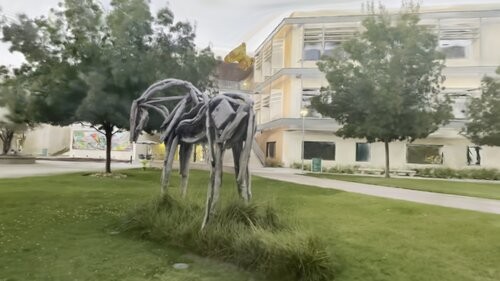}{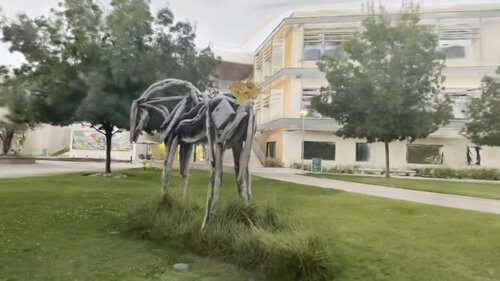}{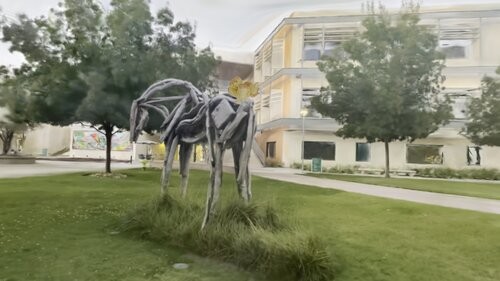}}
		{\textbf{No question provided/possible}}
		{Which individual frame (t5, t6, t7, or t8) is most likely to display an event that occurred before the provided frame sequence (t1, t2, t3, t4)? \opts{t5}{\textbf{t6}}{t7}{t8}}
		
		\vqaSplitEntry{F\_TEMPORAL\_\allowbreak PREDICTION\_MISSING\_IMAGE}
		{Temporal}
		{Event Ordering}
		{}
		{\stripEight{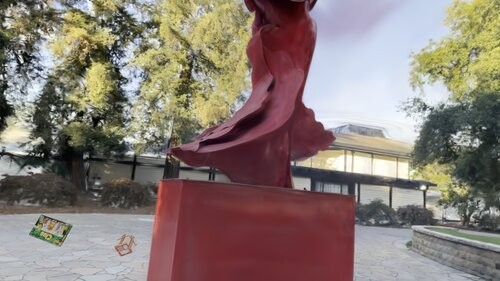}{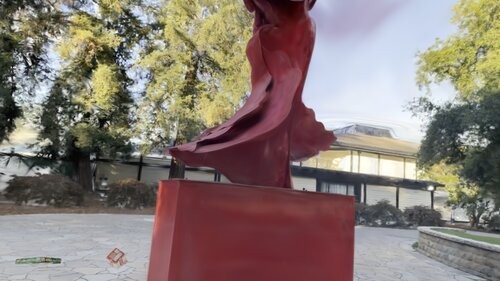}{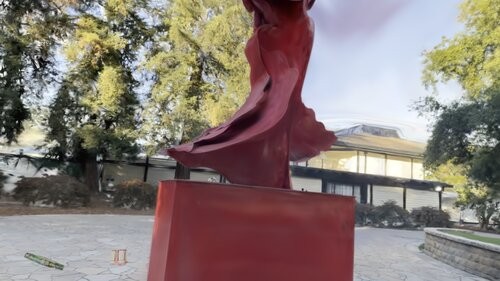}{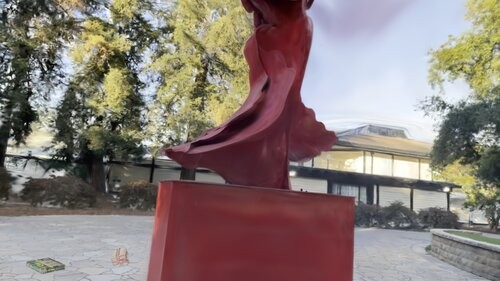}
			{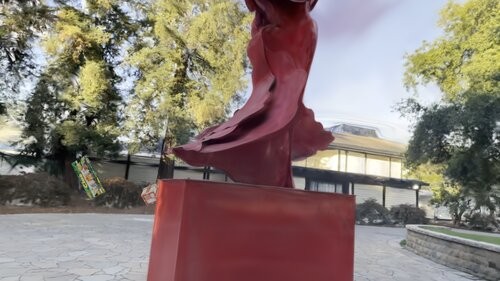}{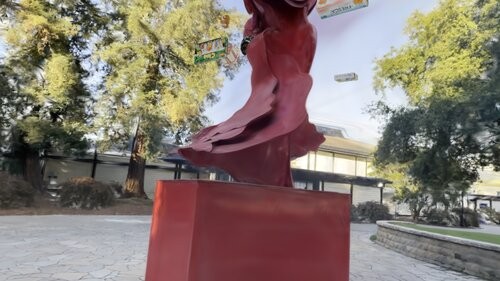}{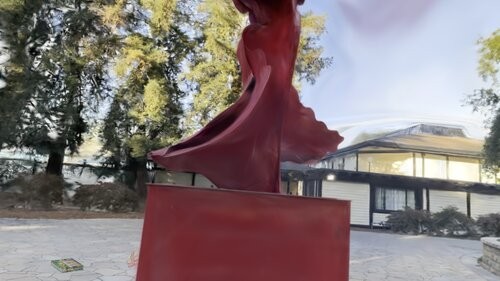}{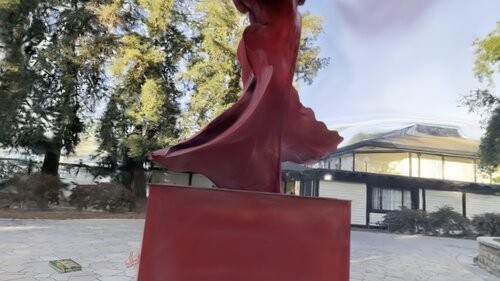}}
		{\textbf{No question provided/possible}}
		{Which individual frame (t5, t6, t7, or t8) is most likely to display an event that occurred during the time span of the provided frame sequence (t1, t2, t3, t4)? \opts{\textbf{t5}}{t6}{t7}{t8}}
		
		\vqaSplitEntry{F\_CAMERA\_MOTION\_DIRECTION}
		{Temporal}
		{Camera Motion}
		{}
		{\stripEight{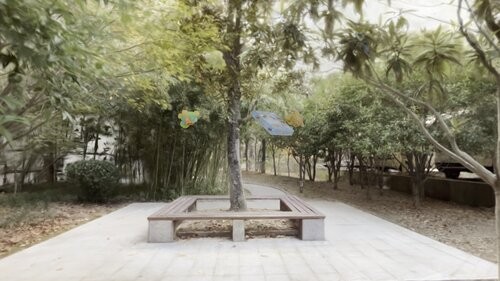}{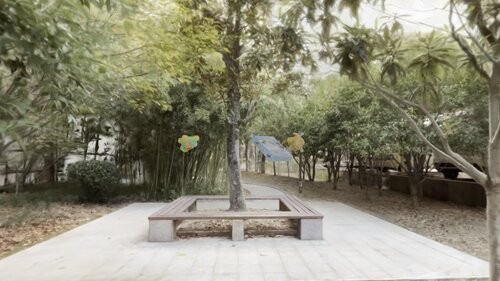}{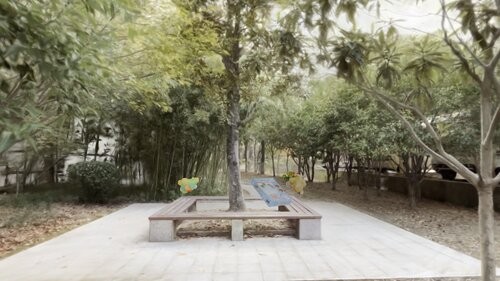}{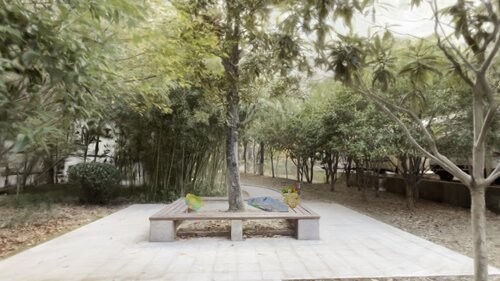}
			{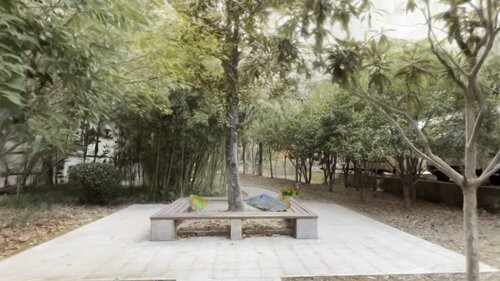}{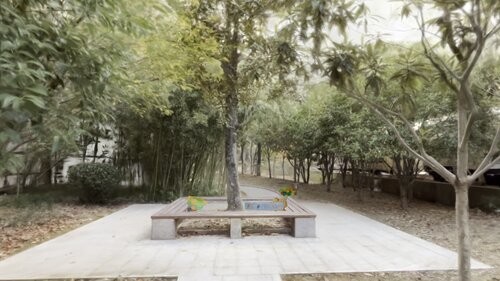}{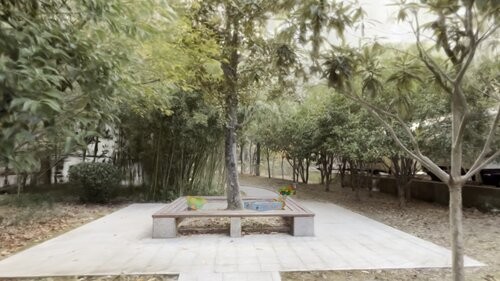}{folder_25089_g/images/image_07.jpg}}
		{\textbf{No question provided/possible}}
		{Across the frame sequence, what is the predominant direction of the camera's motion? \opts{left then up}{backward then down}{\textbf{backward}}{up then down}}
		
		\vqaSplitEntry{F\_CAMERA\_ZOOM\_BEHAVIOR}
		{Temporal}
		{Camera Motion}
		{}
		{\stripEight{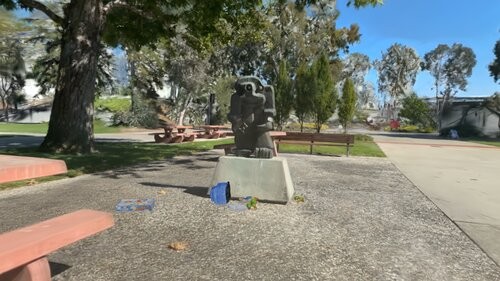}{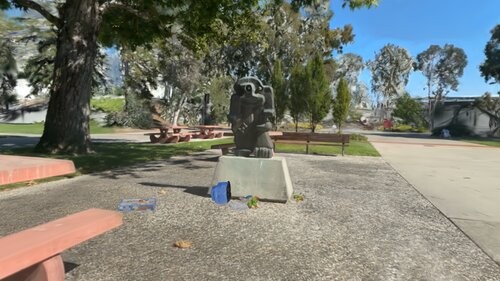}{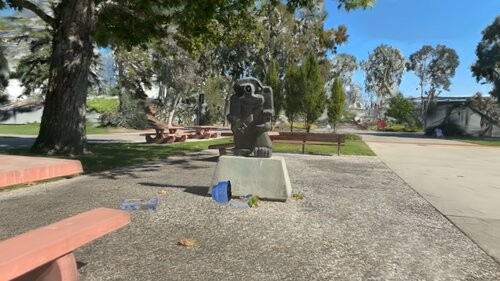}{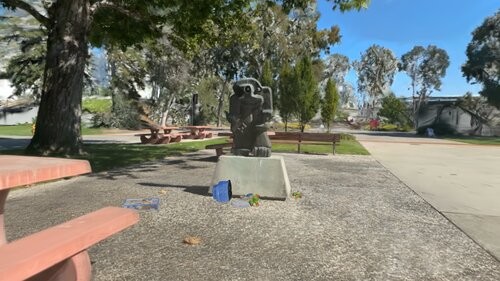}
			{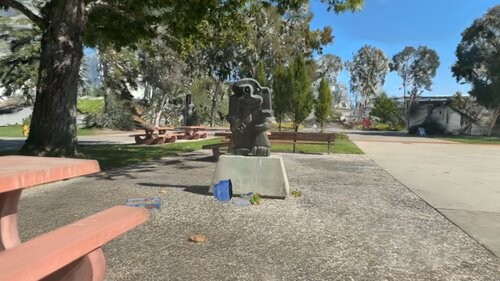}{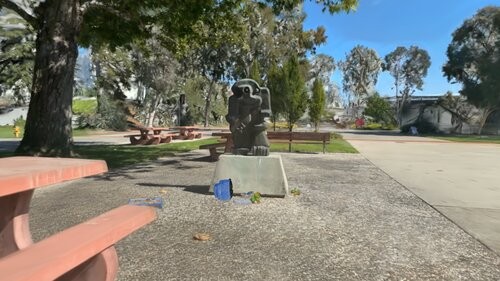}{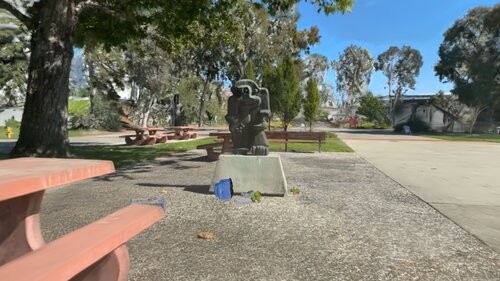}{folder_66191_g/images/image_07.jpg}}
		{\textbf{No question provided/possible}}
		{Across the frame sequence, how does the camera's zoom level change? \opts{\textbf{no zoom}}{zoom out}{zoom out then in}{zoom in then out}}
		
		\vqaSplitEntry{F\_PERSISTENCE\_OBJECT\_PRESENT}
		{Persistence}
		{Object Persistence}
		{}
		{\stripEight{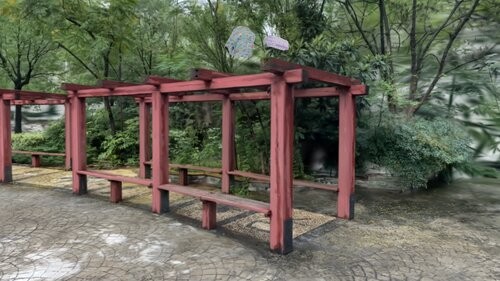}{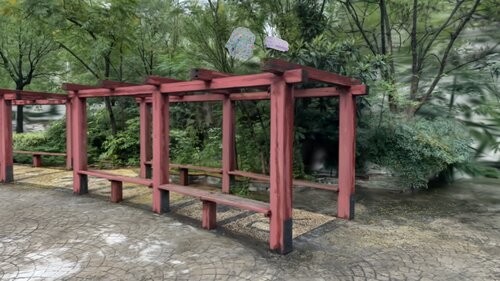}{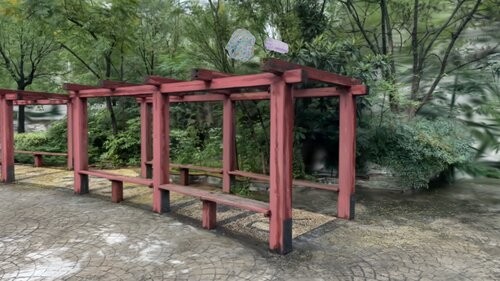}{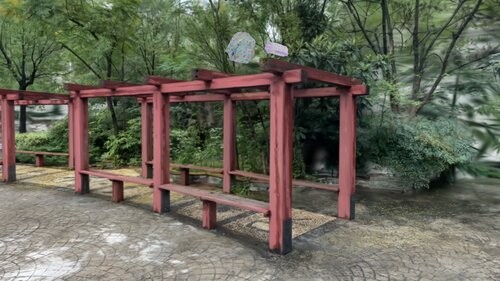}
			{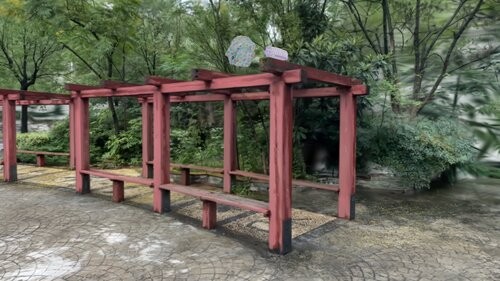}{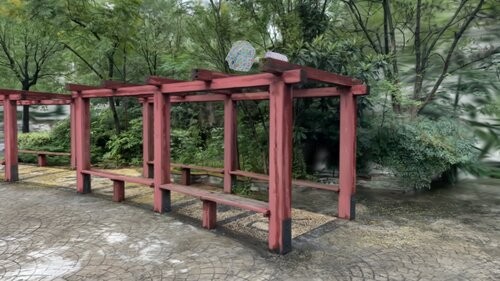}{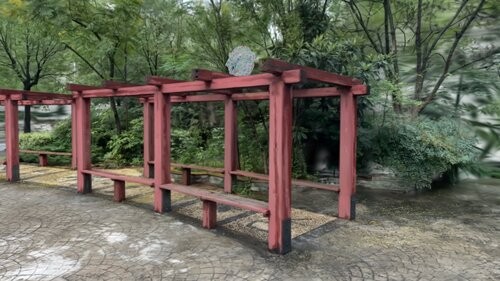}{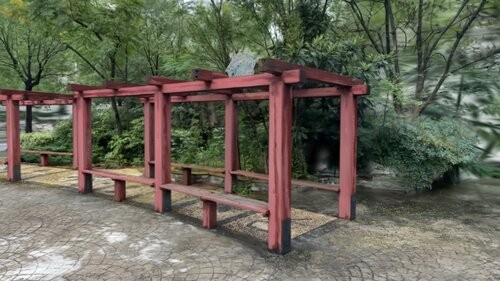}}
		{\textbf{No question provided/possible}}
		{Considering all frames, which object was visible but disappeared in the last frame? \opts{\textbf{white-striped purple pencil case}}{Teenage Mutant Ninja Turtles action figurine}{white soccer shoe}{small blue painted bucket}}
		
		\vqaSplitEntry{F\_PERSISTENCE\_OBJECT\_DISAPPEAR}
		{Persistence}
		{Object Persistence}
		{}
		{\stripEight{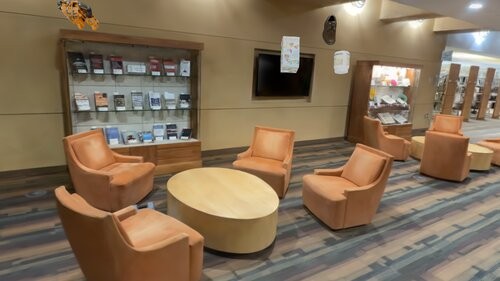}{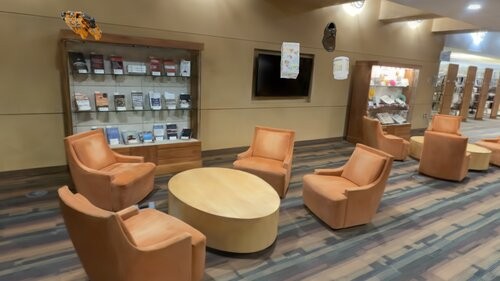}{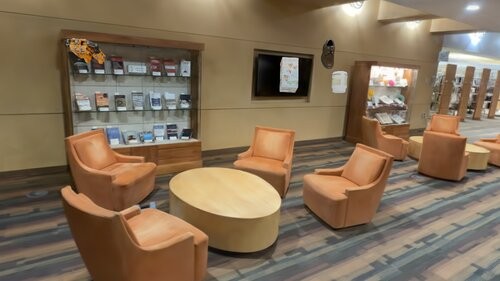}{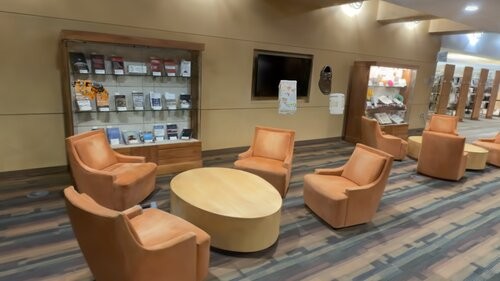}
			{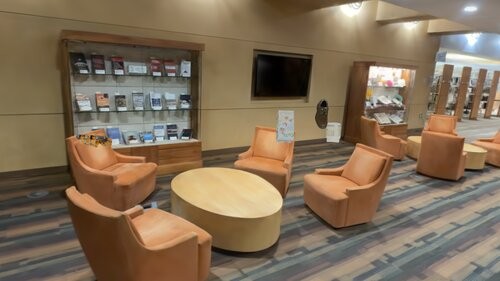}{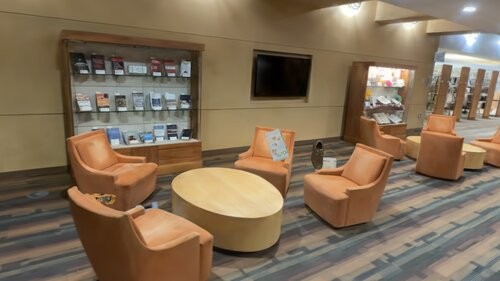}{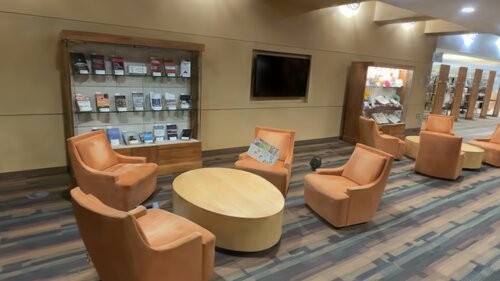}{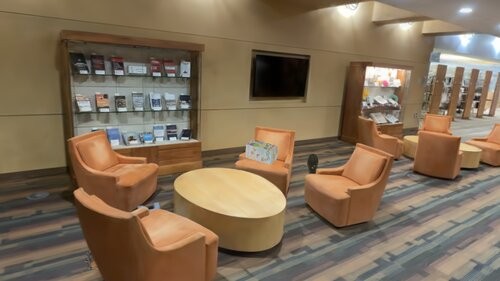}}
		{\textbf{No question provided/possible}}
		{Considering all frames, which visible object disappeared and does not reappear in the last frame? \opts{small blue painted bucket}{\textbf{white jar of proteins}}{Teenage Mutant Ninja Turtles action figurine}{white soccer shoe}}
		
		\vqaSplitEntry{F\_PERSISTENCE\_\allowbreak OBJECT\_TOTAL\_COUNT}
		{Persistence}
		{Object Persistence}
		{}
		{\stripEight{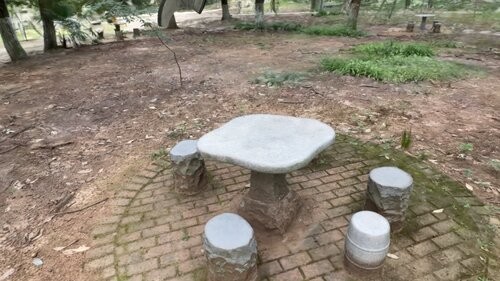}{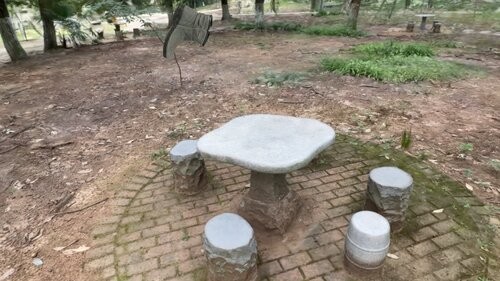}{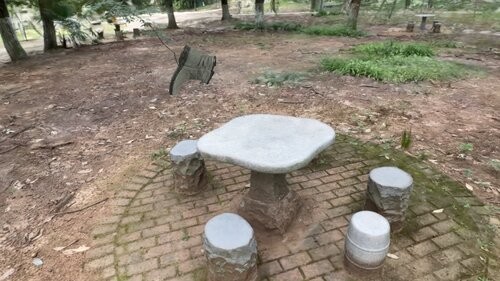}{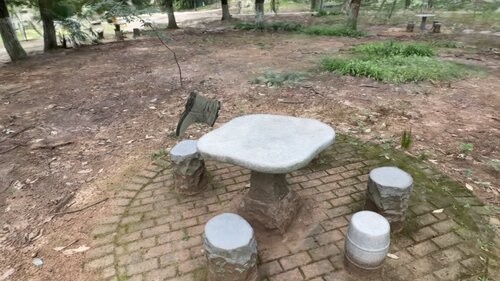}
			{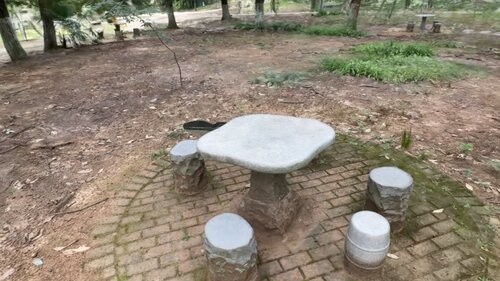}{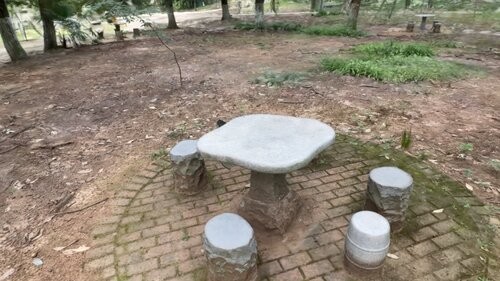}{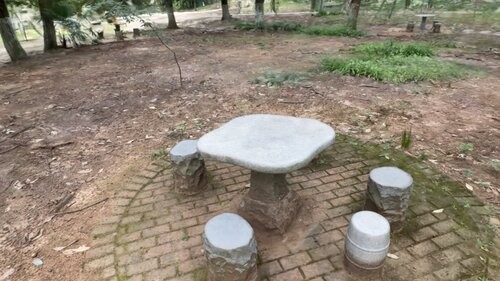}{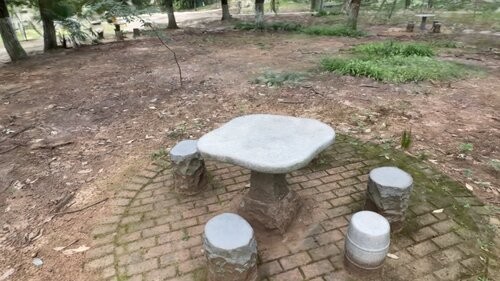}}
		{\textbf{No question provided/possible}}
		{Considering all frames, how many objects are present at the time of the last frame, including those currently hidden or out of frame? \opts{0}{2}{\textbf{1}}{3}}
		
		\vqaSplitEntry{F\_PERSISTENCE\_OBJECT\_\allowbreak TOTAL\_COUNT\_HIDDEN}
		{Persistence}
		{Object Persistence}
		{}
		{\stripEight{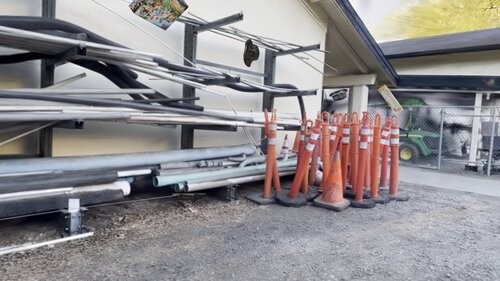}{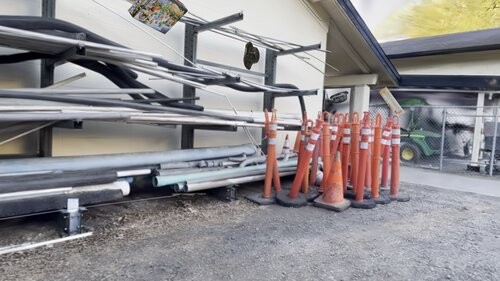}{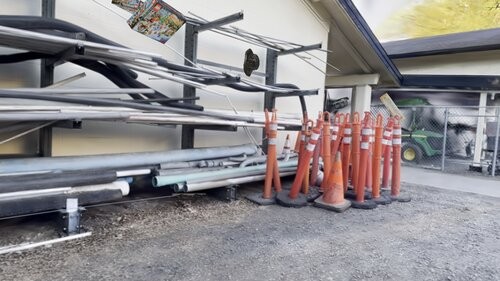}{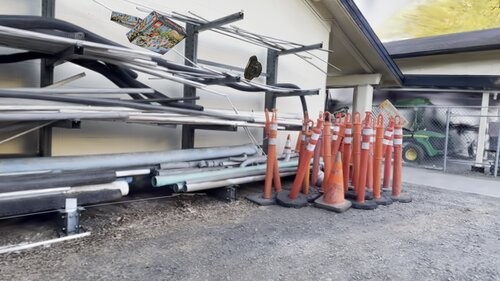}
			{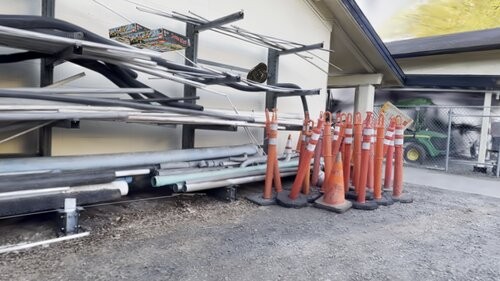}{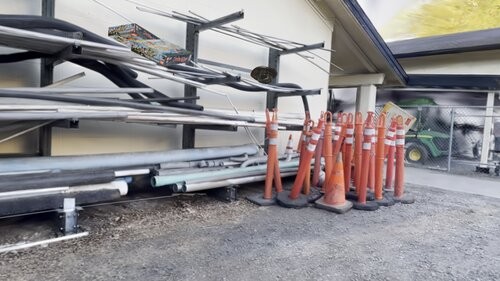}{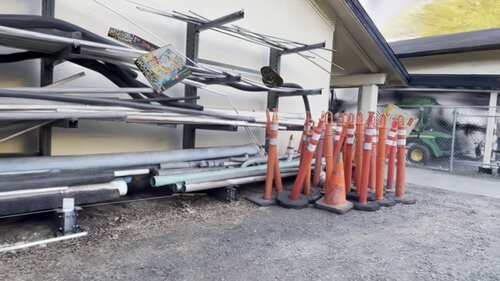}{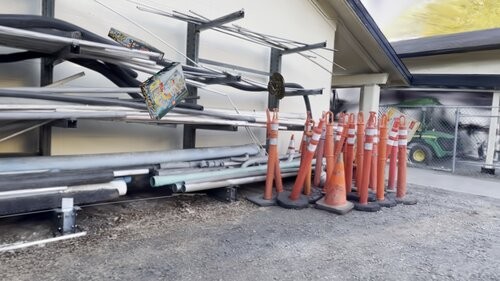}}
		{\textbf{No question provided/possible}}
		{Considering all frames, how many objects are present at the time of the last frame, but not visible? \opts{2}{0}{\textbf{1}}{3}}
		
	\end{longtable}
	}
\endgroup
\twocolumn

\clearpage
\subsection{VQA automation}
\label{app:details_vqa_creation}

The construction of the VQA dataset is a non-trivial process that transforms the templates described in Section~\ref{app:question_taxonomy} into the final set of questions illustrated in Section~\ref{app:vqa_examples}.

The core idea is that each \texttt{question\_id} corresponds to a dedicated function written in python. This function receives as input the simulation data describing the world state, together with optional \texttt{<OBJECT>} parameters required by some templates. The function first checks whether the question is answerable within the given simulation. If the required conditions are satisfied, the correct answer is computed, the corresponding visual evidence is extracted from the simulated frames (either single-frame or multi-frame), and the full metadata associated with the example is stored.

\paragraph{Confounding answer generation.}
In addition to the correct answer, a set of confounding answers is generated in order to construct the final multiple-choice question. The generation strategy depends on the type of answer:

\begin{itemize}
	\item \textbf{Numerical answers.} Confounders are sampled around the correct value. The sampled values are distributed approximately symmetrically around the ground truth with equal spacing, where the spacing varies across questions to avoid predictable patterns. This produces plausible alternatives that remain close to the correct answer while maintaining sufficient separation.
	
	\item \textbf{Object-based answers.} When the answer corresponds to an object category or identity, confounders are sampled from objects that are visible in the scene. If no suitable candidates are available, sampling is performed from the global set of objects present in the dataset.
\end{itemize}

\paragraph{Object visibility criteria.}
A careful definition of object visibility is required, since some objects may appear in the frame with only a few pixels and would not be perceptible to a human observer. An object is therefore considered visible only if it satisfies the following conditions:

\begin{itemize}
	\item the projected area in the image exceeds a minimum threshold of \textbf{2000 pixels};
	\item at least \textbf{31\%} of the object is visible within the image frame;
	d\end{itemize}

These constraints ensure that objects used for both correct answers and confounding answers are visually meaningful.

\paragraph{Frame sampling.}
For single-image questions, a single frame is extracted from the simulation. For multi-frame questions, up to \textbf{eight frames} are sampled. Frames are selected at approximately uniform temporal intervals while preserving short temporal gaps between consecutive frames in order to maintain motion consistency. In practice, the separation between sampled frames ranges between \textbf{one and four frames}, depending on the dynamics of the simulation.

\paragraph{Determinism and parallelization.}
The entire generation pipeline is deterministic: all random operations are controlled through explicit seeding to guarantee reproducibility. Since each simulation and each question type can be processed independently, the pipeline is highly parallelizable and is implemented as a threaded program where questions are generated concurrently across simulations.

\paragraph{Ambiguity filtering.}
During the generation process, certain candidate questions may result in ambiguous or unverifiable answers. This can occur, for example, when multiple answers would be valid, when the relevant information is not sufficiently visible in the scene, or when the simulation state does not provide enough evidence to determine a unique answer. In such cases, the corresponding question is automatically flagged as \emph{unanswerable}. All questions marked in this way are discarded during dataset construction and are therefore not included in the final evaluation set. This filtering step ensures that each retained question has a well-defined and uniquely verifiable ground-truth answer.

\section{Expert-to-novice specification}
\label{app:five_levels}

This section details the expert-to-novice analysis conducted in Sec.~4.4 of the main paper. 

We outline the queries utilized to test model adaptability across different domain expertise. We focus on a carefully curated set of seven physics-based questions. To systematically evaluate how different models handle varying degrees of complexity, each of the seven original questions is expanded into five distinct variations: child, teenager, undergraduate, graduate, and expert. These levels were carefully selected to represent diverse comprehension levels.

\newcommand{\suppquestion}[7]{%
	\rotatebox[origin=c]{90}{\centering \scriptsize \ttfamily #1} & 
	\textbf{\method{} (baseline):} #2 \vspace{0.4em}\newline
	\textbf{Child:} #3 \vspace{0.4em}\newline
	\textbf{Teen:} #4 \vspace{0.4em}\newline
	\textbf{Undergrad:} #5 \vspace{0.4em}\newline
	\textbf{Graduate:} #6 \vspace{0.4em}\newline
	\textbf{Expert:} #7 \\ 
	\hline %
}

\onecolumn
{\begin{longtable}{| c | m{0.9\textwidth} |}
	\hline
	\textbf{ID} & \textbf{Questions} \\
	\hline
	\endfirsthead
	
	\hline
	\textbf{ID} & \textbf{Questions} \\
	\hline
	\endhead
	
	\hline
	\endlastfoot

	\suppquestion{F\_PHYSICS\_PROPERTY\_DENSITY\_OBJECT\_RELATIVE}
	{Considering all frames, which object, visible in the last frame, has the highest effective density?}
	{Looking at all the pictures, if all the visible objects had the same size which one would be the heaviest?}
	{Looking at all the pictures, which object visible in the last frame do you think is the heaviest compared to how big it looks?}
	{Considering the full image sequence, which object visible in the last frame would have the highest mass relative to its volume?}
	{Based on the entire frame sequence, which object visible in the final frame is likely to have the highest effective density (mass per unit volume)?}
	{Considering the entire temporal frame sequence, which visible object visible in the final frame has the highest effective density?}
	
	\suppquestion{F\_PHYSICS\_PROPERTY\_YOUNG\_MODULUS\_OBJECT\_SIMILAR}
	{Considering all frames, which object, visible in the last frame, has a Young's Modulus most similar to that of the <OBJECT>?}
	{Looking at all the pictures, which object you can still see at the end is just as hard to bend or squishe as the <OBJECT>?}
	{Looking at all the pictures, which object visible in the last frame would feel about as stiff or stretchy as the <OBJECT> if you tried to bend it?}
	{Considering the full image sequence, which object visible in the last frame appearas to have a similar stiffness (Young’s modulus) to the <OBJECT>?}
	{Based on the entire frame sequence, which object visible in the final frame would you expect to have a Young’s modulus closest to that of the <OBJECT>?}
	{Considering the entire temporal frame sequence, which visible object visible in the final frame would you infer to have a Young’s modulus most similar to that of the <OBJECT>?}
	
	\suppquestion{F\_PHYSICS\_PROPERTY\_YOUNG\_MODULUS\_OBJECT\_SIMILAR\_NON\_TECHNICAL}
	{Considering all frames, which object, visible in the last frame, has a softness most similar to that of the <OBJECT>?}
	{Looking at all the pictures, which object you can still see at the end feels squishy or hard in the same way as the <OBJECT>?}
	{Looking at all the pictures, which object visible in the last frame would feel about as soft or hard as the <OBJECT>?}
	{Considering the full image sequence, which object visible in the last frame appears to have a similar softness to the <OBJECT>?}
	{Based on the entire frame sequence, which object visible in the final frame would you expect to have a softness most similar to that of the <OBJECT>?}
	{Considering the entire temporal frame sequence, which object visible in the final frame would you infer to have a comparable effective softness to the <OBJECT>, based on its deformation or response to interaction?}
	
	\suppquestion{F\_PHYSICS\_PROPERTY\_YOUNG\_MODULUS\_HIGHEST}
	{Considering all frames, which object, visible in the last frame, exhibits the highest Young's Modulus?}
	{Looking at all the pictures, which object you can still see at the end is the hardest and least bendy?}
	{After watching all the frames, which object visible in the last frame seems the stiffest or hardest to bend?}
	{Considering the full image sequence, which object visible in the last frame appears to be the stiffest?}
	{Based on the entire frame sequence, which object visible in the final frame would you expect to have the highest Young’s modulus?}
	{Considering the entire temporal frame sequence, which object visible in the final frame would you infer to exhibit the highest Young’s modulus, assuming linear elastic behavior under comparable loading?}
	
	\suppquestion{F\_PHYSICS\_PROPERTY\_POISSON\_RATIO\_OBJECT\_SIMILAR}
	{Considering all frames, which visible object has a Poisson ratio most similar to that of the <OBJECT> visible in the last frame?}
	{Looking at all the pictures, which object you can still see at the end is as elastic as the <OBJECT> you see at the end?}
	{After watching all the frames, which object visible in the last frame changes its width in a similar way to the <OBJECT> in the last frame when it is squeezed or stretched?}
	{Considering the full image sequence, which object visible in the last frame shows a similar width change under stretching or compression as the <OBJECT> in the last frame?}
	{Based on the entire frame sequence, which object visible in the final frame would you expect to have a Poisson ratio closest to that of the <OBJECT> visible in the final frame?}
	{Considering the entire temporal frame sequence, which object visible in the final frame would you infer to have a Poisson ratio most similar to that of the <OBJECT> visible in the final frame, based on comparable transverse-to-axial strain behavior?}
	
	\suppquestion{F\_PHYSICS\_PROPERTY\_POISSON\_RATIO\_HIGHEST}
	{Considering all frames, which object, visible in the last frame, exhibits the largest Poisson ratio?}
	{Looking at all the pictures, which object you can still see at the end spreads out the most on the sides when it is squished?}
	{After watching all the frames, which object visible in the last frame gets the widest on the sides when it is squeezed?}
	{Considering the full image sequence, which object visible in the last frame shows the greatest sideways expansion when compressed?}
	{Based on the entire frame sequence, which object visible in the final frame would you expect to have the largest Poisson ratio?}
	{Considering the entire temporal frame sequence, which object visible in the final frame would you infer to exhibit the largest Poisson ratio, based on maximal transverse strain relative to axial strain?}
	
	\suppquestion{F\_PHYSICS\_PROPERTY\_POISSON\_HIGH\_LEVEL}
	{Considering all frames, if the <OBJECT>, visible in the last frame, were compressed vertically, how would its horizontal dimensions change?}
	{Looking at all the pictures, if you squish the <OBJECT> you see at the end from top to bottom, how would it change?}
	{After watching all the frames, if the <OBJECT> in the last frame were pressed down, how would its width change?}
	{Considering the full image sequence, if the <OBJECT> visible in the last frame were compressed vertically, how would its horizontal size change?}
	{Based on the entire frame sequence, if the <OBJECT> visible in the final frame were subjected to vertical compression, how would its horizontal dimensions be expected to respond?}
	{Given the full temporal sequence, if the <OBJECT> visible in the final frame were compressed along the vertical axis, how would its transverse (horizontal) dimensions change as dictated by its Poisson response?}
\end{longtable}}
\twocolumn

\onecolumn
\normalsize
\section{Models specification}
\label{app:modelsspec}

\subsection{Vision-Language Models}
\label{app:models}
\begin{figure*}[h]
	\centering
	\includegraphics[width=1.0\linewidth]{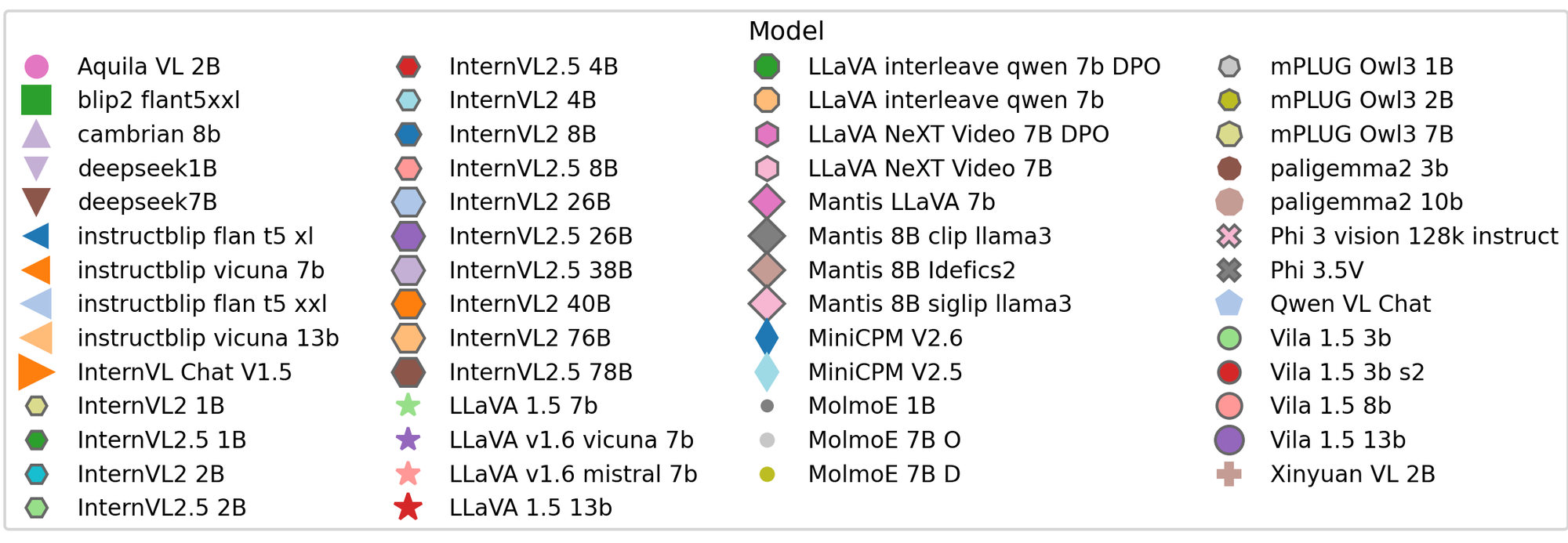}
	\caption{\textbf{Open-source VLM marker listing.} We report here the complete legend of the VLMs markers used in the main paper.}
	\label{fig:vlmmarkers}
\end{figure*}

\cref{fig:vlmmarkers} shows all the markers used in main paper's plots and their corresponding models.

\cref{tab:model_metadata} also provide a detail listing of the 54 open-source VLMs used, spanning across 18 families spanning from single-image models (\eg, cambrian-8b) to multi-image models (\eg, vila-1.5-3b).

\newcommand{\modelmetaRow}[9]{%
	#1 & #2 & #3 & #4 & #5 & #6 & #7 \\ \hline
}

{
\scriptsize
\setlength{\tabcolsep}{2pt}
\renewcommand{\arraystretch}{1.0}
\begin{longtable}{|r|>{\raggedright\arraybackslash}p{0.25\textwidth}|l|l|r|c|>{\raggedright\arraybackslash}p{0.2\textwidth}|}
	\caption{Model metadata.}\label{tab:model_metadata}\\
	\hline
	\textbf{\#} & \textbf{Model ID} & \textbf{Family} & \textbf{type} & \textbf{Params (B)} & \textbf{Year} & \textbf{Licence}\\ \hline
	\endfirsthead
	\hline
	\textbf{\#} & \textbf{Model ID} & \textbf{Family} & \textbf{type} & \textbf{Params (B)} & \textbf{Year} & \textbf{Licence}\\ \hline
	\endhead
	\hline
	\multicolumn{7}{r}{\textit{Continued on next page}} \\
	\endfoot
	\hline
	\endlastfoot
	\modelmetaRow{1}{instructblip-flan-t5-xl}{InstructBlip}{image-only}{4.023}{2023}{mit}{open\_weights}{Salesforce/instructblip-flan-t5-xl}
	\modelmetaRow{2}{instructblip-flan-t5-xxl}{InstructBlip}{image-only}{12.310}{2023}{mit}{open\_weights}{Salesforce/instructblip-flan-t5-xxl}
	\modelmetaRow{3}{instructblip-vicuna-7b}{InstructBlip}{image-only}{7.914}{2023}{other}{open\_weights}{Salesforce/instructblip-vicuna-7b}
	\modelmetaRow{4}{instructblip-vicuna-13b}{InstructBlip}{image-only}{14.192}{2023}{other}{open\_weights}{Salesforce/instructblip-vicuna-13b}
	\modelmetaRow{5}{blip2-flant5xxl}{BLIP2}{image-only}{12.230}{2023}{mit}{open\_weights}{Salesforce/blip2-flan-t5-xxl}
	\modelmetaRow{6}{llava-1.5-7b-hf}{LLaVA}{image-only}{7.063}{2023}{llama2}{open\_weights}{llava-hf/llava-1.5-7b-hf}
	\modelmetaRow{7}{llava-1.5-13b-hf}{LLaVA}{image-only}{13.351}{2023}{llama2}{open\_weights}{llava-hf/llava-1.5-13b-hf}
	\modelmetaRow{8}{llava-v1.6-mistral-7b-hf}{LLaVA}{image-only}{7.567}{2024}{apache-2.0}{open\_weights}{llava-hf/llava-v1.6-mistral-7b-hf}
	\modelmetaRow{9}{llava-v1.6-vicuna-7b-hf}{LLaVA}{image-only}{7.063}{2024}{llama2}{open\_weights}{llava-hf/llava-v1.6-vicuna-7b-hf}
	\modelmetaRow{10}{deepseek1B}{DeepSeekVL}{image-only}{1.975}{2024}{other}{open\_weights}{deepseek-ai/deepseek-vl-1.3b-chat}
	\modelmetaRow{11}{deepseek7B}{DeepSeekVL}{image-only}{7.344}{2024}{other}{open\_weights}{deepseek-ai/deepseek-vl-7b-chat}
	\modelmetaRow{12}{Xinyuan-VL-2B}{XinyuanVL}{image-only}{2.209}{2024}{apache-2.0}{open\_weights}{Cylingo/Xinyuan-VL-2B}
	\modelmetaRow{13}{Aquila-VL-2B}{AquilaVL}{image-only}{2.179}{2024}{apache-2.0}{open\_weights}{BAAI/Aquila-VL-2B-llava-qwen}
	\modelmetaRow{14}{Phi-3-vision-128k-instruct}{Phi}{general}{4.147}{2024}{mit}{open\_weights}{microsoft/Phi-3-vision-128k-instruct}
	\modelmetaRow{15}{Phi-3.5V}{Phi}{general}{4.147}{2024}{mit}{open\_weights}{microsoft/Phi-3.5-vision-instruct}
	\modelmetaRow{16}{mPLUG-Owl3-1B-241014}{Owl3}{general}{0.924}{2024}{apache-2.0}{open\_weights}{mPLUG/mPLUG-Owl3-1B-241014}
	\modelmetaRow{17}{mPLUG-Owl3-2B-241014}{Owl3}{general}{1.977}{2024}{apache-2.0}{open\_weights}{mPLUG/mPLUG-Owl3-2B-241014}
	\modelmetaRow{18}{mPLUG-Owl3-7B-241101}{Owl3}{general}{8.073}{2024}{apache-2.0}{open\_weights}{mPLUG/mPLUG-Owl3-7B-241101}
	\modelmetaRow{19}{MiniCPM-V2}{MiniCPMV}{image-only}{3.435}{2024}{N/A}{open\_weights}{openbmb/MiniCPM-V-2}
	\modelmetaRow{20}{MiniCPM-V2.6}{MiniCPMV}{image-only}{8.099}{2024}{N/A}{open\_weights}{openbmb/MiniCPM-V-2\_6}
	\modelmetaRow{21}{Qwen-VL-Chat}{QwenVLChat}{image-only}{9.600}{2023}{Qwen License}{open\_weights}{Qwen/Qwen-VL-Chat}
	\modelmetaRow{22}{InternVL-Chat-V1-5-quantable}{InternVLChat}{image-only}{25.514}{2024}{mit}{open\_weights}{failspy/InternVL-Chat-V1-5-quantable}
	\modelmetaRow{23}{llava-interleave-qwen-7b-hf}{LLaVAInterleave}{general}{8.141}{2024}{other}{open\_weights}{llava-hf/llava-interleave-qwen-7b-hf}
	\modelmetaRow{24}{llava-interleave-qwen-7b-dpo-hf}{LLaVAInterleave}{general}{8.141}{2024}{other}{open\_weights}{llava-hf/llava-interleave-qwen-7b-dpo-hf}
	\modelmetaRow{25}{vila-1.5-3b}{VILAModel}{general}{3.000}{2024}{cc-by-nc-sa-4.0 (weights); code Apache-2.0}{open\_weights}{Efficient-Large-Model/VILA1.5-3b}
	\modelmetaRow{26}{vila-1.5-3b-s2}{VILAModel}{general}{3.000}{2024}{cc-by-nc-sa-4.0 (weights); code Apache-2.0}{open\_weights}{Efficient-Large-Model/VILA1.5-3b-s2}
	\modelmetaRow{27}{vila-1.5-8b}{VILAModel}{general}{8.000}{2024}{cc-by-nc-sa-4.0 (weights); code Apache-2.0}{open\_weights}{Efficient-Large-Model/Llama-3-VILA1.5-8B}
	\modelmetaRow{28}{vila-1.5-13b}{VILAModel}{general}{13.000}{2024}{cc-by-nc-sa-4.0 (weights); code Apache-2.0}{open\_weights}{Efficient-Large-Model/VILA1.5-13b}
	\modelmetaRow{29}{cambrian-8b}{Cambrian}{image-only}{8.333}{2024}{apache-2.0}{open\_weights}{nyu-visionx/cambrian-8b}
	\modelmetaRow{30}{paligemma2-3b}{PaliGemma2}{image-only}{3.033}{2024}{gemma}{open\_weights}{google/paligemma2-3b-ft-docci-448}
	\modelmetaRow{31}{paligemma2-10b}{PaliGemma2}{image-only}{9.664}{2024}{gemma}{open\_weights}{google/paligemma2-10b-ft-docci-448}
	\modelmetaRow{32}{LLaVA-NeXT-Video-7B-DPO-hf}{LLaVAVideo}{general}{7.063}{2024}{llama2}{open\_weights}{llava-hf/LLaVA-NeXT-Video-7B-DPO-hf}
	\modelmetaRow{33}{LLaVA-NeXT-Video-7B-hf}{LLaVAVideo}{general}{7.063}{2024}{llama2}{open\_weights}{llava-hf/LLaVA-NeXT-Video-7B-hf}
	\modelmetaRow{34}{MolmoE-1B}{Molmo}{image-only}{1.000}{2024}{apache-2.0}{open\_weights}{allenai/MolmoE-1B-0924}
	\modelmetaRow{35}{MolmoE-7B-O}{Molmo}{image-only}{7.665}{2024}{apache-2.0}{open\_weights}{allenai/Molmo-7B-O-0924}
	\modelmetaRow{36}{MolmoE-7B-D}{Molmo}{image-only}{8.021}{2024}{apache-2.0}{open\_weights}{allenai/Molmo-7B-D-0924}
	\modelmetaRow{37}{InternVL2-1B}{InternVLChat2}{general}{0.938}{2024}{mit}{open\_weights}{OpenGVLab/InternVL2-1B}
	\modelmetaRow{38}{InternVL2-2B}{InternVLChat2}{general}{2.206}{2024}{mit}{open\_weights}{OpenGVLab/InternVL2-2B}
	\modelmetaRow{39}{InternVL2-4B}{InternVLChat2}{general}{4.147}{2024}{mit}{open\_weights}{OpenGVLab/InternVL2-4B}
	\modelmetaRow{40}{InternVL2-8B}{InternVLChat2}{general}{8.075}{2024}{mit}{open\_weights}{OpenGVLab/InternVL2-8B}
	\modelmetaRow{41}{InternVL2-26B}{InternVLChat2}{general}{25.514}{2024}{mit}{open\_weights}{OpenGVLab/InternVL2-26B}
	\modelmetaRow{42}{InternVL2-40B}{InternVLChat2}{general}{40.069}{2024}{mit}{open\_weights}{OpenGVLab/InternVL2-40B}
	\modelmetaRow{43}{InternVL2-76B}{InternVLChat2}{general}{76.262}{2024}{llama3}{open\_weights}{OpenGVLab/InternVL2-Llama3-76B}
	\modelmetaRow{44}{InternVL2\_5-1B}{InternVLChat2}{general}{0.938}{2024}{mit}{open\_weights}{OpenGVLab/InternVL2\_5-1B}
	\modelmetaRow{45}{InternVL2\_5-2B}{InternVLChat2}{general}{2.206}{2024}{mit}{open\_weights}{OpenGVLab/InternVL2\_5-2B}
	\modelmetaRow{46}{InternVL2\_5-4B}{InternVLChat2}{general}{3.713}{2024}{mit}{open\_weights}{OpenGVLab/InternVL2\_5-4B}
	\modelmetaRow{47}{InternVL2\_5-8B}{InternVLChat2}{general}{8.075}{2024}{mit}{open\_weights}{OpenGVLab/InternVL2\_5-8B}
	\modelmetaRow{48}{InternVL2\_5-26B}{InternVLChat2}{general}{25.514}{2024}{mit}{open\_weights}{OpenGVLab/InternVL2\_5-26B}
	\modelmetaRow{49}{InternVL2\_5-38B}{InternVLChat2}{general}{38.388}{2024}{mit}{open\_weights}{OpenGVLab/InternVL2\_5-38B}
	\modelmetaRow{50}{InternVL2\_5-78B}{InternVLChat2}{general}{78.408}{2024}{other}{open\_weights}{OpenGVLab/InternVL2\_5-78B}
	\modelmetaRow{51}{Mantis-8B-Idefics2}{Mantis}{general}{8.403}{2024}{apache-2.0}{open\_weights}{TIGER-Lab/Mantis-8B-Idefics2}
	\modelmetaRow{52}{Mantis-llava-7b}{Mantis}{general}{7.063}{2024}{apache-2.0}{open\_weights}{TIGER-Lab/Mantis-llava-7b}
	\modelmetaRow{53}{Mantis-8B-siglip-llama3}{Mantis}{general}{8.480}{2024}{llama3}{open\_weights}{TIGER-Lab/Mantis-8B-siglip-llama3}
	\modelmetaRow{54}{Mantis-8B-clip-llama3}{Mantis}{general}{8.355}{2024}{llama3}{open\_weights}{TIGER-Lab/Mantis-8B-clip-llama3}
    \modelmetaRow{55}{GPT-5.5}{GPT}{general}{unknown}{2026}{}{}{Proprietary}
    \modelmetaRow{56}{Gemini-3.1-flash}{Gemini}{general}{unknown}{2026}{}{}{Proprietary}
\end{longtable}%
}

\subsection{Vision-Foundation Models}
{
In our study, we assess ten prominent VFMs, detailed in~\cref{tbl:vfm_details}.

For collision prediction, the objective is to estimate a binary mask of pixel locations where collisions occur. Performance is evaluated using the F1 score.

For gravity prediction, we ask the model to predict, for each pixel, (i) the gravity direction, evaluated using the Mean Angular Error (mAE), and (ii) the magnitude of the gravity force applied to the corresponding object, evaluated using the absolute Magnitude Error (magE).
We note that the predicted magnitude should depend on both the object’s mass and the strength of the gravitational field in the environment.
The task is non-trivial due to camera rotations and varying object masses.

As motion understanding is a key indicator of physical reasoning~\cite{huang2024vbench}, we additionally investigate the task of scene flow estimation, which is evaluated using the Average Endpoint Error (AEE).
}

\begin{table}[h!]
	\centering
	\begin{tabular}{lccc}
		\toprule
		\textbf{Model} & \textbf{Architecture} & \textbf{Supervision} & \textbf{Dataset} \\
		\midrule
		DeiT III       & ViT-B/16 & Classification & ImageNet-22k \\
		SAM            & ViT-B/16 & Segmentation   & SA-1B        \\
		MiDaS          & ViT-L/16 & Depth          & MIX-6        \\
		\midrule 
		MAE            & ViT-B/16 & SSL            & ImageNet-1k  \\
		DINO           & ViT-B/16 & SSL            & ImageNet-1k  \\
		DINOv2        & ViT-B/14 & SSL            & LVD-142M     \\
		\midrule
		CLIP           & ViT-B/16 & VLM            & WIT-400M     \\
		SigLIP         & ViT-B/16 & VLM            & WebLI        \\
		\midrule
		AM-Radio        & ViT-H/16 & Distillation & DataComp-1B \\
		\midrule
		StableDiffusion& UNet     & Generation     & LAION        \\
		\bottomrule
	\end{tabular}
	\caption{\textbf{Visual Foundation Models}' architecture, supervision types and training datasets}
	\label{tbl:vfm_details}
\end{table}
\twocolumn

{
    \small
    \bibliographystyle{ieeenat_fullname}
    \bibliography{main}
}

\end{document}